%% file: main.tex
\documentclass[10pt,twocolumn,letterpaper]{article}

\usepackage{iccv}
\usepackage{times}
\usepackage{epsfig}
\usepackage{graphicx}
\usepackage{amsmath}
\usepackage{amssymb}
\usepackage{url}
\usepackage[utf8]{inputenc} 
\usepackage[T1]{fontenc}    
\usepackage{booktabs}       
\usepackage{amsfonts}       
\usepackage{nicefrac}       
\usepackage{microtype}      
\usepackage{xcolor}         
\usepackage{hyphenat}
\usepackage{multirow}
\usepackage{tabularx}
\usepackage{float}
\usepackage{colortbl,dcolumn}
\usepackage{adjustbox}
\usepackage{subcaption}
\usepackage{bbm}
\input{math_commands.tex}

\DeclareMathOperator*{\argsort}{arg\,sort}
\usepackage[pagebackref=true,breaklinks=true,letterpaper=true,colorlinks,bookmarks=false]{hyperref}

\iccvfinalcopy 

\def\iccvPaperID{9058} 

\ificcvfinal\fi

\begin{document}

\title{Three New Validators and a Large-Scale Benchmark Ranking\\for Unsupervised Domain Adaptation}

\author{Kevin Musgrave\\
Cornell Tech \\
\and
Serge Belongie\\
University of Copenhagen\\
\and
Ser-Nam Lim\\
Meta AI
}

\maketitle
\ificcvfinal\thispagestyle{empty}\fi

\input{0_Abstract}
\input{1_Intro}
\input{2_Methodology}

\input{3_Results}


\input{supplementary}

\end{document}

%% file: math_commands.tex
\usepackage{amsmath,amsfonts,bm}

\def\eqref#1{equation~\ref{#1}}

\def\1{\bm{1}}

\DeclareMathAlphabet{\mathsfit}{\encodingdefault}{\sfdefault}{m}{sl}
\SetMathAlphabet{\mathsfit}{bold}{\encodingdefault}{\sfdefault}{bx}{n}

\DeclareMathOperator*{\argmax}{arg\,max}

%% file: 0_Abstract.tex
\begin{abstract}
Changes to hyperparameters can have a dramatic effect on model accuracy. Thus, the tuning of hyperparameters plays an important role in optimizing machine-learning models. An integral part of the hyperparameter-tuning process is the evaluation of model checkpoints, which is done through the use of ``validators''. In a supervised setting, these validators evaluate checkpoints by computing accuracy on a validation set that has labels. In contrast, in an unsupervised setting, the validation set has no such labels. Without any labels, it is impossible to compute accuracy, so validators must \textbf{estimate} accuracy instead. But what is the best approach to estimating accuracy? In this paper, we consider this question in the context of unsupervised domain adaptation (UDA). Specifically, we propose three new validators, and we compare and rank them against five other existing validators, on a large dataset of 1,000,000 checkpoints. Extensive experimental results show that two of our proposed validators achieve state-of-the-art performance in various settings. Finally, we find that in many cases, the state-of-the-art is obtained by a simple baseline method. To the best of our knowledge, this is the largest empirical study of UDA validators to date. Code is available at \href{https://www.github.com/KevinMusgrave/powerful-benchmarker}{github.com/KevinMusgrave/powerful-benchmarker}.
\end{abstract}

%% file: 1_Intro.tex
\section{Introduction}
Machine learning models are improving at a dramatic rate as a result of advances in model architectures \cite{He_2016_CVPR, Vaswani2017AttentionIA}, optimization \cite{kingma2014adam, Loshchilov2016SGDRSG}, training algorithms \cite{goodfellow2020generative, ho2020denoising}, and scaling \cite{Brown2020LanguageMA}. One commonality among these advances is that they require hyperparameter tuning to maximize model accuracy. Indeed, hyperparameter tuning is a crucial component of any machine-learning pipeline. 

Consider supervised learning, in which a curated dataset is first divided into training and validation samples. Training samples are used to iteratively update model parameters towards the goal of maximizing accuracy on a specific task. During this process, snapshots (``checkpoints'') of the model are periodically evaluated by measuring their accuracy on the validation samples. Typically, each training run is brought to an end as soon as checkpoint accuracy plateaus. Once training is complete, the best checkpoint is selected as the ``representative'' for the currently-used hyperparameters, and a new training run begins with a new set of hyperparameters. The process of trying out new hyperparameters is repeated as many times as desired, and at the end of this hyperparameter tuning, the best of the best checkpoints is selected as the final model.

Now consider a class of unsupervised learning where the validation samples are unlabeled. This setting is significantly more challenging, because without labels, accuracy on the validation set cannot be directly measured. Instead, only \textit{estimates} of accuracy are possible, where the estimates are expressed as a set of validation scores produced by a ``validator''. Ideally, the validation scores are perfectly correlated with actual accuracy, but in reality, the correlation may be low. A low correlation yields poor estimates of accuracy, resulting in the selection of sub-optimal checkpoints and hyperparameters. To avoid this, researchers and engineers need to use only the most reliable validators. But which validators are the most reliable? That is the question we attempt to answer in this paper, in the context of unsupervised domain adaptation (UDA). During UDA training, the model has access to both labeled ``source'' data and unlabeled ``target'' data. The goal is to maximize accuracy on the unlabeled target data, but it is impossible to directly measure this. Thus, UDA validators are needed to estimate target-domain accuracy.

Unfortunately, this is not a highly-researched subject. Most existing UDA papers are about algorithms, not validators. Furthermore, most of these papers select checkpoints using the ``oracle'' validator \cite{Musgrave2021UnsupervisedDA}, which directly computes target-domain accuracy by accessing target-domain labels. This violates the main assumption of UDA, which is that the target domain does not have labels at all. Hence, the oracle validator cannot be used in real-world applications. Moreover, existing papers that actually do analyze UDA validators, evaluate them on checkpoint sets that are too small and homogeneous to accurately reveal which validators are most reliable. 

In this paper, we present the largest empirical study of UDA validators to date:
\begin{itemize}
\setlength\itemsep{0em}
    \item We introduce three new validators, two of which achieve state-of-the-art performance in various settings.
    \item We benchmark and rank our proposed validators against five existing validators on a large dataset of 1,000,000 checkpoints generated by ten UDA algorithms and 100 hyperparameter settings per algorithm.
\end{itemize}

\noindent The paper is organized as follows: 
\begin{itemize}
 \setlength\itemsep{0em}
    \item Sections \ref{section:intro_existing_validation_methods} and \ref{section:intro_new_validation_methods} provide an overview of existing validators, and the three new validators that we propose. 
    \item Section \ref{section:section_experiment_methodology} explains our experiment methodology.
    \item Section \ref{section:results} presents our experiment results.
    \item The Appendix provides additional explanations, experiment details, and results.
\end{itemize}
\vspace{0.05em}

\subsection{Existing validators}\label{section:intro_existing_validation_methods}

\subsubsection*{Source accuracy} 
This is simply the model's accuracy on the source domain:
\begin{align}
\texttt{Accuracy} = \frac{1}{N}\sum_{i=1}^N{\mathbbm{1}(\argmax(p_i)=y_i)}
\end{align}
\noindent where $\mathbbm{1}$ is the indicator function, $N$ is the size of the source dataset, $p_i$ is the $i$th prediction vector, and $y_i$ is the label for the $i$th dataset sample. The assumption here is that the source and target domains are similar enough that high source-accuracy implies high target-accuracy.

\subsubsection*{Reverse validation \cite{reverseValidation, ganin2016domain}} 
This method consists of two steps. First it trains a model via UDA on the source ($S$) and target ($T$) data, and uses this model to create pseudo-labels for $T$. Next, it trains a reverse model via UDA on $T$ and $S$, where $T$ is the pseudo-labeled target data, and $S$ is the ``unlabeled'' source data. The final score is the accuracy of the reverse model on $S$. One disadvantage of this approach is that it trains two models, doubling the required training time, but still producing only a single usable model. Furthermore, all it does is make it easier to choose between training runs (i.e. for tuning hyperparameters). So selecting the best forward-model checkpoint requires using another validator that can compute scores per checkpoint.

\subsubsection*{Entropy} 
This measures the ``confidence'' of the model:
\begin{align}
\texttt{Entropy} = \frac{1}{N}\sum_{i=1}^{N}H(p_i)
\end{align}
\begin{align}
H(p_i) = -\sum_{j=1}^C p_{ij}\log{p_{ij}}
\end{align}
\noindent where $H(p_i)$ is the entropy of the $i$th prediction vector, $C$ is the number of classes, and $N$ is the size of the target dataset. An accurate model will output prediction vectors that have a single large value corresponding with the correct class for each sample. This produces a low entropy score, indicating high confidence. However, this method fails if a model is incorrectly confident. For example, the model might incorrectly classify all samples in the dataset as belonging to the same class.

\subsubsection*{Deep embedded validation (DEV) \cite{pmlr-v97-you19a}}
This computes a classification loss for every source validation sample, and weights each loss based on the probability that the sample belongs to the target domain. The probability comes from a domain classifier trained on source and target data.
\begin{align}
    \texttt{DEV} = \overline{L} + \eta\overline{W} - \eta
\end{align}
\begin{align}
    \eta = \frac{Cov(L,W)}{Var(W)}
\end{align}
where $L$ contains the weighted loss for each source validation sample, $W$ contains the weight of each loss, and $\overline{L}$ and $\overline{W}$ are the mean of $L$ and $W$ respectively. One practical issue with DEV is that its scores are unbounded. Very large values can occur if $W$ has low variance, or if $L$ and $W$ have high covariance.

\subsubsection*{Proxy risk \cite{chuang2020estimating}}
This method evaluates checkpoints using a ``check'' model. The check model predicts both class label and domain, and is trained on the transfer task using an algorithm such as DANN \cite{ganin2016domain}, with an additional ``disagreement'' loss term on the target samples:
\begin{align}
    D = \frac{1}{B}\sum_{i=1}^B -||x_i - m_i||_2
\end{align}
where $B$ is the batch size, $x_i$ and $m_i$ are the $i$th prediction vector of the checkpoint and the check model respectively, and $||.||_2$ is the L2 norm function. If the check model maintains a low DANN loss, but obtains outputs that differ from the checkpoint, then the checkpoint likely has low accuracy on the target domain. The disadvantage of this method is that it requires training a DANN-like model for every checkpoint, increasing total training time from $O(\texttt{epochs})$, to $O(\texttt{epochs}^2)$.

\subsubsection*{Ensemble-based model selection (EMS) \cite{robbiano2021adversarial}} 
This uses a linear regressor trained on 5 signals: target entropy, target diversity, silhoutte \& Calinski-Harabasz scores on the target features, source accuracy, and time-consistent pseudo-labels. EMS differs from other methods because it requires a dataset of \{signal, ground truth accuracy\} pairs to train the regressor. These pairs have to be collected by training a model on a domain adaptation task that has labeled target data. A drawback of this method is that the regressor may overfit and not generalize to our actual UDA task.

\subsubsection*{Soft neighborhood density (SND) \cite{saito2021tune}}
This computes the entropy of the softmaxed target similarity matrix:
\begin{align}
\texttt{SND} = H(\texttt{softmax}_{\tau}(\widehat{X}))
\end{align}
\begin{align}
X = F^TF 
\end{align}
\noindent where $H$ is the entropy function, $\texttt{softmax}_{\tau}$ is the softmax function with temperature $\tau$, $X$ is the similarity matrix, $F$ is the set of L2 normalized target feature vectors, and $\widehat{X}$ is $X$ with the diagonal entries removed. A high SND score means that each feature is close to many other features, which can indicate good clustering. The caveat of SND is that it assumes the model has not mapped all target features into a single cluster. A single cluster would result in a high SND score, but low accuracy.

\subsection{New validators}\label{section:intro_new_validation_methods}

Here we explain our proposed validators, which include modifications of existing methods.

\subsubsection*{Batch nuclear-norm maximization (BNM) \cite{cui2020towards}}
BNM is a UDA algorithm which aims to generate predictions that are both diverse and confident. It approaches this via singular value decomposition:
\begin{align}
\texttt{BNM} = ||P||_*
\end{align}
\noindent where $P$ is the $N\times C$ prediction matrix ($N$ is the dataset size and $C$ is the number of classes), and $||P||_*$ is the nuclear norm (the sum of the singular values) of $P$. This simple loss function is highly effective at training UDA models, which leads us to wonder if its numerical value is a proxy for target domain accuracy. We propose using BNM as a validator by applying the BNM loss function to all of the prediction vectors of the source and/or target domain. A drawback of BNM is that the computation can be expensive for large datasets with many classes, though fast approximations do exist \cite{Cui2021FastBN}.

\subsubsection*{ClassAMI}
\cite{robbiano2021adversarial} proposed using the silhouette score of the target features clustered with k-means (we call this ``ClassSS''). However, this approach is entirely dependent on the cluster labels. We propose computing the Adjusted Mutual Information (AMI) between cluster labels and the predicted labels, so that the model's predictions are accounted for:
\begin{align}
\texttt{ClassAMI} = \texttt{AMI}(X, \texttt{kmeans}(F).\texttt{labels})
\end{align}
\begin{align}
X_i = \argmax{p_i}
\end{align}
where $X$ is the predicted labels for the target data, $p_i$ is the $i$th prediction vector, and $F$ is the set of target features.

\subsubsection* {DEV with normalization (DEVN)}
One practical concern with DEV is that $\eta$ can become very large if $W$ has low variance, or if $L$ and $W$ have high covariance. To avoid this, we propose max-normalizing the weights:
\begin{align}
W_{max} = V - \overline{V} + 1 
\end{align}
\begin{align}
V = \frac{W}{\max{W}}
\end{align}
In the above equations, $W$ is the vector of unnormalized weights, and $\overline{V}$ is the mean of $V$.
    

%% file: 2_Methodology.tex
\section{Experiment Methodology}\label{section:section_experiment_methodology}

To allow for efficient benchmarking, we created a dataset of feature vectors extracted from model checkpoints, that could be easily loaded and used as input to all validators. At a high level, the benchmarking process consisted of these steps:

\begin{enumerate}
    \item Create a dataset of checkpoints:
    \begin{itemize}
        \item For each UDA algorithm, randomly create 100 hyperparameter settings.
        \item For each hyperparameter setting, train a model on a UDA task for a fixed number of iterations, and save a model checkpoint at regular intervals. Each checkpoint consists of only the features and logits of the model.
    \end{itemize}
    \item Calculate each validator's performance:
    \begin{itemize}
        \item For every checkpoint, compute the validator's score and the target-domain accuracy. (We are able to compute ground-truth accuracy because we have access to the labels. In a real-world application, the target-domain accuracy cannot be computed.)
        \item Compute a rank correlation between the validator's scores and the target-domain accuracies.
    \end{itemize}
\end{enumerate}

In the rest of this section, we will describe the above steps in detail.

\subsection{Creating the dataset of checkpoints}\label{section:methodology_creating_dataset}
We ran experiments on 31 transfer tasks:
\begin{itemize}
    \item \textbf{MNIST}: 1 task between MNIST and MNISTM \cite{ganin2016domain}.
    \item \textbf{Office31} \cite{saenko2010adapting}: 6 tasks between 3 domains (Amazon, DSLR, Webcam).
    \item \textbf{OfficeHome} \cite{venkateswara2017deep}: 12 tasks between 4 domains (Art, Clipart, Product, Real).
    \item \textbf{DomainNet126} \cite{peng2019moment, saito2019semi}: 12 tasks between 4 domains (Clipart, Painting, Real, Sketch).
\end{itemize}

For the MNIST$\rightarrow$MNISTM task, each training run used a LeNet-like model as the trunk, pretrained on MNIST. For Office31, OfficeHome, and DomainNet126, we used a ResNet50 \cite{He_2016_CVPR} pretrained \cite{rw2019timm} on ImageNet \cite{russakovsky2015imagenet}, and finetuned this model on every domain. Then for every task, we started each training run using the model finetuned on the source domain (i.e. the source-only model). We followed this procedure using 10 UDA algorithms (see Table \ref{algorithms_for_experiment}), all implemented in PyTorch \cite{pytorchPaper}.

For each UDA algorithm/task pair, we ran 100 steps of random hyperparameter search using Optuna \cite{optunaPaper}. This full search was run using two different feature layers (the 3rd and 2nd last layers), with the exception of DomainNet126, for which we used just the 2nd-last layer. Each training run lasted for a fixed number of epochs. Features and logits for both source and target datasets were saved at regular intervals, 20 times per training run. The final result was 1,000,000 checkpoints: 10 algorithms * 100 steps of hyperparameter search * 20 checkpoints per training run * (19 tasks * 2 feature layers + 12 tasks * 1 feature layer).

\input{tables/algorithms_for_experiment}

\subsection{Selecting the validators to benchmark}

We benchmarked the validators described in Sections \ref{section:intro_existing_validation_methods} and \ref{section:intro_new_validation_methods}, excluding those that are impractical to apply on a per-checkpoint basis. Computing scores per-checkpoint is preferred because it allows for faster feedback during training, and a greater likelihood of finding the optimal model. As well, it is how checkpoint selection is usually done in the supervised setting. The validation methods we excluded are:
\begin{itemize}
    \item Reverse validation, which is typically applied per training run rather than per checkpoint.
    \item Proxy risk, which requires training a full UDA model per checkpoint. This increases total training-time complexity to $O(\texttt{epochs}^2)$, and is therefore not practical to use on a large scale.
    \item EMS, which requires access to a separate dataset with ground-truth target-labels.
\end{itemize}

Each validator can have multiple variants by changing its parameters. For example, BNM, Entropy, ClassAMI, and ClassSS can be applied to the source, target, or both domains, and DEV, DEVN, and SND can be applied to features or logits. We tested 35 validator variations in total. (See the appendix for details.)

\subsection{Measuring validator performance}

Using our newly-created dataset of checkpoints, we need to measure each validator's checkpoint-ranking ability. An ideal validator will rank highly the checkpoints with the highest ground-truth target-domain accuracies. In other words, an ideal validator will achieve a high \textit{rank correlation} between its validation scores and the ground-truth target-domain accuracies. For this purpose, the Spearman rank correlation is one possible metric. However, the Spearman correlation treats all samples equally, whereas we are more interested in the samples with high validation scores or high target-domain accuracies.

For example, consider a hypothetical set of validation scores that are perfectly correlated with accuracy, with the exception of the highest score that breaks the trend and returns a model with 0\% accuracy. The set of scores with perfect correlation is useless, because ultimately, only the model with the highest validation score is selected. In this example, that model has 0\% accuracy. 

Thus, to account for this type of scenario, we use the weighted Spearman correlation (WSC) \cite{bailey2018weighted} to give more weight to the samples with high validation scores or high target-domain accuracies (see Figure \ref{correlation_comparison}). The weighted Spearman correlation is defined as:

\begin{align}
    \texttt{WSC} &= \frac{\sum_{i=1}^Nw_i(x_i-\widehat{x})(y_i-\widehat{y})}{\sqrt{\sum_{i=1}^N w_i(x_i-\widehat{x})^2 \sum_{i=1}^N w_i(y_i-\widehat{y})^2}} \label{eq:weighted_pearson}\\[5pt]
    \widehat{x} &= \frac{\sum_{i=1}^{N}w_ix_i}{\sum_{i=1}^{N}w_i} \label{eq:weighted_x}\\[5pt]
    \widehat{y} &= \frac{\sum_{i=1}^{N}w_iy_i}{\sum_{i=1}^{N}w_i} \label{eq:weighted_y}
\end{align}

\noindent where $w_i$ is the weight of each pair, $N$ is the number of pairs, and $x_i$ and $y_i$ are the weighted rank of sample $i$ in the $x$ and $y$ variables. Now let's compute the weighted rank for the $i$th sample. In this equation we will use $x$ as the variable, but the same equation applies for the $y$ variable:

\begin{align}
    x_i &= a_i + b_i \label{eq:weighted_rank}\\
    a_i &= \sum_{k=1}^N w_k \mathbbm{1}(\texttt{rank}_k < \texttt{rank}_i) \\
    b_i &= \frac{t+1}{2}\overline{w_i} \\
    \overline{w_i} &= \frac{1}{t}\sum_{k=1}^N w_k \mathbbm{1}(\texttt{rank}_k = \texttt{rank}_i)
\end{align}

\noindent $\mathbbm{1}$ is the indicator function, and $t$ is the number of samples that have the same rank as sample $i$. 

The above equations show how to compute the weighted ranks given a set of weights and ranks, but we still need a method for computing the weights themselves, $w_i$. To emphasize samples with a high validation score or high target-domain accuracy, we set $w_i$ as:

\begin{align}
    w_{i,v} &= \frac{\texttt{rank}(v(x_i))}{\underset{1 \leq k \leq N}{\max}{\texttt{rank}(v(x_k))}} \\
    w_{i,a} &= \frac{\texttt{rank}(a(x_i))}{\underset{1 \leq k \leq N}{\max}{\texttt{rank}(a(x_k))}} \\
    w_i &= \max(w_{i,v}, w_{i,a})^2
\end{align}

where 
\begin{itemize}
    \item $v(x_i)$ and $a(x_i)$ are the validation score and target-domain accuracy of sample $x_i$, respectively.
    \item $\texttt{rank}(.)$ is the integer rank obtained by dense-ranking all values, such that the lowest value has a rank of 1. 
    \item $\underset{1 \leq k \leq N}{\max}{\texttt{rank}(.)}$ is the maximum rank of all $N$ scores.
\end{itemize}

\noindent This formulation satisfies our goal of emphasizing samples with high validation scores or high target-domain accuracies. It does this by making $w$ increase quadratically with increasing validation-score ranks or increasing target-domain accuracy ranks.

\input{figures/correlation_comparison}

To measure a validator's performance across multiple transfer tasks, we use the average weighted Spearman correlation:

\begin{align}\label{eq:average_wsc_across_tasks}
    \texttt{Avg WSC across tasks} = \frac{1}{T} \sum_{i=1}^T \texttt{WSC}(A_i, V_i)
\end{align}
where $A_i$ is the set of target domain accuracies for task $i$, $V_i$ is the set of validator scores for task $i$, and $T$ is the number of tasks.

The comparison of algorithm/validator pairs is a special case where WSC cannot be used. For example, a pair with a high WSC could result in relatively low model accuracy because the algorithm itself performs poorly relative to other algorithms. Thus, for comparing algorithm/validator pairs, we use the average accuracy of each pair's top N training runs (AATN):

\begin{align}
    R_i &= \max_{v\in V_i}{v} \\
    S_i &= \argmax_{v\in V_i}{v} \\
    D &= \argsort(R) \\
    \texttt{Avg Acc of Top N} &= \frac{1}{N}\sum_{i=1}^N \texttt{Acc}(C_{D_i, S_{D_i}}) \label{eq:aatn}
\end{align}
where:
\begin{itemize}
 \setlength\itemsep{0em}
    \item $C_i$ is the $i$th training run, and $C_{i,j}$ is the $j$th checkpoint of $C_i$
    \item $V_i$ is the set of validation scores for $C_i$
    \item $R_i$ is the maximum validation score obtained in $C_i$
    \item $S_i$ is the index of $R_i$ within $C_i$
    \item $D$ contains the indices to sort $R$ in descending order
    \item \texttt{Acc} computes target-domain accuracy
\end{itemize}

Should we use AATN instead of WSC, to measure a validator's performance in general? No, for these reasons:
\begin{itemize}
\item WSC represents the complete behavior of each validator, because it takes all checkpoints into account, not just the top N.
\item Because it relies on more data, WSC is less affected by noise than AATN (see Figure \ref{figure:resilience_to_noise}). Consequently, if we repeated all our experiments, the WSC ranking of validators would remain quite consistent across our experiments. In contrast, the AATN ranking would vary considerably, as it is more affected by noise.
\item AATN requires a choice of N, and that choice can only be arbitrary, because there is no way to predict which N value will prove most useful.
\end{itemize}

\input{figures/resilience_to_noise}

%% file: tables/algorithms_for_experiment.tex
\begin{table}
\centering
\begin{tabular}{cc}
\toprule
Algorithm & Type of algorithm \\
 \midrule
\begin{tabular}{@{}l@{}} ATDOC \cite{liang2021domain} \end{tabular} & 
\begin{tabular}{@{}l@{}} Pseudo labeling \end{tabular} \\
\arrayrulecolor{gray}\hline
\begin{tabular}{@{}l@{}} BNM  \cite{cui2020towards} \\ BSP \cite{chen2019transferability} \end{tabular} & \begin{tabular}{@{}l@{}} SVD loss \end{tabular}\\
\hline
\begin{tabular}{@{}l@{}} CDAN \cite{long2017conditional} \\ DANN \cite{ganin2016domain} \\ GVB \cite{cui2020gradually}  \end{tabular} & 
\begin{tabular}{@{}l@{}} Adversarial \end{tabular}\\
\hline
\begin{tabular}{@{}l@{}} IM \cite{ICML2012Shi_566} \\ MCC \cite{jin2020minimum} \end{tabular} & 
\begin{tabular}{@{}l@{}} Info max \end{tabular}\\
\hline
\begin{tabular}{@{}l@{}} MCD \cite{saito2018maximum} \end{tabular} & 
\begin{tabular}{@{}l@{}} Multiple classifier discrepancy \end{tabular}\\
\hline
\begin{tabular}{@{}l@{}} MMD \cite{long2015learning} \end{tabular} &
\begin{tabular}{@{}l@{}} Feature distance \end{tabular}\\
\bottomrule
\end{tabular}
\caption{The 10 UDA algorithms used to create the dataset of feature vectors.}
\label{algorithms_for_experiment}
\end{table}

%% file: figures/correlation_comparison.tex
\begin{figure*}
     \centering
     \begin{subfigure}[b]{0.22\textwidth}
     \centering
     \includegraphics[width=1.0\textwidth]{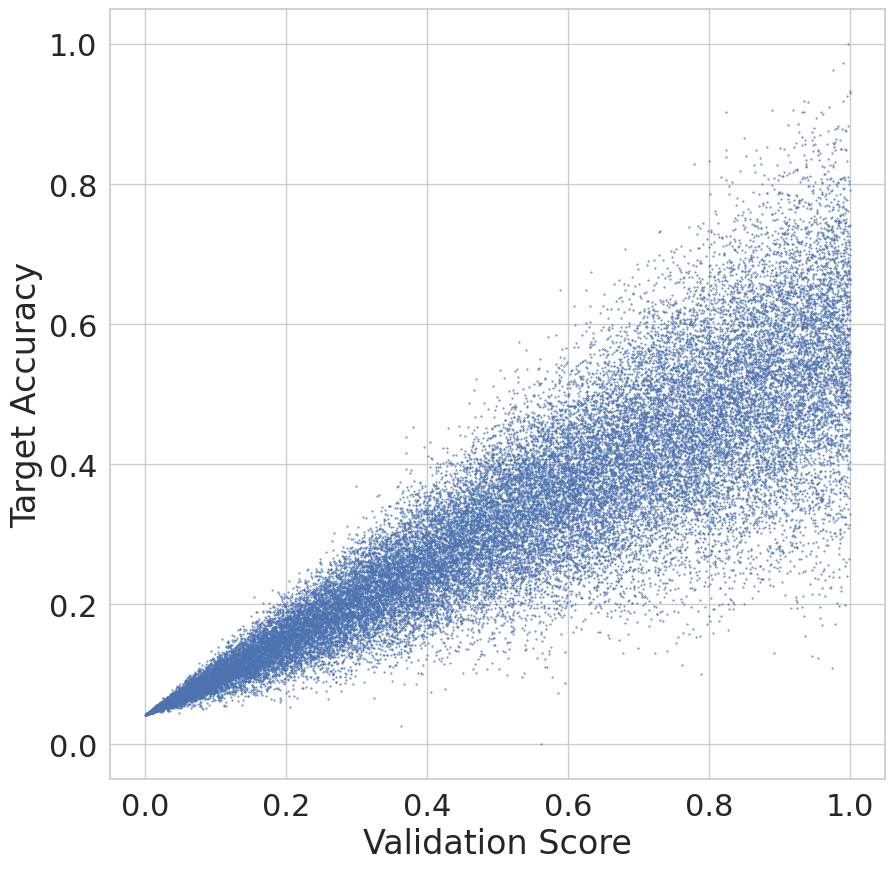}
     \caption{A synthetic example of a poor validator.\\ SC: \textbf{91.6}. WSC: \textbf{67.0}.}
     \label{correlation_comparison:synthetic_bad}
     \end{subfigure}
     \hspace{1em}
    \begin{subfigure}[b]{0.22\textwidth}
     \centering
     \includegraphics[width=1.0\textwidth]{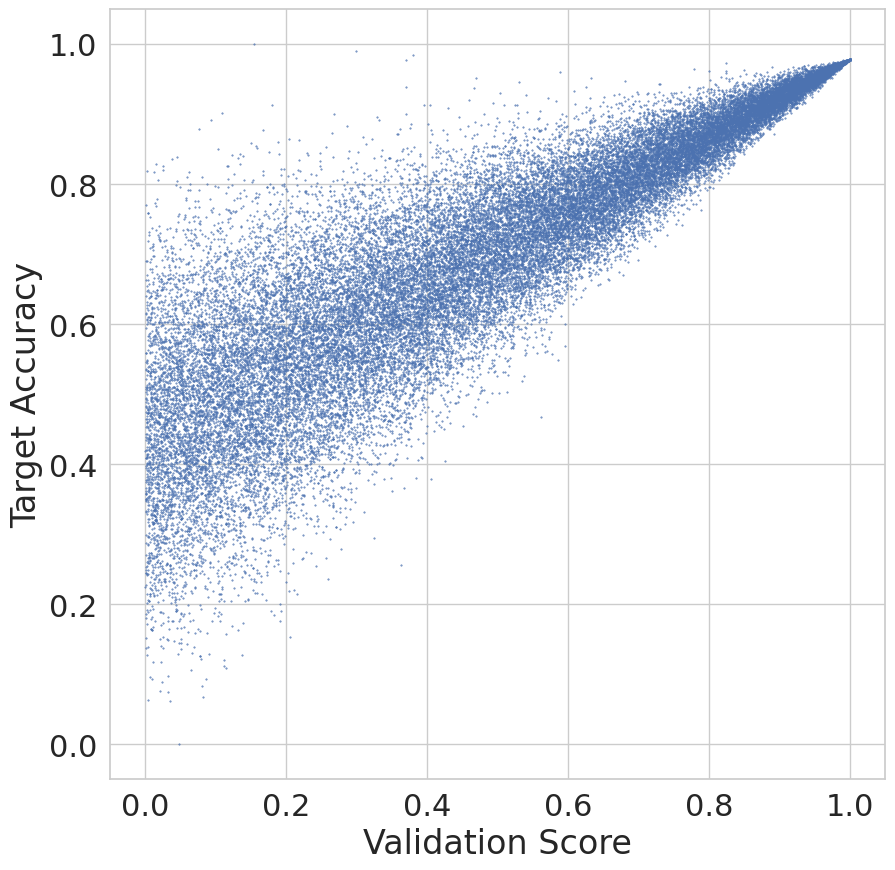}
     \caption{A synthetic example of a better validator.\\ SC: \textbf{91.6}. WSC: \textbf{93.9}.}
     \label{correlation_comparison:synthetic_good}
     \end{subfigure}
     \hspace{1em}
     \begin{subfigure}[b]{0.22\textwidth}
     \centering
     \includegraphics[width=1.0\textwidth]{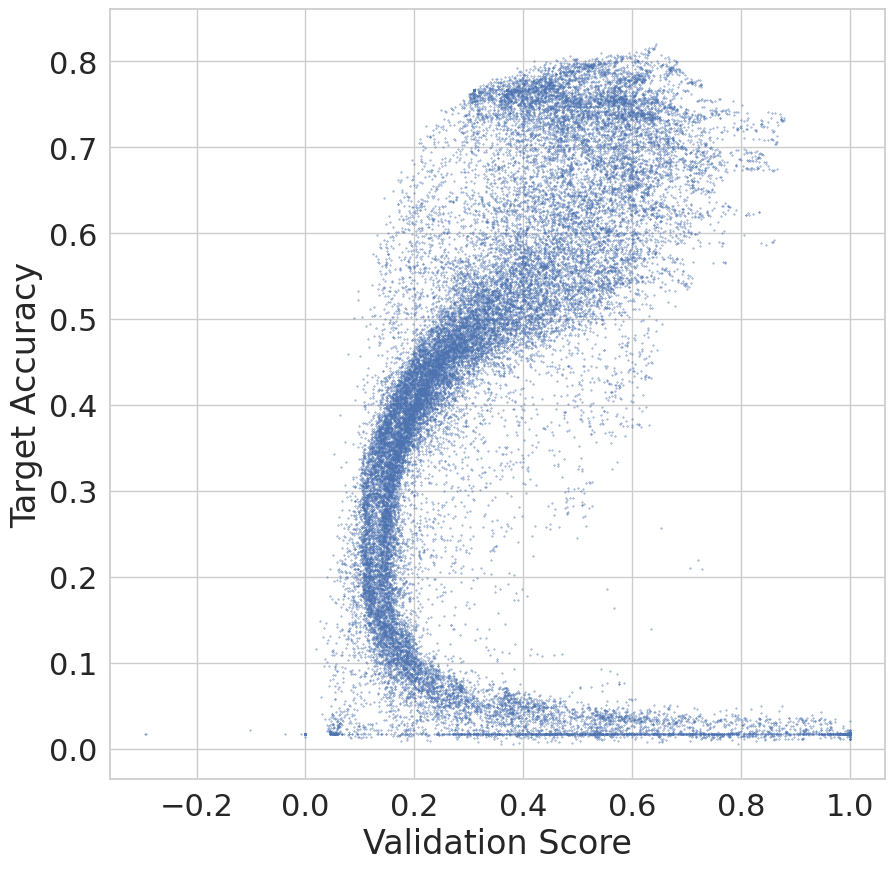}
     \caption{ClassSS validator, OfficeHome Art $\rightarrow$ Real.\\ SC: \textbf{58.5}. WSC: \textbf{0.0}.}
     \label{correlation_comparison:officehome_art_real}
     \end{subfigure}
     \hspace{1em}
    \begin{subfigure}[b]{0.22\textwidth}
     \centering
     \includegraphics[width=1.0\textwidth]{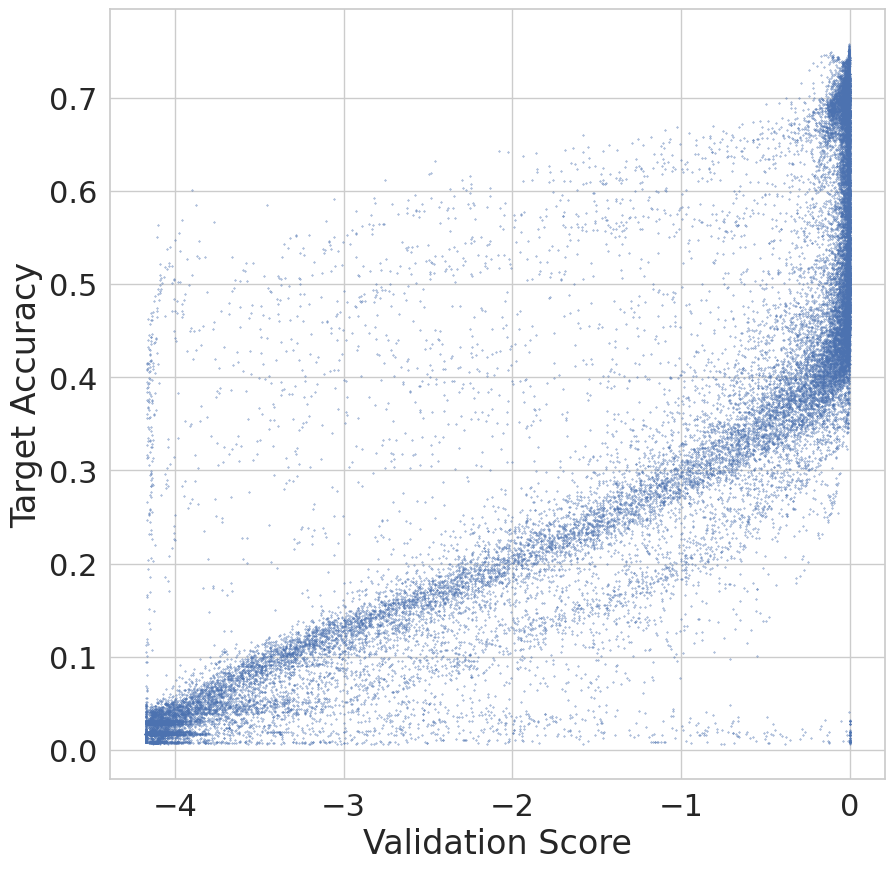}
     \caption{Entropy validator, OfficeHome Real $\rightarrow$ Art.\\ SC: \textbf{84.5}. WSC: \textbf{24.7}.}
     \label{correlation_comparison:officehome_real_product}
     \end{subfigure}
     \caption{These scatter plots show the advantage of the \textit{weighted} Spearman correlation (WSC) over the Spearman correlation (SC). The Spearman correlation gives roughly the same score for Figures \ref{correlation_comparison:synthetic_bad} and \ref{correlation_comparison:synthetic_good}. In contrast, our \textit{weighted} Spearman correlation gives Figure \ref{correlation_comparison:synthetic_bad} a much lower score, because there are many high validation scores corresponding with low accuracies. This means that during checkpoint selection, there is a high chance of selecting a low-accuracy checkpoint. In Figure \ref{correlation_comparison:officehome_art_real}, the worst accuracies correspond with the highest validation scores, and in Figure \ref{correlation_comparison:officehome_real_product}, accuracies ranging from 40\% to 70\% all have roughly the same validation score. The Spearman correlation treats all points equally and produces misleading scores for our purposes. In contrast, our \textit{weighted} Spearman correlation emphasizes the samples with high validation scores, and heavily penalizes these two examples.}
     \label{correlation_comparison}
\end{figure*}

%% file: figures/resilience_to_noise.tex
\begin{figure}
     \centering
     \includegraphics[width=\columnwidth]{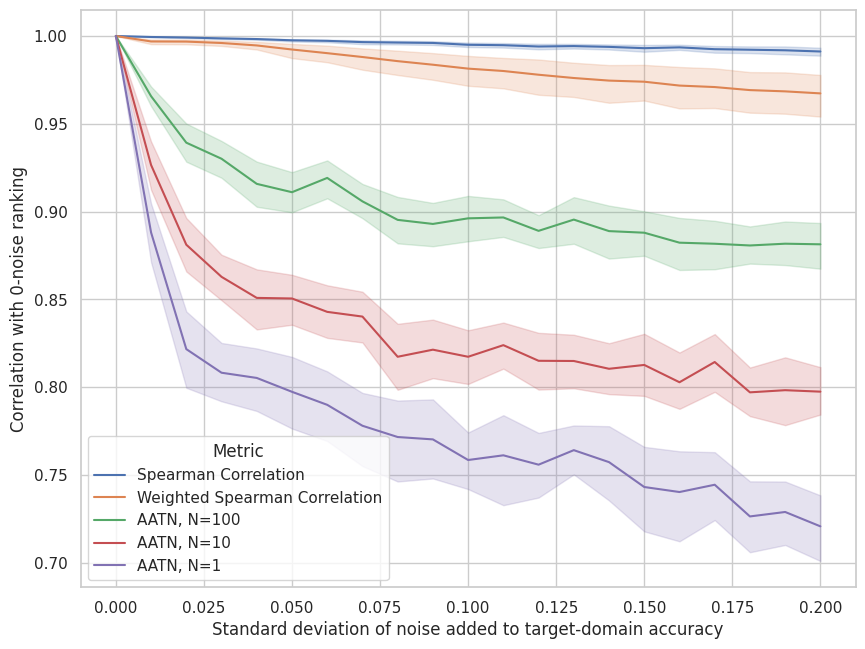}
     \caption{In this plot, the x-axis is the standard deviation of the random noise added to the target-domain accuracies. This represents the variance caused by randomness in experiments. The y-axis is the spearman correlation between the ranking of the validators in the noise-added setting and the ranking of the validators in the original setting. As more noise is added, the rankings obtained by WSC remain highly correlated with the original ranking. In contrast, AATN is significantly less consistent, even when N=100.}
     \label{figure:resilience_to_noise}
\end{figure}

%% file: 3_Results.tex
\section{Results}\label{section:results}

We followed the experiment methodology described in Section \ref{section:section_experiment_methodology} and found the following results: 

\begin{itemize}
 \setlength\itemsep{0em}
\item The top 3 algorithm/validator pairs all use the proposed ClassAMI validator (see Table \ref{best_validator_per_algorithm}).
\item The proposed DEVN validator outperforms DEV on datasets where source validation accuracy performs well (Office31 and OfficeHome). See Figure \ref{office31_officehome_domainnet126_barplots}.
\item The proposed BNM validator is the top validator on datasets where source validation accuracy performs poorly (MNIST and DomainNet126). See Figure \ref{office31_officehome_domainnet126_barplots}.
\item Source validation accuracy (the baseline method) has the highest average performance for six of the UDA algorithms, and two of the datasets (Office31 and OfficeHome). See Figure \ref{office31_officehome_domainnet126_barplots}.
\item SND consistently underperforms all of the other validators. See Figure \ref{office31_officehome_domainnet126_barplots}.
\item When using the oracle validator (a validator that can directly compute target domain accuracy), most UDA algorithms outperform the source-only model (see Table \ref{best_accuracy_per_adapter_5}). However, when using non-oracle validators, accuracy drops by several percentage points on average (see Table \ref{best_validator_per_algorithm}). As a result, the UDA algorithms sometimes only match or even degrade the accuracy of the source-only model (see Table \ref{best_accuracy_per_adapter_ranked_by_score_5}). In these cases, it is better to leave the model as-is, instead of training it using UDA algorithms.
\end{itemize}

\section{Conclusion}
Here are the main takeaways from our benchmark of UDA validators: 

\begin{itemize}
 \setlength\itemsep{0em}
\item Practitioners looking for the best model accuracy should use MCC, BNM, or IM, paired with the ClassAMI validator.
\item Researchers creating new UDA algorithms should use ClassAMI or source validation accuracy as the validator, because they are the best-performing validators on 9 out of 10 UDA algorithms. That said, a new UDA algorithm may require a different kind of validator if it is significantly different from the 10 algorithms that we tested.
\item Validators almost always pick sub-optimal checkpoints, which causes accuracy to drop from the oracle level by several percentage points on average. In some cases, this means that UDA algorithms perform worse than untrained models (see Table \ref{table:best_accuracies}). Thus, there is much room for improvement in the accuracy of UDA validators.
\end{itemize}

To unlock the full potential of UDA algorithms and models, more research is needed to improve validator accuracy and consistency. We hope our large-scale benchmark study, and our three new validators, will serve as a useful reference for future research in this area.

\input{figures/barplots/office31_officehome_domainnet126_barplots}
\input{tables/best_validator_per_algorithm}
\input{tables/best_accuracies}

%% file: figures/barplots/office31_officehome_domainnet126_barplots.tex
\begin{figure}
     \centering
      \begin{subfigure}[b]{0.9\columnwidth}
         \centering
         \includegraphics[width=1.0\textwidth]{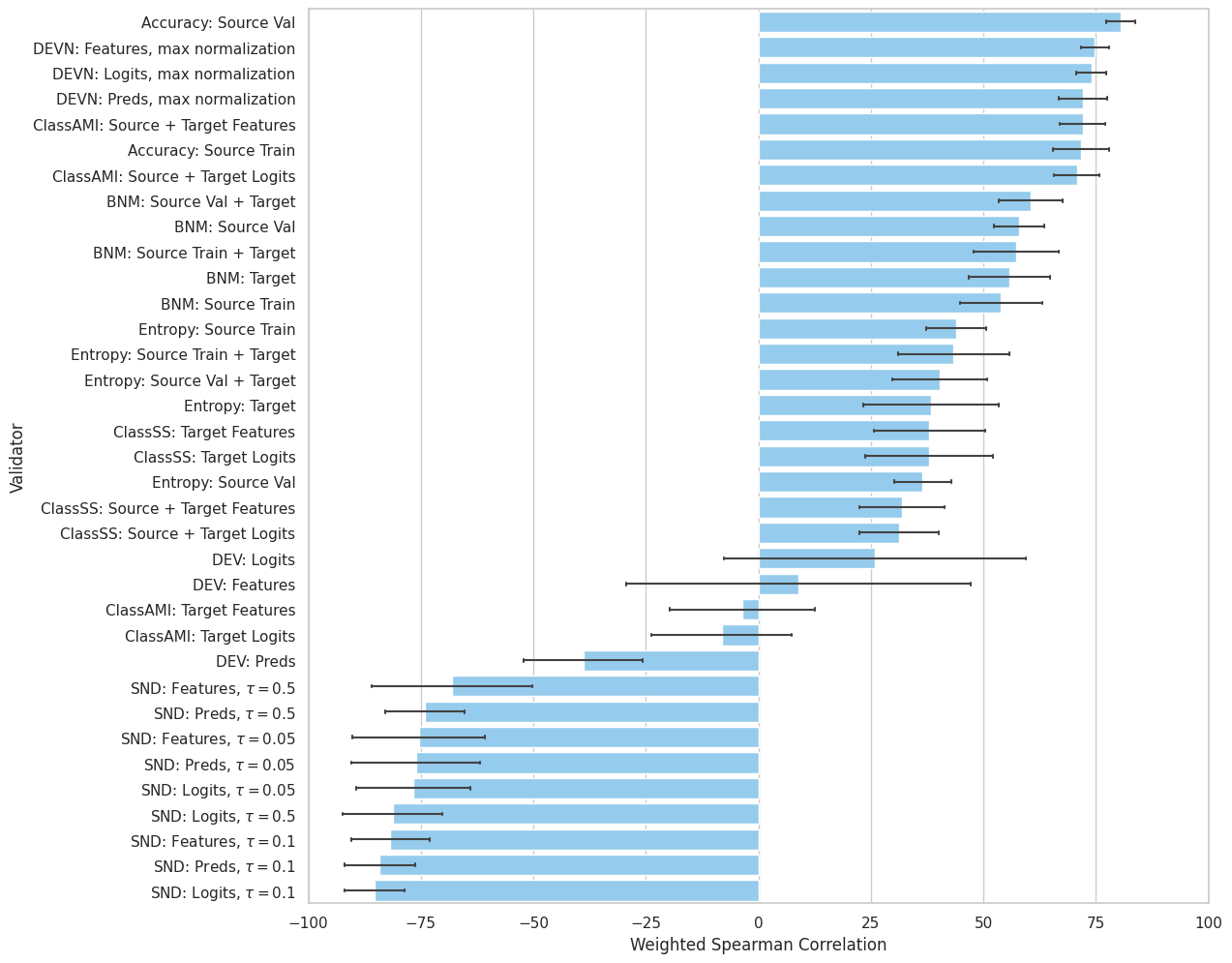}
         \caption{Office31}
     \end{subfigure}
     \\[1ex]
     \begin{subfigure}[b]{0.9\columnwidth}
         \centering
         \includegraphics[width=1.0\textwidth]{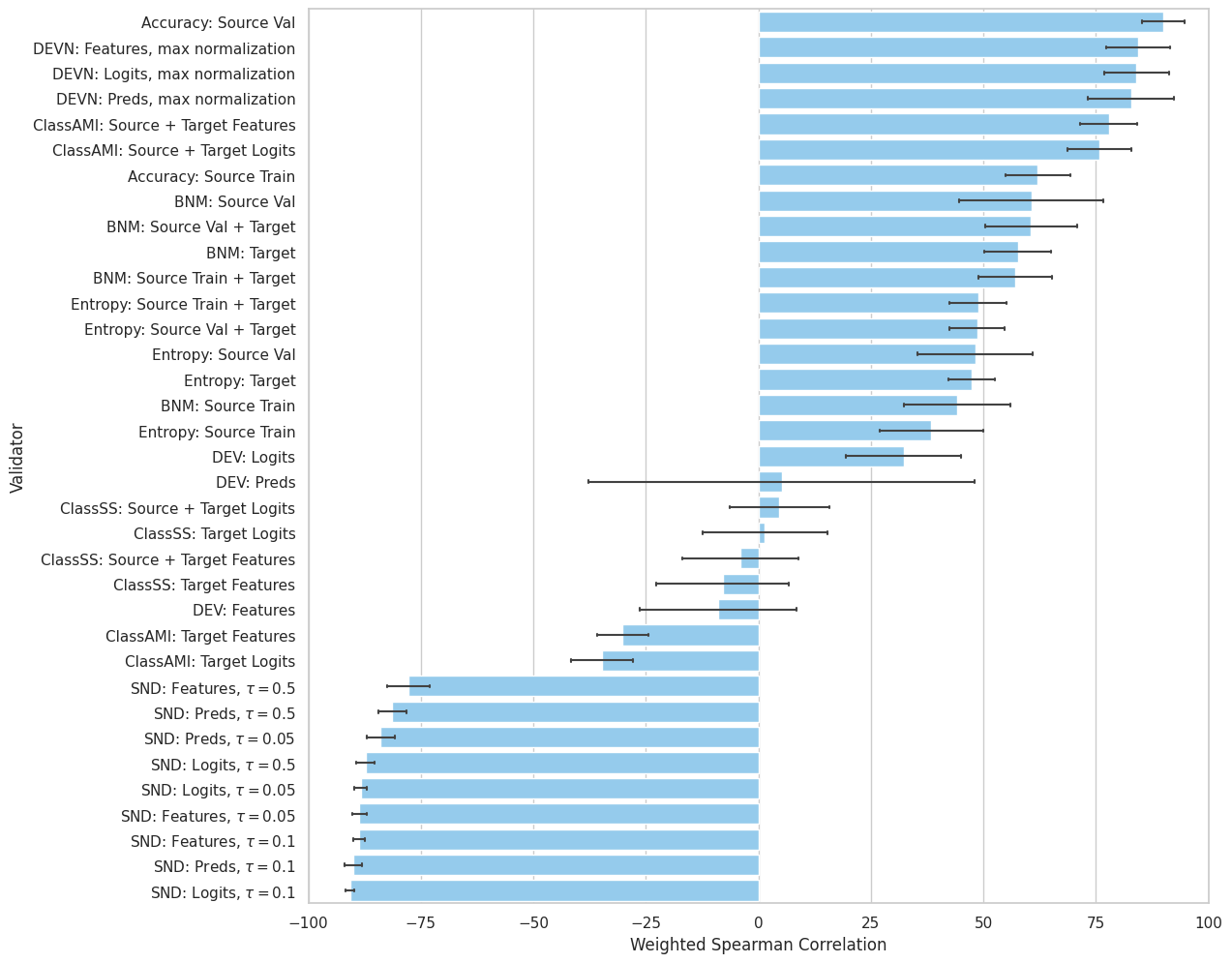}
         \caption{OfficeHome}
     \end{subfigure}
     \\[1ex]
     \begin{subfigure}[b]{0.9\columnwidth}
         \centering
         \includegraphics[width=1.0\textwidth]{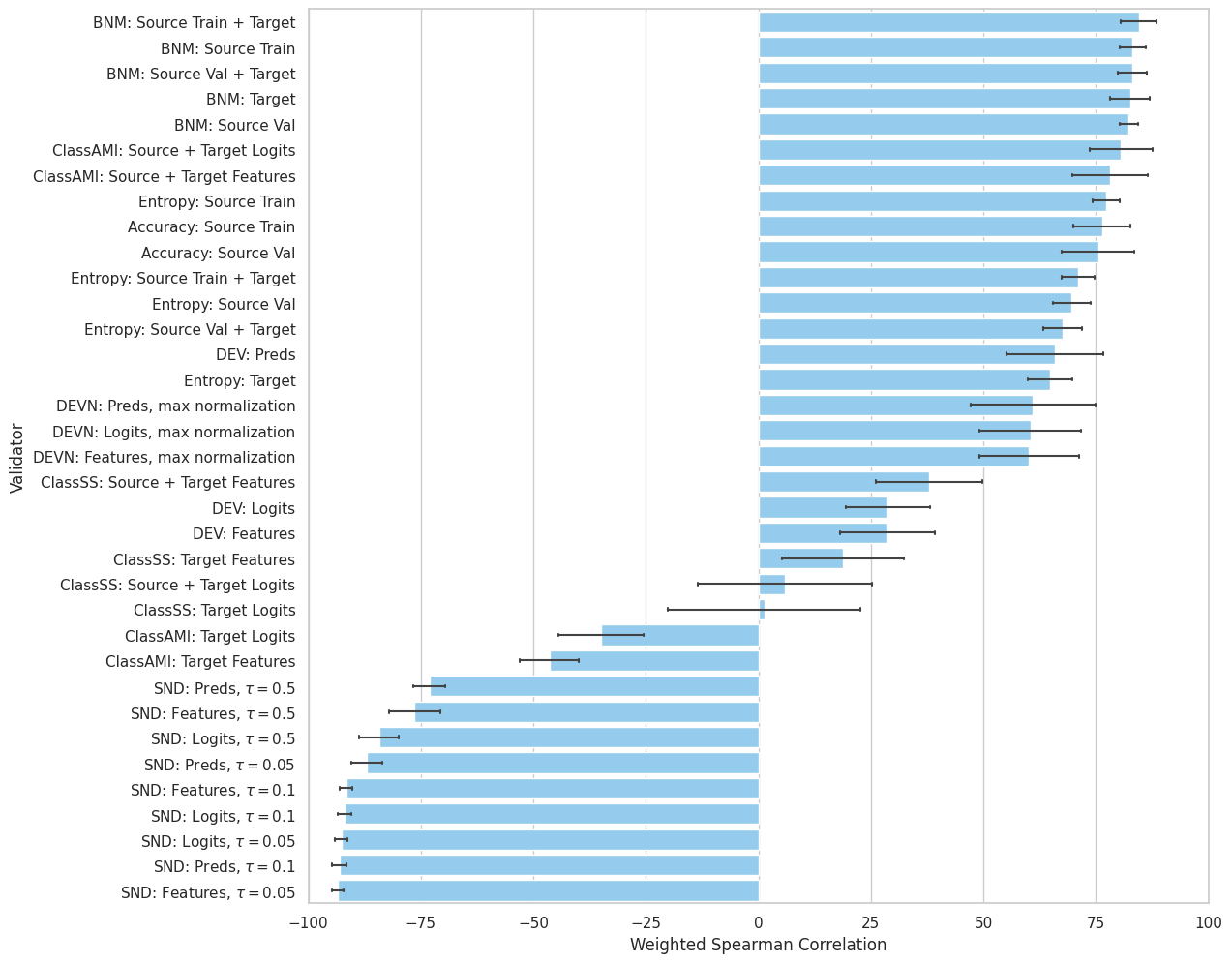}
        \caption{DomainNet126}
     \end{subfigure}
     \caption{Each validator's average WSC across tasks (equation \ref{eq:average_wsc_across_tasks}) within Office31, OfficeHome, and DomainNet126 (see the appendix for MNIST). The error bars represent the standard deviation across transfer tasks. The correlations are computed using the checkpoints of all UDA algorithms.}
    \label{office31_officehome_domainnet126_barplots}
\end{figure}

%% file: tables/best_validator_per_algorithm.tex
\begin{table}
\centering
\resizebox{0.9\columnwidth}{!}{\begin{tabular}{cccll}
\toprule
Algorithm & Validator & Validator Parameters & AATN & Val - Oracle \\
\midrule
MCC & ClassAMI & Source + Target Features & $68.6 \pm 12.6$ & $-2.3 \pm 2.1$ \\
BNM & ClassAMI & Source + Target Features & $66.5 \pm 12.6$ & $-2.7 \pm 2.3$ \\
IM & ClassAMI & Source + Target Features & $66.4 \pm 12.8$ & $-2.5 \pm 1.8$ \\
ATDOC & BNM & Target & $64.5 \pm 14.9$ & $-1.7 \pm 1.3$ \\
DANN & Accuracy & Source Val & $62.2 \pm 16.3$ & $-5.4 \pm 3.4$ \\
CDAN & Accuracy & Source Val & $62.1 \pm 16.1$ & $-5.3 \pm 3.5$ \\
GVB & Accuracy & Source Val & $61.7 \pm 15.9$ & $-6.3 \pm 4.3$ \\
MMD & Accuracy & Source Val & $61.5 \pm 15.7$ & $-5.0 \pm 2.9$ \\
MCD & Accuracy & Source Val & $60.7 \pm 16.5$ & $-4.2 \pm 4.7$ \\
BSP & Accuracy & Source Val & $59.8 \pm 16.4$ & $-3.3 \pm 5.2$ \\
\bottomrule
\end{tabular}}
\caption{This table shows the validator that scores the highest average WSC across Office31, OfficeHome, and DomainNet126, for each UDA algorithm. We exclude MNIST from the calculation because it is an outlier in terms of results, dataset attributes, and model architecture. The AATN column is the average AATN with $N=5$ (eq. \ref{eq:aatn}). The Val - Oracle column is the average drop in AATN when using the listed validator instead of the oracle validator.}
\label{best_validator_per_algorithm}
\end{table}

%% file: tables/best_accuracies.tex
\begin{table*}
    \input{tables/best_accuracy_per_adapter_5}
    \input{tables/best_accuracy_per_adapter_ranked_by_score_5}
    \input{tables/domainnet126_best_accuracy_per_adapter_5}
    \input{tables/domainnet126_best_accuracy_per_adapter_ranked_by_score_5}
 \caption{Even the best validators tend to pick sub-optimal checkpoints, which in many cases causes UDA algorithms to perform worse than untrained models. These tables show the accuracy of UDA algorithms when using the oracle validator (Tables \ref{best_accuracy_per_adapter_5} and \ref{domainnet126_best_accuracy_per_adapter_5}) and non-oracle validators (Tables \ref{best_accuracy_per_adapter_ranked_by_score_5} and \ref{domainnet126_best_accuracy_per_adapter_ranked_by_score_5}). Each value is the average target-domain accuracy of the top 5 training runs, as determined by the validator (see equation \ref{eq:aatn}). Green cells have an average accuracy greater than the source-only model. A stronger green color indicates higher accuracy. The color scheme is shared between the two tables, i.e. the colors in Tables \ref{best_accuracy_per_adapter_ranked_by_score_5} and \ref{domainnet126_best_accuracy_per_adapter_ranked_by_score_5} are on the same scale as in Tables \ref{best_accuracy_per_adapter_5} and \ref{domainnet126_best_accuracy_per_adapter_5} respectively. Bold indicates the highest value per column, per table. Bolding in Tables \ref{best_accuracy_per_adapter_ranked_by_score_5} and \ref{domainnet126_best_accuracy_per_adapter_ranked_by_score_5} is independent of the bolding in Tables \ref{best_accuracy_per_adapter_5} and \ref{domainnet126_best_accuracy_per_adapter_5}.}
\label{table:best_accuracies}
\end{table*}

%% file: tables/best_accuracy_per_adapter_5.tex
\def\bestaccuracyperadapterfiveMM#1{\ifdim#1pt>88.7pt\cellcolor{lime!100}\else\ifdim#1pt>85.3pt\cellcolor{lime!90}\else\ifdim#1pt>81.8pt\cellcolor{lime!80}\else\ifdim#1pt>78.4pt\cellcolor{lime!70}\else\ifdim#1pt>74.9pt\cellcolor{lime!60}\else\ifdim#1pt>71.4pt\cellcolor{lime!50}\else\ifdim#1pt>68.0pt\cellcolor{lime!40}\else\ifdim#1pt>64.5pt\cellcolor{lime!30}\else\ifdim#1pt>61.1pt\cellcolor{lime!20}\else\ifdim#1pt>57.6pt\cellcolor{lime!10}\else\cellcolor{lime!0}\fi\fi\fi\fi\fi\fi\fi\fi\fi\fi#1}

\def\bestaccuracyperadapterfiveAD#1{\ifdim#1pt>92.2pt\cellcolor{lime!100}\else\ifdim#1pt>91.2pt\cellcolor{lime!90}\else\ifdim#1pt>90.1pt\cellcolor{lime!80}\else\ifdim#1pt>89.1pt\cellcolor{lime!70}\else\ifdim#1pt>88.0pt\cellcolor{lime!60}\else\ifdim#1pt>86.9pt\cellcolor{lime!50}\else\ifdim#1pt>85.9pt\cellcolor{lime!40}\else\ifdim#1pt>84.8pt\cellcolor{lime!30}\else\ifdim#1pt>83.8pt\cellcolor{lime!20}\else\ifdim#1pt>82.7pt\cellcolor{lime!10}\else\cellcolor{lime!0}\fi\fi\fi\fi\fi\fi\fi\fi\fi\fi#1}

\def\bestaccuracyperadapterfiveAW#1{\ifdim#1pt>91.6pt\cellcolor{lime!100}\else\ifdim#1pt>90.3pt\cellcolor{lime!90}\else\ifdim#1pt>89.1pt\cellcolor{lime!80}\else\ifdim#1pt>87.8pt\cellcolor{lime!70}\else\ifdim#1pt>86.6pt\cellcolor{lime!60}\else\ifdim#1pt>85.4pt\cellcolor{lime!50}\else\ifdim#1pt>84.1pt\cellcolor{lime!40}\else\ifdim#1pt>82.9pt\cellcolor{lime!30}\else\ifdim#1pt>81.6pt\cellcolor{lime!20}\else\ifdim#1pt>80.4pt\cellcolor{lime!10}\else\cellcolor{lime!0}\fi\fi\fi\fi\fi\fi\fi\fi\fi\fi#1}

\def\bestaccuracyperadapterfiveDA#1{\ifdim#1pt>73.1pt\cellcolor{lime!100}\else\ifdim#1pt>72.7pt\cellcolor{lime!90}\else\ifdim#1pt>72.3pt\cellcolor{lime!80}\else\ifdim#1pt>71.9pt\cellcolor{lime!70}\else\ifdim#1pt>71.5pt\cellcolor{lime!60}\else\ifdim#1pt>71.2pt\cellcolor{lime!50}\else\ifdim#1pt>70.8pt\cellcolor{lime!40}\else\ifdim#1pt>70.4pt\cellcolor{lime!30}\else\ifdim#1pt>70.0pt\cellcolor{lime!20}\else\ifdim#1pt>69.6pt\cellcolor{lime!10}\else\cellcolor{lime!0}\fi\fi\fi\fi\fi\fi\fi\fi\fi\fi#1}

\def\bestaccuracyperadapterfiveDW#1{\ifdim#1pt>98.3pt\cellcolor{lime!100}\else\ifdim#1pt>97.9pt\cellcolor{lime!90}\else\ifdim#1pt>97.5pt\cellcolor{lime!80}\else\ifdim#1pt>97.0pt\cellcolor{lime!70}\else\ifdim#1pt>96.5pt\cellcolor{lime!60}\else\ifdim#1pt>96.1pt\cellcolor{lime!50}\else\ifdim#1pt>95.6pt\cellcolor{lime!40}\else\ifdim#1pt>95.2pt\cellcolor{lime!30}\else\ifdim#1pt>94.8pt\cellcolor{lime!20}\else\ifdim#1pt>94.3pt\cellcolor{lime!10}\else\cellcolor{lime!0}\fi\fi\fi\fi\fi\fi\fi\fi\fi\fi#1}

\def\bestaccuracyperadapterfiveWA#1{\ifdim#1pt>74.8pt\cellcolor{lime!100}\else\ifdim#1pt>74.5pt\cellcolor{lime!90}\else\ifdim#1pt>74.1pt\cellcolor{lime!80}\else\ifdim#1pt>73.7pt\cellcolor{lime!70}\else\ifdim#1pt>73.3pt\cellcolor{lime!60}\else\ifdim#1pt>73.0pt\cellcolor{lime!50}\else\ifdim#1pt>72.6pt\cellcolor{lime!40}\else\ifdim#1pt>72.2pt\cellcolor{lime!30}\else\ifdim#1pt>71.9pt\cellcolor{lime!20}\else\ifdim#1pt>71.5pt\cellcolor{lime!10}\else\cellcolor{lime!0}\fi\fi\fi\fi\fi\fi\fi\fi\fi\fi#1}

\def\bestaccuracyperadapterfiveWD#1{\ifdim#1pt>99.9pt\cellcolor{lime!100}\else\ifdim#1pt>99.8pt\cellcolor{lime!90}\else\ifdim#1pt>99.7pt\cellcolor{lime!80}\else\ifdim#1pt>99.6pt\cellcolor{lime!70}\else\ifdim#1pt>99.5pt\cellcolor{lime!60}\else\ifdim#1pt>99.4pt\cellcolor{lime!50}\else\ifdim#1pt>99.3pt\cellcolor{lime!40}\else\ifdim#1pt>99.2pt\cellcolor{lime!30}\else\ifdim#1pt>99.1pt\cellcolor{lime!20}\else\ifdim#1pt>99.0pt\cellcolor{lime!10}\else\cellcolor{lime!0}\fi\fi\fi\fi\fi\fi\fi\fi\fi\fi#1}

\def\bestaccuracyperadapterfiveAC#1{\ifdim#1pt>53.7pt\cellcolor{lime!100}\else\ifdim#1pt>52.3pt\cellcolor{lime!90}\else\ifdim#1pt>50.9pt\cellcolor{lime!80}\else\ifdim#1pt>49.5pt\cellcolor{lime!70}\else\ifdim#1pt>48.2pt\cellcolor{lime!60}\else\ifdim#1pt>46.8pt\cellcolor{lime!50}\else\ifdim#1pt>45.4pt\cellcolor{lime!40}\else\ifdim#1pt>44.0pt\cellcolor{lime!30}\else\ifdim#1pt>42.6pt\cellcolor{lime!20}\else\ifdim#1pt>41.2pt\cellcolor{lime!10}\else\cellcolor{lime!0}\fi\fi\fi\fi\fi\fi\fi\fi\fi\fi#1}

\def\bestaccuracyperadapterfiveAP#1{\ifdim#1pt>74.1pt\cellcolor{lime!100}\else\ifdim#1pt>73.5pt\cellcolor{lime!90}\else\ifdim#1pt>72.9pt\cellcolor{lime!80}\else\ifdim#1pt>72.3pt\cellcolor{lime!70}\else\ifdim#1pt>71.7pt\cellcolor{lime!60}\else\ifdim#1pt>71.0pt\cellcolor{lime!50}\else\ifdim#1pt>70.4pt\cellcolor{lime!40}\else\ifdim#1pt>69.8pt\cellcolor{lime!30}\else\ifdim#1pt>69.2pt\cellcolor{lime!20}\else\ifdim#1pt>68.6pt\cellcolor{lime!10}\else\cellcolor{lime!0}\fi\fi\fi\fi\fi\fi\fi\fi\fi\fi#1}

\def\bestaccuracyperadapterfiveAR#1{\ifdim#1pt>80.7pt\cellcolor{lime!100}\else\ifdim#1pt>80.2pt\cellcolor{lime!90}\else\ifdim#1pt>79.8pt\cellcolor{lime!80}\else\ifdim#1pt>79.3pt\cellcolor{lime!70}\else\ifdim#1pt>78.9pt\cellcolor{lime!60}\else\ifdim#1pt>78.5pt\cellcolor{lime!50}\else\ifdim#1pt>78.0pt\cellcolor{lime!40}\else\ifdim#1pt>77.6pt\cellcolor{lime!30}\else\ifdim#1pt>77.1pt\cellcolor{lime!20}\else\ifdim#1pt>76.7pt\cellcolor{lime!10}\else\cellcolor{lime!0}\fi\fi\fi\fi\fi\fi\fi\fi\fi\fi#1}

\def\bestaccuracyperadapterfiveCA#1{\ifdim#1pt>68.8pt\cellcolor{lime!100}\else\ifdim#1pt>67.8pt\cellcolor{lime!90}\else\ifdim#1pt>66.9pt\cellcolor{lime!80}\else\ifdim#1pt>65.9pt\cellcolor{lime!70}\else\ifdim#1pt>65.0pt\cellcolor{lime!60}\else\ifdim#1pt>64.0pt\cellcolor{lime!50}\else\ifdim#1pt>63.1pt\cellcolor{lime!40}\else\ifdim#1pt>62.1pt\cellcolor{lime!30}\else\ifdim#1pt>61.2pt\cellcolor{lime!20}\else\ifdim#1pt>60.2pt\cellcolor{lime!10}\else\cellcolor{lime!0}\fi\fi\fi\fi\fi\fi\fi\fi\fi\fi#1}

\def\bestaccuracyperadapterfiveCP#1{\ifdim#1pt>75.1pt\cellcolor{lime!100}\else\ifdim#1pt>74.2pt\cellcolor{lime!90}\else\ifdim#1pt>73.4pt\cellcolor{lime!80}\else\ifdim#1pt>72.6pt\cellcolor{lime!70}\else\ifdim#1pt>71.8pt\cellcolor{lime!60}\else\ifdim#1pt>70.9pt\cellcolor{lime!50}\else\ifdim#1pt>70.1pt\cellcolor{lime!40}\else\ifdim#1pt>69.3pt\cellcolor{lime!30}\else\ifdim#1pt>68.4pt\cellcolor{lime!20}\else\ifdim#1pt>67.6pt\cellcolor{lime!10}\else\cellcolor{lime!0}\fi\fi\fi\fi\fi\fi\fi\fi\fi\fi#1}

\def\bestaccuracyperadapterfiveCR#1{\ifdim#1pt>76.9pt\cellcolor{lime!100}\else\ifdim#1pt>76.2pt\cellcolor{lime!90}\else\ifdim#1pt>75.5pt\cellcolor{lime!80}\else\ifdim#1pt>74.8pt\cellcolor{lime!70}\else\ifdim#1pt>74.0pt\cellcolor{lime!60}\else\ifdim#1pt>73.3pt\cellcolor{lime!50}\else\ifdim#1pt>72.6pt\cellcolor{lime!40}\else\ifdim#1pt>71.9pt\cellcolor{lime!30}\else\ifdim#1pt>71.2pt\cellcolor{lime!20}\else\ifdim#1pt>70.5pt\cellcolor{lime!10}\else\cellcolor{lime!0}\fi\fi\fi\fi\fi\fi\fi\fi\fi\fi#1}

\def\bestaccuracyperadapterfivePA#1{\ifdim#1pt>67.5pt\cellcolor{lime!100}\else\ifdim#1pt>66.6pt\cellcolor{lime!90}\else\ifdim#1pt>65.8pt\cellcolor{lime!80}\else\ifdim#1pt>65.0pt\cellcolor{lime!70}\else\ifdim#1pt>64.2pt\cellcolor{lime!60}\else\ifdim#1pt>63.3pt\cellcolor{lime!50}\else\ifdim#1pt>62.5pt\cellcolor{lime!40}\else\ifdim#1pt>61.7pt\cellcolor{lime!30}\else\ifdim#1pt>60.8pt\cellcolor{lime!20}\else\ifdim#1pt>60.0pt\cellcolor{lime!10}\else\cellcolor{lime!0}\fi\fi\fi\fi\fi\fi\fi\fi\fi\fi#1}

\def\bestaccuracyperadapterfivePC#1{\ifdim#1pt>53.3pt\cellcolor{lime!100}\else\ifdim#1pt>52.2pt\cellcolor{lime!90}\else\ifdim#1pt>51.0pt\cellcolor{lime!80}\else\ifdim#1pt>49.8pt\cellcolor{lime!70}\else\ifdim#1pt>48.6pt\cellcolor{lime!60}\else\ifdim#1pt>47.5pt\cellcolor{lime!50}\else\ifdim#1pt>46.3pt\cellcolor{lime!40}\else\ifdim#1pt>45.1pt\cellcolor{lime!30}\else\ifdim#1pt>44.0pt\cellcolor{lime!20}\else\ifdim#1pt>42.8pt\cellcolor{lime!10}\else\cellcolor{lime!0}\fi\fi\fi\fi\fi\fi\fi\fi\fi\fi#1}

\def\bestaccuracyperadapterfivePR#1{\ifdim#1pt>82.0pt\cellcolor{lime!100}\else\ifdim#1pt>81.3pt\cellcolor{lime!90}\else\ifdim#1pt>80.7pt\cellcolor{lime!80}\else\ifdim#1pt>80.0pt\cellcolor{lime!70}\else\ifdim#1pt>79.4pt\cellcolor{lime!60}\else\ifdim#1pt>78.8pt\cellcolor{lime!50}\else\ifdim#1pt>78.1pt\cellcolor{lime!40}\else\ifdim#1pt>77.5pt\cellcolor{lime!30}\else\ifdim#1pt>76.8pt\cellcolor{lime!20}\else\ifdim#1pt>76.2pt\cellcolor{lime!10}\else\cellcolor{lime!0}\fi\fi\fi\fi\fi\fi\fi\fi\fi\fi#1}

\def\bestaccuracyperadapterfiveRA#1{\ifdim#1pt>74.6pt\cellcolor{lime!100}\else\ifdim#1pt>74.0pt\cellcolor{lime!90}\else\ifdim#1pt>73.3pt\cellcolor{lime!80}\else\ifdim#1pt>72.7pt\cellcolor{lime!70}\else\ifdim#1pt>72.0pt\cellcolor{lime!60}\else\ifdim#1pt>71.4pt\cellcolor{lime!50}\else\ifdim#1pt>70.8pt\cellcolor{lime!40}\else\ifdim#1pt>70.1pt\cellcolor{lime!30}\else\ifdim#1pt>69.5pt\cellcolor{lime!20}\else\ifdim#1pt>68.8pt\cellcolor{lime!10}\else\cellcolor{lime!0}\fi\fi\fi\fi\fi\fi\fi\fi\fi\fi#1}

\def\bestaccuracyperadapterfiveRC#1{\ifdim#1pt>56.8pt\cellcolor{lime!100}\else\ifdim#1pt>55.4pt\cellcolor{lime!90}\else\ifdim#1pt>54.1pt\cellcolor{lime!80}\else\ifdim#1pt>52.7pt\cellcolor{lime!70}\else\ifdim#1pt>51.4pt\cellcolor{lime!60}\else\ifdim#1pt>50.1pt\cellcolor{lime!50}\else\ifdim#1pt>48.7pt\cellcolor{lime!40}\else\ifdim#1pt>47.4pt\cellcolor{lime!30}\else\ifdim#1pt>46.0pt\cellcolor{lime!20}\else\ifdim#1pt>44.7pt\cellcolor{lime!10}\else\cellcolor{lime!0}\fi\fi\fi\fi\fi\fi\fi\fi\fi\fi#1}

\def\bestaccuracyperadapterfiveRP#1{\ifdim#1pt>83.1pt\cellcolor{lime!100}\else\ifdim#1pt>82.6pt\cellcolor{lime!90}\else\ifdim#1pt>82.2pt\cellcolor{lime!80}\else\ifdim#1pt>81.7pt\cellcolor{lime!70}\else\ifdim#1pt>81.3pt\cellcolor{lime!60}\else\ifdim#1pt>80.9pt\cellcolor{lime!50}\else\ifdim#1pt>80.4pt\cellcolor{lime!40}\else\ifdim#1pt>80.0pt\cellcolor{lime!30}\else\ifdim#1pt>79.5pt\cellcolor{lime!20}\else\ifdim#1pt>79.1pt\cellcolor{lime!10}\else\cellcolor{lime!0}\fi\fi\fi\fi\fi\fi\fi\fi\fi\fi#1}

\begin{subtable}{\textwidth}
\centering
\resizebox{0.85\textwidth}{!}{\begin{tabular}{lr|rrrrrr|rrrrrrrrrrrr}
\toprule
 & \multicolumn{1}{c|}{} & \multicolumn{6}{c|}{Office31} & \multicolumn{12}{c}{OfficeHome} \\
 & MM & AD & AW & DA & DW & WA & WD & AC & AP & AR & CA & CP & CR & PA & PC & PR & RA & RC & RP \\
\midrule
Source only & \bestaccuracyperadapterfiveMM{57.6} & \bestaccuracyperadapterfiveAD{82.7} & \bestaccuracyperadapterfiveAW{80.4} & \bestaccuracyperadapterfiveDA{69.6} & \bestaccuracyperadapterfiveDW{94.3} & \bestaccuracyperadapterfiveWA{71.5} & \bestaccuracyperadapterfiveWD{99.0} & \bestaccuracyperadapterfiveAC{41.2} & \bestaccuracyperadapterfiveAP{68.6} & \bestaccuracyperadapterfiveAR{76.7} & \bestaccuracyperadapterfiveCA{60.2} & \bestaccuracyperadapterfiveCP{67.6} & \bestaccuracyperadapterfiveCR{70.5} & \bestaccuracyperadapterfivePA{60.0} & \bestaccuracyperadapterfivePC{42.8} & \bestaccuracyperadapterfivePR{76.2} & \bestaccuracyperadapterfiveRA{68.8} & \bestaccuracyperadapterfiveRC{44.7} & \bestaccuracyperadapterfiveRP{79.1} \\
ATDOC & \bestaccuracyperadapterfiveMM{60.4} & \bestaccuracyperadapterfiveAD{88.8} & \bestaccuracyperadapterfiveAW{84.4} & \bestaccuracyperadapterfiveDA{72.1} & \bestaccuracyperadapterfiveDW{96.6} & \bestaccuracyperadapterfiveWA{72.3} & \bestaccuracyperadapterfiveWD{99.6} & \bestaccuracyperadapterfiveAC{47.5} & \bestaccuracyperadapterfiveAP{72.3} & \bestaccuracyperadapterfiveAR{76.8} & \bestaccuracyperadapterfiveCA{63.9} & \bestaccuracyperadapterfiveCP{74.6} & \bestaccuracyperadapterfiveCR{74.0} & \bestaccuracyperadapterfivePA{65.9} & \bestaccuracyperadapterfivePC{47.9} & \bestaccuracyperadapterfivePR{78.4} & \bestaccuracyperadapterfiveRA{73.6} & \bestaccuracyperadapterfiveRC{52.2} & \bestaccuracyperadapterfiveRP{80.9} \\
BNM & \bestaccuracyperadapterfiveMM{63.3} & \bestaccuracyperadapterfiveAD{89.3} & \bestaccuracyperadapterfiveAW{91.3} & \textbf{\bestaccuracyperadapterfiveDA{73.5}} & \bestaccuracyperadapterfiveDW{97.4} & \bestaccuracyperadapterfiveWA{74.8} & \textbf{\bestaccuracyperadapterfiveWD{100.0}} & \bestaccuracyperadapterfiveAC{52.3} & \bestaccuracyperadapterfiveAP{74.2} & \bestaccuracyperadapterfiveAR{79.9} & \bestaccuracyperadapterfiveCA{67.0} & \bestaccuracyperadapterfiveCP{74.2} & \bestaccuracyperadapterfiveCR{76.7} & \bestaccuracyperadapterfivePA{66.8} & \bestaccuracyperadapterfivePC{51.9} & \bestaccuracyperadapterfivePR{80.4} & \bestaccuracyperadapterfiveRA{72.6} & \bestaccuracyperadapterfiveRC{56.9} & \bestaccuracyperadapterfiveRP{81.9} \\
BSP & \bestaccuracyperadapterfiveMM{57.7} & \bestaccuracyperadapterfiveAD{85.3} & \bestaccuracyperadapterfiveAW{80.5} & \bestaccuracyperadapterfiveDA{69.7} & \bestaccuracyperadapterfiveDW{96.4} & \bestaccuracyperadapterfiveWA{71.5} & \bestaccuracyperadapterfiveWD{99.9} & \bestaccuracyperadapterfiveAC{43.9} & \bestaccuracyperadapterfiveAP{68.6} & \bestaccuracyperadapterfiveAR{76.7} & \bestaccuracyperadapterfiveCA{60.3} & \bestaccuracyperadapterfiveCP{67.6} & \bestaccuracyperadapterfiveCR{70.5} & \bestaccuracyperadapterfivePA{60.3} & \bestaccuracyperadapterfivePC{42.8} & \bestaccuracyperadapterfivePR{76.3} & \bestaccuracyperadapterfiveRA{69.6} & \bestaccuracyperadapterfiveRC{45.9} & \bestaccuracyperadapterfiveRP{79.1} \\
CDAN & \bestaccuracyperadapterfiveMM{91.6} & \bestaccuracyperadapterfiveAD{88.2} & \bestaccuracyperadapterfiveAW{91.3} & \bestaccuracyperadapterfiveDA{72.7} & \bestaccuracyperadapterfiveDW{96.6} & \bestaccuracyperadapterfiveWA{74.1} & \bestaccuracyperadapterfiveWD{99.8} & \bestaccuracyperadapterfiveAC{52.3} & \bestaccuracyperadapterfiveAP{70.9} & \bestaccuracyperadapterfiveAR{77.6} & \bestaccuracyperadapterfiveCA{62.6} & \bestaccuracyperadapterfiveCP{69.0} & \bestaccuracyperadapterfiveCR{72.5} & \bestaccuracyperadapterfivePA{64.0} & \bestaccuracyperadapterfivePC{53.2} & \bestaccuracyperadapterfivePR{79.9} & \bestaccuracyperadapterfiveRA{72.0} & \bestaccuracyperadapterfiveRC{57.1} & \bestaccuracyperadapterfiveRP{81.3} \\
DANN & \textbf{\bestaccuracyperadapterfiveMM{92.2}} & \bestaccuracyperadapterfiveAD{89.6} & \bestaccuracyperadapterfiveAW{92.0} & \bestaccuracyperadapterfiveDA{72.7} & \bestaccuracyperadapterfiveDW{97.1} & \bestaccuracyperadapterfiveWA{74.2} & \bestaccuracyperadapterfiveWD{99.9} & \bestaccuracyperadapterfiveAC{53.2} & \bestaccuracyperadapterfiveAP{71.3} & \bestaccuracyperadapterfiveAR{78.0} & \bestaccuracyperadapterfiveCA{63.0} & \bestaccuracyperadapterfiveCP{69.5} & \bestaccuracyperadapterfiveCR{72.6} & \bestaccuracyperadapterfivePA{64.6} & \bestaccuracyperadapterfivePC{52.6} & \bestaccuracyperadapterfivePR{79.7} & \bestaccuracyperadapterfiveRA{73.1} & \textbf{\bestaccuracyperadapterfiveRC{58.1}} & \bestaccuracyperadapterfiveRP{81.7} \\
GVB & \bestaccuracyperadapterfiveMM{78.8} & \bestaccuracyperadapterfiveAD{90.2} & \bestaccuracyperadapterfiveAW{91.7} & \bestaccuracyperadapterfiveDA{71.5} & \bestaccuracyperadapterfiveDW{95.9} & \bestaccuracyperadapterfiveWA{74.6} & \textbf{\bestaccuracyperadapterfiveWD{100.0}} & \bestaccuracyperadapterfiveAC{52.9} & \bestaccuracyperadapterfiveAP{70.4} & \bestaccuracyperadapterfiveAR{78.3} & \bestaccuracyperadapterfiveCA{65.4} & \bestaccuracyperadapterfiveCP{71.2} & \bestaccuracyperadapterfiveCR{74.5} & \bestaccuracyperadapterfivePA{64.9} & \bestaccuracyperadapterfivePC{53.4} & \bestaccuracyperadapterfivePR{81.1} & \bestaccuracyperadapterfiveRA{74.2} & \bestaccuracyperadapterfiveRC{56.9} & \bestaccuracyperadapterfiveRP{82.3} \\
IM & \bestaccuracyperadapterfiveMM{63.2} & \bestaccuracyperadapterfiveAD{89.1} & \bestaccuracyperadapterfiveAW{91.3} & \bestaccuracyperadapterfiveDA{72.8} & \bestaccuracyperadapterfiveDW{96.7} & \bestaccuracyperadapterfiveWA{75.0} & \bestaccuracyperadapterfiveWD{99.9} & \bestaccuracyperadapterfiveAC{52.7} & \bestaccuracyperadapterfiveAP{73.4} & \bestaccuracyperadapterfiveAR{80.3} & \bestaccuracyperadapterfiveCA{67.0} & \bestaccuracyperadapterfiveCP{74.1} & \bestaccuracyperadapterfiveCR{76.5} & \bestaccuracyperadapterfivePA{66.2} & \bestaccuracyperadapterfivePC{51.3} & \bestaccuracyperadapterfivePR{80.9} & \bestaccuracyperadapterfiveRA{73.1} & \bestaccuracyperadapterfiveRC{56.5} & \bestaccuracyperadapterfiveRP{81.9} \\
MCC & \bestaccuracyperadapterfiveMM{69.0} & \textbf{\bestaccuracyperadapterfiveAD{93.3}} & \textbf{\bestaccuracyperadapterfiveAW{92.8}} & \bestaccuracyperadapterfiveDA{73.2} & \bestaccuracyperadapterfiveDW{97.7} & \textbf{\bestaccuracyperadapterfiveWA{75.2}} & \textbf{\bestaccuracyperadapterfiveWD{100.0}} & \textbf{\bestaccuracyperadapterfiveAC{55.1}} & \textbf{\bestaccuracyperadapterfiveAP{74.7}} & \textbf{\bestaccuracyperadapterfiveAR{81.1}} & \textbf{\bestaccuracyperadapterfiveCA{69.7}} & \textbf{\bestaccuracyperadapterfiveCP{75.9}} & \textbf{\bestaccuracyperadapterfiveCR{77.6}} & \textbf{\bestaccuracyperadapterfivePA{68.3}} & \textbf{\bestaccuracyperadapterfivePC{54.5}} & \textbf{\bestaccuracyperadapterfivePR{82.6}} & \textbf{\bestaccuracyperadapterfiveRA{75.3}} & \bestaccuracyperadapterfiveRC{58.0} & \textbf{\bestaccuracyperadapterfiveRP{83.5}} \\
MCD & \bestaccuracyperadapterfiveMM{68.1} & \bestaccuracyperadapterfiveAD{88.1} & \bestaccuracyperadapterfiveAW{83.3} & \bestaccuracyperadapterfiveDA{69.6} & \textbf{\bestaccuracyperadapterfiveDW{98.8}} & \bestaccuracyperadapterfiveWA{71.5} & \textbf{\bestaccuracyperadapterfiveWD{100.0}} & \bestaccuracyperadapterfiveAC{43.3} & \bestaccuracyperadapterfiveAP{68.8} & \bestaccuracyperadapterfiveAR{76.7} & \bestaccuracyperadapterfiveCA{60.8} & \bestaccuracyperadapterfiveCP{68.5} & \bestaccuracyperadapterfiveCR{70.9} & \bestaccuracyperadapterfivePA{61.8} & \bestaccuracyperadapterfivePC{44.9} & \bestaccuracyperadapterfivePR{77.3} & \bestaccuracyperadapterfiveRA{72.2} & \bestaccuracyperadapterfiveRC{49.4} & \bestaccuracyperadapterfiveRP{80.6} \\
MMD & \bestaccuracyperadapterfiveMM{71.6} & \bestaccuracyperadapterfiveAD{87.8} & \bestaccuracyperadapterfiveAW{88.0} & \bestaccuracyperadapterfiveDA{72.1} & \bestaccuracyperadapterfiveDW{97.3} & \bestaccuracyperadapterfiveWA{72.1} & \textbf{\bestaccuracyperadapterfiveWD{100.0}} & \bestaccuracyperadapterfiveAC{50.5} & \bestaccuracyperadapterfiveAP{71.1} & \bestaccuracyperadapterfiveAR{77.4} & \bestaccuracyperadapterfiveCA{64.2} & \bestaccuracyperadapterfiveCP{69.7} & \bestaccuracyperadapterfiveCR{72.1} & \bestaccuracyperadapterfivePA{64.9} & \bestaccuracyperadapterfivePC{48.7} & \bestaccuracyperadapterfivePR{78.8} & \bestaccuracyperadapterfiveRA{72.3} & \bestaccuracyperadapterfiveRC{52.8} & \bestaccuracyperadapterfiveRP{80.5} \\
\bottomrule
\end{tabular}}
\caption{The accuracy of UDA algorithms when using the oracle validator.}
\label{best_accuracy_per_adapter_5}
\end{subtable}

%% file: tables/best_accuracy_per_adapter_ranked_by_score_5.tex
\def\bestaccuracyperadapterfiveMM#1{\ifdim#1pt>58.6pt\cellcolor{lime!100}\else\ifdim#1pt>58.5pt\cellcolor{lime!90}\else\ifdim#1pt>58.4pt\cellcolor{lime!80}\else\ifdim#1pt>58.3pt\cellcolor{lime!70}\else\ifdim#1pt>58.2pt\cellcolor{lime!60}\else\ifdim#1pt>58.0pt\cellcolor{lime!50}\else\ifdim#1pt>57.9pt\cellcolor{lime!40}\else\ifdim#1pt>57.8pt\cellcolor{lime!30}\else\ifdim#1pt>57.7pt\cellcolor{lime!20}\else\ifdim#1pt>57.6pt\cellcolor{lime!10}\else\cellcolor{lime!0}\fi\fi\fi\fi\fi\fi\fi\fi\fi\fi#1}

\def\bestaccuracyperadapterfiveAD#1{\ifdim#1pt>85.2pt\cellcolor{lime!100}\else\ifdim#1pt>84.9pt\cellcolor{lime!90}\else\ifdim#1pt>84.7pt\cellcolor{lime!80}\else\ifdim#1pt>84.4pt\cellcolor{lime!70}\else\ifdim#1pt>84.1pt\cellcolor{lime!60}\else\ifdim#1pt>83.8pt\cellcolor{lime!50}\else\ifdim#1pt>83.5pt\cellcolor{lime!40}\else\ifdim#1pt>83.3pt\cellcolor{lime!30}\else\ifdim#1pt>83.0pt\cellcolor{lime!20}\else\ifdim#1pt>82.7pt\cellcolor{lime!10}\else\cellcolor{lime!0}\fi\fi\fi\fi\fi\fi\fi\fi\fi\fi#1}

\def\bestaccuracyperadapterfiveAW#1{\ifdim#1pt>86.8pt\cellcolor{lime!100}\else\ifdim#1pt>86.1pt\cellcolor{lime!90}\else\ifdim#1pt>85.4pt\cellcolor{lime!80}\else\ifdim#1pt>84.7pt\cellcolor{lime!70}\else\ifdim#1pt>84.0pt\cellcolor{lime!60}\else\ifdim#1pt>83.2pt\cellcolor{lime!50}\else\ifdim#1pt>82.5pt\cellcolor{lime!40}\else\ifdim#1pt>81.8pt\cellcolor{lime!30}\else\ifdim#1pt>81.1pt\cellcolor{lime!20}\else\ifdim#1pt>80.4pt\cellcolor{lime!10}\else\cellcolor{lime!0}\fi\fi\fi\fi\fi\fi\fi\fi\fi\fi#1}

\def\bestaccuracyperadapterfiveDA#1{\ifdim#1pt>71.5pt\cellcolor{lime!100}\else\ifdim#1pt>71.3pt\cellcolor{lime!90}\else\ifdim#1pt>71.1pt\cellcolor{lime!80}\else\ifdim#1pt>70.9pt\cellcolor{lime!70}\else\ifdim#1pt>70.7pt\cellcolor{lime!60}\else\ifdim#1pt>70.4pt\cellcolor{lime!50}\else\ifdim#1pt>70.2pt\cellcolor{lime!40}\else\ifdim#1pt>70.0pt\cellcolor{lime!30}\else\ifdim#1pt>69.8pt\cellcolor{lime!20}\else\ifdim#1pt>69.6pt\cellcolor{lime!10}\else\cellcolor{lime!0}\fi\fi\fi\fi\fi\fi\fi\fi\fi\fi#1}

\def\bestaccuracyperadapterfiveDW#1{\ifdim#1pt>95.6pt\cellcolor{lime!100}\else\ifdim#1pt>95.4pt\cellcolor{lime!90}\else\ifdim#1pt>95.3pt\cellcolor{lime!80}\else\ifdim#1pt>95.1pt\cellcolor{lime!70}\else\ifdim#1pt>95.0pt\cellcolor{lime!60}\else\ifdim#1pt>94.9pt\cellcolor{lime!50}\else\ifdim#1pt>94.7pt\cellcolor{lime!40}\else\ifdim#1pt>94.6pt\cellcolor{lime!30}\else\ifdim#1pt>94.4pt\cellcolor{lime!20}\else\ifdim#1pt>94.3pt\cellcolor{lime!10}\else\cellcolor{lime!0}\fi\fi\fi\fi\fi\fi\fi\fi\fi\fi#1}

\def\bestaccuracyperadapterfiveWA#1{\ifdim#1pt>72.7pt\cellcolor{lime!100}\else\ifdim#1pt>72.5pt\cellcolor{lime!90}\else\ifdim#1pt>72.4pt\cellcolor{lime!80}\else\ifdim#1pt>72.3pt\cellcolor{lime!70}\else\ifdim#1pt>72.2pt\cellcolor{lime!60}\else\ifdim#1pt>72.0pt\cellcolor{lime!50}\else\ifdim#1pt>71.9pt\cellcolor{lime!40}\else\ifdim#1pt>71.8pt\cellcolor{lime!30}\else\ifdim#1pt>71.6pt\cellcolor{lime!20}\else\ifdim#1pt>71.5pt\cellcolor{lime!10}\else\cellcolor{lime!0}\fi\fi\fi\fi\fi\fi\fi\fi\fi\fi#1}

\def\bestaccuracyperadapterfiveWD#1{\ifdim#1pt>99.9pt\cellcolor{lime!100}\else\ifdim#1pt>99.8pt\cellcolor{lime!90}\else\ifdim#1pt>99.7pt\cellcolor{lime!80}\else\ifdim#1pt>99.6pt\cellcolor{lime!70}\else\ifdim#1pt>99.5pt\cellcolor{lime!60}\else\ifdim#1pt>99.4pt\cellcolor{lime!50}\else\ifdim#1pt>99.3pt\cellcolor{lime!40}\else\ifdim#1pt>99.2pt\cellcolor{lime!30}\else\ifdim#1pt>99.1pt\cellcolor{lime!20}\else\ifdim#1pt>99.0pt\cellcolor{lime!10}\else\cellcolor{lime!0}\fi\fi\fi\fi\fi\fi\fi\fi\fi\fi#1}

\def\bestaccuracyperadapterfiveAC#1{\ifdim#1pt>51.2pt\cellcolor{lime!100}\else\ifdim#1pt>50.1pt\cellcolor{lime!90}\else\ifdim#1pt>49.0pt\cellcolor{lime!80}\else\ifdim#1pt>47.9pt\cellcolor{lime!70}\else\ifdim#1pt>46.8pt\cellcolor{lime!60}\else\ifdim#1pt>45.6pt\cellcolor{lime!50}\else\ifdim#1pt>44.5pt\cellcolor{lime!40}\else\ifdim#1pt>43.4pt\cellcolor{lime!30}\else\ifdim#1pt>42.3pt\cellcolor{lime!20}\else\ifdim#1pt>41.2pt\cellcolor{lime!10}\else\cellcolor{lime!0}\fi\fi\fi\fi\fi\fi\fi\fi\fi\fi#1}

\def\bestaccuracyperadapterfiveAP#1{\ifdim#1pt>72.4pt\cellcolor{lime!100}\else\ifdim#1pt>72.0pt\cellcolor{lime!90}\else\ifdim#1pt>71.5pt\cellcolor{lime!80}\else\ifdim#1pt>71.1pt\cellcolor{lime!70}\else\ifdim#1pt>70.7pt\cellcolor{lime!60}\else\ifdim#1pt>70.3pt\cellcolor{lime!50}\else\ifdim#1pt>69.9pt\cellcolor{lime!40}\else\ifdim#1pt>69.4pt\cellcolor{lime!30}\else\ifdim#1pt>69.0pt\cellcolor{lime!20}\else\ifdim#1pt>68.6pt\cellcolor{lime!10}\else\cellcolor{lime!0}\fi\fi\fi\fi\fi\fi\fi\fi\fi\fi#1}

\def\bestaccuracyperadapterfiveAR#1{\ifdim#1pt>79.8pt\cellcolor{lime!100}\else\ifdim#1pt>79.4pt\cellcolor{lime!90}\else\ifdim#1pt>79.1pt\cellcolor{lime!80}\else\ifdim#1pt>78.7pt\cellcolor{lime!70}\else\ifdim#1pt>78.4pt\cellcolor{lime!60}\else\ifdim#1pt>78.1pt\cellcolor{lime!50}\else\ifdim#1pt>77.7pt\cellcolor{lime!40}\else\ifdim#1pt>77.4pt\cellcolor{lime!30}\else\ifdim#1pt>77.0pt\cellcolor{lime!20}\else\ifdim#1pt>76.7pt\cellcolor{lime!10}\else\cellcolor{lime!0}\fi\fi\fi\fi\fi\fi\fi\fi\fi\fi#1}

\def\bestaccuracyperadapterfiveCA#1{\ifdim#1pt>62.5pt\cellcolor{lime!100}\else\ifdim#1pt>62.3pt\cellcolor{lime!90}\else\ifdim#1pt>62.0pt\cellcolor{lime!80}\else\ifdim#1pt>61.8pt\cellcolor{lime!70}\else\ifdim#1pt>61.5pt\cellcolor{lime!60}\else\ifdim#1pt>61.2pt\cellcolor{lime!50}\else\ifdim#1pt>61.0pt\cellcolor{lime!40}\else\ifdim#1pt>60.7pt\cellcolor{lime!30}\else\ifdim#1pt>60.5pt\cellcolor{lime!20}\else\ifdim#1pt>60.2pt\cellcolor{lime!10}\else\cellcolor{lime!0}\fi\fi\fi\fi\fi\fi\fi\fi\fi\fi#1}

\def\bestaccuracyperadapterfiveCP#1{\ifdim#1pt>72.5pt\cellcolor{lime!100}\else\ifdim#1pt>72.0pt\cellcolor{lime!90}\else\ifdim#1pt>71.4pt\cellcolor{lime!80}\else\ifdim#1pt>70.9pt\cellcolor{lime!70}\else\ifdim#1pt>70.3pt\cellcolor{lime!60}\else\ifdim#1pt>69.8pt\cellcolor{lime!50}\else\ifdim#1pt>69.2pt\cellcolor{lime!40}\else\ifdim#1pt>68.7pt\cellcolor{lime!30}\else\ifdim#1pt>68.1pt\cellcolor{lime!20}\else\ifdim#1pt>67.6pt\cellcolor{lime!10}\else\cellcolor{lime!0}\fi\fi\fi\fi\fi\fi\fi\fi\fi\fi#1}

\def\bestaccuracyperadapterfiveCR#1{\ifdim#1pt>74.5pt\cellcolor{lime!100}\else\ifdim#1pt>74.0pt\cellcolor{lime!90}\else\ifdim#1pt>73.6pt\cellcolor{lime!80}\else\ifdim#1pt>73.1pt\cellcolor{lime!70}\else\ifdim#1pt>72.7pt\cellcolor{lime!60}\else\ifdim#1pt>72.3pt\cellcolor{lime!50}\else\ifdim#1pt>71.8pt\cellcolor{lime!40}\else\ifdim#1pt>71.4pt\cellcolor{lime!30}\else\ifdim#1pt>70.9pt\cellcolor{lime!20}\else\ifdim#1pt>70.5pt\cellcolor{lime!10}\else\cellcolor{lime!0}\fi\fi\fi\fi\fi\fi\fi\fi\fi\fi#1}

\def\bestaccuracyperadapterfivePA#1{\ifdim#1pt>64.0pt\cellcolor{lime!100}\else\ifdim#1pt>63.6pt\cellcolor{lime!90}\else\ifdim#1pt>63.1pt\cellcolor{lime!80}\else\ifdim#1pt>62.7pt\cellcolor{lime!70}\else\ifdim#1pt>62.2pt\cellcolor{lime!60}\else\ifdim#1pt>61.8pt\cellcolor{lime!50}\else\ifdim#1pt>61.4pt\cellcolor{lime!40}\else\ifdim#1pt>60.9pt\cellcolor{lime!30}\else\ifdim#1pt>60.5pt\cellcolor{lime!20}\else\ifdim#1pt>60.0pt\cellcolor{lime!10}\else\cellcolor{lime!0}\fi\fi\fi\fi\fi\fi\fi\fi\fi\fi#1}

\def\bestaccuracyperadapterfivePC#1{\ifdim#1pt>52.2pt\cellcolor{lime!100}\else\ifdim#1pt>51.1pt\cellcolor{lime!90}\else\ifdim#1pt>50.1pt\cellcolor{lime!80}\else\ifdim#1pt>49.0pt\cellcolor{lime!70}\else\ifdim#1pt>48.0pt\cellcolor{lime!60}\else\ifdim#1pt>47.0pt\cellcolor{lime!50}\else\ifdim#1pt>45.9pt\cellcolor{lime!40}\else\ifdim#1pt>44.9pt\cellcolor{lime!30}\else\ifdim#1pt>43.8pt\cellcolor{lime!20}\else\ifdim#1pt>42.8pt\cellcolor{lime!10}\else\cellcolor{lime!0}\fi\fi\fi\fi\fi\fi\fi\fi\fi\fi#1}

\def\bestaccuracyperadapterfivePR#1{\ifdim#1pt>80.2pt\cellcolor{lime!100}\else\ifdim#1pt>79.7pt\cellcolor{lime!90}\else\ifdim#1pt>79.3pt\cellcolor{lime!80}\else\ifdim#1pt>78.8pt\cellcolor{lime!70}\else\ifdim#1pt>78.4pt\cellcolor{lime!60}\else\ifdim#1pt>78.0pt\cellcolor{lime!50}\else\ifdim#1pt>77.5pt\cellcolor{lime!40}\else\ifdim#1pt>77.1pt\cellcolor{lime!30}\else\ifdim#1pt>76.6pt\cellcolor{lime!20}\else\ifdim#1pt>76.2pt\cellcolor{lime!10}\else\cellcolor{lime!0}\fi\fi\fi\fi\fi\fi\fi\fi\fi\fi#1}

\def\bestaccuracyperadapterfiveRA#1{\ifdim#1pt>71.0pt\cellcolor{lime!100}\else\ifdim#1pt>70.8pt\cellcolor{lime!90}\else\ifdim#1pt>70.5pt\cellcolor{lime!80}\else\ifdim#1pt>70.3pt\cellcolor{lime!70}\else\ifdim#1pt>70.0pt\cellcolor{lime!60}\else\ifdim#1pt>69.8pt\cellcolor{lime!50}\else\ifdim#1pt>69.5pt\cellcolor{lime!40}\else\ifdim#1pt>69.3pt\cellcolor{lime!30}\else\ifdim#1pt>69.0pt\cellcolor{lime!20}\else\ifdim#1pt>68.8pt\cellcolor{lime!10}\else\cellcolor{lime!0}\fi\fi\fi\fi\fi\fi\fi\fi\fi\fi#1}

\def\bestaccuracyperadapterfiveRC#1{\ifdim#1pt>55.5pt\cellcolor{lime!100}\else\ifdim#1pt>54.3pt\cellcolor{lime!90}\else\ifdim#1pt>53.1pt\cellcolor{lime!80}\else\ifdim#1pt>51.9pt\cellcolor{lime!70}\else\ifdim#1pt>50.7pt\cellcolor{lime!60}\else\ifdim#1pt>49.5pt\cellcolor{lime!50}\else\ifdim#1pt>48.3pt\cellcolor{lime!40}\else\ifdim#1pt>47.1pt\cellcolor{lime!30}\else\ifdim#1pt>45.9pt\cellcolor{lime!20}\else\ifdim#1pt>44.7pt\cellcolor{lime!10}\else\cellcolor{lime!0}\fi\fi\fi\fi\fi\fi\fi\fi\fi\fi#1}

\def\bestaccuracyperadapterfiveRP#1{\ifdim#1pt>80.4pt\cellcolor{lime!100}\else\ifdim#1pt>80.3pt\cellcolor{lime!90}\else\ifdim#1pt>80.1pt\cellcolor{lime!80}\else\ifdim#1pt>80.0pt\cellcolor{lime!70}\else\ifdim#1pt>79.8pt\cellcolor{lime!60}\else\ifdim#1pt>79.7pt\cellcolor{lime!50}\else\ifdim#1pt>79.5pt\cellcolor{lime!40}\else\ifdim#1pt>79.4pt\cellcolor{lime!30}\else\ifdim#1pt>79.2pt\cellcolor{lime!20}\else\ifdim#1pt>79.1pt\cellcolor{lime!10}\else\cellcolor{lime!0}\fi\fi\fi\fi\fi\fi\fi\fi\fi\fi#1}

\begin{subtable}{\textwidth}
\centering
\resizebox{0.85\textwidth}{!}{\begin{tabular}{lr|rrrrrr|rrrrrrrrrrrr}
\toprule
 & \multicolumn{1}{c|}{} & \multicolumn{6}{c|}{Office31} & \multicolumn{12}{c}{OfficeHome} \\
 & MM & AD & AW & DA & DW & WA & WD & AC & AP & AR & CA & CP & CR & PA & PC & PR & RA & RC & RP \\
\midrule
Source only & \bestaccuracyperadapterfiveMM{57.6} & \bestaccuracyperadapterfiveAD{82.7} & \bestaccuracyperadapterfiveAW{80.4} & \bestaccuracyperadapterfiveDA{69.6} & \bestaccuracyperadapterfiveDW{94.3} & \bestaccuracyperadapterfiveWA{71.5} & \bestaccuracyperadapterfiveWD{99.0} & \bestaccuracyperadapterfiveAC{41.2} & \bestaccuracyperadapterfiveAP{68.6} & \bestaccuracyperadapterfiveAR{76.7} & \bestaccuracyperadapterfiveCA{60.2} & \bestaccuracyperadapterfiveCP{67.6} & \bestaccuracyperadapterfiveCR{70.5} & \bestaccuracyperadapterfivePA{60.0} & \bestaccuracyperadapterfivePC{42.8} & \bestaccuracyperadapterfivePR{76.2} & \bestaccuracyperadapterfiveRA{68.8} & \bestaccuracyperadapterfiveRC{44.7} & \bestaccuracyperadapterfiveRP{79.1} \\
ATDOC & \textbf{\bestaccuracyperadapterfiveMM{58.7}} & \textbf{\bestaccuracyperadapterfiveAD{85.5}} & \bestaccuracyperadapterfiveAW{83.8} & \bestaccuracyperadapterfiveDA{68.8} & \textbf{\bestaccuracyperadapterfiveDW{95.7}} & \bestaccuracyperadapterfiveWA{69.5} & \bestaccuracyperadapterfiveWD{97.8} & \bestaccuracyperadapterfiveAC{44.6} & \bestaccuracyperadapterfiveAP{71.6} & \bestaccuracyperadapterfiveAR{75.1} & \bestaccuracyperadapterfiveCA{60.1} & \textbf{\bestaccuracyperadapterfiveCP{73.1}} & \bestaccuracyperadapterfiveCR{70.6} & \textbf{\bestaccuracyperadapterfivePA{64.5}} & \bestaccuracyperadapterfivePC{42.4} & \bestaccuracyperadapterfivePR{76.6} & \bestaccuracyperadapterfiveRA{71.0} & \bestaccuracyperadapterfiveRC{49.5} & \bestaccuracyperadapterfiveRP{79.7} \\
BNM & \bestaccuracyperadapterfiveMM{24.8} & \bestaccuracyperadapterfiveAD{81.6} & \bestaccuracyperadapterfiveAW{82.6} & \textbf{\bestaccuracyperadapterfiveDA{71.7}} & \bestaccuracyperadapterfiveDW{94.3} & \textbf{\bestaccuracyperadapterfiveWA{72.8}} & \bestaccuracyperadapterfiveWD{96.3} & \bestaccuracyperadapterfiveAC{49.5} & \bestaccuracyperadapterfiveAP{68.0} & \bestaccuracyperadapterfiveAR{77.5} & \bestaccuracyperadapterfiveCA{60.2} & \bestaccuracyperadapterfiveCP{69.6} & \bestaccuracyperadapterfiveCR{73.8} & \bestaccuracyperadapterfivePA{61.6} & \bestaccuracyperadapterfivePC{50.5} & \bestaccuracyperadapterfivePR{78.7} & \bestaccuracyperadapterfiveRA{67.4} & \bestaccuracyperadapterfiveRC{55.8} & \bestaccuracyperadapterfiveRP{78.6} \\
BSP & \bestaccuracyperadapterfiveMM{18.2} & \bestaccuracyperadapterfiveAD{82.7} & \bestaccuracyperadapterfiveAW{80.3} & \bestaccuracyperadapterfiveDA{69.3} & \bestaccuracyperadapterfiveDW{94.2} & \bestaccuracyperadapterfiveWA{43.8} & \bestaccuracyperadapterfiveWD{96.8} & \bestaccuracyperadapterfiveAC{41.2} & \bestaccuracyperadapterfiveAP{68.6} & \bestaccuracyperadapterfiveAR{76.6} & \bestaccuracyperadapterfiveCA{56.7} & \bestaccuracyperadapterfiveCP{58.5} & \bestaccuracyperadapterfiveCR{65.2} & \bestaccuracyperadapterfivePA{54.7} & \bestaccuracyperadapterfivePC{35.7} & \bestaccuracyperadapterfivePR{74.8} & \bestaccuracyperadapterfiveRA{67.3} & \bestaccuracyperadapterfiveRC{44.7} & \bestaccuracyperadapterfiveRP{79.1} \\
CDAN & \bestaccuracyperadapterfiveMM{57.6} & \bestaccuracyperadapterfiveAD{82.2} & \bestaccuracyperadapterfiveAW{87.0} & \bestaccuracyperadapterfiveDA{69.6} & \bestaccuracyperadapterfiveDW{94.3} & \bestaccuracyperadapterfiveWA{68.5} & \bestaccuracyperadapterfiveWD{95.3} & \bestaccuracyperadapterfiveAC{41.2} & \bestaccuracyperadapterfiveAP{68.6} & \bestaccuracyperadapterfiveAR{76.4} & \bestaccuracyperadapterfiveCA{56.4} & \bestaccuracyperadapterfiveCP{67.9} & \bestaccuracyperadapterfiveCR{68.3} & \bestaccuracyperadapterfivePA{62.2} & \bestaccuracyperadapterfivePC{50.0} & \bestaccuracyperadapterfivePR{78.8} & \bestaccuracyperadapterfiveRA{70.2} & \bestaccuracyperadapterfiveRC{44.7} & \bestaccuracyperadapterfiveRP{79.1} \\
DANN & \bestaccuracyperadapterfiveMM{57.6} & \bestaccuracyperadapterfiveAD{84.1} & \textbf{\bestaccuracyperadapterfiveAW{87.5}} & \bestaccuracyperadapterfiveDA{69.5} & \bestaccuracyperadapterfiveDW{94.4} & \bestaccuracyperadapterfiveWA{68.4} & \bestaccuracyperadapterfiveWD{97.5} & \bestaccuracyperadapterfiveAC{41.2} & \bestaccuracyperadapterfiveAP{68.6} & \bestaccuracyperadapterfiveAR{76.2} & \bestaccuracyperadapterfiveCA{56.7} & \bestaccuracyperadapterfiveCP{66.2} & \bestaccuracyperadapterfiveCR{70.4} & \bestaccuracyperadapterfivePA{60.8} & \bestaccuracyperadapterfivePC{49.9} & \bestaccuracyperadapterfivePR{77.9} & \bestaccuracyperadapterfiveRA{69.8} & \bestaccuracyperadapterfiveRC{44.7} & \bestaccuracyperadapterfiveRP{79.1} \\
GVB & \bestaccuracyperadapterfiveMM{57.6} & \bestaccuracyperadapterfiveAD{82.3} & \bestaccuracyperadapterfiveAW{84.4} & \bestaccuracyperadapterfiveDA{69.5} & \bestaccuracyperadapterfiveDW{94.3} & \bestaccuracyperadapterfiveWA{54.9} & \bestaccuracyperadapterfiveWD{94.9} & \bestaccuracyperadapterfiveAC{41.2} & \bestaccuracyperadapterfiveAP{68.6} & \bestaccuracyperadapterfiveAR{76.7} & \bestaccuracyperadapterfiveCA{60.4} & \bestaccuracyperadapterfiveCP{64.2} & \bestaccuracyperadapterfiveCR{70.6} & \bestaccuracyperadapterfivePA{61.9} & \bestaccuracyperadapterfivePC{50.0} & \bestaccuracyperadapterfivePR{78.5} & \textbf{\bestaccuracyperadapterfiveRA{71.3}} & \bestaccuracyperadapterfiveRC{44.7} & \bestaccuracyperadapterfiveRP{79.1} \\
IM & \bestaccuracyperadapterfiveMM{22.0} & \bestaccuracyperadapterfiveAD{81.4} & \bestaccuracyperadapterfiveAW{85.9} & \bestaccuracyperadapterfiveDA{70.3} & \bestaccuracyperadapterfiveDW{94.2} & \bestaccuracyperadapterfiveWA{72.7} & \bestaccuracyperadapterfiveWD{96.5} & \bestaccuracyperadapterfiveAC{51.9} & \bestaccuracyperadapterfiveAP{68.9} & \bestaccuracyperadapterfiveAR{78.2} & \bestaccuracyperadapterfiveCA{61.5} & \bestaccuracyperadapterfiveCP{69.4} & \bestaccuracyperadapterfiveCR{73.2} & \bestaccuracyperadapterfivePA{62.9} & \bestaccuracyperadapterfivePC{49.6} & \bestaccuracyperadapterfivePR{78.7} & \bestaccuracyperadapterfiveRA{69.0} & \bestaccuracyperadapterfiveRC{55.2} & \bestaccuracyperadapterfiveRP{77.9} \\
MCC & \bestaccuracyperadapterfiveMM{49.9} & \bestaccuracyperadapterfiveAD{84.9} & \textbf{\bestaccuracyperadapterfiveAW{87.5}} & \bestaccuracyperadapterfiveDA{69.7} & \bestaccuracyperadapterfiveDW{95.3} & \bestaccuracyperadapterfiveWA{72.6} & \bestaccuracyperadapterfiveWD{99.3} & \textbf{\bestaccuracyperadapterfiveAC{52.3}} & \textbf{\bestaccuracyperadapterfiveAP{72.8}} & \textbf{\bestaccuracyperadapterfiveAR{80.1}} & \textbf{\bestaccuracyperadapterfiveCA{62.8}} & \bestaccuracyperadapterfiveCP{71.8} & \textbf{\bestaccuracyperadapterfiveCR{74.9}} & \bestaccuracyperadapterfivePA{63.3} & \textbf{\bestaccuracyperadapterfivePC{53.2}} & \textbf{\bestaccuracyperadapterfivePR{80.6}} & \bestaccuracyperadapterfiveRA{69.2} & \textbf{\bestaccuracyperadapterfiveRC{56.7}} & \textbf{\bestaccuracyperadapterfiveRP{80.6}} \\
MCD & \bestaccuracyperadapterfiveMM{57.4} & \bestaccuracyperadapterfiveAD{82.7} & \bestaccuracyperadapterfiveAW{80.3} & \bestaccuracyperadapterfiveDA{69.4} & \bestaccuracyperadapterfiveDW{94.3} & \bestaccuracyperadapterfiveWA{45.4} & \textbf{\bestaccuracyperadapterfiveWD{100.0}} & \bestaccuracyperadapterfiveAC{41.2} & \bestaccuracyperadapterfiveAP{68.6} & \bestaccuracyperadapterfiveAR{76.7} & \bestaccuracyperadapterfiveCA{56.6} & \bestaccuracyperadapterfiveCP{66.3} & \bestaccuracyperadapterfiveCR{68.6} & \bestaccuracyperadapterfivePA{60.6} & \bestaccuracyperadapterfivePC{41.6} & \bestaccuracyperadapterfivePR{74.4} & \bestaccuracyperadapterfiveRA{68.8} & \bestaccuracyperadapterfiveRC{44.7} & \bestaccuracyperadapterfiveRP{79.1} \\
MMD & \bestaccuracyperadapterfiveMM{57.6} & \bestaccuracyperadapterfiveAD{79.7} & \bestaccuracyperadapterfiveAW{79.9} & \bestaccuracyperadapterfiveDA{69.6} & \bestaccuracyperadapterfiveDW{94.3} & \bestaccuracyperadapterfiveWA{61.5} & \bestaccuracyperadapterfiveWD{96.5} & \bestaccuracyperadapterfiveAC{41.2} & \bestaccuracyperadapterfiveAP{69.0} & \bestaccuracyperadapterfiveAR{76.8} & \bestaccuracyperadapterfiveCA{57.9} & \bestaccuracyperadapterfiveCP{65.4} & \bestaccuracyperadapterfiveCR{69.2} & \bestaccuracyperadapterfivePA{62.6} & \bestaccuracyperadapterfivePC{45.8} & \bestaccuracyperadapterfivePR{76.8} & \bestaccuracyperadapterfiveRA{70.7} & \bestaccuracyperadapterfiveRC{44.7} & \bestaccuracyperadapterfiveRP{79.1} \\
\bottomrule
\end{tabular}}
\caption{The accuracy of UDA algorithms when using the algorithm/validator pairs shown in Table \ref{best_validator_per_algorithm}.}
\label{best_accuracy_per_adapter_ranked_by_score_5}
\end{subtable}

%% file: tables/domainnet126_best_accuracy_per_adapter_5.tex
\def\bestaccuracyperadapterfiveDCP#1{\ifdim#1pt>46.7pt\cellcolor{lime!100}\else\ifdim#1pt>45.5pt\cellcolor{lime!90}\else\ifdim#1pt>44.4pt\cellcolor{lime!80}\else\ifdim#1pt>43.2pt\cellcolor{lime!70}\else\ifdim#1pt>42.1pt\cellcolor{lime!60}\else\ifdim#1pt>41.0pt\cellcolor{lime!50}\else\ifdim#1pt>39.8pt\cellcolor{lime!40}\else\ifdim#1pt>38.7pt\cellcolor{lime!30}\else\ifdim#1pt>37.5pt\cellcolor{lime!20}\else\ifdim#1pt>36.4pt\cellcolor{lime!10}\else\cellcolor{lime!0}\fi\fi\fi\fi\fi\fi\fi\fi\fi\fi#1}

\def\bestaccuracyperadapterfiveDCR#1{\ifdim#1pt>56.1pt\cellcolor{lime!100}\else\ifdim#1pt>54.7pt\cellcolor{lime!90}\else\ifdim#1pt>53.4pt\cellcolor{lime!80}\else\ifdim#1pt>52.1pt\cellcolor{lime!70}\else\ifdim#1pt>50.8pt\cellcolor{lime!60}\else\ifdim#1pt>49.4pt\cellcolor{lime!50}\else\ifdim#1pt>48.1pt\cellcolor{lime!40}\else\ifdim#1pt>46.8pt\cellcolor{lime!30}\else\ifdim#1pt>45.4pt\cellcolor{lime!20}\else\ifdim#1pt>44.1pt\cellcolor{lime!10}\else\cellcolor{lime!0}\fi\fi\fi\fi\fi\fi\fi\fi\fi\fi#1}

\def\bestaccuracyperadapterfiveDCS#1{\ifdim#1pt>55.8pt\cellcolor{lime!100}\else\ifdim#1pt>54.8pt\cellcolor{lime!90}\else\ifdim#1pt>53.8pt\cellcolor{lime!80}\else\ifdim#1pt>52.8pt\cellcolor{lime!70}\else\ifdim#1pt>51.8pt\cellcolor{lime!60}\else\ifdim#1pt>50.9pt\cellcolor{lime!50}\else\ifdim#1pt>49.9pt\cellcolor{lime!40}\else\ifdim#1pt>48.9pt\cellcolor{lime!30}\else\ifdim#1pt>47.9pt\cellcolor{lime!20}\else\ifdim#1pt>46.9pt\cellcolor{lime!10}\else\cellcolor{lime!0}\fi\fi\fi\fi\fi\fi\fi\fi\fi\fi#1}

\def\bestaccuracyperadapterfiveDPC#1{\ifdim#1pt>61.3pt\cellcolor{lime!100}\else\ifdim#1pt>60.2pt\cellcolor{lime!90}\else\ifdim#1pt>59.0pt\cellcolor{lime!80}\else\ifdim#1pt>57.9pt\cellcolor{lime!70}\else\ifdim#1pt>56.7pt\cellcolor{lime!60}\else\ifdim#1pt>55.5pt\cellcolor{lime!50}\else\ifdim#1pt>54.4pt\cellcolor{lime!40}\else\ifdim#1pt>53.2pt\cellcolor{lime!30}\else\ifdim#1pt>52.1pt\cellcolor{lime!20}\else\ifdim#1pt>50.9pt\cellcolor{lime!10}\else\cellcolor{lime!0}\fi\fi\fi\fi\fi\fi\fi\fi\fi\fi#1}

\def\bestaccuracyperadapterfiveDPR#1{\ifdim#1pt>69.3pt\cellcolor{lime!100}\else\ifdim#1pt>68.6pt\cellcolor{lime!90}\else\ifdim#1pt>68.0pt\cellcolor{lime!80}\else\ifdim#1pt>67.3pt\cellcolor{lime!70}\else\ifdim#1pt>66.6pt\cellcolor{lime!60}\else\ifdim#1pt>65.9pt\cellcolor{lime!50}\else\ifdim#1pt>65.2pt\cellcolor{lime!40}\else\ifdim#1pt>64.6pt\cellcolor{lime!30}\else\ifdim#1pt>63.9pt\cellcolor{lime!20}\else\ifdim#1pt>63.2pt\cellcolor{lime!10}\else\cellcolor{lime!0}\fi\fi\fi\fi\fi\fi\fi\fi\fi\fi#1}

\def\bestaccuracyperadapterfiveDPS#1{\ifdim#1pt>57.7pt\cellcolor{lime!100}\else\ifdim#1pt>56.4pt\cellcolor{lime!90}\else\ifdim#1pt>55.0pt\cellcolor{lime!80}\else\ifdim#1pt>53.7pt\cellcolor{lime!70}\else\ifdim#1pt>52.3pt\cellcolor{lime!60}\else\ifdim#1pt>50.9pt\cellcolor{lime!50}\else\ifdim#1pt>49.6pt\cellcolor{lime!40}\else\ifdim#1pt>48.2pt\cellcolor{lime!30}\else\ifdim#1pt>46.9pt\cellcolor{lime!20}\else\ifdim#1pt>45.5pt\cellcolor{lime!10}\else\cellcolor{lime!0}\fi\fi\fi\fi\fi\fi\fi\fi\fi\fi#1}

\def\bestaccuracyperadapterfiveDRC#1{\ifdim#1pt>66.4pt\cellcolor{lime!100}\else\ifdim#1pt>65.4pt\cellcolor{lime!90}\else\ifdim#1pt>64.3pt\cellcolor{lime!80}\else\ifdim#1pt>63.3pt\cellcolor{lime!70}\else\ifdim#1pt>62.2pt\cellcolor{lime!60}\else\ifdim#1pt>61.1pt\cellcolor{lime!50}\else\ifdim#1pt>60.1pt\cellcolor{lime!40}\else\ifdim#1pt>59.0pt\cellcolor{lime!30}\else\ifdim#1pt>58.0pt\cellcolor{lime!20}\else\ifdim#1pt>56.9pt\cellcolor{lime!10}\else\cellcolor{lime!0}\fi\fi\fi\fi\fi\fi\fi\fi\fi\fi#1}

\def\bestaccuracyperadapterfiveDRP#1{\ifdim#1pt>66.7pt\cellcolor{lime!100}\else\ifdim#1pt>66.1pt\cellcolor{lime!90}\else\ifdim#1pt>65.5pt\cellcolor{lime!80}\else\ifdim#1pt>64.9pt\cellcolor{lime!70}\else\ifdim#1pt>64.3pt\cellcolor{lime!60}\else\ifdim#1pt>63.8pt\cellcolor{lime!50}\else\ifdim#1pt>63.2pt\cellcolor{lime!40}\else\ifdim#1pt>62.6pt\cellcolor{lime!30}\else\ifdim#1pt>62.0pt\cellcolor{lime!20}\else\ifdim#1pt>61.4pt\cellcolor{lime!10}\else\cellcolor{lime!0}\fi\fi\fi\fi\fi\fi\fi\fi\fi\fi#1}

\def\bestaccuracyperadapterfiveDRS#1{\ifdim#1pt>60.2pt\cellcolor{lime!100}\else\ifdim#1pt>58.9pt\cellcolor{lime!90}\else\ifdim#1pt>57.5pt\cellcolor{lime!80}\else\ifdim#1pt>56.1pt\cellcolor{lime!70}\else\ifdim#1pt>54.8pt\cellcolor{lime!60}\else\ifdim#1pt>53.4pt\cellcolor{lime!50}\else\ifdim#1pt>52.0pt\cellcolor{lime!40}\else\ifdim#1pt>50.6pt\cellcolor{lime!30}\else\ifdim#1pt>49.3pt\cellcolor{lime!20}\else\ifdim#1pt>47.9pt\cellcolor{lime!10}\else\cellcolor{lime!0}\fi\fi\fi\fi\fi\fi\fi\fi\fi\fi#1}

\def\bestaccuracyperadapterfiveDSC#1{\ifdim#1pt>65.9pt\cellcolor{lime!100}\else\ifdim#1pt>64.8pt\cellcolor{lime!90}\else\ifdim#1pt>63.8pt\cellcolor{lime!80}\else\ifdim#1pt>62.7pt\cellcolor{lime!70}\else\ifdim#1pt>61.7pt\cellcolor{lime!60}\else\ifdim#1pt>60.6pt\cellcolor{lime!50}\else\ifdim#1pt>59.5pt\cellcolor{lime!40}\else\ifdim#1pt>58.5pt\cellcolor{lime!30}\else\ifdim#1pt>57.5pt\cellcolor{lime!20}\else\ifdim#1pt>56.4pt\cellcolor{lime!10}\else\cellcolor{lime!0}\fi\fi\fi\fi\fi\fi\fi\fi\fi\fi#1}

\def\bestaccuracyperadapterfiveDSP#1{\ifdim#1pt>57.8pt\cellcolor{lime!100}\else\ifdim#1pt>56.6pt\cellcolor{lime!90}\else\ifdim#1pt>55.3pt\cellcolor{lime!80}\else\ifdim#1pt>54.1pt\cellcolor{lime!70}\else\ifdim#1pt>52.9pt\cellcolor{lime!60}\else\ifdim#1pt>51.7pt\cellcolor{lime!50}\else\ifdim#1pt>50.5pt\cellcolor{lime!40}\else\ifdim#1pt>49.2pt\cellcolor{lime!30}\else\ifdim#1pt>48.0pt\cellcolor{lime!20}\else\ifdim#1pt>46.8pt\cellcolor{lime!10}\else\cellcolor{lime!0}\fi\fi\fi\fi\fi\fi\fi\fi\fi\fi#1}

\def\bestaccuracyperadapterfiveDSR#1{\ifdim#1pt>61.6pt\cellcolor{lime!100}\else\ifdim#1pt>59.9pt\cellcolor{lime!90}\else\ifdim#1pt>58.3pt\cellcolor{lime!80}\else\ifdim#1pt>56.6pt\cellcolor{lime!70}\else\ifdim#1pt>55.0pt\cellcolor{lime!60}\else\ifdim#1pt>53.4pt\cellcolor{lime!50}\else\ifdim#1pt>51.7pt\cellcolor{lime!40}\else\ifdim#1pt>50.1pt\cellcolor{lime!30}\else\ifdim#1pt>48.4pt\cellcolor{lime!20}\else\ifdim#1pt>46.8pt\cellcolor{lime!10}\else\cellcolor{lime!0}\fi\fi\fi\fi\fi\fi\fi\fi\fi\fi#1}

\begin{subtable}{\textwidth}
\centering
\resizebox{0.55\textwidth}{!}{\begin{tabular}{lrrrrrrrrrrrr}
\toprule
 & CP & CR & CS & PC & PR & PS & RC & RP & RS & SC & SP & SR \\
\midrule
Source only & \bestaccuracyperadapterfiveDCP{36.4} & \bestaccuracyperadapterfiveDCR{44.1} & \bestaccuracyperadapterfiveDCS{46.9} & \bestaccuracyperadapterfiveDPC{50.9} & \bestaccuracyperadapterfiveDPR{63.2} & \bestaccuracyperadapterfiveDPS{45.5} & \bestaccuracyperadapterfiveDRC{56.9} & \bestaccuracyperadapterfiveDRP{61.4} & \bestaccuracyperadapterfiveDRS{47.9} & \bestaccuracyperadapterfiveDSC{56.4} & \bestaccuracyperadapterfiveDSP{46.8} & \bestaccuracyperadapterfiveDSR{46.8} \\
ATDOC & \bestaccuracyperadapterfiveDCP{41.4} & \bestaccuracyperadapterfiveDCR{46.9} & \bestaccuracyperadapterfiveDCS{50.0} & \bestaccuracyperadapterfiveDPC{56.2} & \bestaccuracyperadapterfiveDPR{65.1} & \bestaccuracyperadapterfiveDPS{54.6} & \bestaccuracyperadapterfiveDRC{60.2} & \bestaccuracyperadapterfiveDRP{64.6} & \bestaccuracyperadapterfiveDRS{53.6} & \bestaccuracyperadapterfiveDSC{62.3} & \bestaccuracyperadapterfiveDSP{53.4} & \bestaccuracyperadapterfiveDSR{54.5} \\
BNM & \bestaccuracyperadapterfiveDCP{45.5} & \bestaccuracyperadapterfiveDCR{56.3} & \bestaccuracyperadapterfiveDCS{54.9} & \bestaccuracyperadapterfiveDPC{59.6} & \bestaccuracyperadapterfiveDPR{68.9} & \bestaccuracyperadapterfiveDPS{57.6} & \bestaccuracyperadapterfiveDRC{64.4} & \bestaccuracyperadapterfiveDRP{65.1} & \bestaccuracyperadapterfiveDRS{58.3} & \bestaccuracyperadapterfiveDSC{64.8} & \bestaccuracyperadapterfiveDSP{57.1} & \bestaccuracyperadapterfiveDSR{61.2} \\
BSP & \bestaccuracyperadapterfiveDCP{38.5} & \bestaccuracyperadapterfiveDCR{45.2} & \bestaccuracyperadapterfiveDCS{46.9} & \bestaccuracyperadapterfiveDPC{51.8} & \bestaccuracyperadapterfiveDPR{63.2} & \bestaccuracyperadapterfiveDPS{51.2} & \bestaccuracyperadapterfiveDRC{57.9} & \bestaccuracyperadapterfiveDRP{61.4} & \bestaccuracyperadapterfiveDRS{49.1} & \bestaccuracyperadapterfiveDSC{60.1} & \bestaccuracyperadapterfiveDSP{50.3} & \bestaccuracyperadapterfiveDSR{50.4} \\
CDAN & \bestaccuracyperadapterfiveDCP{44.8} & \bestaccuracyperadapterfiveDCR{53.6} & \bestaccuracyperadapterfiveDCS{52.9} & \bestaccuracyperadapterfiveDPC{56.0} & \bestaccuracyperadapterfiveDPR{64.6} & \bestaccuracyperadapterfiveDPS{55.7} & \bestaccuracyperadapterfiveDRC{62.6} & \bestaccuracyperadapterfiveDRP{63.5} & \bestaccuracyperadapterfiveDRS{58.5} & \bestaccuracyperadapterfiveDSC{62.0} & \bestaccuracyperadapterfiveDSP{55.3} & \bestaccuracyperadapterfiveDSR{58.3} \\
DANN & \bestaccuracyperadapterfiveDCP{44.6} & \bestaccuracyperadapterfiveDCR{53.8} & \bestaccuracyperadapterfiveDCS{52.4} & \bestaccuracyperadapterfiveDPC{56.4} & \bestaccuracyperadapterfiveDPR{64.6} & \bestaccuracyperadapterfiveDPS{56.0} & \bestaccuracyperadapterfiveDRC{62.2} & \bestaccuracyperadapterfiveDRP{63.6} & \bestaccuracyperadapterfiveDRS{57.7} & \bestaccuracyperadapterfiveDSC{62.2} & \bestaccuracyperadapterfiveDSP{54.9} & \bestaccuracyperadapterfiveDSR{58.0} \\
GVB & \bestaccuracyperadapterfiveDCP{44.5} & \bestaccuracyperadapterfiveDCR{53.5} & \bestaccuracyperadapterfiveDCS{53.1} & \bestaccuracyperadapterfiveDPC{56.0} & \bestaccuracyperadapterfiveDPR{63.9} & \bestaccuracyperadapterfiveDPS{57.0} & \bestaccuracyperadapterfiveDRC{64.4} & \bestaccuracyperadapterfiveDRP{64.0} & \bestaccuracyperadapterfiveDRS{59.8} & \bestaccuracyperadapterfiveDSC{61.5} & \bestaccuracyperadapterfiveDSP{55.1} & \bestaccuracyperadapterfiveDSR{58.2} \\
IM & \bestaccuracyperadapterfiveDCP{45.2} & \bestaccuracyperadapterfiveDCR{55.1} & \bestaccuracyperadapterfiveDCS{54.6} & \bestaccuracyperadapterfiveDPC{59.1} & \bestaccuracyperadapterfiveDPR{68.9} & \bestaccuracyperadapterfiveDPS{57.6} & \bestaccuracyperadapterfiveDRC{63.7} & \bestaccuracyperadapterfiveDRP{64.6} & \bestaccuracyperadapterfiveDRS{58.0} & \bestaccuracyperadapterfiveDSC{64.7} & \bestaccuracyperadapterfiveDSP{56.2} & \bestaccuracyperadapterfiveDSR{60.8} \\
MCC & \textbf{\bestaccuracyperadapterfiveDCP{47.8}} & \textbf{\bestaccuracyperadapterfiveDCR{57.4}} & \textbf{\bestaccuracyperadapterfiveDCS{56.8}} & \textbf{\bestaccuracyperadapterfiveDPC{62.5}} & \textbf{\bestaccuracyperadapterfiveDPR{70.0}} & \textbf{\bestaccuracyperadapterfiveDPS{59.1}} & \textbf{\bestaccuracyperadapterfiveDRC{67.5}} & \textbf{\bestaccuracyperadapterfiveDRP{67.3}} & \textbf{\bestaccuracyperadapterfiveDRS{61.6}} & \textbf{\bestaccuracyperadapterfiveDSC{66.9}} & \textbf{\bestaccuracyperadapterfiveDSP{59.0}} & \textbf{\bestaccuracyperadapterfiveDSR{63.2}} \\
MCD & \bestaccuracyperadapterfiveDCP{40.8} & \bestaccuracyperadapterfiveDCR{47.9} & \bestaccuracyperadapterfiveDCS{49.7} & \bestaccuracyperadapterfiveDPC{55.2} & \bestaccuracyperadapterfiveDPR{64.6} & \bestaccuracyperadapterfiveDPS{53.7} & \bestaccuracyperadapterfiveDRC{61.8} & \bestaccuracyperadapterfiveDRP{64.6} & \bestaccuracyperadapterfiveDRS{53.1} & \bestaccuracyperadapterfiveDSC{61.5} & \bestaccuracyperadapterfiveDSP{54.1} & \bestaccuracyperadapterfiveDSR{54.9} \\
MMD & \bestaccuracyperadapterfiveDCP{42.4} & \bestaccuracyperadapterfiveDCR{49.2} & \bestaccuracyperadapterfiveDCS{50.9} & \bestaccuracyperadapterfiveDPC{57.3} & \bestaccuracyperadapterfiveDPR{65.2} & \bestaccuracyperadapterfiveDPS{55.7} & \bestaccuracyperadapterfiveDRC{61.9} & \bestaccuracyperadapterfiveDRP{64.3} & \bestaccuracyperadapterfiveDRS{56.1} & \bestaccuracyperadapterfiveDSC{62.3} & \bestaccuracyperadapterfiveDSP{54.4} & \bestaccuracyperadapterfiveDSR{54.8} \\
\bottomrule
\end{tabular}}
\caption{The accuracy of UDA algorithms on DomainNet126, when using the oracle validator.}
\label{domainnet126_best_accuracy_per_adapter_5}
\end{subtable}

%% file: tables/domainnet126_best_accuracy_per_adapter_ranked_by_score_5.tex
\def\bestaccuracyperadapterfiveDCP#1{\ifdim#1pt>46.3pt\cellcolor{lime!100}\else\ifdim#1pt>45.2pt\cellcolor{lime!90}\else\ifdim#1pt>44.1pt\cellcolor{lime!80}\else\ifdim#1pt>43.0pt\cellcolor{lime!70}\else\ifdim#1pt>41.9pt\cellcolor{lime!60}\else\ifdim#1pt>40.8pt\cellcolor{lime!50}\else\ifdim#1pt>39.7pt\cellcolor{lime!40}\else\ifdim#1pt>38.6pt\cellcolor{lime!30}\else\ifdim#1pt>37.5pt\cellcolor{lime!20}\else\ifdim#1pt>36.4pt\cellcolor{lime!10}\else\cellcolor{lime!0}\fi\fi\fi\fi\fi\fi\fi\fi\fi\fi#1}

\def\bestaccuracyperadapterfiveDCR#1{\ifdim#1pt>55.2pt\cellcolor{lime!100}\else\ifdim#1pt>53.9pt\cellcolor{lime!90}\else\ifdim#1pt>52.7pt\cellcolor{lime!80}\else\ifdim#1pt>51.5pt\cellcolor{lime!70}\else\ifdim#1pt>50.2pt\cellcolor{lime!60}\else\ifdim#1pt>49.0pt\cellcolor{lime!50}\else\ifdim#1pt>47.8pt\cellcolor{lime!40}\else\ifdim#1pt>46.6pt\cellcolor{lime!30}\else\ifdim#1pt>45.3pt\cellcolor{lime!20}\else\ifdim#1pt>44.1pt\cellcolor{lime!10}\else\cellcolor{lime!0}\fi\fi\fi\fi\fi\fi\fi\fi\fi\fi#1}

\def\bestaccuracyperadapterfiveDCS#1{\ifdim#1pt>55.4pt\cellcolor{lime!100}\else\ifdim#1pt>54.4pt\cellcolor{lime!90}\else\ifdim#1pt>53.5pt\cellcolor{lime!80}\else\ifdim#1pt>52.5pt\cellcolor{lime!70}\else\ifdim#1pt>51.6pt\cellcolor{lime!60}\else\ifdim#1pt>50.7pt\cellcolor{lime!50}\else\ifdim#1pt>49.7pt\cellcolor{lime!40}\else\ifdim#1pt>48.8pt\cellcolor{lime!30}\else\ifdim#1pt>47.8pt\cellcolor{lime!20}\else\ifdim#1pt>46.9pt\cellcolor{lime!10}\else\cellcolor{lime!0}\fi\fi\fi\fi\fi\fi\fi\fi\fi\fi#1}

\def\bestaccuracyperadapterfiveDPC#1{\ifdim#1pt>60.6pt\cellcolor{lime!100}\else\ifdim#1pt>59.5pt\cellcolor{lime!90}\else\ifdim#1pt>58.5pt\cellcolor{lime!80}\else\ifdim#1pt>57.4pt\cellcolor{lime!70}\else\ifdim#1pt>56.3pt\cellcolor{lime!60}\else\ifdim#1pt>55.2pt\cellcolor{lime!50}\else\ifdim#1pt>54.1pt\cellcolor{lime!40}\else\ifdim#1pt>53.1pt\cellcolor{lime!30}\else\ifdim#1pt>52.0pt\cellcolor{lime!20}\else\ifdim#1pt>50.9pt\cellcolor{lime!10}\else\cellcolor{lime!0}\fi\fi\fi\fi\fi\fi\fi\fi\fi\fi#1}

\def\bestaccuracyperadapterfiveDPR#1{\ifdim#1pt>68.6pt\cellcolor{lime!100}\else\ifdim#1pt>68.0pt\cellcolor{lime!90}\else\ifdim#1pt>67.4pt\cellcolor{lime!80}\else\ifdim#1pt>66.8pt\cellcolor{lime!70}\else\ifdim#1pt>66.2pt\cellcolor{lime!60}\else\ifdim#1pt>65.6pt\cellcolor{lime!50}\else\ifdim#1pt>65.0pt\cellcolor{lime!40}\else\ifdim#1pt>64.4pt\cellcolor{lime!30}\else\ifdim#1pt>63.8pt\cellcolor{lime!20}\else\ifdim#1pt>63.2pt\cellcolor{lime!10}\else\cellcolor{lime!0}\fi\fi\fi\fi\fi\fi\fi\fi\fi\fi#1}

\def\bestaccuracyperadapterfiveDPS#1{\ifdim#1pt>57.1pt\cellcolor{lime!100}\else\ifdim#1pt>55.8pt\cellcolor{lime!90}\else\ifdim#1pt>54.5pt\cellcolor{lime!80}\else\ifdim#1pt>53.2pt\cellcolor{lime!70}\else\ifdim#1pt>52.0pt\cellcolor{lime!60}\else\ifdim#1pt>50.7pt\cellcolor{lime!50}\else\ifdim#1pt>49.4pt\cellcolor{lime!40}\else\ifdim#1pt>48.1pt\cellcolor{lime!30}\else\ifdim#1pt>46.8pt\cellcolor{lime!20}\else\ifdim#1pt>45.5pt\cellcolor{lime!10}\else\cellcolor{lime!0}\fi\fi\fi\fi\fi\fi\fi\fi\fi\fi#1}

\def\bestaccuracyperadapterfiveDRC#1{\ifdim#1pt>65.5pt\cellcolor{lime!100}\else\ifdim#1pt>64.6pt\cellcolor{lime!90}\else\ifdim#1pt>63.6pt\cellcolor{lime!80}\else\ifdim#1pt>62.7pt\cellcolor{lime!70}\else\ifdim#1pt>61.7pt\cellcolor{lime!60}\else\ifdim#1pt>60.7pt\cellcolor{lime!50}\else\ifdim#1pt>59.8pt\cellcolor{lime!40}\else\ifdim#1pt>58.8pt\cellcolor{lime!30}\else\ifdim#1pt>57.9pt\cellcolor{lime!20}\else\ifdim#1pt>56.9pt\cellcolor{lime!10}\else\cellcolor{lime!0}\fi\fi\fi\fi\fi\fi\fi\fi\fi\fi#1}

\def\bestaccuracyperadapterfiveDRP#1{\ifdim#1pt>66.3pt\cellcolor{lime!100}\else\ifdim#1pt>65.7pt\cellcolor{lime!90}\else\ifdim#1pt>65.2pt\cellcolor{lime!80}\else\ifdim#1pt>64.6pt\cellcolor{lime!70}\else\ifdim#1pt>64.1pt\cellcolor{lime!60}\else\ifdim#1pt>63.6pt\cellcolor{lime!50}\else\ifdim#1pt>63.0pt\cellcolor{lime!40}\else\ifdim#1pt>62.5pt\cellcolor{lime!30}\else\ifdim#1pt>61.9pt\cellcolor{lime!20}\else\ifdim#1pt>61.4pt\cellcolor{lime!10}\else\cellcolor{lime!0}\fi\fi\fi\fi\fi\fi\fi\fi\fi\fi#1}

\def\bestaccuracyperadapterfiveDRS#1{\ifdim#1pt>59.4pt\cellcolor{lime!100}\else\ifdim#1pt>58.1pt\cellcolor{lime!90}\else\ifdim#1pt>56.9pt\cellcolor{lime!80}\else\ifdim#1pt>55.6pt\cellcolor{lime!70}\else\ifdim#1pt>54.3pt\cellcolor{lime!60}\else\ifdim#1pt>53.0pt\cellcolor{lime!50}\else\ifdim#1pt>51.7pt\cellcolor{lime!40}\else\ifdim#1pt>50.5pt\cellcolor{lime!30}\else\ifdim#1pt>49.2pt\cellcolor{lime!20}\else\ifdim#1pt>47.9pt\cellcolor{lime!10}\else\cellcolor{lime!0}\fi\fi\fi\fi\fi\fi\fi\fi\fi\fi#1}

\def\bestaccuracyperadapterfiveDSC#1{\ifdim#1pt>65.7pt\cellcolor{lime!100}\else\ifdim#1pt>64.6pt\cellcolor{lime!90}\else\ifdim#1pt>63.6pt\cellcolor{lime!80}\else\ifdim#1pt>62.6pt\cellcolor{lime!70}\else\ifdim#1pt>61.5pt\cellcolor{lime!60}\else\ifdim#1pt>60.5pt\cellcolor{lime!50}\else\ifdim#1pt>59.5pt\cellcolor{lime!40}\else\ifdim#1pt>58.5pt\cellcolor{lime!30}\else\ifdim#1pt>57.4pt\cellcolor{lime!20}\else\ifdim#1pt>56.4pt\cellcolor{lime!10}\else\cellcolor{lime!0}\fi\fi\fi\fi\fi\fi\fi\fi\fi\fi#1}

\def\bestaccuracyperadapterfiveDSP#1{\ifdim#1pt>57.2pt\cellcolor{lime!100}\else\ifdim#1pt>56.1pt\cellcolor{lime!90}\else\ifdim#1pt>54.9pt\cellcolor{lime!80}\else\ifdim#1pt>53.8pt\cellcolor{lime!70}\else\ifdim#1pt>52.6pt\cellcolor{lime!60}\else\ifdim#1pt>51.4pt\cellcolor{lime!50}\else\ifdim#1pt>50.3pt\cellcolor{lime!40}\else\ifdim#1pt>49.1pt\cellcolor{lime!30}\else\ifdim#1pt>48.0pt\cellcolor{lime!20}\else\ifdim#1pt>46.8pt\cellcolor{lime!10}\else\cellcolor{lime!0}\fi\fi\fi\fi\fi\fi\fi\fi\fi\fi#1}

\def\bestaccuracyperadapterfiveDSR#1{\ifdim#1pt>60.2pt\cellcolor{lime!100}\else\ifdim#1pt>58.7pt\cellcolor{lime!90}\else\ifdim#1pt>57.2pt\cellcolor{lime!80}\else\ifdim#1pt>55.7pt\cellcolor{lime!70}\else\ifdim#1pt>54.2pt\cellcolor{lime!60}\else\ifdim#1pt>52.8pt\cellcolor{lime!50}\else\ifdim#1pt>51.3pt\cellcolor{lime!40}\else\ifdim#1pt>49.8pt\cellcolor{lime!30}\else\ifdim#1pt>48.3pt\cellcolor{lime!20}\else\ifdim#1pt>46.8pt\cellcolor{lime!10}\else\cellcolor{lime!0}\fi\fi\fi\fi\fi\fi\fi\fi\fi\fi#1}

\begin{subtable}{\textwidth}
\centering
\resizebox{0.55\textwidth}{!}{\begin{tabular}{lrrrrrrrrrrrr}
\toprule
 & CP & CR & CS & PC & PR & PS & RC & RP & RS & SC & SP & SR \\
\midrule
Source only & \bestaccuracyperadapterfiveDCP{36.4} & \bestaccuracyperadapterfiveDCR{44.1} & \bestaccuracyperadapterfiveDCS{46.9} & \bestaccuracyperadapterfiveDPC{50.9} & \bestaccuracyperadapterfiveDPR{63.2} & \bestaccuracyperadapterfiveDPS{45.5} & \bestaccuracyperadapterfiveDRC{56.9} & \bestaccuracyperadapterfiveDRP{61.4} & \bestaccuracyperadapterfiveDRS{47.9} & \bestaccuracyperadapterfiveDSC{56.4} & \bestaccuracyperadapterfiveDSP{46.8} & \bestaccuracyperadapterfiveDSR{46.8} \\
ATDOC & \bestaccuracyperadapterfiveDCP{40.7} & \bestaccuracyperadapterfiveDCR{46.7} & \bestaccuracyperadapterfiveDCS{49.8} & \bestaccuracyperadapterfiveDPC{55.6} & \bestaccuracyperadapterfiveDPR{62.8} & \bestaccuracyperadapterfiveDPS{53.3} & \bestaccuracyperadapterfiveDRC{59.7} & \bestaccuracyperadapterfiveDRP{63.9} & \bestaccuracyperadapterfiveDRS{53.1} & \bestaccuracyperadapterfiveDSC{61.8} & \bestaccuracyperadapterfiveDSP{53.4} & \bestaccuracyperadapterfiveDSR{53.6} \\
BNM & \bestaccuracyperadapterfiveDCP{44.6} & \bestaccuracyperadapterfiveDCR{55.9} & \bestaccuracyperadapterfiveDCS{54.4} & \bestaccuracyperadapterfiveDPC{58.6} & \bestaccuracyperadapterfiveDPR{68.7} & \bestaccuracyperadapterfiveDPS{55.8} & \bestaccuracyperadapterfiveDRC{63.7} & \bestaccuracyperadapterfiveDRP{63.8} & \bestaccuracyperadapterfiveDRS{57.8} & \bestaccuracyperadapterfiveDSC{63.9} & \bestaccuracyperadapterfiveDSP{56.0} & \bestaccuracyperadapterfiveDSR{60.1} \\
BSP & \bestaccuracyperadapterfiveDCP{36.4} & \bestaccuracyperadapterfiveDCR{44.1} & \bestaccuracyperadapterfiveDCS{46.9} & \bestaccuracyperadapterfiveDPC{50.9} & \bestaccuracyperadapterfiveDPR{63.2} & \bestaccuracyperadapterfiveDPS{45.5} & \bestaccuracyperadapterfiveDRC{56.9} & \bestaccuracyperadapterfiveDRP{61.4} & \bestaccuracyperadapterfiveDRS{47.9} & \bestaccuracyperadapterfiveDSC{56.4} & \bestaccuracyperadapterfiveDSP{46.8} & \bestaccuracyperadapterfiveDSR{46.7} \\
CDAN & \bestaccuracyperadapterfiveDCP{36.4} & \bestaccuracyperadapterfiveDCR{44.1} & \bestaccuracyperadapterfiveDCS{46.9} & \bestaccuracyperadapterfiveDPC{50.9} & \bestaccuracyperadapterfiveDPR{63.2} & \bestaccuracyperadapterfiveDPS{45.5} & \bestaccuracyperadapterfiveDRC{56.9} & \bestaccuracyperadapterfiveDRP{61.4} & \bestaccuracyperadapterfiveDRS{47.9} & \bestaccuracyperadapterfiveDSC{56.4} & \bestaccuracyperadapterfiveDSP{46.8} & \bestaccuracyperadapterfiveDSR{46.7} \\
DANN & \bestaccuracyperadapterfiveDCP{36.4} & \bestaccuracyperadapterfiveDCR{44.1} & \bestaccuracyperadapterfiveDCS{46.9} & \bestaccuracyperadapterfiveDPC{50.9} & \bestaccuracyperadapterfiveDPR{63.2} & \bestaccuracyperadapterfiveDPS{45.5} & \bestaccuracyperadapterfiveDRC{56.9} & \bestaccuracyperadapterfiveDRP{61.4} & \bestaccuracyperadapterfiveDRS{47.9} & \bestaccuracyperadapterfiveDSC{56.4} & \bestaccuracyperadapterfiveDSP{46.8} & \bestaccuracyperadapterfiveDSR{46.7} \\
GVB & \bestaccuracyperadapterfiveDCP{36.4} & \bestaccuracyperadapterfiveDCR{44.1} & \bestaccuracyperadapterfiveDCS{46.9} & \bestaccuracyperadapterfiveDPC{50.9} & \bestaccuracyperadapterfiveDPR{63.2} & \bestaccuracyperadapterfiveDPS{45.5} & \bestaccuracyperadapterfiveDRC{56.9} & \bestaccuracyperadapterfiveDRP{61.4} & \bestaccuracyperadapterfiveDRS{47.9} & \bestaccuracyperadapterfiveDSC{56.4} & \bestaccuracyperadapterfiveDSP{46.8} & \bestaccuracyperadapterfiveDSR{46.7} \\
IM & \bestaccuracyperadapterfiveDCP{44.7} & \bestaccuracyperadapterfiveDCR{52.2} & \bestaccuracyperadapterfiveDCS{54.1} & \bestaccuracyperadapterfiveDPC{57.1} & \bestaccuracyperadapterfiveDPR{68.7} & \bestaccuracyperadapterfiveDPS{55.8} & \bestaccuracyperadapterfiveDRC{63.2} & \bestaccuracyperadapterfiveDRP{63.1} & \bestaccuracyperadapterfiveDRS{57.5} & \bestaccuracyperadapterfiveDSC{64.0} & \bestaccuracyperadapterfiveDSP{55.7} & \bestaccuracyperadapterfiveDSR{59.8} \\
MCC & \textbf{\bestaccuracyperadapterfiveDCP{47.4}} & \textbf{\bestaccuracyperadapterfiveDCR{56.4}} & \textbf{\bestaccuracyperadapterfiveDCS{56.3}} & \textbf{\bestaccuracyperadapterfiveDPC{61.7}} & \textbf{\bestaccuracyperadapterfiveDPR{69.2}} & \textbf{\bestaccuracyperadapterfiveDPS{58.4}} & \textbf{\bestaccuracyperadapterfiveDRC{66.5}} & \textbf{\bestaccuracyperadapterfiveDRP{66.8}} & \textbf{\bestaccuracyperadapterfiveDRS{60.7}} & \textbf{\bestaccuracyperadapterfiveDSC{66.7}} & \textbf{\bestaccuracyperadapterfiveDSP{58.4}} & \textbf{\bestaccuracyperadapterfiveDSR{61.7}} \\
MCD & \bestaccuracyperadapterfiveDCP{36.4} & \bestaccuracyperadapterfiveDCR{44.1} & \bestaccuracyperadapterfiveDCS{46.9} & \bestaccuracyperadapterfiveDPC{50.9} & \bestaccuracyperadapterfiveDPR{63.2} & \bestaccuracyperadapterfiveDPS{45.5} & \bestaccuracyperadapterfiveDRC{56.9} & \bestaccuracyperadapterfiveDRP{61.4} & \bestaccuracyperadapterfiveDRS{47.9} & \bestaccuracyperadapterfiveDSC{56.4} & \bestaccuracyperadapterfiveDSP{46.8} & \bestaccuracyperadapterfiveDSR{46.7} \\
MMD & \bestaccuracyperadapterfiveDCP{36.4} & \bestaccuracyperadapterfiveDCR{44.1} & \bestaccuracyperadapterfiveDCS{46.9} & \bestaccuracyperadapterfiveDPC{50.9} & \bestaccuracyperadapterfiveDPR{63.2} & \bestaccuracyperadapterfiveDPS{45.5} & \bestaccuracyperadapterfiveDRC{56.9} & \bestaccuracyperadapterfiveDRP{61.4} & \bestaccuracyperadapterfiveDRS{47.9} & \bestaccuracyperadapterfiveDSC{56.4} & \bestaccuracyperadapterfiveDSP{46.8} & \bestaccuracyperadapterfiveDSR{46.7} \\
\bottomrule
\end{tabular}}
\caption{The accuracy of UDA algorithms on DomainNet126, when using the algorithm/validator pairs shown in Table \ref{best_validator_per_algorithm}.}
\label{domainnet126_best_accuracy_per_adapter_ranked_by_score_5}
\end{subtable}

%% file: supplementary.tex

\appendix
\onecolumn
\clearpage
\section{Validator parameters explained}\label{section:validator_parameters}
\input{supp_tables/validator_parameters_explained}
\input{supp_tables/validator_parameters}

\clearpage
\section{Training methodology details}\label{supp_experiment_details_scatter_plots}
\input{supp_tables/models_table}
\input{supp_tables/various_experiment_settings_table}
\input{supp_tables/hyperparameter_search_table}
\input{supp_tables/hyperparameter_descriptions}

\clearpage
\section{Scatter plots}
\input{supp_figures/ClassAMI_vs_ClassSS}
\input{supp_figures/DEVBinary_normalization_effect}
\input{supp_figures/SND_examples}
\input{supp_figures/domainnet126_accuracy_low_wsc}
\input{supp_figures/mcc_classami_examples}
\input{supp_figures/atdoc_bnm_examples}
\input{supp_figures/dann_accuracy_examples}

\clearpage
\section{MNIST correlation bar plot}
\input{supp_figures/mnist_barplot}

\clearpage
\section{Correlation tables}

\input{supp_tables/supplementary_table_of_contents}

\subsubsection*{What the green coloring means}
For all tables, the green coloring indicates better performance. The greener the cell color, the better the performance, compared to the Source Val Accuracy validator. The best value per column is bolded. The Mean and Std columns are the mean and standard deviation of all task or algorithm columns. A high mean and low standard deviation reflect good performance.

\subsection{Weighted Spearman Correlation for Office31 and OfficeHome}\label{section:weighted_spearman_correlation_tables_office}

\input{supp_tables/office31_officehome/weighted_spearman_0.0_src_threshold}
\input{supp_tables/office31_officehome/weighted_spearman_0.0_src_threshold_per_adapter_ATDOC}
\input{supp_tables/office31_officehome/weighted_spearman_0.0_src_threshold_per_adapter_BNM}
\input{supp_tables/office31_officehome/weighted_spearman_0.0_src_threshold_per_adapter_BSP}
\input{supp_tables/office31_officehome/weighted_spearman_0.0_src_threshold_per_adapter_CDAN}
\input{supp_tables/office31_officehome/weighted_spearman_0.0_src_threshold_per_adapter_DANN}
\input{supp_tables/office31_officehome/weighted_spearman_0.0_src_threshold_per_adapter_GVB}
\input{supp_tables/office31_officehome/weighted_spearman_0.0_src_threshold_per_adapter_IM}
\input{supp_tables/office31_officehome/weighted_spearman_0.0_src_threshold_per_adapter_MCC}
\input{supp_tables/office31_officehome/weighted_spearman_0.0_src_threshold_per_adapter_MCD}
\input{supp_tables/office31_officehome/weighted_spearman_0.0_src_threshold_per_adapter_MMD}

\clearpage
\subsection{Weighted Spearman Correlation for DomainNet126}\label{section:weighted_spearman_correlation_tables_domainnet126}

\input{supp_tables/domainnet126/weighted_spearman_0.0_src_threshold}
\input{supp_tables/domainnet126/weighted_spearman_0.0_src_threshold_per_adapter_ATDOC}
\input{supp_tables/domainnet126/weighted_spearman_0.0_src_threshold_per_adapter_BNM}
\input{supp_tables/domainnet126/weighted_spearman_0.0_src_threshold_per_adapter_BSP}
\input{supp_tables/domainnet126/weighted_spearman_0.0_src_threshold_per_adapter_CDAN}
\input{supp_tables/domainnet126/weighted_spearman_0.0_src_threshold_per_adapter_DANN}
\input{supp_tables/domainnet126/weighted_spearman_0.0_src_threshold_per_adapter_GVB}
\input{supp_tables/domainnet126/weighted_spearman_0.0_src_threshold_per_adapter_IM}
\input{supp_tables/domainnet126/weighted_spearman_0.0_src_threshold_per_adapter_MCC}
\input{supp_tables/domainnet126/weighted_spearman_0.0_src_threshold_per_adapter_MCD}
\input{supp_tables/domainnet126/weighted_spearman_0.0_src_threshold_per_adapter_MMD}

\clearpage
\subsection{Weighted Spearman Correlation for MNIST}\label{section:weighted_spearman_correlation_tables_mnist}
\input{supp_tables/mnist/weighted_spearman_0.0_src_threshold}
\input{supp_tables/mnist/weighted_spearman_0.0_src_threshold_per_adapter}

\clearpage
\section{Summary of UDA Algorithms}\label{summary_of_uda_algorithms}

\noindent The goal of unsupervised domain adaptation (UDA) is to adapt a model trained on labeled source data, for use on unlabeled target data. Applications of UDA include:

\begin{itemize}
    \item semantic segmentation \cite{toldo2020unsupervised}
    \item object detection \cite{oza2021unsupervised}
    \item natural language processing \cite{ramponi-plank-2020-neural}
\end{itemize}

\noindent There are also other types of domain adaptation, including:

\begin{itemize}
    \item semi-supervised \cite{saito2019semi}
    \item multi-source \cite{peng2019moment}
    \item partial \cite{panareda2017open, Saito_2018_ECCV, cao2018partial}
    \item universal \cite{you2019universal}
    \item source-free \cite{liang2020shot}
\end{itemize}

\noindent In this paper, we focus on UDA for image classification, because it is well-studied and often used as a foundation for other domain adaptation subfields. Here we provide a summary of UDA algorithms by category:

\begin{itemize}
    \item \textbf{Adversarial} methods use a GAN where the generator outputs feature vectors. The discriminator's goal is to correctly classify features as coming from the source or target domain, while the generator tries to minimize the discriminator's accuracy. Examples:
    \begin{itemize}
        \item DANN \cite{ganin2016domain}
        \item Domain Confusion \cite{tzeng2015simultaneous}
        \item ADDA \cite{tzeng2017adversarial}
        \item CDAN \cite{long2017conditional}
        \item VADA \cite{shu2018dirt}
    \end{itemize}
    \item \textbf{Feature distance losses} encourage source and target features to have similar distributions. Examples:
    \begin{itemize}
        \item MMD \cite{long2015learning}
        \item CORAL \cite{sun2016return}
        \item JMMD \cite{long2017deep}
    \end{itemize}
    \item \textbf{Maximum classifier discrepancy} methods use a generator and multiple classifiers in an adversarial setup. The classifiers' goal is to maximize the difference between their prediction vectors (i.e. after softmax) for the target domain data, while the generator's goal is to minimize this discrepancy. Examples:
    \begin{itemize}
        \item MCD \cite{saito2018maximum}
        \item SWD \cite{lee2019sliced}
        \item STAR \cite{lu2020stochastic}
    \end{itemize}
    \item \textbf{Information maximization} methods use the entropy or mutual information of prediction vectors. Examples:
    \begin{itemize}
        \item ITL \cite{ICML2012Shi_566}
        \item MCC \cite{jin2020minimum}
        \item SENTRY \cite{prabhu2021sentry}
    \end{itemize}
    \item \textbf{SVD losses} apply singular value decomposition to the source and/or target features. Examples:
    \begin{itemize}
        \item BSP \cite{chen2019transferability}
        \item BNM \cite{cui2020towards}
    \end{itemize}
    \item \textbf{Image generation} methods use a decoder model to generate source/target -like images from feature vectors, usually as part of of an adversarial method. Examples:
    \begin{itemize}
        \item DRCN \cite{ghifary2016deep}
        \item GTA \cite{sankaranarayanan2018generate}
    \end{itemize}
    \item \textbf{Pseudo-labeling} methods generate labels for the unlabeled target-domain data, to transform the problem from unsupervised to supervised. This is also known as self-supervised learning. Examples:
    \begin{itemize}
        \item ATDA \cite{saito2017asymmetric}
        \item ATDOC \cite{liang2021domain}
    \end{itemize}
    \item \textbf{Mixup augmentations} create training data and features that are a blend between source and target domains. Examples:
    \begin{itemize}
        \item DM-ADA \cite{xu2020adversarial}
        \item DMRL \cite{wu2020dual}
    \end{itemize}
    \item Other notable methods that are more difficult to categorize include:
    \begin{itemize}
        \item RTN \cite{long2016unsupervised}
        \item AFN \cite{xu2019larger}
        \item DSBN \cite{chang2019domain}
        \item SymNets \cite{zhang2019domain}
        \item GVB \cite{cui2020gradually}
    \end{itemize}
\end{itemize}

\clearpage
\section{UDA validators explained}
\subsection{Definitions}
\begin{itemize}
    \item \textbf{Model}: a function that receives some input (e.g. photographic images), and returns a label for each item in that input.
    \item \textbf{Domain adaptation}: a type of machine-learning algorithm that repurposes existing models to work in new domains (a.k.a. target domains). For example, the existing model might work on photographs of food, whereas the target domain contains drawings of food.
    \item \textbf{Unsupervised domain adaptation (UDA)}: a type of domain adaptation where the target-domain does not have any existing class labels.
    \item \textbf{Validator}: a function that evaluates how closely a model's output reflects certain attributes of the dataset, such as labels. The validator will return a quality score. Ideally, the quality score will indicate how similar the model's output is to the dataset attributes.
    \item \textbf{UDA validator}: a validator that estimates target domain accuracy, without having access to target labels. An effective UDA validator is one that reliably estimates target-domain accuracy. For example, a higher score returned by an effective UDA validator will reliably indicate that the target-domain accuracy is high.
    \item \textbf{Target-domain accuracy}: a model's accuracy in the target domain.
    \item \textbf{Oracle validator}: a validator that has access to existing target labels and is therefore able to directly compute target-domain accuracy. In UDA, no target labels are available, so the oracle validator cannot be used. When target labels are available, they should be used during training, as this will improve the model's target-domain accuracy. This type of training is known as semi-supervised or supervised domain adaptation. On the other hand, when target labels are not available, UDA is the only training method possible, and non-oracle validators must be used.
\end{itemize}

\subsection{What are validators, and why are they important?}
In this paper, we compare the performance of various validators. Validators are functions that are used to evaluate the accuracy of machine-learning models, or in the case of this paper, unsupervised domain-adaptation models. This kind of research is essential, for the following reasons. 

To date, most UDA papers have focused on improvements to the training procedure (algorithm), with the goal of maximizing target-domain accuracy. These papers tend to use the oracle validator to evaluate their models, which is useful only when target labels are available. In contrast, when target labels are not available, the oracle validator cannot be used. In that case, UDA validators are the only viable choice.

Unfortunately, UDA validators produce scores that are not 100\% correlated with target-domain accuracy. For example, in an extreme case, the UDA validator could return a high score for a low-accuracy model, and a low score for a high-accuracy model. Even in less extreme scenarios, the score might mislead the user into selecting a model that is not the most accurate one available. Yet achieving the highest possible accuracy is crucial in most application scenarios.

\subsection{How validators are used}
Here is what a typical model-selection workflow looks like:

\begin{enumerate}
    \item Select a UDA algorithm that you think will train your model effectively.
    \item Set the hyperparameters either arbitrarily, or by using a hyperparameter optimizer. The UDA algorithm and hyperparameters will determine how your model is trained.
    \item Use the UDA algorithm to train your model for an arbitrary amount of time, and save a version (checkpoint) of the model at arbitrary regular intervals.
    \item To evaluate each checkpoint, employ whatever UDA validator you think will correlate well with target-domain accuracy. The goal is to obtain validation scores that are as accurate as possible. 
    \item Keep the checkpoint with the highest validation score. Discard all other checkpoints.
    \item Repeat steps 2-5 an arbitrary number of times, or until the best validation scores start to plateau.
\end{enumerate}

At the end of this procedure, the model with the highest validation score will typically be deployed in some application. (For example, the model might be used on a smartphone to classify images of food.) 

\subsection{Why validators are an important area of research}
The above model-selection workflow optimizes for a high validation score, but the goal is to have a model with high target-domain accuracy. UDA validation scores are not perfectly correlated with target-domain accuracy. Thus, the model with the highest validation score might have sub-optimal target-domain accuracy. The lower the correlation, the more likely that a sub-optimal model will be selected inadvertently.

Our research shows that existing UDA validators have considerable room for improvement. For example, the model with the best validation score often has low target-domain accuracy. In other words, the model that gets chosen for deployment actually performs poorly, even though the validator indicates that it performs well. Until UDA validators are able to produce more reliable results, it will be difficult to determine which models have the highest target-domain accuracy. As long as this is the case, the full potential of UDA algorithms will be unrealized.

Despite this fact, there are far more papers on UDA algorithms than on UDA validators. Yet validators have much more room for improvement. A UDA algorithm paper might improve target domain accuracy (as computed by an oracle) by a single percentage point, from 89\% to 90\% for example. But the checkpoint with the highest validation score might have only 70\% target-domain accuracy. Thus, the validator has a much larger effect on accuracy than the choice of UDA algorithm. Hence, research into UDA validators is crucial.

\clearpage
\twocolumn
{\small
\bibliographystyle{ieee_fullname}
\bibliography{main}
}

%% file: supp_tables/validator_parameters_explained.tex
\begin{table}[H]
\caption{Each validator has its own settings. In the other tables and figures in this paper, we use short descriptions to indicate what settings are used. This table explains what those short descriptions mean. Note that we L2-normalized the inputs to ClassSS because it performed worse with un-normalized inputs.}
\label{supp_validator_parameters_explained}
\centering
\resizebox{\textwidth}{!}{\begin{tabular}{lll}
\toprule
Validator & Parameters & Explanation\\
\midrule
\multirow[c]{2}{*}{Accuracy} & Source Train & \texttt{Accuracy(Source train predictions)} \\
 & Source Val & \texttt{Accuracy(Source validation predictions)} \\
\cline{1-3}
\multirow[c]{5}{*}{BNM} & Source Train & \texttt{BNM(Source train predictions)} \\
 & Source Train + Target & \texttt{BNM(Source train predictions)} + \texttt{BNM(Target predictions)} \\
 & Source Val & \texttt{BNM(Source validation predictions)} \\
 & Source Val + Target & \texttt{BNM(Source validation predictions)} + \texttt{BNM(Target predictions)} \\
 & Target & \texttt{BNM(Target predictions)} \\
\cline{1-3}
\multirow[c]{4}{*}{ClassAMI} & Source + Target Features & \texttt{ClassAMI(concat(Source train features, Target features))} \\
 & Source + Target Logits & \texttt{ClassAMI(concat(Source train logits, Target logits))} \\
 & Target Features & \texttt{ClassAMI(Target features)} \\
 & Target Logits & \texttt{ClassAMI(Target logits)} \\
\cline{1-3}
\multirow[c]{4}{*}{ClassSS} & Source + Target Features & \texttt{ClassSS(concat(Source train normalized features, Target normalized features))} \\
 & Source + Target Logits & \texttt{ClassSS(concat(Source train normalized logits, Target normalized logits))} \\
 & Target Features & \texttt{ClassSS(Target normalized features)} \\
 & Target Logits & \texttt{ClassSS(Target normalized logits)} \\
\cline{1-3}
\multirow[c]{3}{*}{DEV} & Features & The discriminator is trained on feature vectors. \\
 & Logits & The discriminator is trained on logits. \\
 & Preds & The discriminator is trained on prediction vectors. \\
\cline{1-3}
\multirow[c]{3}{*}{DEVN} & Features, max normalization & The discriminator is trained on feature vectors. The sample weights are max-normalized. \\
 & Logits, max normalization & The discriminator is trained on logits. The sample weights are max-normalized. \\
 & Preds, max normalization & The discriminator is trained on prediction vectors. The sample weights are max-normalized. \\
\cline{1-3}
\multirow[c]{5}{*}{Entropy} & Source Train & \texttt{Entropy(Source train predictions)} \\
 & Source Train + Target & \texttt{Entropy(Source train predictions)} + \texttt{Entropy(Target predictions)} \\
 & Source Val & \texttt{Entropy(Source validation predictions)} \\
 & Source Val + Target & \texttt{Entropy(Source validation predictions)} + \texttt{Entropy(Target predictions)} \\
 & Target & \texttt{Entropy(Target predictions)} \\
\cline{1-3}
\multirow[c]{9}{*}{SND} & Features, $\tau=0.05$ & The similarity matrix is derived from target features. Softmax temperature is 0.05. \\
 & Features, $\tau=0.1$ & The similarity matrix is derived from target features. Softmax temperature is 0.1. \\
 & Features, $\tau=0.5$ & The similarity matrix is derived from target features. Softmax temperature is 0.5. \\
 & Logits, $\tau=0.05$ & The similarity matrix is derived from target logits. Softmax temperature is 0.05. \\
 & Logits, $\tau=0.1$ & The similarity matrix is derived from target logits. Softmax temperature is 0.1 \\
 & Logits, $\tau=0.5$ & The similarity matrix is derived from target logits. Softmax temperature is 0.5 \\
 & Preds, $\tau=0.05$ & The similarity matrix is derived from target predictions. Softmax temperature is 0.05 \\
 & Preds, $\tau=0.1$ & The similarity matrix is derived from target predictions. Softmax temperature is 0.1 \\
 & Preds, $\tau=0.5$ & The similarity matrix is derived from target predictions. Softmax temperature is 0.5 \\
\cline{1-3}
\bottomrule
\end{tabular}}
\end{table}

%% file: supp_tables/validator_parameters.tex
\begin{table}[H]
\caption{The validator parameters categorized by function.}
\label{validator_parameters}
\centering
\begin{tabular}{r|l}
\toprule
Validators & Parameters \\
 \midrule
Accuracy, BNM, Entropy & Dataset splits \\
ClassAMI, ClassSS & Dataset splits, Choice of feature vector \\
DEV, DEVN & Choice of feature vector \\
SND & Choice of feature vector, softmax temperature \\
\bottomrule
\end{tabular}
\end{table}

%% file: supp_tables/models_table.tex
\begin{table}[H]
\caption{A list of the models used in our experiments. We created the dataset of model outputs using two feature layers: FL3 and FL6. Every checkpoint contains the feature layer (FL3 or FL6) and the logits. The discriminator is used only for adversarial methods. It receives the feature layer as input, but keeps the same depth regardless of feature layer.}
\label{supp_models_table}
\centering
\begin{tabular}{l|ll}
\toprule
 & Layers & Feature name \\
 \midrule
  \begin{tabular}{@{}l@{}} Trunk \end{tabular} &
 \begin{tabular}{@{}l@{}} \texttt{LeNet} or \texttt{ResNet50} \end{tabular} & 
  \begin{tabular}{@{}l@{}}  \end{tabular} \\
 \arrayrulecolor{gray}\hline
 \begin{tabular}{@{}l@{}} Classifier \end{tabular} &
 \begin{tabular}{@{}l@{}} \texttt{Linear(256)} \\ \texttt{ReLU()} \\ \texttt{Dropout(0.5)} \\ \texttt{Linear(128)} \\ \texttt{ReLU()} \\ \texttt{Dropout(0.5)} \\ \texttt{Linear(num\_cls)} \\ \texttt{Softmax()} \end{tabular} & 
  \begin{tabular}{@{}l@{}} \\ \\ \texttt{FL3} \\ \\ \\ \texttt{FL6} \\ \texttt{Logits} \\ \texttt{Preds} \\ \end{tabular} \\
 \hline
 \begin{tabular}{@{}l@{}} Discriminator \end{tabular} &
 \begin{tabular}{@{}l@{}} \texttt{Linear(2048)} \\ \texttt{ReLU()} \\ \texttt{Linear(2048)} \\ \texttt{ReLU()} \\ \texttt{Linear(1)} \end{tabular} &
 \\
 \arrayrulecolor{black}\bottomrule
\end{tabular}
\end{table}

%% file: supp_tables/various_experiment_settings_table.tex
\begin{table}[H]
\caption{A list of various experiment settings. The learning rate (lr) is one of the hyperparameters. The same lr is used by trunk, classifier, and discriminator.}
\label{supp_various_experiment_settings}
\centering
\begin{tabular}{l|l} 
 \toprule
 Category & Settings \\
 \midrule
 \begin{tabular}{@{}l@{}} Optimizer \end{tabular} &
 \begin{tabular}{@{}l@{}} Adam \cite{kingma2014adam} \\ Weight decay of \texttt{1e-4} \\ lr $\in$ \texttt{log([1e-5,0.1])} \end{tabular} \\
 \arrayrulecolor{gray}\hline
  \begin{tabular}{@{}l@{}} LR scheduler \end{tabular} &
 \begin{tabular}{@{}l@{}} One Cycle \cite{onelr2019} \\ 5\% warmup period \\ $\text{lr}_{init} = \text{lr}_{max} / 100$ \\$\text{lr}_{final} = 0$ \\ Cosine annealing \end{tabular} \\
 \hline
 Batch size & 64 source + 64 target \\
 \hline
 \begin{tabular}{@{}l@{}} Epochs / checkpoint interval \end{tabular} &
 \begin{tabular}{@{}l@{}} Digits: 100 / 5 \\ Office31 (W and D as target): 2000 / 100 \\ Office31 (A as target): 200 / 10 \\ OfficeHome: 200 / 10 \\ DomainNet126: 40 / 2 \end{tabular} \\
 \hline
  \begin{tabular}{@{}l@{}} Training image transforms \end{tabular} &
 \begin{tabular}{@{}l@{}} \texttt{Resize(256)} \\ \texttt{RandomCrop(224)} \\ \texttt{RandomHorizontalFlip()} \\ \texttt{Normalize()} \end{tabular} \\
 \hline
  \begin{tabular}{@{}l@{}} Validation image transforms \end{tabular} &
 \begin{tabular}{@{}l@{}} \texttt{Resize(256)} \\ \texttt{CenterCrop(224)} \\ \texttt{Normalize()} \end{tabular} \\
 \hline
   \begin{tabular}{@{}l@{}} MNIST image transforms \end{tabular} &
 \begin{tabular}{@{}l@{}} \texttt{Resize(32)} \\ \texttt{GrayscaleToRGB()} \\ \texttt{Normalize()} \end{tabular} \\
 \arrayrulecolor{black}\bottomrule
\end{tabular}
\end{table}

%% file: supp_tables/hyperparameter_search_table.tex
\begin{table}[H]
\caption{A list of the hyperparameter search settings used in the experiment.}
\label{supp_hyperparameter_search}
\centering
\begin{tabular}{c c c} 
 \toprule
 Algorithm & Hyperparameter & Search space \\ [0.5ex] 
 \midrule
 \begin{tabular}{@{}c@{}} ATDOC \end{tabular} &
\begin{tabular}{@{}c@{}} $\lambda_{atdoc}$ \\ $k_{atdoc}$ \\ $\lambda_L$ \end{tabular} &
\begin{tabular}{@{}c@{}} \texttt{[0,1]} \\ \texttt{int([5, 25], step=5)} \\ \texttt{[0,1]} \end{tabular} \\
 \hline
 \begin{tabular}{@{}c@{}} BNM \end{tabular} &
\begin{tabular}{@{}c@{}} $\lambda_{bnm}$ \\ $\lambda_L$ \end{tabular} &
\begin{tabular}{@{}c@{}} \texttt{[0,1]} \\ \texttt{[0,1]} \end{tabular} \\
\hline
\begin{tabular}{@{}c@{}} BSP \end{tabular} &
\begin{tabular}{@{}c@{}} $\lambda_{bsp}$ \\ $\lambda_L$ \end{tabular} &
\begin{tabular}{@{}c@{}} \texttt{log([1e-6,1])} \\ \texttt{[0,1]} \end{tabular} \\
 \hline
 \begin{tabular}{@{}c@{}} CDAN \\ \end{tabular} &
\begin{tabular}{@{}c@{}} $\lambda_D$ \\ $\lambda_G$ \\ $\lambda_L$ \end{tabular} &
\begin{tabular}{@{}c@{}} \texttt{[0,1]} \\ \texttt{[0,1]} \\ \texttt{[0,1]} \end{tabular} \\
 \hline
 \begin{tabular}{@{}c@{}} DANN \end{tabular} &
\begin{tabular}{@{}c@{}} $\lambda_{D}$ \\ $\lambda_{grl}$ \\ $\lambda_{L}$  \end{tabular} &
\begin{tabular}{@{}c@{}} \texttt{[0,1]} \\ \texttt{log([0.1,10])} \\ \texttt{[0,1]} \end{tabular} \\
 \hline
 \begin{tabular}{@{}c@{}} GVB \end{tabular} &
\begin{tabular}{@{}c@{}} $\lambda_{D}$ \\ $\lambda_{B_G}$ \\ $\lambda_{B_D}$ \\ $\lambda_{grl}$ \end{tabular} &
\begin{tabular}{@{}c@{}} \texttt{[0,1]} \\ \texttt{[0,1]} \\ \texttt{[0,1]} \\ \texttt{log([0.1,10])} \end{tabular} \\
 \hline
\begin{tabular}{@{}c@{}} IM \end{tabular} &
\begin{tabular}{@{}c@{}} $\lambda_{imax}$ \\$\lambda_{L}$  \end{tabular} &
\begin{tabular}{@{}c@{}} \texttt{[0,1]} \\ \texttt{[0,1]} \end{tabular} \\
 \hline

\begin{tabular}{@{}c@{}}  MMD \end{tabular} &
\begin{tabular}{@{}c@{}} $\lambda_F$ \\ $\lambda_L$ \\ $\gamma_{exp}$ \end{tabular} &
\begin{tabular}{@{}c@{}} \texttt{[0,1]} \\ \texttt{[0,1]} \\ \texttt{int([1,8])} \end{tabular} \\
 \hline
 \begin{tabular}{@{}c@{}} MCC \end{tabular} &
\begin{tabular}{@{}c@{}} $\lambda_{mcc}$ \\ $T_{mcc}$ \\ $\lambda_L$ \end{tabular} &
\begin{tabular}{@{}c@{}} \texttt{[0,1]} \\ \texttt{[0.2,5]} \\ \texttt{[0,1]} \end{tabular} \\
\hline
\begin{tabular}{@{}c@{}} MCD \end{tabular} &
\begin{tabular}{@{}c@{}} $N_{mcd}$ \\ $\lambda_{L}$ \\ $\lambda_{disc}$ \\ \end{tabular} &
\begin{tabular}{@{}c@{}} \texttt{int([1,10])} \\ \texttt{[0,1]} \\ \texttt{[0,1]} \\ \end{tabular} \\

 \arrayrulecolor{black}\bottomrule
\end{tabular}
\end{table}

%% file: supp_tables/hyperparameter_descriptions.tex
\begin{table}[H]
\caption{Description of every hyperparameter that is mentioned in Table \ref{supp_hyperparameter_search}.}
\label{supp_hyperparameter_description}
\centering
\begin{tabularx}{\textwidth}{>{\raggedright\arraybackslash\hsize=.6\hsize}X|>{\raggedright\arraybackslash\hsize=1.4\hsize}X}
\toprule
Hyperparameter & Description \\
\midrule
$\lambda_{atdoc}$ & ATDOC loss weight \\
\hline
$\lambda_{bnm}$ & BNM loss weight \\
\hline
$\lambda_{bsp}$ & BSP loss weight \\
\hline
$\lambda_{disc}$ & Classifier discrepancy loss weight for MCD \\
\hline
$\lambda_{grl}$ & Gradient reversal weight, i.e. gradients are multiplied by $-\lambda_{grl}$ \\
\hline
$\lambda_{imax}$ & Information maximization loss weight \\
\hline
$\lambda_{mcc}$ & MCC loss weight \\
\hline
$\lambda_{B_G}$ & Generator bridge loss weight for GVB \\
\hline
$\lambda_{B_D}$ & Discriminator bridge loss weight for GVB \\
\hline
$\lambda_D$ & Discriminator loss weight\\
\arrayrulecolor{gray}\hline
$\lambda_{F}$ & Feature distance loss weight \\
\hline
$\lambda_G$ & Generator loss weight\\
\hline
$\lambda_{L}$ & Source classification loss weight \\
\hline
$\gamma_{exp}$ & Exponent of the bandwidth multiplier for MMD. For example, if $\gamma_{exp}=2$, then the bandwidths used will be $\{2^{-2}x, 2^{-1}x, 2^{0}x, 2^{1}x, 2^{2}x\}$, where $x$ is the base bandwidth. \\
\hline
$k_{atdoc}$ & Number of nearest neighbors to retrieve for computing pseudolabels in ATDOC \\
\hline
$N_{mcd}$ & Number of times the MCD generator is updated per batch \\
\hline
$T_{mcc}$ & Softmax temperature used by MCC \\
\arrayrulecolor{black}\bottomrule
\end{tabularx}
\end{table}

%% file: supp_figures/ClassAMI_vs_ClassSS.tex
\begin{figure}[H]
     \centering
     \begin{subfigure}[b]{0.4\textwidth}
         \centering
         \includegraphics[width=1.0\textwidth]{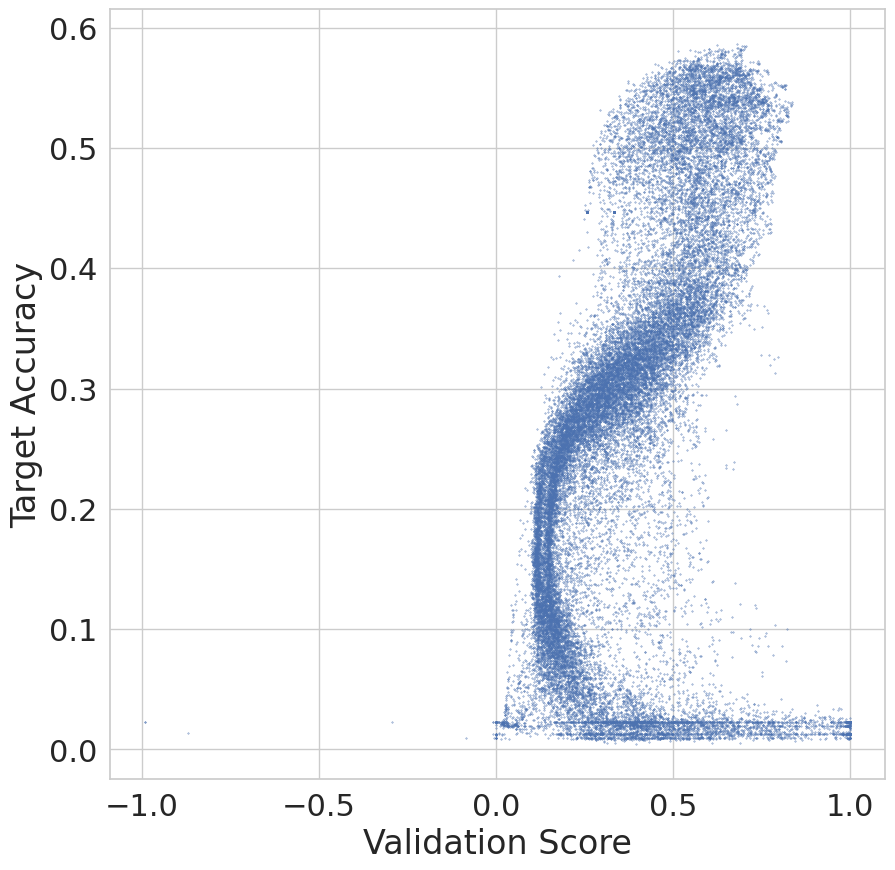}
         \caption{ClassSS}
         \label{}
     \end{subfigure}
     \hspace{2em}
     \begin{subfigure}[b]{0.4\textwidth}
         \centering
         \includegraphics[width=1.0\textwidth]{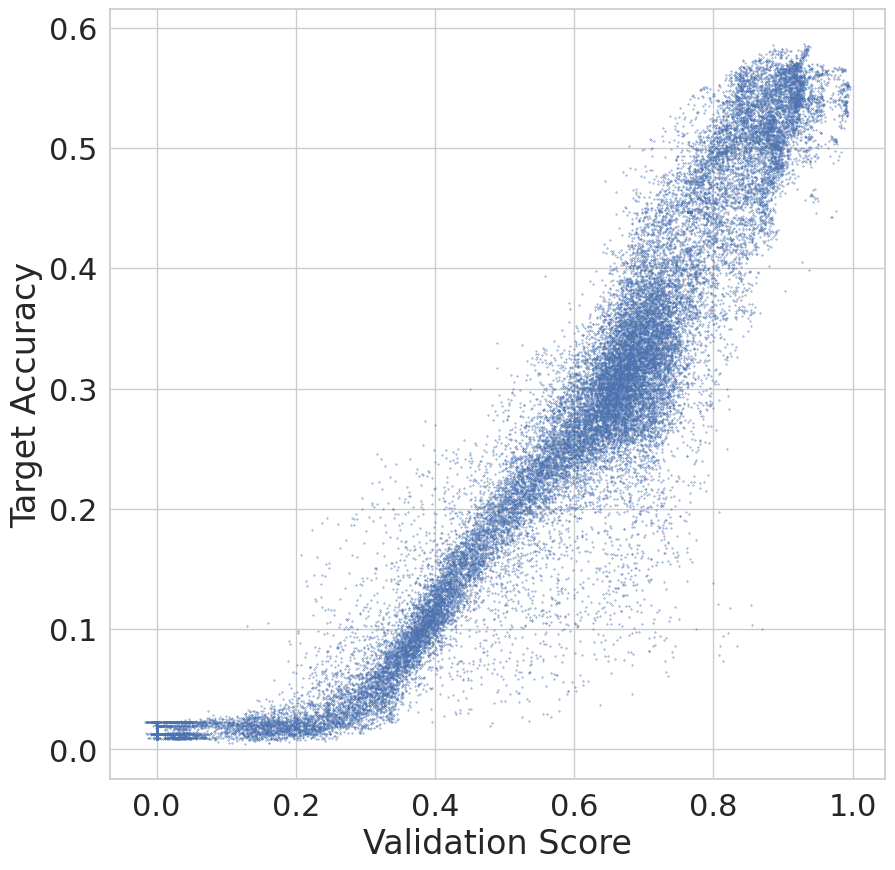}
         \caption{ClassAMI}
         \label{}
     \end{subfigure}     
     \caption{We used the Adjusted Mutual Information (ClassAMI) instead of the Silhouette Score (ClassSS) to achieve a significant improvement for the class clustering validation method. These plots are for the OfficeHome Real $\rightarrow$ Clipart task.}
    \label{ClassAMI_vs_ClassSS}
\end{figure}

%% file: supp_figures/DEVBinary_normalization_effect.tex
\begin{figure}[H]
     \centering
     \begin{subfigure}[t]{0.4\textwidth}
         \centering
         \includegraphics[width=1.0\textwidth]{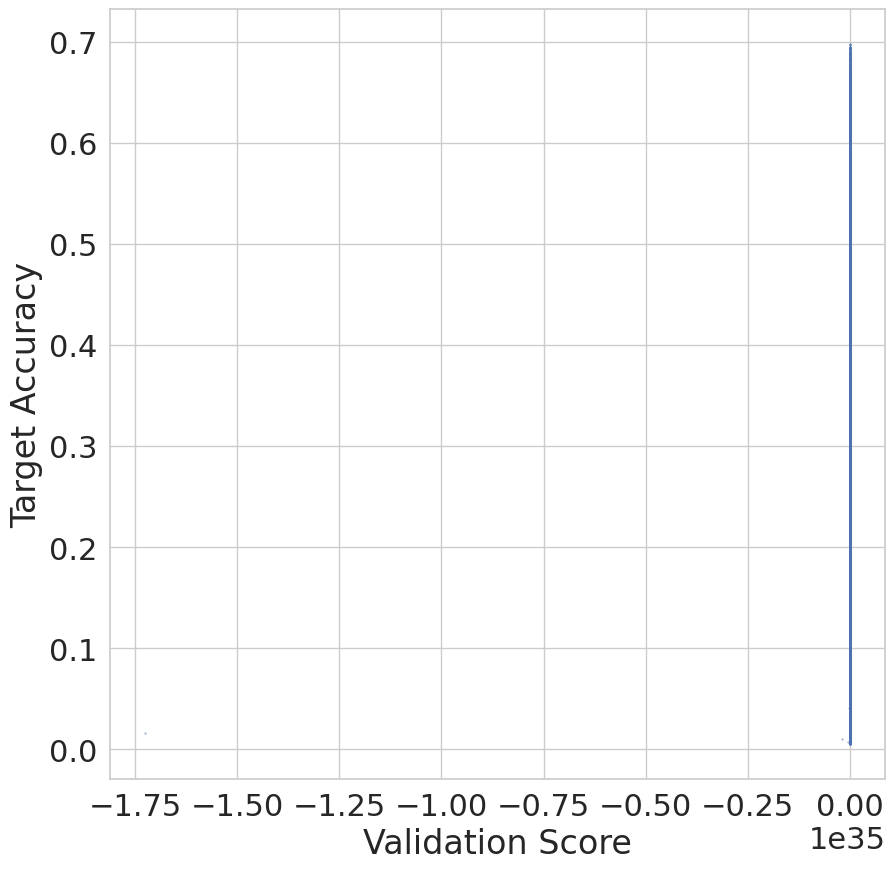}
         \caption{DEV}
         \label{DEVBinary_layer_features_normalization_None}
     \end{subfigure}
     \hspace{1em}
     \begin{subfigure}[t]{0.4\textwidth}
         \centering
         \includegraphics[width=1.0\textwidth]{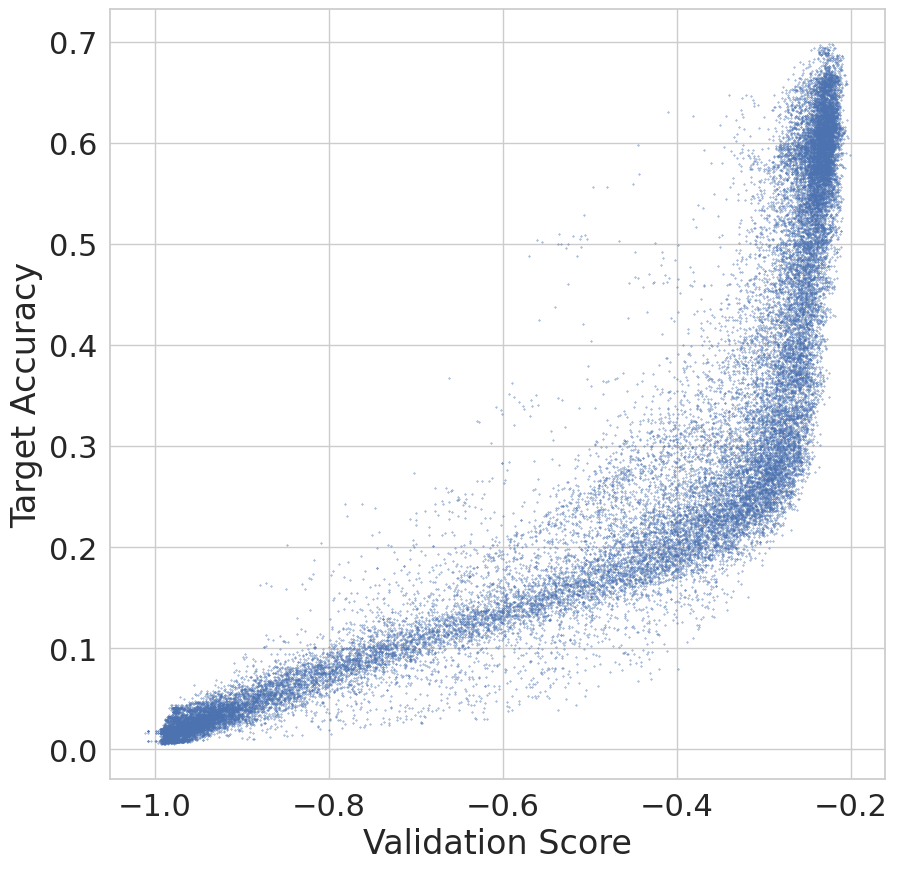}
         \caption{DEVN}
         \label{DEVBinary_layer_features_normalization_max}
     \end{subfigure}
     \caption{DEV can produce scores approaching infinity (Figure \ref{DEVBinary_layer_features_normalization_None}). Our proposed method, DEVN, fixes this problem by normalizing the sample weights. (Figure \ref{DEVBinary_layer_features_normalization_max}). These plots are for the OfficeHome Clipart $\rightarrow$ Art task.}
    \label{DEVBinary_normalization_effect}
\end{figure}

%% file: supp_figures/SND_examples.tex
\begin{figure}[H]
     \centering
     \begin{subfigure}[b]{0.4\textwidth}
         \centering
         \includegraphics[width=1.0\textwidth]{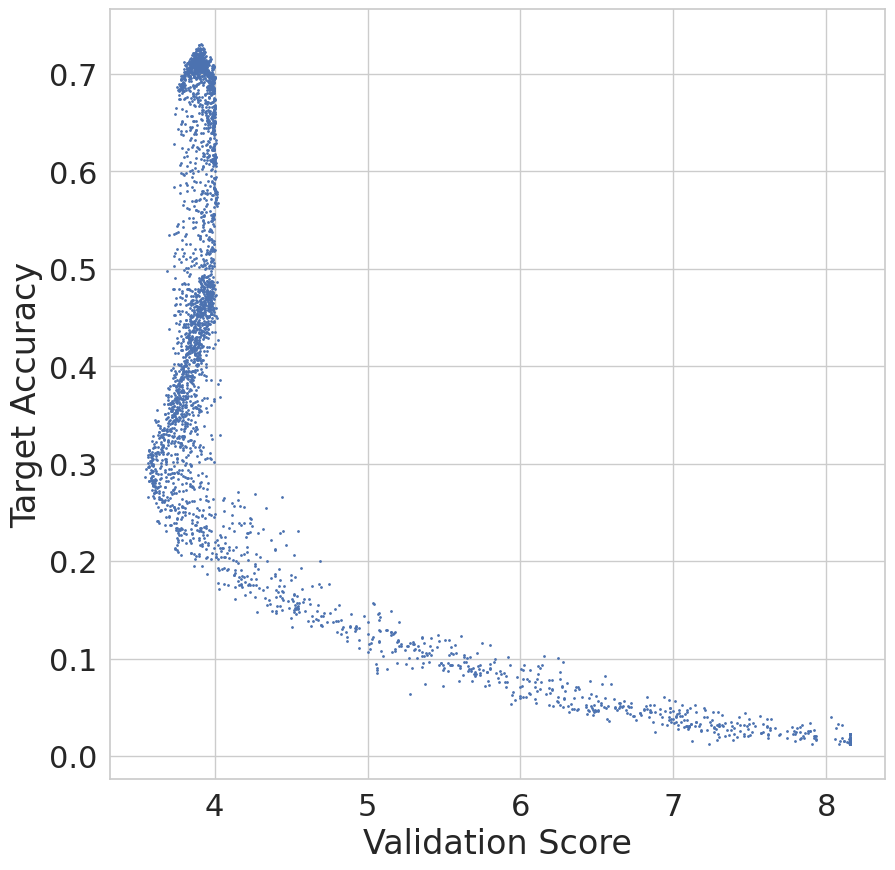}
         \caption{SND applied to the checkpoints of the CDAN algorithm, on the OfficeHome Clipart $\rightarrow$ Real task. The weighted Spearman correlation is -81.9. This is SND applied to the prediction vectors, with $\tau = 0.05$.}
         \label{}
     \end{subfigure}
     \hspace{2em}
     \begin{subfigure}[b]{0.4\textwidth}
         \centering
         \includegraphics[width=1.0\textwidth]{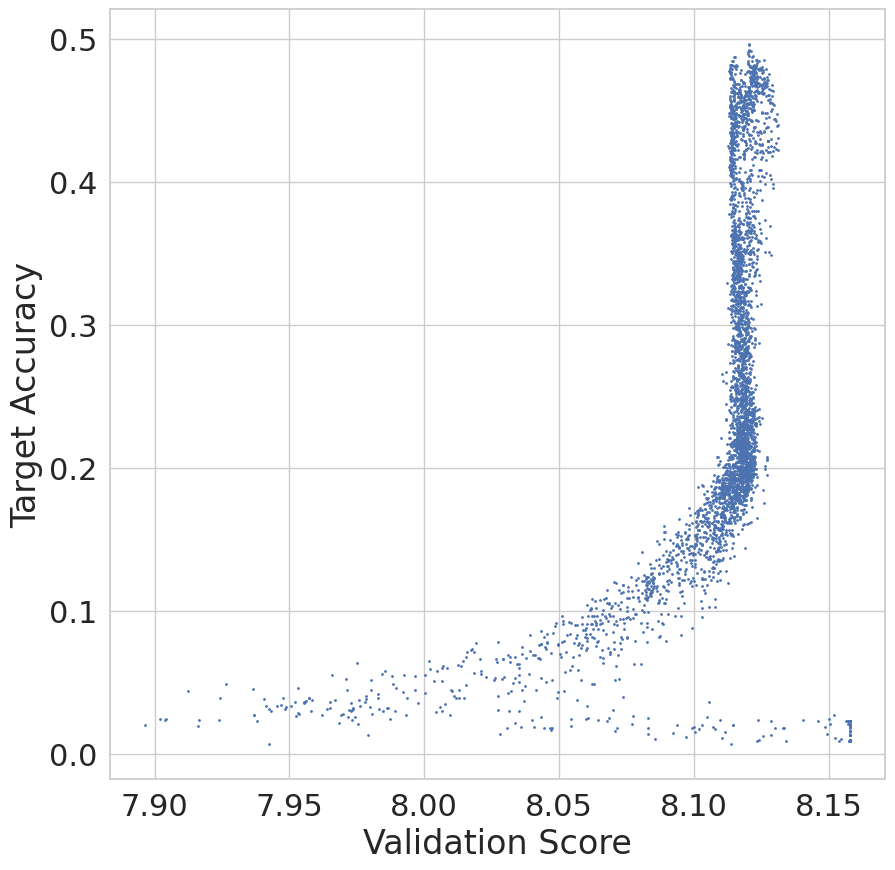}
         \caption{SND applied to the checkpoints of the MMD algorithm, on the OfficeHome Product $\rightarrow$ Clipart task. The weighted Spearman correlation is -3.1. This is SND applied to the feature vectors, with $\tau = 0.5$.}
         \label{}
     \end{subfigure}
     \caption{Examples of the SND validator being a poor predictor of accuracy.}
    \label{figure:SND_examples}
\end{figure}

%% file: supp_figures/domainnet126_accuracy_low_wsc.tex
\begin{figure}[H]
     \centering
     \begin{subfigure}[b]{0.4\textwidth}
         \centering
         \includegraphics[width=1.0\textwidth]{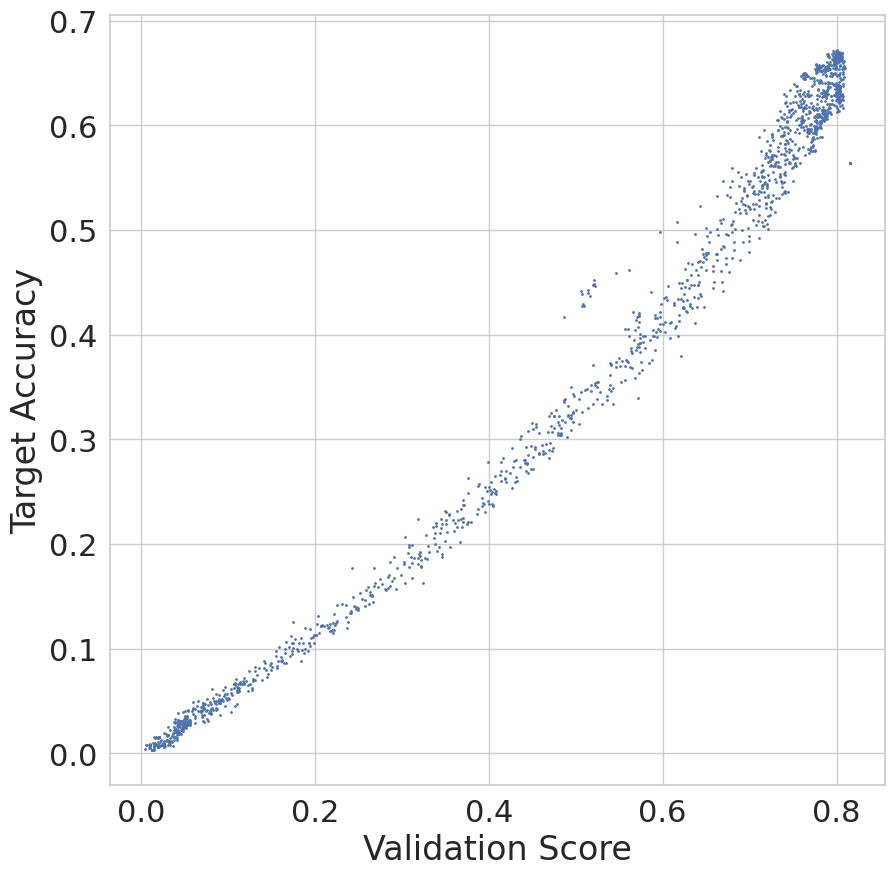}
         \caption{DomainNet126 Sketch $\rightarrow$ Clipart, MCC algorithm, Accuracy (Source Val) validator, WSC = 53.0}
         \label{}
     \end{subfigure}
     \hspace{2em}
     \begin{subfigure}[b]{0.4\textwidth}
         \centering
         \includegraphics[width=1.0\textwidth]{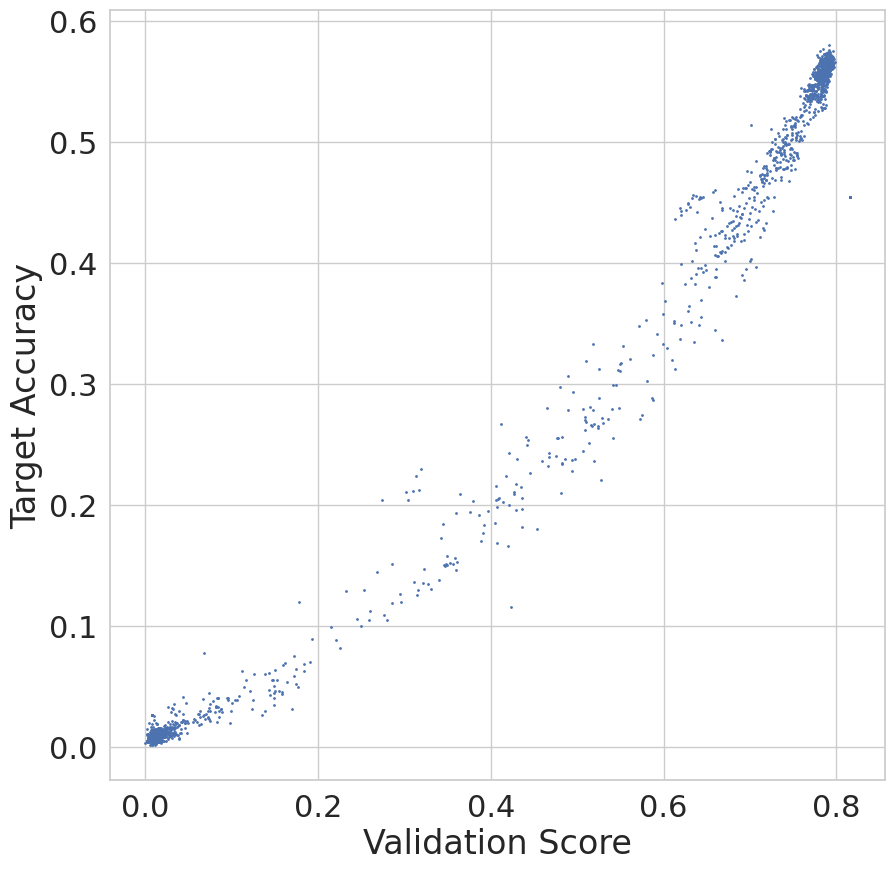}
         \caption{DomainNet126 Painting $\rightarrow$ Sketch, IM Algorithm, Accuracy (Source Val) validator, WSC = 38.3}
         \label{}
     \end{subfigure}     
     \caption{What causes the low WSC for the Accuracy (Source Val) validator on DomainNet126 tasks? Although source validation accuracy ranks most checkpoints correctly, it incorrectly ranks the untrained model as the best. In the above plots, the untrained model is represented by the dot furthest to the right.}
    \label{domainnet126_accuracy_low_wsc}
\end{figure}

%% file: supp_figures/mcc_classami_examples.tex
\begin{figure}[H]
     \centering
     \begin{subfigure}[b]{0.4\textwidth}
         \centering
         \includegraphics[width=1.0\textwidth]{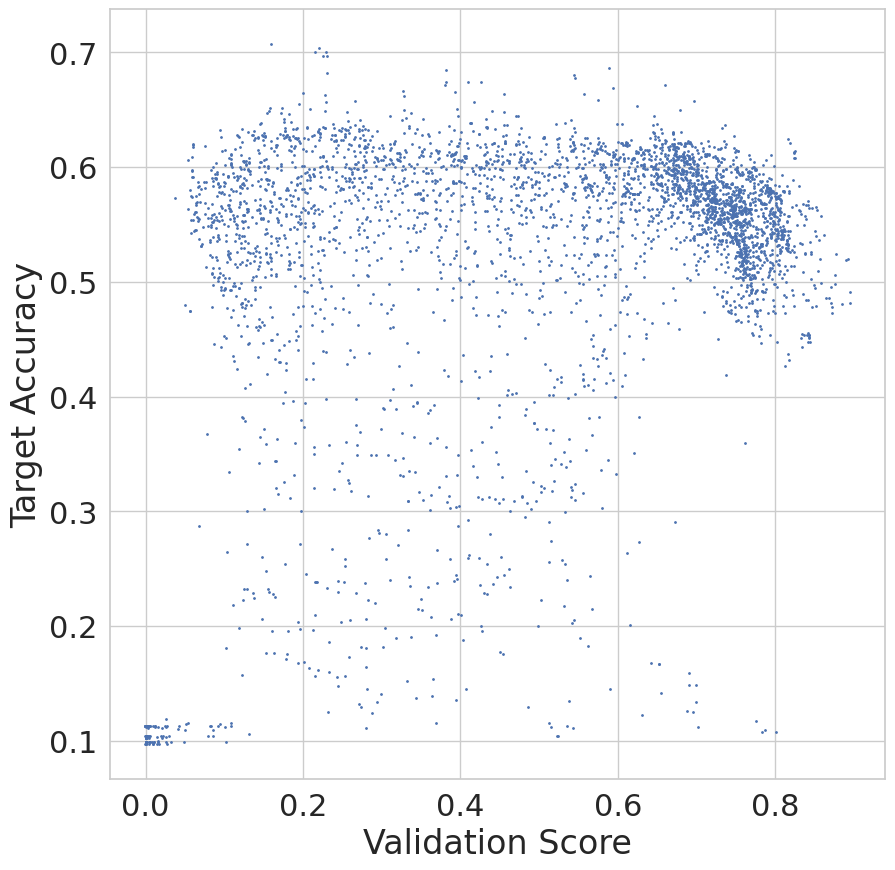}
         \caption{MNIST $\rightarrow$ MNISTM, WSC = -55.4}
         \label{}
     \end{subfigure}
     \hspace{2em}
     \begin{subfigure}[b]{0.4\textwidth}
         \centering
         \includegraphics[width=1.0\textwidth]{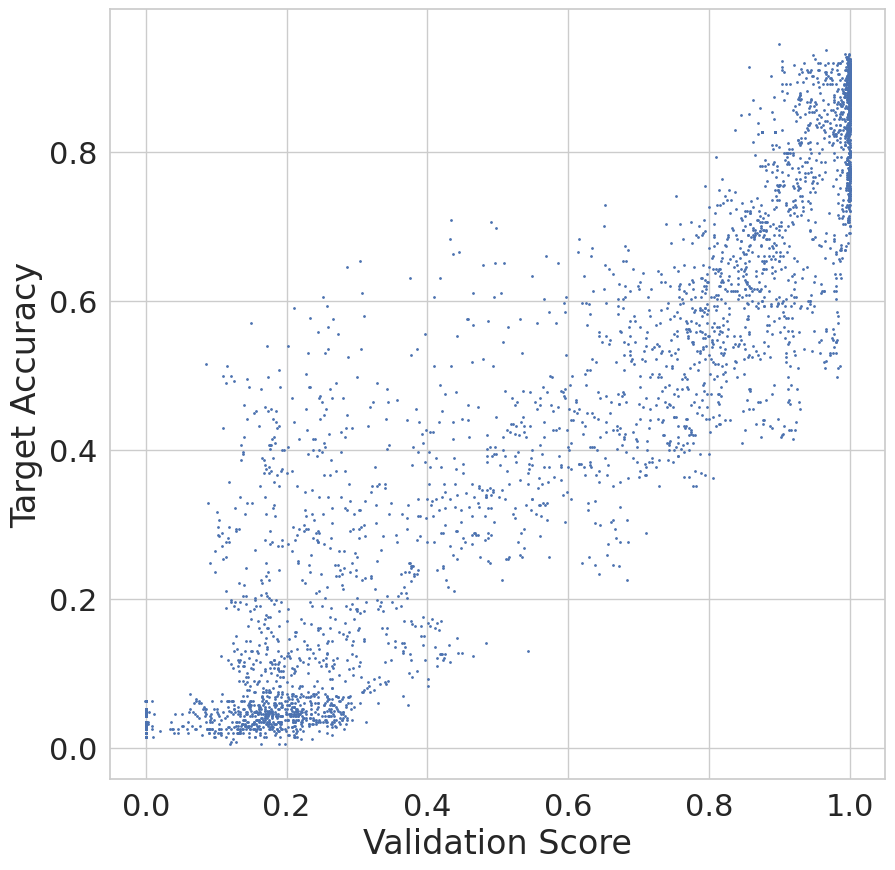}
         \caption{Office31 Amazon $\rightarrow$ DSLR, WSC = 72.7}
         \label{}
     \end{subfigure}   
     \\[3ex]
      \begin{subfigure}[b]{0.4\textwidth}
         \centering
         \includegraphics[width=1.0\textwidth]{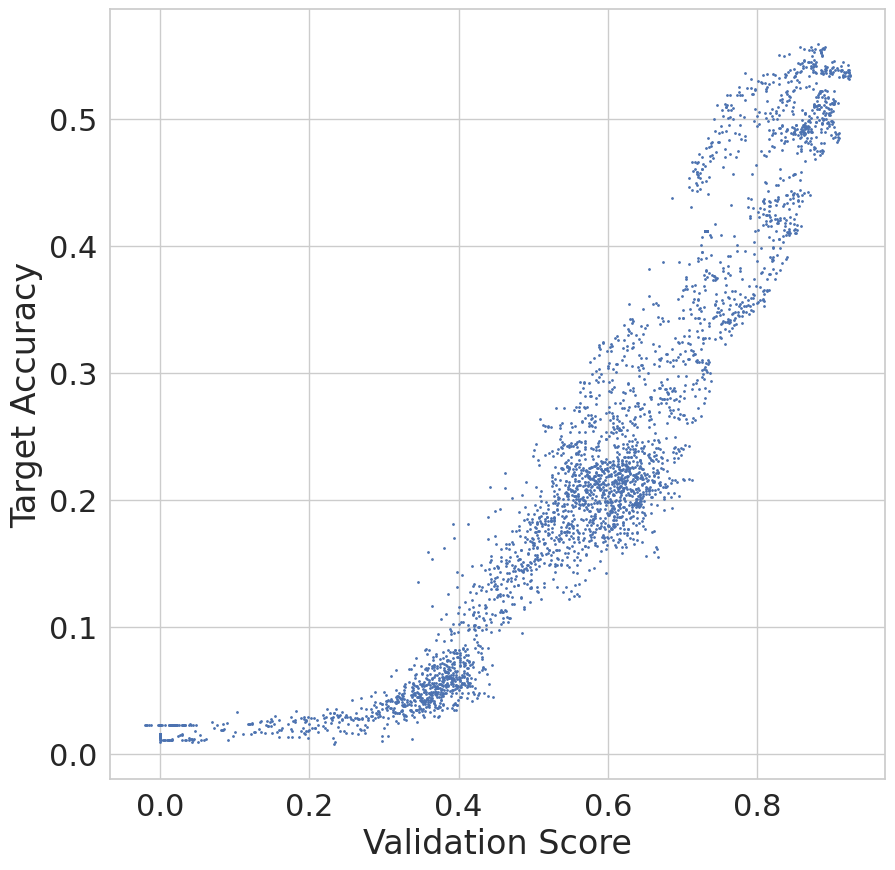}
         \caption{OfficeHome Art $\rightarrow$ Clipart, WSC = 90.1}
         \label{}
     \end{subfigure}
     \hspace{2em}
     \begin{subfigure}[b]{0.4\textwidth}
         \centering
         \includegraphics[width=1.0\textwidth]{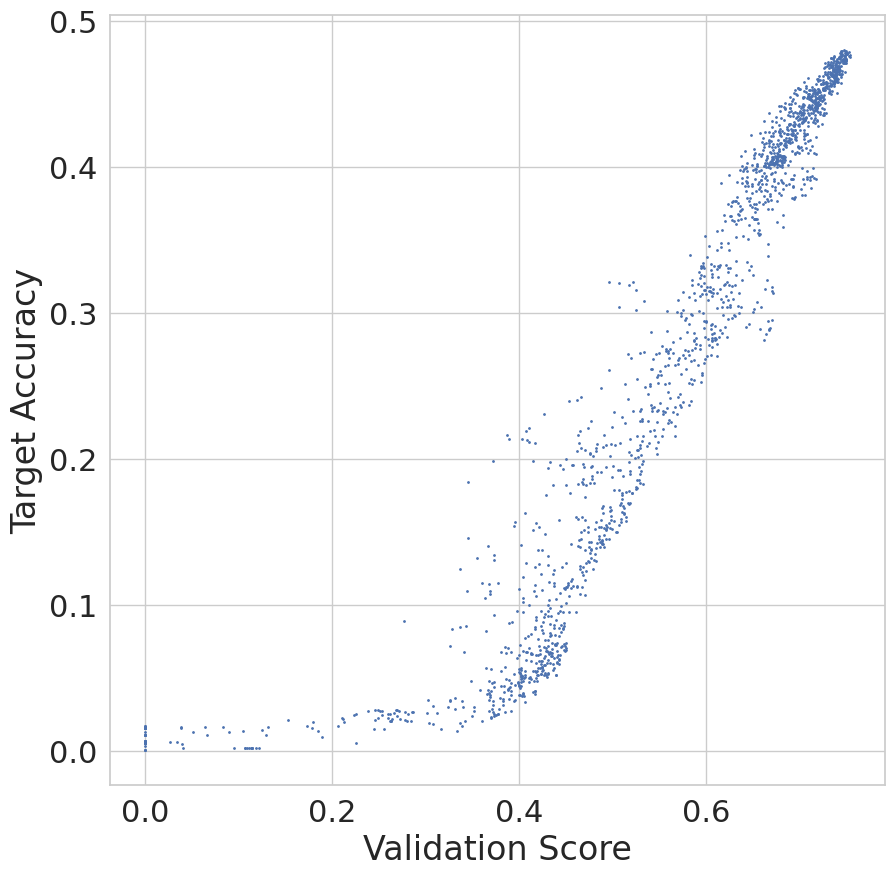}
         \caption{DomainNet126 Clipart $\rightarrow$ Painting, WSC = 95.1}
         \label{}
     \end{subfigure}    
     \caption{These plots show the checkpoints of the MCC algorithm, using the ClassAMI validator, on a task from each dataset. ClassAMI is the best validator for MCC, as measured by average WSC across tasks (excluding the MNIST task). MCC/ClassAMI is the best performing algorithm/validator pair as measured by AATN across tasks (excluding the MNIST task). Note how MNIST is an outlier in terms of results, dataset attributes, and model architecture.}
    \label{mcc_classami_examples}
\end{figure}

%% file: supp_figures/atdoc_bnm_examples.tex
\begin{figure}[H]
     \centering
     \begin{subfigure}[b]{0.4\textwidth}
         \centering
         \includegraphics[width=1.0\textwidth]{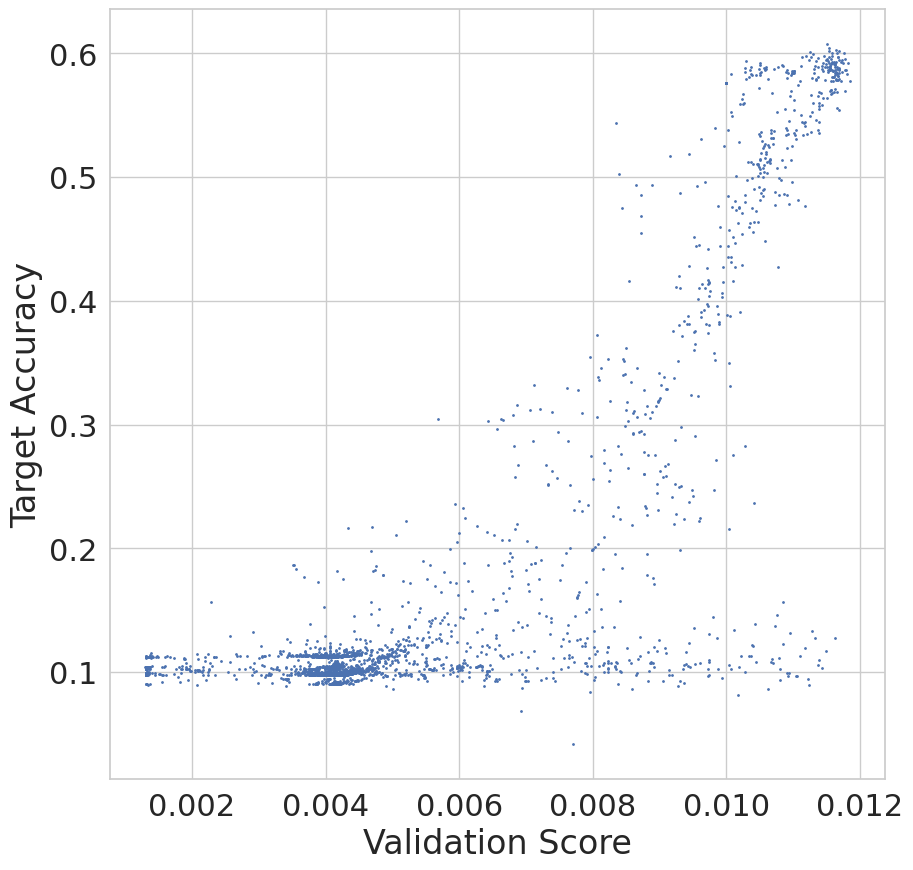}
         \caption{MNIST $\rightarrow$ MNISTM, WSC = 78.0}
         \label{}
     \end{subfigure}
     \hspace{2em}
     \begin{subfigure}[b]{0.4\textwidth}
         \centering
         \includegraphics[width=1.0\textwidth]{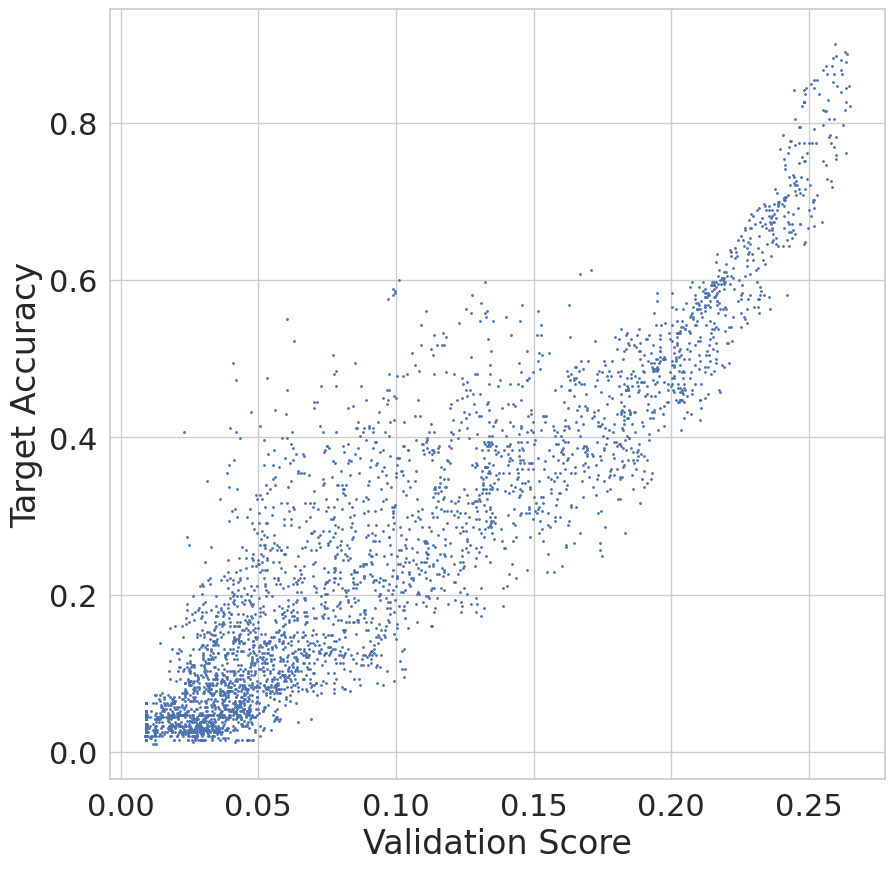}
         \caption{Office31 Amazon $\rightarrow$ DSLR, WSC = 93.7}
         \label{}
     \end{subfigure}   
     \\[3ex]
      \begin{subfigure}[b]{0.4\textwidth}
         \centering
         \includegraphics[width=1.0\textwidth]{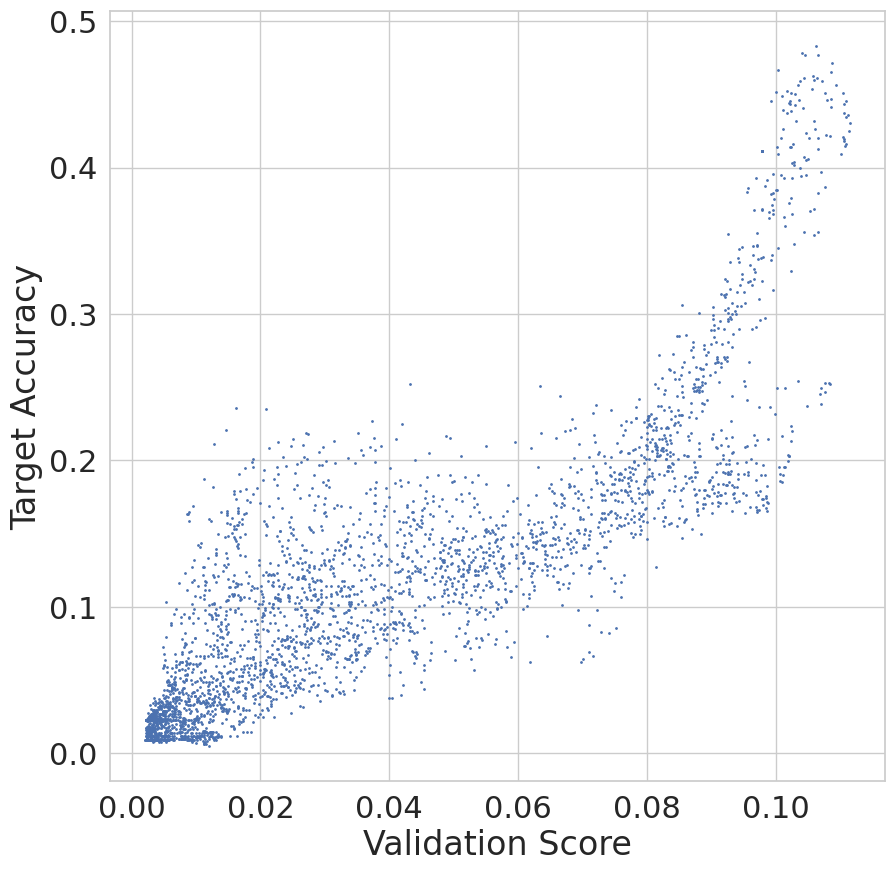}
         \caption{OfficeHome Art $\rightarrow$ Clipart, WSC = 88.7}
         \label{}
     \end{subfigure}
     \hspace{2em}
     \begin{subfigure}[b]{0.4\textwidth}
         \centering
         \includegraphics[width=1.0\textwidth]{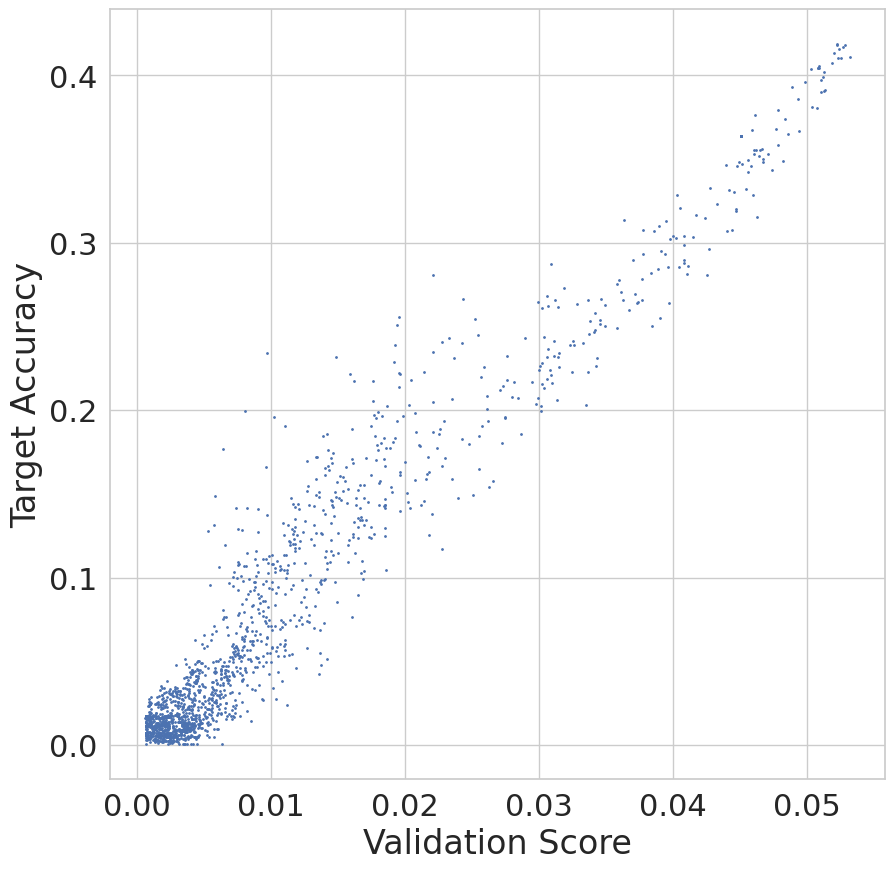}
         \caption{DomainNet126 Clipart $\rightarrow$ Painting, WSC = 96.7}
         \label{}
     \end{subfigure}    
     \caption{These plots show the checkpoints of the ATDOC algorithm, using the BNM validator, on a task from each dataset. BNM is the best validator for ATDOC, as measured by average WSC across tasks (excluding the MNIST task). ATDOC/BNM is the 4th-best performing algorithm/validator pair as measured by AATN across tasks (excluding the MNIST task).}
    \label{atdoc_bnm_examples}
\end{figure}

%% file: supp_figures/dann_accuracy_examples.tex
\begin{figure}[H]
     \centering
     \begin{subfigure}[b]{0.4\textwidth}
         \centering
         \includegraphics[width=1.0\textwidth]{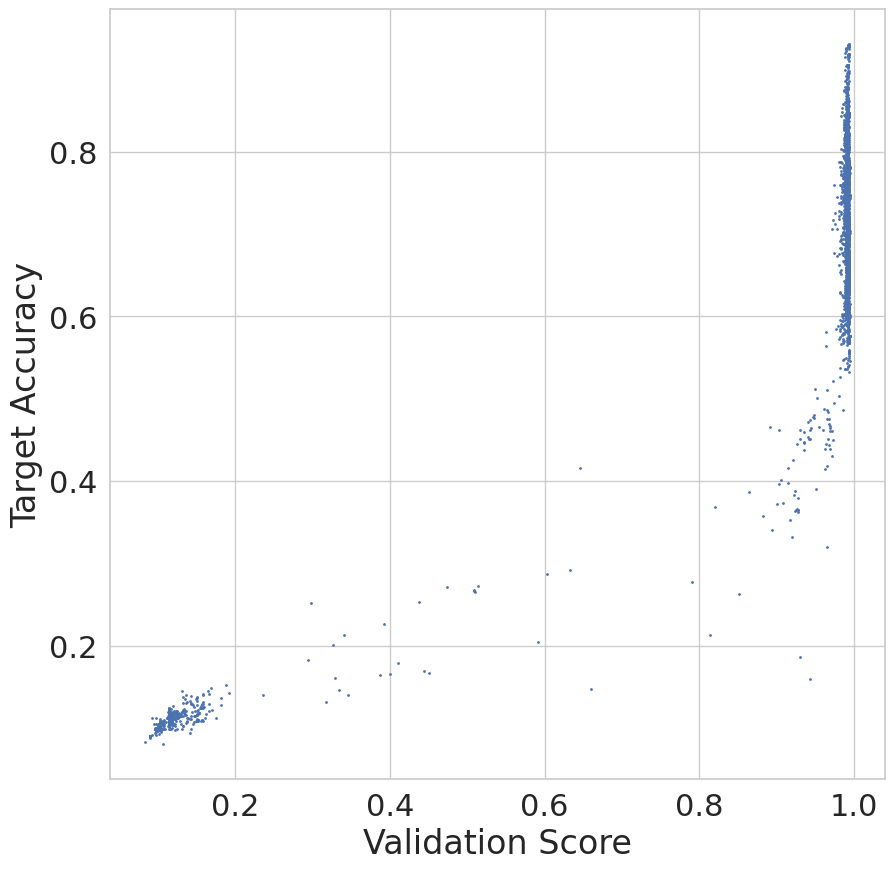}
         \caption{MNIST $\rightarrow$ MNISTM, WSC = -23.6}
         \label{}
     \end{subfigure}
     \hspace{2em}
     \begin{subfigure}[b]{0.4\textwidth}
         \centering
         \includegraphics[width=1.0\textwidth]{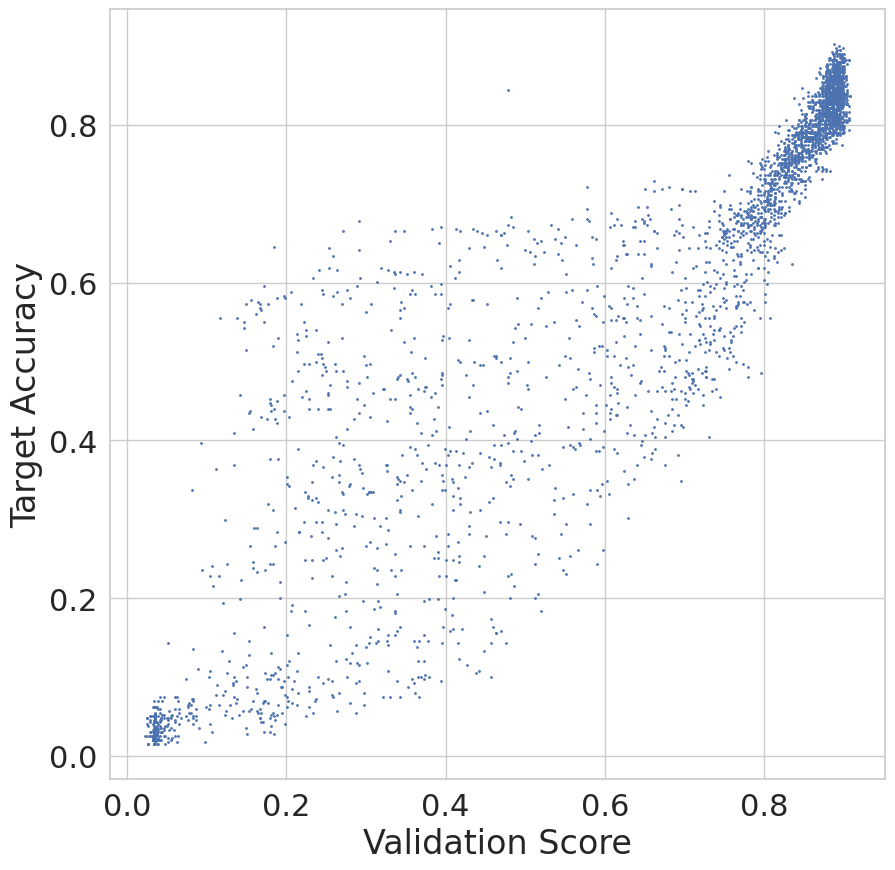}
         \caption{Office31 Amazon $\rightarrow$ DSLR, WSC = 84.4}
         \label{}
     \end{subfigure}   
     \\[3ex]
      \begin{subfigure}[b]{0.4\textwidth}
         \centering
         \includegraphics[width=1.0\textwidth]{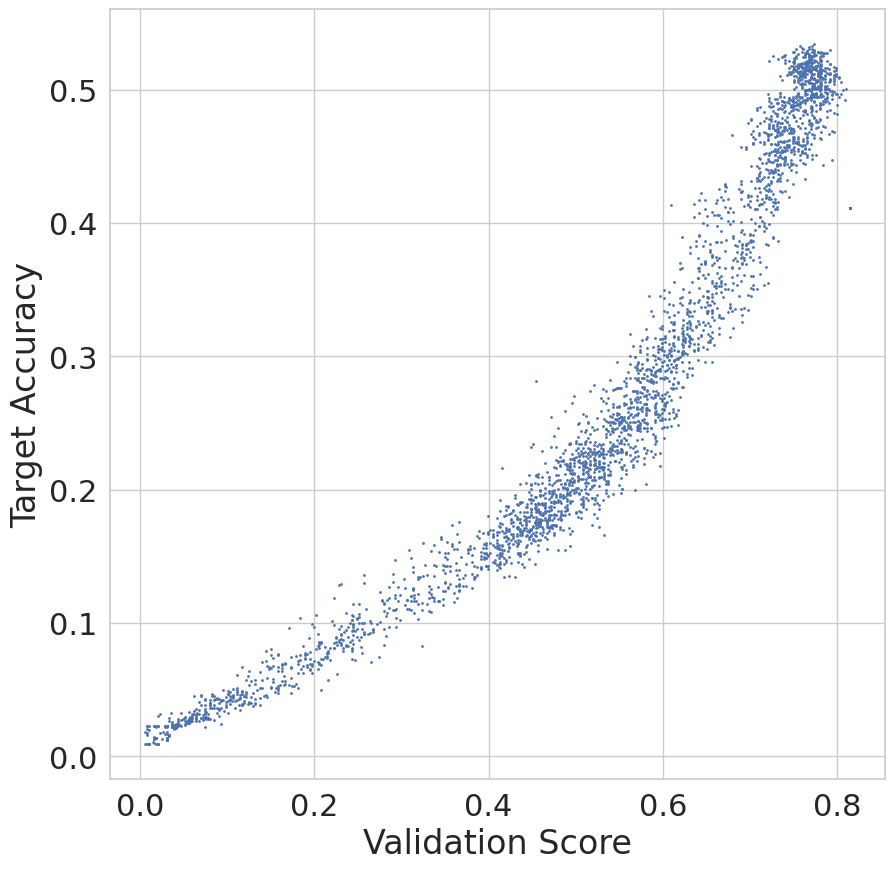}
         \caption{OfficeHome Art $\rightarrow$ Clipart, WSC = 78.5}
         \label{}
     \end{subfigure}
     \hspace{2em}
     \begin{subfigure}[b]{0.4\textwidth}
         \centering
         \includegraphics[width=1.0\textwidth]{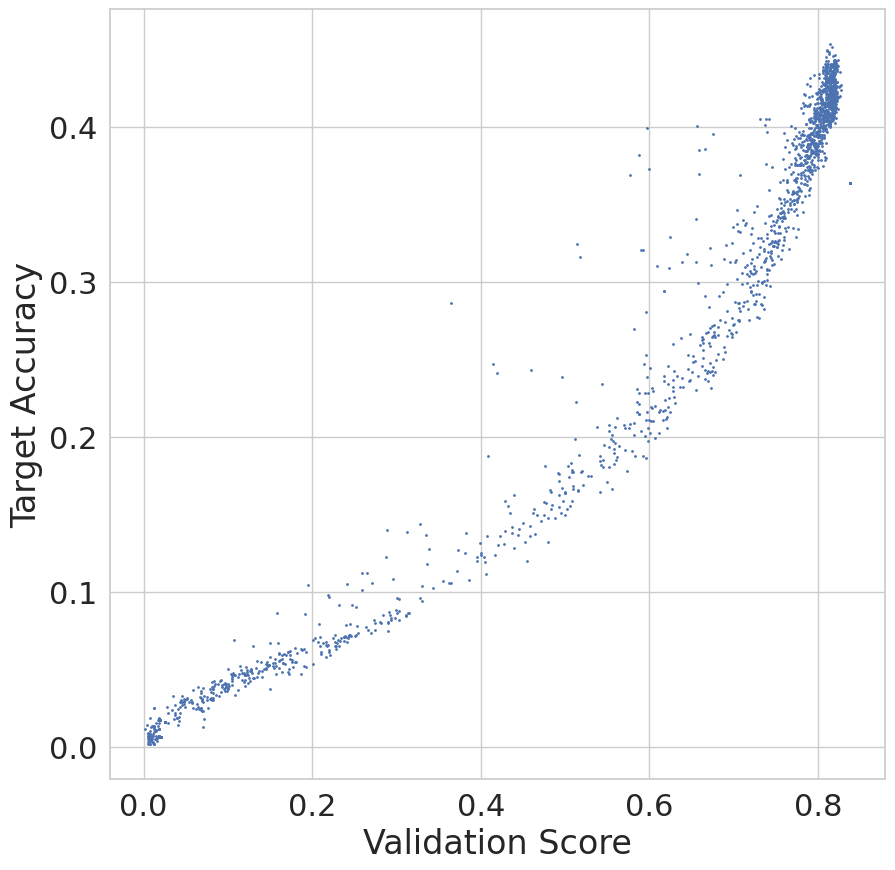}
         \caption{DomainNet126 Clipart $\rightarrow$ Painting, WSC = 53.5}
         \label{}
     \end{subfigure}    
     \caption{These plots show the checkpoints of the DANN algorithm, using the Accuracy (Source Val) validator, on a task from each dataset. Accuracy (Source Val) is the best validator for DANN, as measured by average WSC across tasks (excluding the MNIST task). DANN/Accuracy is the 5th-best performing algorithm/validator pair as measured by AATN across tasks (excluding the MNIST task). Note how the untrained (source-only) model is ranked the highest for both the OfficeHome and DomainNet126 tasks.}
    \label{dann_accuracy_examples}
\end{figure}

%% file: supp_figures/mnist_barplot.tex
\begin{figure}[H]
     \centering
     \centering
     \includegraphics[width=1.0\textwidth]{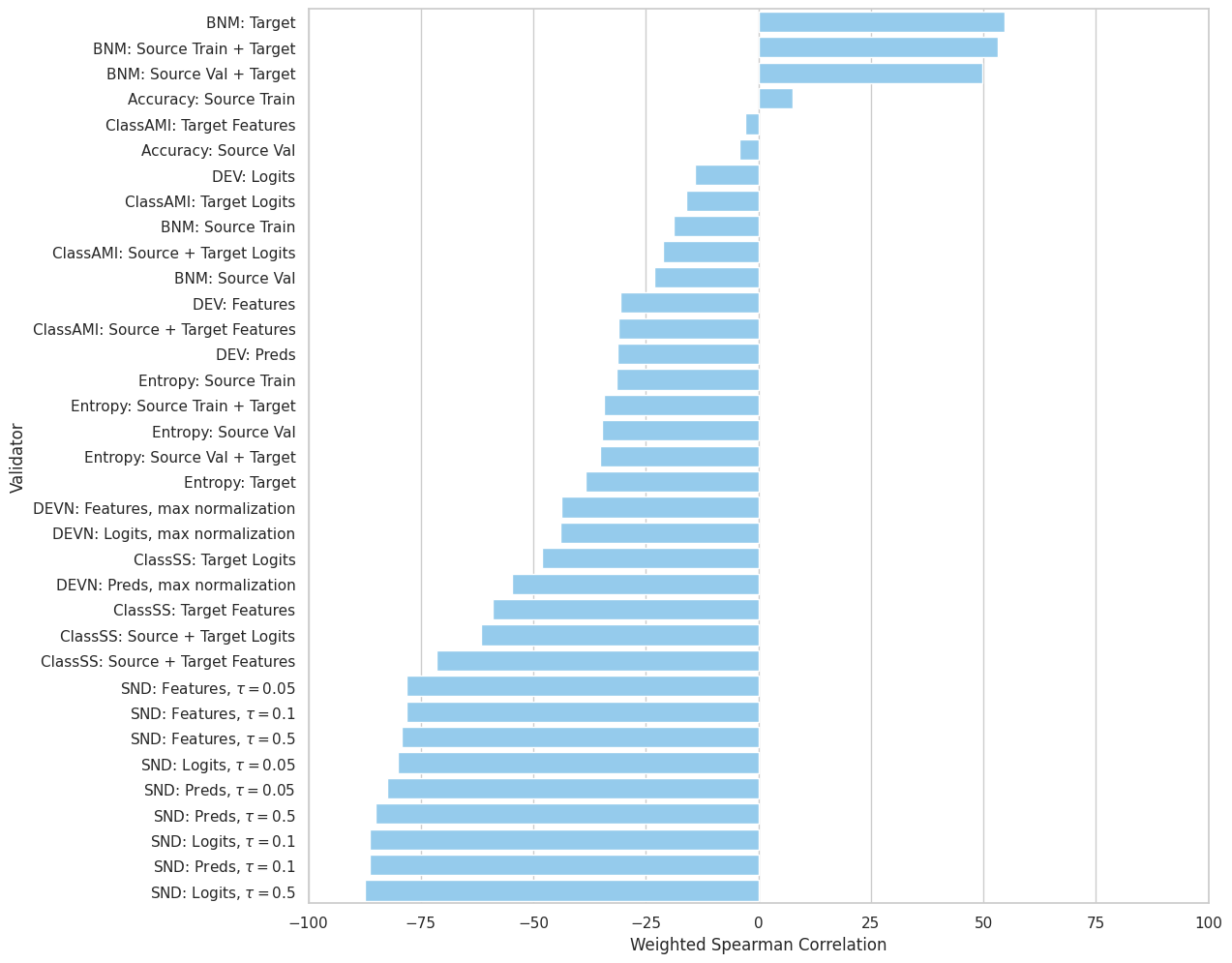}
     \caption{Each validator's WSC on the MNIST $\rightarrow$ MNISTM transfer task. The correlations are computed using the checkpoints of all UDA algorithms.}
    \label{mnist_barplot}
\end{figure}

%% file: supp_tables/supplementary_table_of_contents.tex
\begin{table}[H]
\caption{Summary of tables that show the weighted Spearman correlation of the benchmarked validators.}
\label{supplementary_table_of_contents}
\centering\begin{tabular}{cccc}
\toprule
Algorithm & Office31 \& OfficeHome & DomainNet126 & MNIST$\rightarrow$MNISTM (MM) \\
 \midrule
All combined & Table \ref{office_weighted_spearman_0.0_src_threshold} & Table \ref{domainnet126_weighted_spearman_0.0_src_threshold} & Table \ref{mnist_weighted_spearman_0.0_src_threshold} \\
ATDOC & Table \ref{office_weighted_spearman_0.0_src_threshold_per_adapter_ATDOC} & Table \ref{domainnet126_weighted_spearman_0.0_src_threshold_per_adapter_ATDOC} & Table \ref{mnist_weighted_spearman_0.0_src_threshold_per_adapter} \\
BNM & Table \ref{office_weighted_spearman_0.0_src_threshold_per_adapter_BNM} & Table \ref{domainnet126_weighted_spearman_0.0_src_threshold_per_adapter_BNM} & Table \ref{mnist_weighted_spearman_0.0_src_threshold_per_adapter} \\
BSP & Table \ref{office_weighted_spearman_0.0_src_threshold_per_adapter_BSP} & Table \ref{domainnet126_weighted_spearman_0.0_src_threshold_per_adapter_BSP} & Table \ref{mnist_weighted_spearman_0.0_src_threshold_per_adapter} \\
CDAN & Table \ref{office_weighted_spearman_0.0_src_threshold_per_adapter_CDAN} & Table \ref{domainnet126_weighted_spearman_0.0_src_threshold_per_adapter_CDAN} & Table \ref{mnist_weighted_spearman_0.0_src_threshold_per_adapter} \\
DANN & Table \ref{office_weighted_spearman_0.0_src_threshold_per_adapter_DANN} & Table \ref{domainnet126_weighted_spearman_0.0_src_threshold_per_adapter_DANN} & Table \ref{mnist_weighted_spearman_0.0_src_threshold_per_adapter} \\
GVB & Table \ref{office_weighted_spearman_0.0_src_threshold_per_adapter_GVB} & Table \ref{domainnet126_weighted_spearman_0.0_src_threshold_per_adapter_GVB} & Table \ref{mnist_weighted_spearman_0.0_src_threshold_per_adapter} \\
IM & Table \ref{office_weighted_spearman_0.0_src_threshold_per_adapter_IM} & Table \ref{domainnet126_weighted_spearman_0.0_src_threshold_per_adapter_IM} & Table \ref{mnist_weighted_spearman_0.0_src_threshold_per_adapter} \\
MCC & Table \ref{office_weighted_spearman_0.0_src_threshold_per_adapter_MCC} & Table \ref{domainnet126_weighted_spearman_0.0_src_threshold_per_adapter_MCC} & Table \ref{mnist_weighted_spearman_0.0_src_threshold_per_adapter} \\
MCD & Table \ref{office_weighted_spearman_0.0_src_threshold_per_adapter_MCD} & Table \ref{domainnet126_weighted_spearman_0.0_src_threshold_per_adapter_MCD} & Table \ref{mnist_weighted_spearman_0.0_src_threshold_per_adapter} \\
MMD & Table \ref{office_weighted_spearman_0.0_src_threshold_per_adapter_MMD} & Table \ref{domainnet126_weighted_spearman_0.0_src_threshold_per_adapter_MMD} & Table \ref{mnist_weighted_spearman_0.0_src_threshold_per_adapter} \\
\bottomrule
\end{tabular}
\end{table}

%% file: supp_tables/office31_officehome/weighted_spearman_0.0_src_threshold.tex
\def\weightedspearmanzerozerosrcthresholdAD#1{\ifdim#1pt>82.5pt\cellcolor{lime!100}\else\ifdim#1pt>82.5pt\cellcolor{lime!90}\else\ifdim#1pt>82.5pt\cellcolor{lime!80}\else\ifdim#1pt>82.5pt\cellcolor{lime!70}\else\ifdim#1pt>82.5pt\cellcolor{lime!60}\else\ifdim#1pt>82.5pt\cellcolor{lime!50}\else\ifdim#1pt>82.5pt\cellcolor{lime!40}\else\ifdim#1pt>82.5pt\cellcolor{lime!30}\else\ifdim#1pt>82.5pt\cellcolor{lime!20}\else\ifdim#1pt>82.5pt\cellcolor{lime!10}\else\cellcolor{lime!0}\fi\fi\fi\fi\fi\fi\fi\fi\fi\fi#1}

\def\weightedspearmanzerozerosrcthresholdAW#1{\ifdim#1pt>83.5pt\cellcolor{lime!100}\else\ifdim#1pt>83.5pt\cellcolor{lime!90}\else\ifdim#1pt>83.5pt\cellcolor{lime!80}\else\ifdim#1pt>83.5pt\cellcolor{lime!70}\else\ifdim#1pt>83.5pt\cellcolor{lime!60}\else\ifdim#1pt>83.5pt\cellcolor{lime!50}\else\ifdim#1pt>83.5pt\cellcolor{lime!40}\else\ifdim#1pt>83.5pt\cellcolor{lime!30}\else\ifdim#1pt>83.5pt\cellcolor{lime!20}\else\ifdim#1pt>83.5pt\cellcolor{lime!10}\else\cellcolor{lime!0}\fi\fi\fi\fi\fi\fi\fi\fi\fi\fi#1}

\def\weightedspearmanzerozerosrcthresholdDA#1{\ifdim#1pt>81.1pt\cellcolor{lime!100}\else\ifdim#1pt>81.1pt\cellcolor{lime!90}\else\ifdim#1pt>81.1pt\cellcolor{lime!80}\else\ifdim#1pt>81.1pt\cellcolor{lime!70}\else\ifdim#1pt>81.1pt\cellcolor{lime!60}\else\ifdim#1pt>81.1pt\cellcolor{lime!50}\else\ifdim#1pt>81.1pt\cellcolor{lime!40}\else\ifdim#1pt>81.1pt\cellcolor{lime!30}\else\ifdim#1pt>81.1pt\cellcolor{lime!20}\else\ifdim#1pt>81.1pt\cellcolor{lime!10}\else\cellcolor{lime!0}\fi\fi\fi\fi\fi\fi\fi\fi\fi\fi#1}

\def\weightedspearmanzerozerosrcthresholdDW#1{\ifdim#1pt>80.2pt\cellcolor{lime!100}\else\ifdim#1pt>80.2pt\cellcolor{lime!90}\else\ifdim#1pt>80.2pt\cellcolor{lime!80}\else\ifdim#1pt>80.2pt\cellcolor{lime!70}\else\ifdim#1pt>80.2pt\cellcolor{lime!60}\else\ifdim#1pt>80.2pt\cellcolor{lime!50}\else\ifdim#1pt>80.2pt\cellcolor{lime!40}\else\ifdim#1pt>80.2pt\cellcolor{lime!30}\else\ifdim#1pt>80.2pt\cellcolor{lime!20}\else\ifdim#1pt>80.2pt\cellcolor{lime!10}\else\cellcolor{lime!0}\fi\fi\fi\fi\fi\fi\fi\fi\fi\fi#1}

\def\weightedspearmanzerozerosrcthresholdWA#1{\ifdim#1pt>81.8pt\cellcolor{lime!100}\else\ifdim#1pt>81.8pt\cellcolor{lime!90}\else\ifdim#1pt>81.8pt\cellcolor{lime!80}\else\ifdim#1pt>81.8pt\cellcolor{lime!70}\else\ifdim#1pt>81.8pt\cellcolor{lime!60}\else\ifdim#1pt>81.8pt\cellcolor{lime!50}\else\ifdim#1pt>81.8pt\cellcolor{lime!40}\else\ifdim#1pt>81.8pt\cellcolor{lime!30}\else\ifdim#1pt>81.8pt\cellcolor{lime!20}\else\ifdim#1pt>81.8pt\cellcolor{lime!10}\else\cellcolor{lime!0}\fi\fi\fi\fi\fi\fi\fi\fi\fi\fi#1}

\def\weightedspearmanzerozerosrcthresholdWD#1{\ifdim#1pt>73.7pt\cellcolor{lime!100}\else\ifdim#1pt>73.7pt\cellcolor{lime!90}\else\ifdim#1pt>73.7pt\cellcolor{lime!80}\else\ifdim#1pt>73.7pt\cellcolor{lime!70}\else\ifdim#1pt>73.7pt\cellcolor{lime!60}\else\ifdim#1pt>73.7pt\cellcolor{lime!50}\else\ifdim#1pt>73.7pt\cellcolor{lime!40}\else\ifdim#1pt>73.7pt\cellcolor{lime!30}\else\ifdim#1pt>73.7pt\cellcolor{lime!20}\else\ifdim#1pt>73.7pt\cellcolor{lime!10}\else\cellcolor{lime!0}\fi\fi\fi\fi\fi\fi\fi\fi\fi\fi#1}

\def\weightedspearmanzerozerosrcthresholdAC#1{\ifdim#1pt>86.9pt\cellcolor{lime!100}\else\ifdim#1pt>86.9pt\cellcolor{lime!90}\else\ifdim#1pt>86.9pt\cellcolor{lime!80}\else\ifdim#1pt>86.9pt\cellcolor{lime!70}\else\ifdim#1pt>86.9pt\cellcolor{lime!60}\else\ifdim#1pt>86.9pt\cellcolor{lime!50}\else\ifdim#1pt>86.9pt\cellcolor{lime!40}\else\ifdim#1pt>86.9pt\cellcolor{lime!30}\else\ifdim#1pt>86.9pt\cellcolor{lime!20}\else\ifdim#1pt>86.9pt\cellcolor{lime!10}\else\cellcolor{lime!0}\fi\fi\fi\fi\fi\fi\fi\fi\fi\fi#1}

\def\weightedspearmanzerozerosrcthresholdAP#1{\ifdim#1pt>94.5pt\cellcolor{lime!100}\else\ifdim#1pt>94.5pt\cellcolor{lime!90}\else\ifdim#1pt>94.5pt\cellcolor{lime!80}\else\ifdim#1pt>94.5pt\cellcolor{lime!70}\else\ifdim#1pt>94.5pt\cellcolor{lime!60}\else\ifdim#1pt>94.5pt\cellcolor{lime!50}\else\ifdim#1pt>94.5pt\cellcolor{lime!40}\else\ifdim#1pt>94.5pt\cellcolor{lime!30}\else\ifdim#1pt>94.5pt\cellcolor{lime!20}\else\ifdim#1pt>94.5pt\cellcolor{lime!10}\else\cellcolor{lime!0}\fi\fi\fi\fi\fi\fi\fi\fi\fi\fi#1}

\def\weightedspearmanzerozerosrcthresholdAR#1{\ifdim#1pt>96.4pt\cellcolor{lime!100}\else\ifdim#1pt>96.4pt\cellcolor{lime!90}\else\ifdim#1pt>96.4pt\cellcolor{lime!80}\else\ifdim#1pt>96.4pt\cellcolor{lime!70}\else\ifdim#1pt>96.4pt\cellcolor{lime!60}\else\ifdim#1pt>96.4pt\cellcolor{lime!50}\else\ifdim#1pt>96.4pt\cellcolor{lime!40}\else\ifdim#1pt>96.4pt\cellcolor{lime!30}\else\ifdim#1pt>96.4pt\cellcolor{lime!20}\else\ifdim#1pt>96.4pt\cellcolor{lime!10}\else\cellcolor{lime!0}\fi\fi\fi\fi\fi\fi\fi\fi\fi\fi#1}

\def\weightedspearmanzerozerosrcthresholdCA#1{\ifdim#1pt>80.7pt\cellcolor{lime!100}\else\ifdim#1pt>80.7pt\cellcolor{lime!90}\else\ifdim#1pt>80.7pt\cellcolor{lime!80}\else\ifdim#1pt>80.7pt\cellcolor{lime!70}\else\ifdim#1pt>80.7pt\cellcolor{lime!60}\else\ifdim#1pt>80.7pt\cellcolor{lime!50}\else\ifdim#1pt>80.7pt\cellcolor{lime!40}\else\ifdim#1pt>80.7pt\cellcolor{lime!30}\else\ifdim#1pt>80.7pt\cellcolor{lime!20}\else\ifdim#1pt>80.7pt\cellcolor{lime!10}\else\cellcolor{lime!0}\fi\fi\fi\fi\fi\fi\fi\fi\fi\fi#1}

\def\weightedspearmanzerozerosrcthresholdCP#1{\ifdim#1pt>85.4pt\cellcolor{lime!100}\else\ifdim#1pt>85.4pt\cellcolor{lime!90}\else\ifdim#1pt>85.4pt\cellcolor{lime!80}\else\ifdim#1pt>85.4pt\cellcolor{lime!70}\else\ifdim#1pt>85.4pt\cellcolor{lime!60}\else\ifdim#1pt>85.4pt\cellcolor{lime!50}\else\ifdim#1pt>85.4pt\cellcolor{lime!40}\else\ifdim#1pt>85.4pt\cellcolor{lime!30}\else\ifdim#1pt>85.4pt\cellcolor{lime!20}\else\ifdim#1pt>85.4pt\cellcolor{lime!10}\else\cellcolor{lime!0}\fi\fi\fi\fi\fi\fi\fi\fi\fi\fi#1}

\def\weightedspearmanzerozerosrcthresholdCR#1{\ifdim#1pt>86.0pt\cellcolor{lime!100}\else\ifdim#1pt>86.0pt\cellcolor{lime!90}\else\ifdim#1pt>86.0pt\cellcolor{lime!80}\else\ifdim#1pt>86.0pt\cellcolor{lime!70}\else\ifdim#1pt>86.0pt\cellcolor{lime!60}\else\ifdim#1pt>86.0pt\cellcolor{lime!50}\else\ifdim#1pt>86.0pt\cellcolor{lime!40}\else\ifdim#1pt>86.0pt\cellcolor{lime!30}\else\ifdim#1pt>86.0pt\cellcolor{lime!20}\else\ifdim#1pt>86.0pt\cellcolor{lime!10}\else\cellcolor{lime!0}\fi\fi\fi\fi\fi\fi\fi\fi\fi\fi#1}

\def\weightedspearmanzerozerosrcthresholdPA#1{\ifdim#1pt>92.2pt\cellcolor{lime!100}\else\ifdim#1pt>92.2pt\cellcolor{lime!90}\else\ifdim#1pt>92.2pt\cellcolor{lime!80}\else\ifdim#1pt>92.2pt\cellcolor{lime!70}\else\ifdim#1pt>92.2pt\cellcolor{lime!60}\else\ifdim#1pt>92.2pt\cellcolor{lime!50}\else\ifdim#1pt>92.2pt\cellcolor{lime!40}\else\ifdim#1pt>92.2pt\cellcolor{lime!30}\else\ifdim#1pt>92.2pt\cellcolor{lime!20}\else\ifdim#1pt>92.2pt\cellcolor{lime!10}\else\cellcolor{lime!0}\fi\fi\fi\fi\fi\fi\fi\fi\fi\fi#1}

\def\weightedspearmanzerozerosrcthresholdPC#1{\ifdim#1pt>90.4pt\cellcolor{lime!100}\else\ifdim#1pt>90.4pt\cellcolor{lime!90}\else\ifdim#1pt>90.4pt\cellcolor{lime!80}\else\ifdim#1pt>90.4pt\cellcolor{lime!70}\else\ifdim#1pt>90.4pt\cellcolor{lime!60}\else\ifdim#1pt>90.4pt\cellcolor{lime!50}\else\ifdim#1pt>90.4pt\cellcolor{lime!40}\else\ifdim#1pt>90.4pt\cellcolor{lime!30}\else\ifdim#1pt>90.4pt\cellcolor{lime!20}\else\ifdim#1pt>90.4pt\cellcolor{lime!10}\else\cellcolor{lime!0}\fi\fi\fi\fi\fi\fi\fi\fi\fi\fi#1}

\def\weightedspearmanzerozerosrcthresholdPR#1{\ifdim#1pt>94.8pt\cellcolor{lime!100}\else\ifdim#1pt>94.8pt\cellcolor{lime!90}\else\ifdim#1pt>94.8pt\cellcolor{lime!80}\else\ifdim#1pt>94.8pt\cellcolor{lime!70}\else\ifdim#1pt>94.8pt\cellcolor{lime!60}\else\ifdim#1pt>94.8pt\cellcolor{lime!50}\else\ifdim#1pt>94.8pt\cellcolor{lime!40}\else\ifdim#1pt>94.8pt\cellcolor{lime!30}\else\ifdim#1pt>94.8pt\cellcolor{lime!20}\else\ifdim#1pt>94.8pt\cellcolor{lime!10}\else\cellcolor{lime!0}\fi\fi\fi\fi\fi\fi\fi\fi\fi\fi#1}

\def\weightedspearmanzerozerosrcthresholdRA#1{\ifdim#1pt>92.3pt\cellcolor{lime!100}\else\ifdim#1pt>92.3pt\cellcolor{lime!90}\else\ifdim#1pt>92.3pt\cellcolor{lime!80}\else\ifdim#1pt>92.3pt\cellcolor{lime!70}\else\ifdim#1pt>92.3pt\cellcolor{lime!60}\else\ifdim#1pt>92.3pt\cellcolor{lime!50}\else\ifdim#1pt>92.3pt\cellcolor{lime!40}\else\ifdim#1pt>92.3pt\cellcolor{lime!30}\else\ifdim#1pt>92.3pt\cellcolor{lime!20}\else\ifdim#1pt>92.3pt\cellcolor{lime!10}\else\cellcolor{lime!0}\fi\fi\fi\fi\fi\fi\fi\fi\fi\fi#1}

\def\weightedspearmanzerozerosrcthresholdRC#1{\ifdim#1pt>89.0pt\cellcolor{lime!100}\else\ifdim#1pt>88.6pt\cellcolor{lime!90}\else\ifdim#1pt>88.1pt\cellcolor{lime!80}\else\ifdim#1pt>87.7pt\cellcolor{lime!70}\else\ifdim#1pt>87.2pt\cellcolor{lime!60}\else\ifdim#1pt>86.7pt\cellcolor{lime!50}\else\ifdim#1pt>86.3pt\cellcolor{lime!40}\else\ifdim#1pt>85.8pt\cellcolor{lime!30}\else\ifdim#1pt>85.4pt\cellcolor{lime!20}\else\ifdim#1pt>84.9pt\cellcolor{lime!10}\else\cellcolor{lime!0}\fi\fi\fi\fi\fi\fi\fi\fi\fi\fi#1}

\def\weightedspearmanzerozerosrcthresholdRP#1{\ifdim#1pt>94.0pt\cellcolor{lime!100}\else\ifdim#1pt>94.0pt\cellcolor{lime!90}\else\ifdim#1pt>94.0pt\cellcolor{lime!80}\else\ifdim#1pt>94.0pt\cellcolor{lime!70}\else\ifdim#1pt>94.0pt\cellcolor{lime!60}\else\ifdim#1pt>94.0pt\cellcolor{lime!50}\else\ifdim#1pt>94.0pt\cellcolor{lime!40}\else\ifdim#1pt>94.0pt\cellcolor{lime!30}\else\ifdim#1pt>94.0pt\cellcolor{lime!20}\else\ifdim#1pt>94.0pt\cellcolor{lime!10}\else\cellcolor{lime!0}\fi\fi\fi\fi\fi\fi\fi\fi\fi\fi#1}

\def\weightedspearmanzerozerosrcthresholdMean#1{\ifdim#1pt>86.7pt\cellcolor{lime!100}\else\ifdim#1pt>86.7pt\cellcolor{lime!90}\else\ifdim#1pt>86.7pt\cellcolor{lime!80}\else\ifdim#1pt>86.7pt\cellcolor{lime!70}\else\ifdim#1pt>86.7pt\cellcolor{lime!60}\else\ifdim#1pt>86.7pt\cellcolor{lime!50}\else\ifdim#1pt>86.7pt\cellcolor{lime!40}\else\ifdim#1pt>86.7pt\cellcolor{lime!30}\else\ifdim#1pt>86.7pt\cellcolor{lime!20}\else\ifdim#1pt>86.7pt\cellcolor{lime!10}\else\cellcolor{lime!0}\fi\fi\fi\fi\fi\fi\fi\fi\fi\fi#1}

\def\weightedspearmanzerozerosrcthresholdStd#1{\ifdim#1pt<4.9pt\cellcolor{lime!100}\else\ifdim#1pt<5.0pt\cellcolor{lime!90}\else\ifdim#1pt<5.2pt\cellcolor{lime!80}\else\ifdim#1pt<5.3pt\cellcolor{lime!70}\else\ifdim#1pt<5.5pt\cellcolor{lime!60}\else\ifdim#1pt<5.6pt\cellcolor{lime!50}\else\ifdim#1pt<5.8pt\cellcolor{lime!40}\else\ifdim#1pt<5.9pt\cellcolor{lime!30}\else\ifdim#1pt<6.0pt\cellcolor{lime!20}\else\ifdim#1pt<6.2pt\cellcolor{lime!10}\else\cellcolor{lime!0}\fi\fi\fi\fi\fi\fi\fi\fi\fi\fi#1}

\begin{table}[H]
\centering
\caption{The weighted Spearman correlation of each validator/task pair, using the checkpoints of all algorithms.}
\label{office_weighted_spearman_0.0_src_threshold}
\resizebox{!}{0.21\textheight}{
\begin{tabular}{llrrrrrrrrrrrrrrrrrrrr}
\toprule
 &  & AD & AW & DA & DW & WA & WD & AC & AP & AR & CA & CP & CR & PA & PC & PR & RA & RC & RP & Mean & Std \\
\midrule
\multirow[c]{2}{*}{Accuracy} & Source Train & \weightedspearmanzerozerosrcthresholdAD{66.3} & \weightedspearmanzerozerosrcthresholdAW{71.1} & \weightedspearmanzerozerosrcthresholdDA{79.2} & \weightedspearmanzerozerosrcthresholdDW{67.3} & \weightedspearmanzerozerosrcthresholdWA{81.2} & \weightedspearmanzerozerosrcthresholdWD{65.2} & \weightedspearmanzerozerosrcthresholdAC{61.6} & \weightedspearmanzerozerosrcthresholdAP{63.5} & \weightedspearmanzerozerosrcthresholdAR{56.1} & \weightedspearmanzerozerosrcthresholdCA{52.7} & \weightedspearmanzerozerosrcthresholdCP{54.0} & \weightedspearmanzerozerosrcthresholdCR{55.5} & \weightedspearmanzerozerosrcthresholdPA{71.8} & \weightedspearmanzerozerosrcthresholdPC{69.6} & \weightedspearmanzerozerosrcthresholdPR{72.5} & \weightedspearmanzerozerosrcthresholdRA{54.7} & \weightedspearmanzerozerosrcthresholdRC{70.8} & \weightedspearmanzerozerosrcthresholdRP{62.5} & \weightedspearmanzerozerosrcthresholdMean{65.3} & \weightedspearmanzerozerosrcthresholdStd{8.3} \\
\bottomrule
\end{tabular}}
\end{table}

%% file: supp_tables/office31_officehome/weighted_spearman_0.0_src_threshold_per_adapter_ATDOC.tex
\def\weightedspearmanzerozerosrcthresholdperadapterAD#1{\ifdim#1pt>93.6pt\cellcolor{lime!100}\else\ifdim#1pt>93.0pt\cellcolor{lime!90}\else\ifdim#1pt>92.3pt\cellcolor{lime!80}\else\ifdim#1pt>91.6pt\cellcolor{lime!70}\else\ifdim#1pt>90.9pt\cellcolor{lime!60}\else\ifdim#1pt>90.3pt\cellcolor{lime!50}\else\ifdim#1pt>89.6pt\cellcolor{lime!40}\else\ifdim#1pt>88.9pt\cellcolor{lime!30}\else\ifdim#1pt>88.3pt\cellcolor{lime!20}\else\ifdim#1pt>87.6pt\cellcolor{lime!10}\else\cellcolor{lime!0}\fi\fi\fi\fi\fi\fi\fi\fi\fi\fi#1}

\def\weightedspearmanzerozerosrcthresholdperadapterAW#1{\ifdim#1pt>93.7pt\cellcolor{lime!100}\else\ifdim#1pt>93.5pt\cellcolor{lime!90}\else\ifdim#1pt>93.4pt\cellcolor{lime!80}\else\ifdim#1pt>93.2pt\cellcolor{lime!70}\else\ifdim#1pt>93.0pt\cellcolor{lime!60}\else\ifdim#1pt>92.8pt\cellcolor{lime!50}\else\ifdim#1pt>92.6pt\cellcolor{lime!40}\else\ifdim#1pt>92.5pt\cellcolor{lime!30}\else\ifdim#1pt>92.3pt\cellcolor{lime!20}\else\ifdim#1pt>92.1pt\cellcolor{lime!10}\else\cellcolor{lime!0}\fi\fi\fi\fi\fi\fi\fi\fi\fi\fi#1}

\def\weightedspearmanzerozerosrcthresholdperadapterDA#1{\ifdim#1pt>90.5pt\cellcolor{lime!100}\else\ifdim#1pt>90.3pt\cellcolor{lime!90}\else\ifdim#1pt>90.2pt\cellcolor{lime!80}\else\ifdim#1pt>90.1pt\cellcolor{lime!70}\else\ifdim#1pt>89.9pt\cellcolor{lime!60}\else\ifdim#1pt>89.8pt\cellcolor{lime!50}\else\ifdim#1pt>89.7pt\cellcolor{lime!40}\else\ifdim#1pt>89.6pt\cellcolor{lime!30}\else\ifdim#1pt>89.4pt\cellcolor{lime!20}\else\ifdim#1pt>89.3pt\cellcolor{lime!10}\else\cellcolor{lime!0}\fi\fi\fi\fi\fi\fi\fi\fi\fi\fi#1}

\def\weightedspearmanzerozerosrcthresholdperadapterDW#1{\ifdim#1pt>93.4pt\cellcolor{lime!100}\else\ifdim#1pt>93.2pt\cellcolor{lime!90}\else\ifdim#1pt>93.1pt\cellcolor{lime!80}\else\ifdim#1pt>93.0pt\cellcolor{lime!70}\else\ifdim#1pt>92.8pt\cellcolor{lime!60}\else\ifdim#1pt>92.7pt\cellcolor{lime!50}\else\ifdim#1pt>92.6pt\cellcolor{lime!40}\else\ifdim#1pt>92.5pt\cellcolor{lime!30}\else\ifdim#1pt>92.3pt\cellcolor{lime!20}\else\ifdim#1pt>92.2pt\cellcolor{lime!10}\else\cellcolor{lime!0}\fi\fi\fi\fi\fi\fi\fi\fi\fi\fi#1}

\def\weightedspearmanzerozerosrcthresholdperadapterWA#1{\ifdim#1pt>86.2pt\cellcolor{lime!100}\else\ifdim#1pt>85.6pt\cellcolor{lime!90}\else\ifdim#1pt>85.1pt\cellcolor{lime!80}\else\ifdim#1pt>84.5pt\cellcolor{lime!70}\else\ifdim#1pt>83.9pt\cellcolor{lime!60}\else\ifdim#1pt>83.3pt\cellcolor{lime!50}\else\ifdim#1pt>82.7pt\cellcolor{lime!40}\else\ifdim#1pt>82.2pt\cellcolor{lime!30}\else\ifdim#1pt>81.6pt\cellcolor{lime!20}\else\ifdim#1pt>81.0pt\cellcolor{lime!10}\else\cellcolor{lime!0}\fi\fi\fi\fi\fi\fi\fi\fi\fi\fi#1}

\def\weightedspearmanzerozerosrcthresholdperadapterWD#1{\ifdim#1pt>92.4pt\cellcolor{lime!100}\else\ifdim#1pt>92.2pt\cellcolor{lime!90}\else\ifdim#1pt>91.9pt\cellcolor{lime!80}\else\ifdim#1pt>91.6pt\cellcolor{lime!70}\else\ifdim#1pt>91.3pt\cellcolor{lime!60}\else\ifdim#1pt>91.1pt\cellcolor{lime!50}\else\ifdim#1pt>90.8pt\cellcolor{lime!40}\else\ifdim#1pt>90.5pt\cellcolor{lime!30}\else\ifdim#1pt>90.3pt\cellcolor{lime!20}\else\ifdim#1pt>90.0pt\cellcolor{lime!10}\else\cellcolor{lime!0}\fi\fi\fi\fi\fi\fi\fi\fi\fi\fi#1}

\def\weightedspearmanzerozerosrcthresholdperadapterAC#1{\ifdim#1pt>94.7pt\cellcolor{lime!100}\else\ifdim#1pt>94.4pt\cellcolor{lime!90}\else\ifdim#1pt>94.2pt\cellcolor{lime!80}\else\ifdim#1pt>94.0pt\cellcolor{lime!70}\else\ifdim#1pt>93.8pt\cellcolor{lime!60}\else\ifdim#1pt>93.5pt\cellcolor{lime!50}\else\ifdim#1pt>93.3pt\cellcolor{lime!40}\else\ifdim#1pt>93.1pt\cellcolor{lime!30}\else\ifdim#1pt>92.8pt\cellcolor{lime!20}\else\ifdim#1pt>92.6pt\cellcolor{lime!10}\else\cellcolor{lime!0}\fi\fi\fi\fi\fi\fi\fi\fi\fi\fi#1}

\def\weightedspearmanzerozerosrcthresholdperadapterAP#1{\ifdim#1pt>94.9pt\cellcolor{lime!100}\else\ifdim#1pt>94.8pt\cellcolor{lime!90}\else\ifdim#1pt>94.6pt\cellcolor{lime!80}\else\ifdim#1pt>94.5pt\cellcolor{lime!70}\else\ifdim#1pt>94.3pt\cellcolor{lime!60}\else\ifdim#1pt>94.1pt\cellcolor{lime!50}\else\ifdim#1pt>94.0pt\cellcolor{lime!40}\else\ifdim#1pt>93.8pt\cellcolor{lime!30}\else\ifdim#1pt>93.7pt\cellcolor{lime!20}\else\ifdim#1pt>93.5pt\cellcolor{lime!10}\else\cellcolor{lime!0}\fi\fi\fi\fi\fi\fi\fi\fi\fi\fi#1}

\def\weightedspearmanzerozerosrcthresholdperadapterAR#1{\ifdim#1pt>94.6pt\cellcolor{lime!100}\else\ifdim#1pt>94.6pt\cellcolor{lime!90}\else\ifdim#1pt>94.5pt\cellcolor{lime!80}\else\ifdim#1pt>94.5pt\cellcolor{lime!70}\else\ifdim#1pt>94.4pt\cellcolor{lime!60}\else\ifdim#1pt>94.3pt\cellcolor{lime!50}\else\ifdim#1pt>94.3pt\cellcolor{lime!40}\else\ifdim#1pt>94.2pt\cellcolor{lime!30}\else\ifdim#1pt>94.2pt\cellcolor{lime!20}\else\ifdim#1pt>94.1pt\cellcolor{lime!10}\else\cellcolor{lime!0}\fi\fi\fi\fi\fi\fi\fi\fi\fi\fi#1}

\def\weightedspearmanzerozerosrcthresholdperadapterCA#1{\ifdim#1pt>95.1pt\cellcolor{lime!100}\else\ifdim#1pt>94.2pt\cellcolor{lime!90}\else\ifdim#1pt>93.2pt\cellcolor{lime!80}\else\ifdim#1pt>92.3pt\cellcolor{lime!70}\else\ifdim#1pt>91.3pt\cellcolor{lime!60}\else\ifdim#1pt>90.3pt\cellcolor{lime!50}\else\ifdim#1pt>89.4pt\cellcolor{lime!40}\else\ifdim#1pt>88.4pt\cellcolor{lime!30}\else\ifdim#1pt>87.5pt\cellcolor{lime!20}\else\ifdim#1pt>86.5pt\cellcolor{lime!10}\else\cellcolor{lime!0}\fi\fi\fi\fi\fi\fi\fi\fi\fi\fi#1}

\def\weightedspearmanzerozerosrcthresholdperadapterCP#1{\ifdim#1pt>94.6pt\cellcolor{lime!100}\else\ifdim#1pt>93.3pt\cellcolor{lime!90}\else\ifdim#1pt>92.1pt\cellcolor{lime!80}\else\ifdim#1pt>90.8pt\cellcolor{lime!70}\else\ifdim#1pt>89.5pt\cellcolor{lime!60}\else\ifdim#1pt>88.2pt\cellcolor{lime!50}\else\ifdim#1pt>86.9pt\cellcolor{lime!40}\else\ifdim#1pt>85.7pt\cellcolor{lime!30}\else\ifdim#1pt>84.4pt\cellcolor{lime!20}\else\ifdim#1pt>83.1pt\cellcolor{lime!10}\else\cellcolor{lime!0}\fi\fi\fi\fi\fi\fi\fi\fi\fi\fi#1}

\def\weightedspearmanzerozerosrcthresholdperadapterCR#1{\ifdim#1pt>94.1pt\cellcolor{lime!100}\else\ifdim#1pt>93.0pt\cellcolor{lime!90}\else\ifdim#1pt>91.8pt\cellcolor{lime!80}\else\ifdim#1pt>90.7pt\cellcolor{lime!70}\else\ifdim#1pt>89.6pt\cellcolor{lime!60}\else\ifdim#1pt>88.5pt\cellcolor{lime!50}\else\ifdim#1pt>87.4pt\cellcolor{lime!40}\else\ifdim#1pt>86.2pt\cellcolor{lime!30}\else\ifdim#1pt>85.1pt\cellcolor{lime!20}\else\ifdim#1pt>84.0pt\cellcolor{lime!10}\else\cellcolor{lime!0}\fi\fi\fi\fi\fi\fi\fi\fi\fi\fi#1}

\def\weightedspearmanzerozerosrcthresholdperadapterPA#1{\ifdim#1pt>96.2pt\cellcolor{lime!100}\else\ifdim#1pt>95.8pt\cellcolor{lime!90}\else\ifdim#1pt>95.4pt\cellcolor{lime!80}\else\ifdim#1pt>94.9pt\cellcolor{lime!70}\else\ifdim#1pt>94.5pt\cellcolor{lime!60}\else\ifdim#1pt>94.0pt\cellcolor{lime!50}\else\ifdim#1pt>93.5pt\cellcolor{lime!40}\else\ifdim#1pt>93.1pt\cellcolor{lime!30}\else\ifdim#1pt>92.7pt\cellcolor{lime!20}\else\ifdim#1pt>92.2pt\cellcolor{lime!10}\else\cellcolor{lime!0}\fi\fi\fi\fi\fi\fi\fi\fi\fi\fi#1}

\def\weightedspearmanzerozerosrcthresholdperadapterPC#1{\ifdim#1pt>91.2pt\cellcolor{lime!100}\else\ifdim#1pt>90.1pt\cellcolor{lime!90}\else\ifdim#1pt>88.9pt\cellcolor{lime!80}\else\ifdim#1pt>87.7pt\cellcolor{lime!70}\else\ifdim#1pt>86.6pt\cellcolor{lime!60}\else\ifdim#1pt>85.4pt\cellcolor{lime!50}\else\ifdim#1pt>84.2pt\cellcolor{lime!40}\else\ifdim#1pt>83.0pt\cellcolor{lime!30}\else\ifdim#1pt>81.9pt\cellcolor{lime!20}\else\ifdim#1pt>80.7pt\cellcolor{lime!10}\else\cellcolor{lime!0}\fi\fi\fi\fi\fi\fi\fi\fi\fi\fi#1}

\def\weightedspearmanzerozerosrcthresholdperadapterPR#1{\ifdim#1pt>97.2pt\cellcolor{lime!100}\else\ifdim#1pt>96.7pt\cellcolor{lime!90}\else\ifdim#1pt>96.1pt\cellcolor{lime!80}\else\ifdim#1pt>95.5pt\cellcolor{lime!70}\else\ifdim#1pt>94.9pt\cellcolor{lime!60}\else\ifdim#1pt>94.4pt\cellcolor{lime!50}\else\ifdim#1pt>93.8pt\cellcolor{lime!40}\else\ifdim#1pt>93.2pt\cellcolor{lime!30}\else\ifdim#1pt>92.7pt\cellcolor{lime!20}\else\ifdim#1pt>92.1pt\cellcolor{lime!10}\else\cellcolor{lime!0}\fi\fi\fi\fi\fi\fi\fi\fi\fi\fi#1}

\def\weightedspearmanzerozerosrcthresholdperadapterRA#1{\ifdim#1pt>93.9pt\cellcolor{lime!100}\else\ifdim#1pt>93.9pt\cellcolor{lime!90}\else\ifdim#1pt>93.9pt\cellcolor{lime!80}\else\ifdim#1pt>93.9pt\cellcolor{lime!70}\else\ifdim#1pt>93.9pt\cellcolor{lime!60}\else\ifdim#1pt>93.9pt\cellcolor{lime!50}\else\ifdim#1pt>93.9pt\cellcolor{lime!40}\else\ifdim#1pt>93.9pt\cellcolor{lime!30}\else\ifdim#1pt>93.9pt\cellcolor{lime!20}\else\ifdim#1pt>93.9pt\cellcolor{lime!10}\else\cellcolor{lime!0}\fi\fi\fi\fi\fi\fi\fi\fi\fi\fi#1}

\def\weightedspearmanzerozerosrcthresholdperadapterRC#1{\ifdim#1pt>91.9pt\cellcolor{lime!100}\else\ifdim#1pt>91.8pt\cellcolor{lime!90}\else\ifdim#1pt>91.7pt\cellcolor{lime!80}\else\ifdim#1pt>91.6pt\cellcolor{lime!70}\else\ifdim#1pt>91.5pt\cellcolor{lime!60}\else\ifdim#1pt>91.5pt\cellcolor{lime!50}\else\ifdim#1pt>91.4pt\cellcolor{lime!40}\else\ifdim#1pt>91.3pt\cellcolor{lime!30}\else\ifdim#1pt>91.2pt\cellcolor{lime!20}\else\ifdim#1pt>91.1pt\cellcolor{lime!10}\else\cellcolor{lime!0}\fi\fi\fi\fi\fi\fi\fi\fi\fi\fi#1}

\def\weightedspearmanzerozerosrcthresholdperadapterRP#1{\ifdim#1pt>95.7pt\cellcolor{lime!100}\else\ifdim#1pt>95.6pt\cellcolor{lime!90}\else\ifdim#1pt>95.5pt\cellcolor{lime!80}\else\ifdim#1pt>95.5pt\cellcolor{lime!70}\else\ifdim#1pt>95.5pt\cellcolor{lime!60}\else\ifdim#1pt>95.4pt\cellcolor{lime!50}\else\ifdim#1pt>95.4pt\cellcolor{lime!40}\else\ifdim#1pt>95.3pt\cellcolor{lime!30}\else\ifdim#1pt>95.2pt\cellcolor{lime!20}\else\ifdim#1pt>95.2pt\cellcolor{lime!10}\else\cellcolor{lime!0}\fi\fi\fi\fi\fi\fi\fi\fi\fi\fi#1}

\def\weightedspearmanzerozerosrcthresholdperadapterMean#1{\ifdim#1pt>92.1pt\cellcolor{lime!100}\else\ifdim#1pt>91.8pt\cellcolor{lime!90}\else\ifdim#1pt>91.5pt\cellcolor{lime!80}\else\ifdim#1pt>91.2pt\cellcolor{lime!70}\else\ifdim#1pt>91.0pt\cellcolor{lime!60}\else\ifdim#1pt>90.7pt\cellcolor{lime!50}\else\ifdim#1pt>90.4pt\cellcolor{lime!40}\else\ifdim#1pt>90.1pt\cellcolor{lime!30}\else\ifdim#1pt>89.8pt\cellcolor{lime!20}\else\ifdim#1pt>89.5pt\cellcolor{lime!10}\else\cellcolor{lime!0}\fi\fi\fi\fi\fi\fi\fi\fi\fi\fi#1}

\def\weightedspearmanzerozerosrcthresholdperadapterStd#1{\ifdim#1pt<2.5pt\cellcolor{lime!100}\else\ifdim#1pt<2.7pt\cellcolor{lime!90}\else\ifdim#1pt<3.0pt\cellcolor{lime!80}\else\ifdim#1pt<3.2pt\cellcolor{lime!70}\else\ifdim#1pt<3.4pt\cellcolor{lime!60}\else\ifdim#1pt<3.6pt\cellcolor{lime!50}\else\ifdim#1pt<3.8pt\cellcolor{lime!40}\else\ifdim#1pt<4.1pt\cellcolor{lime!30}\else\ifdim#1pt<4.3pt\cellcolor{lime!20}\else\ifdim#1pt<4.5pt\cellcolor{lime!10}\else\cellcolor{lime!0}\fi\fi\fi\fi\fi\fi\fi\fi\fi\fi#1}

\begin{table}[H]
\centering
\caption{The weighted Spearman correlation of each validator/task pair for \textbf{ATDOC}.}
\label{office_weighted_spearman_0.0_src_threshold_per_adapter_ATDOC}
\resizebox{!}{0.22\textheight}{
\begin{tabular}{llrrrrrrrrrrrrrrrrrrrr}
\toprule
 &  & AD & AW & DA & DW & WA & WD & AC & AP & AR & CA & CP & CR & PA & PC & PR & RA & RC & RP & Mean & Std \\
\midrule
\multirow[c]{2}{*}{Accuracy} & Source Train & \weightedspearmanzerozerosrcthresholdperadapterAD{72.4} & \weightedspearmanzerozerosrcthresholdperadapterAW{75.9} & \weightedspearmanzerozerosrcthresholdperadapterDA{82.8} & \weightedspearmanzerozerosrcthresholdperadapterDW{83.5} & \weightedspearmanzerozerosrcthresholdperadapterWA{78.3} & \weightedspearmanzerozerosrcthresholdperadapterWD{89.7} & \weightedspearmanzerozerosrcthresholdperadapterAC{61.6} & \weightedspearmanzerozerosrcthresholdperadapterAP{64.8} & \weightedspearmanzerozerosrcthresholdperadapterAR{48.6} & \weightedspearmanzerozerosrcthresholdperadapterCA{70.8} & \weightedspearmanzerozerosrcthresholdperadapterCP{40.3} & \weightedspearmanzerozerosrcthresholdperadapterCR{57.2} & \weightedspearmanzerozerosrcthresholdperadapterPA{88.6} & \weightedspearmanzerozerosrcthresholdperadapterPC{63.9} & \weightedspearmanzerozerosrcthresholdperadapterPR{80.1} & \weightedspearmanzerozerosrcthresholdperadapterRA{43.2} & \weightedspearmanzerozerosrcthresholdperadapterRC{64.0} & \weightedspearmanzerozerosrcthresholdperadapterRP{66.6} & \weightedspearmanzerozerosrcthresholdperadapterMean{68.5} & \weightedspearmanzerozerosrcthresholdperadapterStd{14.2} \\
\bottomrule
\end{tabular}}
\end{table}

%% file: supp_tables/office31_officehome/weighted_spearman_0.0_src_threshold_per_adapter_BNM.tex
\def\weightedspearmanzerozerosrcthresholdperadapterAD#1{\ifdim#1pt>80.2pt\cellcolor{lime!100}\else\ifdim#1pt>80.2pt\cellcolor{lime!90}\else\ifdim#1pt>80.2pt\cellcolor{lime!80}\else\ifdim#1pt>80.2pt\cellcolor{lime!70}\else\ifdim#1pt>80.2pt\cellcolor{lime!60}\else\ifdim#1pt>80.2pt\cellcolor{lime!50}\else\ifdim#1pt>80.2pt\cellcolor{lime!40}\else\ifdim#1pt>80.2pt\cellcolor{lime!30}\else\ifdim#1pt>80.2pt\cellcolor{lime!20}\else\ifdim#1pt>80.2pt\cellcolor{lime!10}\else\cellcolor{lime!0}\fi\fi\fi\fi\fi\fi\fi\fi\fi\fi#1}

\def\weightedspearmanzerozerosrcthresholdperadapterAW#1{\ifdim#1pt>88.9pt\cellcolor{lime!100}\else\ifdim#1pt>88.9pt\cellcolor{lime!90}\else\ifdim#1pt>88.9pt\cellcolor{lime!80}\else\ifdim#1pt>88.9pt\cellcolor{lime!70}\else\ifdim#1pt>88.9pt\cellcolor{lime!60}\else\ifdim#1pt>88.9pt\cellcolor{lime!50}\else\ifdim#1pt>88.9pt\cellcolor{lime!40}\else\ifdim#1pt>88.9pt\cellcolor{lime!30}\else\ifdim#1pt>88.9pt\cellcolor{lime!20}\else\ifdim#1pt>88.9pt\cellcolor{lime!10}\else\cellcolor{lime!0}\fi\fi\fi\fi\fi\fi\fi\fi\fi\fi#1}

\def\weightedspearmanzerozerosrcthresholdperadapterDA#1{\ifdim#1pt>85.0pt\cellcolor{lime!100}\else\ifdim#1pt>85.0pt\cellcolor{lime!90}\else\ifdim#1pt>84.9pt\cellcolor{lime!80}\else\ifdim#1pt>84.8pt\cellcolor{lime!70}\else\ifdim#1pt>84.8pt\cellcolor{lime!60}\else\ifdim#1pt>84.7pt\cellcolor{lime!50}\else\ifdim#1pt>84.6pt\cellcolor{lime!40}\else\ifdim#1pt>84.5pt\cellcolor{lime!30}\else\ifdim#1pt>84.5pt\cellcolor{lime!20}\else\ifdim#1pt>84.4pt\cellcolor{lime!10}\else\cellcolor{lime!0}\fi\fi\fi\fi\fi\fi\fi\fi\fi\fi#1}

\def\weightedspearmanzerozerosrcthresholdperadapterDW#1{\ifdim#1pt>81.4pt\cellcolor{lime!100}\else\ifdim#1pt>81.4pt\cellcolor{lime!90}\else\ifdim#1pt>81.4pt\cellcolor{lime!80}\else\ifdim#1pt>81.4pt\cellcolor{lime!70}\else\ifdim#1pt>81.4pt\cellcolor{lime!60}\else\ifdim#1pt>81.4pt\cellcolor{lime!50}\else\ifdim#1pt>81.4pt\cellcolor{lime!40}\else\ifdim#1pt>81.4pt\cellcolor{lime!30}\else\ifdim#1pt>81.4pt\cellcolor{lime!20}\else\ifdim#1pt>81.4pt\cellcolor{lime!10}\else\cellcolor{lime!0}\fi\fi\fi\fi\fi\fi\fi\fi\fi\fi#1}

\def\weightedspearmanzerozerosrcthresholdperadapterWA#1{\ifdim#1pt>88.0pt\cellcolor{lime!100}\else\ifdim#1pt>87.8pt\cellcolor{lime!90}\else\ifdim#1pt>87.7pt\cellcolor{lime!80}\else\ifdim#1pt>87.5pt\cellcolor{lime!70}\else\ifdim#1pt>87.3pt\cellcolor{lime!60}\else\ifdim#1pt>87.1pt\cellcolor{lime!50}\else\ifdim#1pt>86.9pt\cellcolor{lime!40}\else\ifdim#1pt>86.8pt\cellcolor{lime!30}\else\ifdim#1pt>86.6pt\cellcolor{lime!20}\else\ifdim#1pt>86.4pt\cellcolor{lime!10}\else\cellcolor{lime!0}\fi\fi\fi\fi\fi\fi\fi\fi\fi\fi#1}

\def\weightedspearmanzerozerosrcthresholdperadapterWD#1{\ifdim#1pt>63.7pt\cellcolor{lime!100}\else\ifdim#1pt>63.7pt\cellcolor{lime!90}\else\ifdim#1pt>63.7pt\cellcolor{lime!80}\else\ifdim#1pt>63.7pt\cellcolor{lime!70}\else\ifdim#1pt>63.7pt\cellcolor{lime!60}\else\ifdim#1pt>63.7pt\cellcolor{lime!50}\else\ifdim#1pt>63.7pt\cellcolor{lime!40}\else\ifdim#1pt>63.7pt\cellcolor{lime!30}\else\ifdim#1pt>63.7pt\cellcolor{lime!20}\else\ifdim#1pt>63.7pt\cellcolor{lime!10}\else\cellcolor{lime!0}\fi\fi\fi\fi\fi\fi\fi\fi\fi\fi#1}

\def\weightedspearmanzerozerosrcthresholdperadapterAC#1{\ifdim#1pt>84.2pt\cellcolor{lime!100}\else\ifdim#1pt>84.2pt\cellcolor{lime!90}\else\ifdim#1pt>84.2pt\cellcolor{lime!80}\else\ifdim#1pt>84.2pt\cellcolor{lime!70}\else\ifdim#1pt>84.2pt\cellcolor{lime!60}\else\ifdim#1pt>84.2pt\cellcolor{lime!50}\else\ifdim#1pt>84.2pt\cellcolor{lime!40}\else\ifdim#1pt>84.2pt\cellcolor{lime!30}\else\ifdim#1pt>84.2pt\cellcolor{lime!20}\else\ifdim#1pt>84.2pt\cellcolor{lime!10}\else\cellcolor{lime!0}\fi\fi\fi\fi\fi\fi\fi\fi\fi\fi#1}

\def\weightedspearmanzerozerosrcthresholdperadapterAP#1{\ifdim#1pt>93.0pt\cellcolor{lime!100}\else\ifdim#1pt>93.0pt\cellcolor{lime!90}\else\ifdim#1pt>93.0pt\cellcolor{lime!80}\else\ifdim#1pt>93.0pt\cellcolor{lime!70}\else\ifdim#1pt>93.0pt\cellcolor{lime!60}\else\ifdim#1pt>93.0pt\cellcolor{lime!50}\else\ifdim#1pt>93.0pt\cellcolor{lime!40}\else\ifdim#1pt>93.0pt\cellcolor{lime!30}\else\ifdim#1pt>93.0pt\cellcolor{lime!20}\else\ifdim#1pt>93.0pt\cellcolor{lime!10}\else\cellcolor{lime!0}\fi\fi\fi\fi\fi\fi\fi\fi\fi\fi#1}

\def\weightedspearmanzerozerosrcthresholdperadapterAR#1{\ifdim#1pt>95.4pt\cellcolor{lime!100}\else\ifdim#1pt>95.4pt\cellcolor{lime!90}\else\ifdim#1pt>95.4pt\cellcolor{lime!80}\else\ifdim#1pt>95.4pt\cellcolor{lime!70}\else\ifdim#1pt>95.4pt\cellcolor{lime!60}\else\ifdim#1pt>95.4pt\cellcolor{lime!50}\else\ifdim#1pt>95.4pt\cellcolor{lime!40}\else\ifdim#1pt>95.4pt\cellcolor{lime!30}\else\ifdim#1pt>95.4pt\cellcolor{lime!20}\else\ifdim#1pt>95.4pt\cellcolor{lime!10}\else\cellcolor{lime!0}\fi\fi\fi\fi\fi\fi\fi\fi\fi\fi#1}

\def\weightedspearmanzerozerosrcthresholdperadapterCA#1{\ifdim#1pt>83.8pt\cellcolor{lime!100}\else\ifdim#1pt>83.8pt\cellcolor{lime!90}\else\ifdim#1pt>83.8pt\cellcolor{lime!80}\else\ifdim#1pt>83.8pt\cellcolor{lime!70}\else\ifdim#1pt>83.8pt\cellcolor{lime!60}\else\ifdim#1pt>83.8pt\cellcolor{lime!50}\else\ifdim#1pt>83.8pt\cellcolor{lime!40}\else\ifdim#1pt>83.8pt\cellcolor{lime!30}\else\ifdim#1pt>83.8pt\cellcolor{lime!20}\else\ifdim#1pt>83.8pt\cellcolor{lime!10}\else\cellcolor{lime!0}\fi\fi\fi\fi\fi\fi\fi\fi\fi\fi#1}

\def\weightedspearmanzerozerosrcthresholdperadapterCP#1{\ifdim#1pt>86.1pt\cellcolor{lime!100}\else\ifdim#1pt>86.1pt\cellcolor{lime!90}\else\ifdim#1pt>86.1pt\cellcolor{lime!80}\else\ifdim#1pt>86.1pt\cellcolor{lime!70}\else\ifdim#1pt>86.1pt\cellcolor{lime!60}\else\ifdim#1pt>86.1pt\cellcolor{lime!50}\else\ifdim#1pt>86.1pt\cellcolor{lime!40}\else\ifdim#1pt>86.1pt\cellcolor{lime!30}\else\ifdim#1pt>86.1pt\cellcolor{lime!20}\else\ifdim#1pt>86.1pt\cellcolor{lime!10}\else\cellcolor{lime!0}\fi\fi\fi\fi\fi\fi\fi\fi\fi\fi#1}

\def\weightedspearmanzerozerosrcthresholdperadapterCR#1{\ifdim#1pt>88.6pt\cellcolor{lime!100}\else\ifdim#1pt>88.6pt\cellcolor{lime!90}\else\ifdim#1pt>88.6pt\cellcolor{lime!80}\else\ifdim#1pt>88.6pt\cellcolor{lime!70}\else\ifdim#1pt>88.6pt\cellcolor{lime!60}\else\ifdim#1pt>88.6pt\cellcolor{lime!50}\else\ifdim#1pt>88.6pt\cellcolor{lime!40}\else\ifdim#1pt>88.6pt\cellcolor{lime!30}\else\ifdim#1pt>88.6pt\cellcolor{lime!20}\else\ifdim#1pt>88.6pt\cellcolor{lime!10}\else\cellcolor{lime!0}\fi\fi\fi\fi\fi\fi\fi\fi\fi\fi#1}

\def\weightedspearmanzerozerosrcthresholdperadapterPA#1{\ifdim#1pt>92.6pt\cellcolor{lime!100}\else\ifdim#1pt>92.6pt\cellcolor{lime!90}\else\ifdim#1pt>92.6pt\cellcolor{lime!80}\else\ifdim#1pt>92.6pt\cellcolor{lime!70}\else\ifdim#1pt>92.6pt\cellcolor{lime!60}\else\ifdim#1pt>92.6pt\cellcolor{lime!50}\else\ifdim#1pt>92.6pt\cellcolor{lime!40}\else\ifdim#1pt>92.6pt\cellcolor{lime!30}\else\ifdim#1pt>92.6pt\cellcolor{lime!20}\else\ifdim#1pt>92.6pt\cellcolor{lime!10}\else\cellcolor{lime!0}\fi\fi\fi\fi\fi\fi\fi\fi\fi\fi#1}

\def\weightedspearmanzerozerosrcthresholdperadapterPC#1{\ifdim#1pt>91.8pt\cellcolor{lime!100}\else\ifdim#1pt>91.8pt\cellcolor{lime!90}\else\ifdim#1pt>91.8pt\cellcolor{lime!80}\else\ifdim#1pt>91.8pt\cellcolor{lime!70}\else\ifdim#1pt>91.8pt\cellcolor{lime!60}\else\ifdim#1pt>91.8pt\cellcolor{lime!50}\else\ifdim#1pt>91.8pt\cellcolor{lime!40}\else\ifdim#1pt>91.8pt\cellcolor{lime!30}\else\ifdim#1pt>91.8pt\cellcolor{lime!20}\else\ifdim#1pt>91.8pt\cellcolor{lime!10}\else\cellcolor{lime!0}\fi\fi\fi\fi\fi\fi\fi\fi\fi\fi#1}

\def\weightedspearmanzerozerosrcthresholdperadapterPR#1{\ifdim#1pt>95.1pt\cellcolor{lime!100}\else\ifdim#1pt>95.1pt\cellcolor{lime!90}\else\ifdim#1pt>95.1pt\cellcolor{lime!80}\else\ifdim#1pt>95.1pt\cellcolor{lime!70}\else\ifdim#1pt>95.1pt\cellcolor{lime!60}\else\ifdim#1pt>95.1pt\cellcolor{lime!50}\else\ifdim#1pt>95.1pt\cellcolor{lime!40}\else\ifdim#1pt>95.1pt\cellcolor{lime!30}\else\ifdim#1pt>95.1pt\cellcolor{lime!20}\else\ifdim#1pt>95.1pt\cellcolor{lime!10}\else\cellcolor{lime!0}\fi\fi\fi\fi\fi\fi\fi\fi\fi\fi#1}

\def\weightedspearmanzerozerosrcthresholdperadapterRA#1{\ifdim#1pt>93.0pt\cellcolor{lime!100}\else\ifdim#1pt>93.0pt\cellcolor{lime!90}\else\ifdim#1pt>93.0pt\cellcolor{lime!80}\else\ifdim#1pt>93.0pt\cellcolor{lime!70}\else\ifdim#1pt>93.0pt\cellcolor{lime!60}\else\ifdim#1pt>93.0pt\cellcolor{lime!50}\else\ifdim#1pt>93.0pt\cellcolor{lime!40}\else\ifdim#1pt>93.0pt\cellcolor{lime!30}\else\ifdim#1pt>93.0pt\cellcolor{lime!20}\else\ifdim#1pt>93.0pt\cellcolor{lime!10}\else\cellcolor{lime!0}\fi\fi\fi\fi\fi\fi\fi\fi\fi\fi#1}

\def\weightedspearmanzerozerosrcthresholdperadapterRC#1{\ifdim#1pt>91.5pt\cellcolor{lime!100}\else\ifdim#1pt>90.9pt\cellcolor{lime!90}\else\ifdim#1pt>90.2pt\cellcolor{lime!80}\else\ifdim#1pt>89.6pt\cellcolor{lime!70}\else\ifdim#1pt>88.9pt\cellcolor{lime!60}\else\ifdim#1pt>88.2pt\cellcolor{lime!50}\else\ifdim#1pt>87.6pt\cellcolor{lime!40}\else\ifdim#1pt>86.9pt\cellcolor{lime!30}\else\ifdim#1pt>86.3pt\cellcolor{lime!20}\else\ifdim#1pt>85.6pt\cellcolor{lime!10}\else\cellcolor{lime!0}\fi\fi\fi\fi\fi\fi\fi\fi\fi\fi#1}

\def\weightedspearmanzerozerosrcthresholdperadapterRP#1{\ifdim#1pt>95.2pt\cellcolor{lime!100}\else\ifdim#1pt>95.2pt\cellcolor{lime!90}\else\ifdim#1pt>95.2pt\cellcolor{lime!80}\else\ifdim#1pt>95.2pt\cellcolor{lime!70}\else\ifdim#1pt>95.2pt\cellcolor{lime!60}\else\ifdim#1pt>95.2pt\cellcolor{lime!50}\else\ifdim#1pt>95.2pt\cellcolor{lime!40}\else\ifdim#1pt>95.2pt\cellcolor{lime!30}\else\ifdim#1pt>95.2pt\cellcolor{lime!20}\else\ifdim#1pt>95.2pt\cellcolor{lime!10}\else\cellcolor{lime!0}\fi\fi\fi\fi\fi\fi\fi\fi\fi\fi#1}

\def\weightedspearmanzerozerosrcthresholdperadapterMean#1{\ifdim#1pt>87.2pt\cellcolor{lime!100}\else\ifdim#1pt>87.2pt\cellcolor{lime!90}\else\ifdim#1pt>87.2pt\cellcolor{lime!80}\else\ifdim#1pt>87.2pt\cellcolor{lime!70}\else\ifdim#1pt>87.2pt\cellcolor{lime!60}\else\ifdim#1pt>87.2pt\cellcolor{lime!50}\else\ifdim#1pt>87.2pt\cellcolor{lime!40}\else\ifdim#1pt>87.2pt\cellcolor{lime!30}\else\ifdim#1pt>87.2pt\cellcolor{lime!20}\else\ifdim#1pt>87.2pt\cellcolor{lime!10}\else\cellcolor{lime!0}\fi\fi\fi\fi\fi\fi\fi\fi\fi\fi#1}

\def\weightedspearmanzerozerosrcthresholdperadapterStd#1{\ifdim#1pt<7.4pt\cellcolor{lime!100}\else\ifdim#1pt<7.4pt\cellcolor{lime!90}\else\ifdim#1pt<7.4pt\cellcolor{lime!80}\else\ifdim#1pt<7.4pt\cellcolor{lime!70}\else\ifdim#1pt<7.4pt\cellcolor{lime!60}\else\ifdim#1pt<7.4pt\cellcolor{lime!50}\else\ifdim#1pt<7.4pt\cellcolor{lime!40}\else\ifdim#1pt<7.4pt\cellcolor{lime!30}\else\ifdim#1pt<7.4pt\cellcolor{lime!20}\else\ifdim#1pt<7.4pt\cellcolor{lime!10}\else\cellcolor{lime!0}\fi\fi\fi\fi\fi\fi\fi\fi\fi\fi#1}

\begin{table}[H]
\centering
\caption{The weighted Spearman correlation of each validator/task pair for \textbf{BNM}.}
\label{office_weighted_spearman_0.0_src_threshold_per_adapter_BNM}
\resizebox{!}{0.22\textheight}{
\begin{tabular}{llrrrrrrrrrrrrrrrrrrrr}
\toprule
 &  & AD & AW & DA & DW & WA & WD & AC & AP & AR & CA & CP & CR & PA & PC & PR & RA & RC & RP & Mean & Std \\
\midrule
\multirow[c]{2}{*}{Accuracy} & Source Train & \weightedspearmanzerozerosrcthresholdperadapterAD{65.3} & \weightedspearmanzerozerosrcthresholdperadapterAW{70.4} & \textbf{\weightedspearmanzerozerosrcthresholdperadapterDA{85.1}} & \weightedspearmanzerozerosrcthresholdperadapterDW{64.6} & \weightedspearmanzerozerosrcthresholdperadapterWA{82.1} & \weightedspearmanzerozerosrcthresholdperadapterWD{53.8} & \weightedspearmanzerozerosrcthresholdperadapterAC{64.7} & \weightedspearmanzerozerosrcthresholdperadapterAP{69.1} & \weightedspearmanzerozerosrcthresholdperadapterAR{57.9} & \weightedspearmanzerozerosrcthresholdperadapterCA{49.2} & \weightedspearmanzerozerosrcthresholdperadapterCP{51.7} & \weightedspearmanzerozerosrcthresholdperadapterCR{49.7} & \weightedspearmanzerozerosrcthresholdperadapterPA{67.8} & \weightedspearmanzerozerosrcthresholdperadapterPC{66.1} & \weightedspearmanzerozerosrcthresholdperadapterPR{61.3} & \weightedspearmanzerozerosrcthresholdperadapterRA{49.3} & \weightedspearmanzerozerosrcthresholdperadapterRC{67.4} & \weightedspearmanzerozerosrcthresholdperadapterRP{49.3} & \weightedspearmanzerozerosrcthresholdperadapterMean{62.5} & \weightedspearmanzerozerosrcthresholdperadapterStd{10.5} \\
\bottomrule
\end{tabular}}
\end{table}

%% file: supp_tables/office31_officehome/weighted_spearman_0.0_src_threshold_per_adapter_BSP.tex
\def\weightedspearmanzerozerosrcthresholdperadapterAD#1{\ifdim#1pt>87.9pt\cellcolor{lime!100}\else\ifdim#1pt>87.9pt\cellcolor{lime!90}\else\ifdim#1pt>87.8pt\cellcolor{lime!80}\else\ifdim#1pt>87.7pt\cellcolor{lime!70}\else\ifdim#1pt>87.7pt\cellcolor{lime!60}\else\ifdim#1pt>87.6pt\cellcolor{lime!50}\else\ifdim#1pt>87.5pt\cellcolor{lime!40}\else\ifdim#1pt>87.4pt\cellcolor{lime!30}\else\ifdim#1pt>87.4pt\cellcolor{lime!20}\else\ifdim#1pt>87.3pt\cellcolor{lime!10}\else\cellcolor{lime!0}\fi\fi\fi\fi\fi\fi\fi\fi\fi\fi#1}

\def\weightedspearmanzerozerosrcthresholdperadapterAW#1{\ifdim#1pt>87.9pt\cellcolor{lime!100}\else\ifdim#1pt>87.9pt\cellcolor{lime!90}\else\ifdim#1pt>87.9pt\cellcolor{lime!80}\else\ifdim#1pt>87.9pt\cellcolor{lime!70}\else\ifdim#1pt>87.9pt\cellcolor{lime!60}\else\ifdim#1pt>87.9pt\cellcolor{lime!50}\else\ifdim#1pt>87.9pt\cellcolor{lime!40}\else\ifdim#1pt>87.9pt\cellcolor{lime!30}\else\ifdim#1pt>87.9pt\cellcolor{lime!20}\else\ifdim#1pt>87.9pt\cellcolor{lime!10}\else\cellcolor{lime!0}\fi\fi\fi\fi\fi\fi\fi\fi\fi\fi#1}

\def\weightedspearmanzerozerosrcthresholdperadapterDA#1{\ifdim#1pt>73.0pt\cellcolor{lime!100}\else\ifdim#1pt>73.0pt\cellcolor{lime!90}\else\ifdim#1pt>73.0pt\cellcolor{lime!80}\else\ifdim#1pt>73.0pt\cellcolor{lime!70}\else\ifdim#1pt>73.0pt\cellcolor{lime!60}\else\ifdim#1pt>73.0pt\cellcolor{lime!50}\else\ifdim#1pt>73.0pt\cellcolor{lime!40}\else\ifdim#1pt>73.0pt\cellcolor{lime!30}\else\ifdim#1pt>73.0pt\cellcolor{lime!20}\else\ifdim#1pt>73.0pt\cellcolor{lime!10}\else\cellcolor{lime!0}\fi\fi\fi\fi\fi\fi\fi\fi\fi\fi#1}

\def\weightedspearmanzerozerosrcthresholdperadapterDW#1{\ifdim#1pt>85.3pt\cellcolor{lime!100}\else\ifdim#1pt>85.3pt\cellcolor{lime!90}\else\ifdim#1pt>85.3pt\cellcolor{lime!80}\else\ifdim#1pt>85.3pt\cellcolor{lime!70}\else\ifdim#1pt>85.3pt\cellcolor{lime!60}\else\ifdim#1pt>85.3pt\cellcolor{lime!50}\else\ifdim#1pt>85.3pt\cellcolor{lime!40}\else\ifdim#1pt>85.3pt\cellcolor{lime!30}\else\ifdim#1pt>85.3pt\cellcolor{lime!20}\else\ifdim#1pt>85.3pt\cellcolor{lime!10}\else\cellcolor{lime!0}\fi\fi\fi\fi\fi\fi\fi\fi\fi\fi#1}

\def\weightedspearmanzerozerosrcthresholdperadapterWA#1{\ifdim#1pt>66.7pt\cellcolor{lime!100}\else\ifdim#1pt>66.4pt\cellcolor{lime!90}\else\ifdim#1pt>66.2pt\cellcolor{lime!80}\else\ifdim#1pt>65.9pt\cellcolor{lime!70}\else\ifdim#1pt>65.6pt\cellcolor{lime!60}\else\ifdim#1pt>65.3pt\cellcolor{lime!50}\else\ifdim#1pt>65.0pt\cellcolor{lime!40}\else\ifdim#1pt>64.8pt\cellcolor{lime!30}\else\ifdim#1pt>64.5pt\cellcolor{lime!20}\else\ifdim#1pt>64.2pt\cellcolor{lime!10}\else\cellcolor{lime!0}\fi\fi\fi\fi\fi\fi\fi\fi\fi\fi#1}

\def\weightedspearmanzerozerosrcthresholdperadapterWD#1{\ifdim#1pt>83.2pt\cellcolor{lime!100}\else\ifdim#1pt>83.2pt\cellcolor{lime!90}\else\ifdim#1pt>83.2pt\cellcolor{lime!80}\else\ifdim#1pt>83.2pt\cellcolor{lime!70}\else\ifdim#1pt>83.2pt\cellcolor{lime!60}\else\ifdim#1pt>83.2pt\cellcolor{lime!50}\else\ifdim#1pt>83.2pt\cellcolor{lime!40}\else\ifdim#1pt>83.2pt\cellcolor{lime!30}\else\ifdim#1pt>83.2pt\cellcolor{lime!20}\else\ifdim#1pt>83.2pt\cellcolor{lime!10}\else\cellcolor{lime!0}\fi\fi\fi\fi\fi\fi\fi\fi\fi\fi#1}

\def\weightedspearmanzerozerosrcthresholdperadapterAC#1{\ifdim#1pt>96.7pt\cellcolor{lime!100}\else\ifdim#1pt>96.7pt\cellcolor{lime!90}\else\ifdim#1pt>96.7pt\cellcolor{lime!80}\else\ifdim#1pt>96.7pt\cellcolor{lime!70}\else\ifdim#1pt>96.7pt\cellcolor{lime!60}\else\ifdim#1pt>96.7pt\cellcolor{lime!50}\else\ifdim#1pt>96.7pt\cellcolor{lime!40}\else\ifdim#1pt>96.7pt\cellcolor{lime!30}\else\ifdim#1pt>96.7pt\cellcolor{lime!20}\else\ifdim#1pt>96.7pt\cellcolor{lime!10}\else\cellcolor{lime!0}\fi\fi\fi\fi\fi\fi\fi\fi\fi\fi#1}

\def\weightedspearmanzerozerosrcthresholdperadapterAP#1{\ifdim#1pt>97.8pt\cellcolor{lime!100}\else\ifdim#1pt>97.7pt\cellcolor{lime!90}\else\ifdim#1pt>97.7pt\cellcolor{lime!80}\else\ifdim#1pt>97.7pt\cellcolor{lime!70}\else\ifdim#1pt>97.7pt\cellcolor{lime!60}\else\ifdim#1pt>97.6pt\cellcolor{lime!50}\else\ifdim#1pt>97.6pt\cellcolor{lime!40}\else\ifdim#1pt>97.6pt\cellcolor{lime!30}\else\ifdim#1pt>97.5pt\cellcolor{lime!20}\else\ifdim#1pt>97.5pt\cellcolor{lime!10}\else\cellcolor{lime!0}\fi\fi\fi\fi\fi\fi\fi\fi\fi\fi#1}

\def\weightedspearmanzerozerosrcthresholdperadapterAR#1{\ifdim#1pt>97.6pt\cellcolor{lime!100}\else\ifdim#1pt>97.6pt\cellcolor{lime!90}\else\ifdim#1pt>97.5pt\cellcolor{lime!80}\else\ifdim#1pt>97.5pt\cellcolor{lime!70}\else\ifdim#1pt>97.4pt\cellcolor{lime!60}\else\ifdim#1pt>97.3pt\cellcolor{lime!50}\else\ifdim#1pt>97.3pt\cellcolor{lime!40}\else\ifdim#1pt>97.2pt\cellcolor{lime!30}\else\ifdim#1pt>97.2pt\cellcolor{lime!20}\else\ifdim#1pt>97.1pt\cellcolor{lime!10}\else\cellcolor{lime!0}\fi\fi\fi\fi\fi\fi\fi\fi\fi\fi#1}

\def\weightedspearmanzerozerosrcthresholdperadapterCA#1{\ifdim#1pt>78.4pt\cellcolor{lime!100}\else\ifdim#1pt>77.5pt\cellcolor{lime!90}\else\ifdim#1pt>76.5pt\cellcolor{lime!80}\else\ifdim#1pt>75.6pt\cellcolor{lime!70}\else\ifdim#1pt>74.7pt\cellcolor{lime!60}\else\ifdim#1pt>73.8pt\cellcolor{lime!50}\else\ifdim#1pt>72.9pt\cellcolor{lime!40}\else\ifdim#1pt>71.9pt\cellcolor{lime!30}\else\ifdim#1pt>71.0pt\cellcolor{lime!20}\else\ifdim#1pt>70.1pt\cellcolor{lime!10}\else\cellcolor{lime!0}\fi\fi\fi\fi\fi\fi\fi\fi\fi\fi#1}

\def\weightedspearmanzerozerosrcthresholdperadapterCP#1{\ifdim#1pt>82.0pt\cellcolor{lime!100}\else\ifdim#1pt>81.5pt\cellcolor{lime!90}\else\ifdim#1pt>81.1pt\cellcolor{lime!80}\else\ifdim#1pt>80.7pt\cellcolor{lime!70}\else\ifdim#1pt>80.2pt\cellcolor{lime!60}\else\ifdim#1pt>79.8pt\cellcolor{lime!50}\else\ifdim#1pt>79.4pt\cellcolor{lime!40}\else\ifdim#1pt>79.0pt\cellcolor{lime!30}\else\ifdim#1pt>78.5pt\cellcolor{lime!20}\else\ifdim#1pt>78.1pt\cellcolor{lime!10}\else\cellcolor{lime!0}\fi\fi\fi\fi\fi\fi\fi\fi\fi\fi#1}

\def\weightedspearmanzerozerosrcthresholdperadapterCR#1{\ifdim#1pt>84.0pt\cellcolor{lime!100}\else\ifdim#1pt>83.7pt\cellcolor{lime!90}\else\ifdim#1pt>83.3pt\cellcolor{lime!80}\else\ifdim#1pt>83.0pt\cellcolor{lime!70}\else\ifdim#1pt>82.7pt\cellcolor{lime!60}\else\ifdim#1pt>82.4pt\cellcolor{lime!50}\else\ifdim#1pt>82.1pt\cellcolor{lime!40}\else\ifdim#1pt>81.7pt\cellcolor{lime!30}\else\ifdim#1pt>81.4pt\cellcolor{lime!20}\else\ifdim#1pt>81.1pt\cellcolor{lime!10}\else\cellcolor{lime!0}\fi\fi\fi\fi\fi\fi\fi\fi\fi\fi#1}

\def\weightedspearmanzerozerosrcthresholdperadapterPA#1{\ifdim#1pt>92.5pt\cellcolor{lime!100}\else\ifdim#1pt>92.2pt\cellcolor{lime!90}\else\ifdim#1pt>91.9pt\cellcolor{lime!80}\else\ifdim#1pt>91.6pt\cellcolor{lime!70}\else\ifdim#1pt>91.3pt\cellcolor{lime!60}\else\ifdim#1pt>91.1pt\cellcolor{lime!50}\else\ifdim#1pt>90.8pt\cellcolor{lime!40}\else\ifdim#1pt>90.5pt\cellcolor{lime!30}\else\ifdim#1pt>90.2pt\cellcolor{lime!20}\else\ifdim#1pt>89.9pt\cellcolor{lime!10}\else\cellcolor{lime!0}\fi\fi\fi\fi\fi\fi\fi\fi\fi\fi#1}

\def\weightedspearmanzerozerosrcthresholdperadapterPC#1{\ifdim#1pt>90.9pt\cellcolor{lime!100}\else\ifdim#1pt>90.7pt\cellcolor{lime!90}\else\ifdim#1pt>90.6pt\cellcolor{lime!80}\else\ifdim#1pt>90.5pt\cellcolor{lime!70}\else\ifdim#1pt>90.3pt\cellcolor{lime!60}\else\ifdim#1pt>90.2pt\cellcolor{lime!50}\else\ifdim#1pt>90.1pt\cellcolor{lime!40}\else\ifdim#1pt>90.0pt\cellcolor{lime!30}\else\ifdim#1pt>89.8pt\cellcolor{lime!20}\else\ifdim#1pt>89.7pt\cellcolor{lime!10}\else\cellcolor{lime!0}\fi\fi\fi\fi\fi\fi\fi\fi\fi\fi#1}

\def\weightedspearmanzerozerosrcthresholdperadapterPR#1{\ifdim#1pt>94.9pt\cellcolor{lime!100}\else\ifdim#1pt>94.7pt\cellcolor{lime!90}\else\ifdim#1pt>94.6pt\cellcolor{lime!80}\else\ifdim#1pt>94.4pt\cellcolor{lime!70}\else\ifdim#1pt>94.2pt\cellcolor{lime!60}\else\ifdim#1pt>94.0pt\cellcolor{lime!50}\else\ifdim#1pt>93.8pt\cellcolor{lime!40}\else\ifdim#1pt>93.7pt\cellcolor{lime!30}\else\ifdim#1pt>93.5pt\cellcolor{lime!20}\else\ifdim#1pt>93.3pt\cellcolor{lime!10}\else\cellcolor{lime!0}\fi\fi\fi\fi\fi\fi\fi\fi\fi\fi#1}

\def\weightedspearmanzerozerosrcthresholdperadapterRA#1{\ifdim#1pt>97.1pt\cellcolor{lime!100}\else\ifdim#1pt>97.0pt\cellcolor{lime!90}\else\ifdim#1pt>97.0pt\cellcolor{lime!80}\else\ifdim#1pt>96.9pt\cellcolor{lime!70}\else\ifdim#1pt>96.9pt\cellcolor{lime!60}\else\ifdim#1pt>96.9pt\cellcolor{lime!50}\else\ifdim#1pt>96.8pt\cellcolor{lime!40}\else\ifdim#1pt>96.8pt\cellcolor{lime!30}\else\ifdim#1pt>96.7pt\cellcolor{lime!20}\else\ifdim#1pt>96.7pt\cellcolor{lime!10}\else\cellcolor{lime!0}\fi\fi\fi\fi\fi\fi\fi\fi\fi\fi#1}

\def\weightedspearmanzerozerosrcthresholdperadapterRC#1{\ifdim#1pt>93.2pt\cellcolor{lime!100}\else\ifdim#1pt>93.0pt\cellcolor{lime!90}\else\ifdim#1pt>92.9pt\cellcolor{lime!80}\else\ifdim#1pt>92.7pt\cellcolor{lime!70}\else\ifdim#1pt>92.5pt\cellcolor{lime!60}\else\ifdim#1pt>92.3pt\cellcolor{lime!50}\else\ifdim#1pt>92.1pt\cellcolor{lime!40}\else\ifdim#1pt>92.0pt\cellcolor{lime!30}\else\ifdim#1pt>91.8pt\cellcolor{lime!20}\else\ifdim#1pt>91.6pt\cellcolor{lime!10}\else\cellcolor{lime!0}\fi\fi\fi\fi\fi\fi\fi\fi\fi\fi#1}

\def\weightedspearmanzerozerosrcthresholdperadapterRP#1{\ifdim#1pt>97.3pt\cellcolor{lime!100}\else\ifdim#1pt>97.1pt\cellcolor{lime!90}\else\ifdim#1pt>97.0pt\cellcolor{lime!80}\else\ifdim#1pt>96.8pt\cellcolor{lime!70}\else\ifdim#1pt>96.6pt\cellcolor{lime!60}\else\ifdim#1pt>96.4pt\cellcolor{lime!50}\else\ifdim#1pt>96.2pt\cellcolor{lime!40}\else\ifdim#1pt>96.1pt\cellcolor{lime!30}\else\ifdim#1pt>95.9pt\cellcolor{lime!20}\else\ifdim#1pt>95.7pt\cellcolor{lime!10}\else\cellcolor{lime!0}\fi\fi\fi\fi\fi\fi\fi\fi\fi\fi#1}

\def\weightedspearmanzerozerosrcthresholdperadapterMean#1{\ifdim#1pt>86.6pt\cellcolor{lime!100}\else\ifdim#1pt>86.6pt\cellcolor{lime!90}\else\ifdim#1pt>86.6pt\cellcolor{lime!80}\else\ifdim#1pt>86.6pt\cellcolor{lime!70}\else\ifdim#1pt>86.6pt\cellcolor{lime!60}\else\ifdim#1pt>86.6pt\cellcolor{lime!50}\else\ifdim#1pt>86.6pt\cellcolor{lime!40}\else\ifdim#1pt>86.6pt\cellcolor{lime!30}\else\ifdim#1pt>86.6pt\cellcolor{lime!20}\else\ifdim#1pt>86.6pt\cellcolor{lime!10}\else\cellcolor{lime!0}\fi\fi\fi\fi\fi\fi\fi\fi\fi\fi#1}

\def\weightedspearmanzerozerosrcthresholdperadapterStd#1{\ifdim#1pt<3.8pt\cellcolor{lime!100}\else\ifdim#1pt<4.4pt\cellcolor{lime!90}\else\ifdim#1pt<5.1pt\cellcolor{lime!80}\else\ifdim#1pt<5.7pt\cellcolor{lime!70}\else\ifdim#1pt<6.4pt\cellcolor{lime!60}\else\ifdim#1pt<7.1pt\cellcolor{lime!50}\else\ifdim#1pt<7.7pt\cellcolor{lime!40}\else\ifdim#1pt<8.4pt\cellcolor{lime!30}\else\ifdim#1pt<9.0pt\cellcolor{lime!20}\else\ifdim#1pt<9.7pt\cellcolor{lime!10}\else\cellcolor{lime!0}\fi\fi\fi\fi\fi\fi\fi\fi\fi\fi#1}

\begin{table}[H]
\centering
\caption{The weighted Spearman correlation of each validator/task pair for \textbf{BSP}.}
\label{office_weighted_spearman_0.0_src_threshold_per_adapter_BSP}
\resizebox{!}{0.22\textheight}{
\begin{tabular}{llrrrrrrrrrrrrrrrrrrrr}
\toprule
 &  & AD & AW & DA & DW & WA & WD & AC & AP & AR & CA & CP & CR & PA & PC & PR & RA & RC & RP & Mean & Std \\
\midrule
\multirow[c]{2}{*}{Accuracy} & Source Train & \weightedspearmanzerozerosrcthresholdperadapterAD{67.6} & \weightedspearmanzerozerosrcthresholdperadapterAW{52.8} & \weightedspearmanzerozerosrcthresholdperadapterDA{62.8} & \weightedspearmanzerozerosrcthresholdperadapterDW{75.4} & \textbf{\weightedspearmanzerozerosrcthresholdperadapterWA{67.0}} & \weightedspearmanzerozerosrcthresholdperadapterWD{79.2} & \weightedspearmanzerozerosrcthresholdperadapterAC{38.0} & \weightedspearmanzerozerosrcthresholdperadapterAP{35.6} & \weightedspearmanzerozerosrcthresholdperadapterAR{37.2} & \weightedspearmanzerozerosrcthresholdperadapterCA{31.4} & \weightedspearmanzerozerosrcthresholdperadapterCP{20.3} & \weightedspearmanzerozerosrcthresholdperadapterCR{30.7} & \weightedspearmanzerozerosrcthresholdperadapterPA{67.9} & \weightedspearmanzerozerosrcthresholdperadapterPC{36.0} & \weightedspearmanzerozerosrcthresholdperadapterPR{52.0} & \weightedspearmanzerozerosrcthresholdperadapterRA{39.4} & \weightedspearmanzerozerosrcthresholdperadapterRC{50.1} & \weightedspearmanzerozerosrcthresholdperadapterRP{40.8} & \weightedspearmanzerozerosrcthresholdperadapterMean{49.1} & \weightedspearmanzerozerosrcthresholdperadapterStd{16.8} \\
\bottomrule
\end{tabular}}
\end{table}

%% file: supp_tables/office31_officehome/weighted_spearman_0.0_src_threshold_per_adapter_CDAN.tex
\def\weightedspearmanzerozerosrcthresholdperadapterAD#1{\ifdim#1pt>84.4pt\cellcolor{lime!100}\else\ifdim#1pt>84.4pt\cellcolor{lime!90}\else\ifdim#1pt>84.4pt\cellcolor{lime!80}\else\ifdim#1pt>84.4pt\cellcolor{lime!70}\else\ifdim#1pt>84.4pt\cellcolor{lime!60}\else\ifdim#1pt>84.4pt\cellcolor{lime!50}\else\ifdim#1pt>84.4pt\cellcolor{lime!40}\else\ifdim#1pt>84.4pt\cellcolor{lime!30}\else\ifdim#1pt>84.4pt\cellcolor{lime!20}\else\ifdim#1pt>84.4pt\cellcolor{lime!10}\else\cellcolor{lime!0}\fi\fi\fi\fi\fi\fi\fi\fi\fi\fi#1}

\def\weightedspearmanzerozerosrcthresholdperadapterAW#1{\ifdim#1pt>82.1pt\cellcolor{lime!100}\else\ifdim#1pt>82.1pt\cellcolor{lime!90}\else\ifdim#1pt>82.1pt\cellcolor{lime!80}\else\ifdim#1pt>82.1pt\cellcolor{lime!70}\else\ifdim#1pt>82.1pt\cellcolor{lime!60}\else\ifdim#1pt>82.1pt\cellcolor{lime!50}\else\ifdim#1pt>82.1pt\cellcolor{lime!40}\else\ifdim#1pt>82.1pt\cellcolor{lime!30}\else\ifdim#1pt>82.1pt\cellcolor{lime!20}\else\ifdim#1pt>82.1pt\cellcolor{lime!10}\else\cellcolor{lime!0}\fi\fi\fi\fi\fi\fi\fi\fi\fi\fi#1}

\def\weightedspearmanzerozerosrcthresholdperadapterDA#1{\ifdim#1pt>80.1pt\cellcolor{lime!100}\else\ifdim#1pt>80.1pt\cellcolor{lime!90}\else\ifdim#1pt>80.1pt\cellcolor{lime!80}\else\ifdim#1pt>80.1pt\cellcolor{lime!70}\else\ifdim#1pt>80.1pt\cellcolor{lime!60}\else\ifdim#1pt>80.1pt\cellcolor{lime!50}\else\ifdim#1pt>80.1pt\cellcolor{lime!40}\else\ifdim#1pt>80.1pt\cellcolor{lime!30}\else\ifdim#1pt>80.1pt\cellcolor{lime!20}\else\ifdim#1pt>80.1pt\cellcolor{lime!10}\else\cellcolor{lime!0}\fi\fi\fi\fi\fi\fi\fi\fi\fi\fi#1}

\def\weightedspearmanzerozerosrcthresholdperadapterDW#1{\ifdim#1pt>79.9pt\cellcolor{lime!100}\else\ifdim#1pt>79.9pt\cellcolor{lime!90}\else\ifdim#1pt>79.9pt\cellcolor{lime!80}\else\ifdim#1pt>79.9pt\cellcolor{lime!70}\else\ifdim#1pt>79.9pt\cellcolor{lime!60}\else\ifdim#1pt>79.9pt\cellcolor{lime!50}\else\ifdim#1pt>79.9pt\cellcolor{lime!40}\else\ifdim#1pt>79.9pt\cellcolor{lime!30}\else\ifdim#1pt>79.9pt\cellcolor{lime!20}\else\ifdim#1pt>79.9pt\cellcolor{lime!10}\else\cellcolor{lime!0}\fi\fi\fi\fi\fi\fi\fi\fi\fi\fi#1}

\def\weightedspearmanzerozerosrcthresholdperadapterWA#1{\ifdim#1pt>85.6pt\cellcolor{lime!100}\else\ifdim#1pt>85.6pt\cellcolor{lime!90}\else\ifdim#1pt>85.5pt\cellcolor{lime!80}\else\ifdim#1pt>85.5pt\cellcolor{lime!70}\else\ifdim#1pt>85.5pt\cellcolor{lime!60}\else\ifdim#1pt>85.5pt\cellcolor{lime!50}\else\ifdim#1pt>85.5pt\cellcolor{lime!40}\else\ifdim#1pt>85.4pt\cellcolor{lime!30}\else\ifdim#1pt>85.4pt\cellcolor{lime!20}\else\ifdim#1pt>85.4pt\cellcolor{lime!10}\else\cellcolor{lime!0}\fi\fi\fi\fi\fi\fi\fi\fi\fi\fi#1}

\def\weightedspearmanzerozerosrcthresholdperadapterWD#1{\ifdim#1pt>66.4pt\cellcolor{lime!100}\else\ifdim#1pt>66.4pt\cellcolor{lime!90}\else\ifdim#1pt>66.3pt\cellcolor{lime!80}\else\ifdim#1pt>66.3pt\cellcolor{lime!70}\else\ifdim#1pt>66.3pt\cellcolor{lime!60}\else\ifdim#1pt>66.3pt\cellcolor{lime!50}\else\ifdim#1pt>66.3pt\cellcolor{lime!40}\else\ifdim#1pt>66.2pt\cellcolor{lime!30}\else\ifdim#1pt>66.2pt\cellcolor{lime!20}\else\ifdim#1pt>66.2pt\cellcolor{lime!10}\else\cellcolor{lime!0}\fi\fi\fi\fi\fi\fi\fi\fi\fi\fi#1}

\def\weightedspearmanzerozerosrcthresholdperadapterAC#1{\ifdim#1pt>88.1pt\cellcolor{lime!100}\else\ifdim#1pt>87.4pt\cellcolor{lime!90}\else\ifdim#1pt>86.7pt\cellcolor{lime!80}\else\ifdim#1pt>86.0pt\cellcolor{lime!70}\else\ifdim#1pt>85.2pt\cellcolor{lime!60}\else\ifdim#1pt>84.5pt\cellcolor{lime!50}\else\ifdim#1pt>83.8pt\cellcolor{lime!40}\else\ifdim#1pt>83.1pt\cellcolor{lime!30}\else\ifdim#1pt>82.4pt\cellcolor{lime!20}\else\ifdim#1pt>81.7pt\cellcolor{lime!10}\else\cellcolor{lime!0}\fi\fi\fi\fi\fi\fi\fi\fi\fi\fi#1}

\def\weightedspearmanzerozerosrcthresholdperadapterAP#1{\ifdim#1pt>97.7pt\cellcolor{lime!100}\else\ifdim#1pt>97.7pt\cellcolor{lime!90}\else\ifdim#1pt>97.7pt\cellcolor{lime!80}\else\ifdim#1pt>97.7pt\cellcolor{lime!70}\else\ifdim#1pt>97.7pt\cellcolor{lime!60}\else\ifdim#1pt>97.7pt\cellcolor{lime!50}\else\ifdim#1pt>97.7pt\cellcolor{lime!40}\else\ifdim#1pt>97.7pt\cellcolor{lime!30}\else\ifdim#1pt>97.7pt\cellcolor{lime!20}\else\ifdim#1pt>97.7pt\cellcolor{lime!10}\else\cellcolor{lime!0}\fi\fi\fi\fi\fi\fi\fi\fi\fi\fi#1}

\def\weightedspearmanzerozerosrcthresholdperadapterAR#1{\ifdim#1pt>93.0pt\cellcolor{lime!100}\else\ifdim#1pt>93.0pt\cellcolor{lime!90}\else\ifdim#1pt>93.0pt\cellcolor{lime!80}\else\ifdim#1pt>93.0pt\cellcolor{lime!70}\else\ifdim#1pt>93.0pt\cellcolor{lime!60}\else\ifdim#1pt>93.0pt\cellcolor{lime!50}\else\ifdim#1pt>93.0pt\cellcolor{lime!40}\else\ifdim#1pt>93.0pt\cellcolor{lime!30}\else\ifdim#1pt>93.0pt\cellcolor{lime!20}\else\ifdim#1pt>93.0pt\cellcolor{lime!10}\else\cellcolor{lime!0}\fi\fi\fi\fi\fi\fi\fi\fi\fi\fi#1}

\def\weightedspearmanzerozerosrcthresholdperadapterCA#1{\ifdim#1pt>66.7pt\cellcolor{lime!100}\else\ifdim#1pt>66.7pt\cellcolor{lime!90}\else\ifdim#1pt>66.7pt\cellcolor{lime!80}\else\ifdim#1pt>66.7pt\cellcolor{lime!70}\else\ifdim#1pt>66.7pt\cellcolor{lime!60}\else\ifdim#1pt>66.7pt\cellcolor{lime!50}\else\ifdim#1pt>66.7pt\cellcolor{lime!40}\else\ifdim#1pt>66.7pt\cellcolor{lime!30}\else\ifdim#1pt>66.7pt\cellcolor{lime!20}\else\ifdim#1pt>66.7pt\cellcolor{lime!10}\else\cellcolor{lime!0}\fi\fi\fi\fi\fi\fi\fi\fi\fi\fi#1}

\def\weightedspearmanzerozerosrcthresholdperadapterCP#1{\ifdim#1pt>85.2pt\cellcolor{lime!100}\else\ifdim#1pt>85.2pt\cellcolor{lime!90}\else\ifdim#1pt>85.2pt\cellcolor{lime!80}\else\ifdim#1pt>85.2pt\cellcolor{lime!70}\else\ifdim#1pt>85.2pt\cellcolor{lime!60}\else\ifdim#1pt>85.2pt\cellcolor{lime!50}\else\ifdim#1pt>85.2pt\cellcolor{lime!40}\else\ifdim#1pt>85.2pt\cellcolor{lime!30}\else\ifdim#1pt>85.2pt\cellcolor{lime!20}\else\ifdim#1pt>85.2pt\cellcolor{lime!10}\else\cellcolor{lime!0}\fi\fi\fi\fi\fi\fi\fi\fi\fi\fi#1}

\def\weightedspearmanzerozerosrcthresholdperadapterCR#1{\ifdim#1pt>87.7pt\cellcolor{lime!100}\else\ifdim#1pt>87.7pt\cellcolor{lime!90}\else\ifdim#1pt>87.7pt\cellcolor{lime!80}\else\ifdim#1pt>87.7pt\cellcolor{lime!70}\else\ifdim#1pt>87.7pt\cellcolor{lime!60}\else\ifdim#1pt>87.7pt\cellcolor{lime!50}\else\ifdim#1pt>87.7pt\cellcolor{lime!40}\else\ifdim#1pt>87.7pt\cellcolor{lime!30}\else\ifdim#1pt>87.7pt\cellcolor{lime!20}\else\ifdim#1pt>87.7pt\cellcolor{lime!10}\else\cellcolor{lime!0}\fi\fi\fi\fi\fi\fi\fi\fi\fi\fi#1}

\def\weightedspearmanzerozerosrcthresholdperadapterPA#1{\ifdim#1pt>94.2pt\cellcolor{lime!100}\else\ifdim#1pt>94.2pt\cellcolor{lime!90}\else\ifdim#1pt>94.2pt\cellcolor{lime!80}\else\ifdim#1pt>94.2pt\cellcolor{lime!70}\else\ifdim#1pt>94.2pt\cellcolor{lime!60}\else\ifdim#1pt>94.2pt\cellcolor{lime!50}\else\ifdim#1pt>94.2pt\cellcolor{lime!40}\else\ifdim#1pt>94.2pt\cellcolor{lime!30}\else\ifdim#1pt>94.2pt\cellcolor{lime!20}\else\ifdim#1pt>94.2pt\cellcolor{lime!10}\else\cellcolor{lime!0}\fi\fi\fi\fi\fi\fi\fi\fi\fi\fi#1}

\def\weightedspearmanzerozerosrcthresholdperadapterPC#1{\ifdim#1pt>90.5pt\cellcolor{lime!100}\else\ifdim#1pt>90.5pt\cellcolor{lime!90}\else\ifdim#1pt>90.5pt\cellcolor{lime!80}\else\ifdim#1pt>90.5pt\cellcolor{lime!70}\else\ifdim#1pt>90.5pt\cellcolor{lime!60}\else\ifdim#1pt>90.5pt\cellcolor{lime!50}\else\ifdim#1pt>90.5pt\cellcolor{lime!40}\else\ifdim#1pt>90.5pt\cellcolor{lime!30}\else\ifdim#1pt>90.5pt\cellcolor{lime!20}\else\ifdim#1pt>90.5pt\cellcolor{lime!10}\else\cellcolor{lime!0}\fi\fi\fi\fi\fi\fi\fi\fi\fi\fi#1}

\def\weightedspearmanzerozerosrcthresholdperadapterPR#1{\ifdim#1pt>96.5pt\cellcolor{lime!100}\else\ifdim#1pt>96.5pt\cellcolor{lime!90}\else\ifdim#1pt>96.5pt\cellcolor{lime!80}\else\ifdim#1pt>96.5pt\cellcolor{lime!70}\else\ifdim#1pt>96.5pt\cellcolor{lime!60}\else\ifdim#1pt>96.5pt\cellcolor{lime!50}\else\ifdim#1pt>96.5pt\cellcolor{lime!40}\else\ifdim#1pt>96.5pt\cellcolor{lime!30}\else\ifdim#1pt>96.5pt\cellcolor{lime!20}\else\ifdim#1pt>96.5pt\cellcolor{lime!10}\else\cellcolor{lime!0}\fi\fi\fi\fi\fi\fi\fi\fi\fi\fi#1}

\def\weightedspearmanzerozerosrcthresholdperadapterRA#1{\ifdim#1pt>92.9pt\cellcolor{lime!100}\else\ifdim#1pt>92.9pt\cellcolor{lime!90}\else\ifdim#1pt>92.9pt\cellcolor{lime!80}\else\ifdim#1pt>92.9pt\cellcolor{lime!70}\else\ifdim#1pt>92.9pt\cellcolor{lime!60}\else\ifdim#1pt>92.9pt\cellcolor{lime!50}\else\ifdim#1pt>92.9pt\cellcolor{lime!40}\else\ifdim#1pt>92.9pt\cellcolor{lime!30}\else\ifdim#1pt>92.9pt\cellcolor{lime!20}\else\ifdim#1pt>92.9pt\cellcolor{lime!10}\else\cellcolor{lime!0}\fi\fi\fi\fi\fi\fi\fi\fi\fi\fi#1}

\def\weightedspearmanzerozerosrcthresholdperadapterRC#1{\ifdim#1pt>89.4pt\cellcolor{lime!100}\else\ifdim#1pt>88.5pt\cellcolor{lime!90}\else\ifdim#1pt>87.7pt\cellcolor{lime!80}\else\ifdim#1pt>86.8pt\cellcolor{lime!70}\else\ifdim#1pt>85.9pt\cellcolor{lime!60}\else\ifdim#1pt>85.0pt\cellcolor{lime!50}\else\ifdim#1pt>84.1pt\cellcolor{lime!40}\else\ifdim#1pt>83.3pt\cellcolor{lime!30}\else\ifdim#1pt>82.4pt\cellcolor{lime!20}\else\ifdim#1pt>81.5pt\cellcolor{lime!10}\else\cellcolor{lime!0}\fi\fi\fi\fi\fi\fi\fi\fi\fi\fi#1}

\def\weightedspearmanzerozerosrcthresholdperadapterRP#1{\ifdim#1pt>95.0pt\cellcolor{lime!100}\else\ifdim#1pt>95.0pt\cellcolor{lime!90}\else\ifdim#1pt>95.0pt\cellcolor{lime!80}\else\ifdim#1pt>95.0pt\cellcolor{lime!70}\else\ifdim#1pt>95.0pt\cellcolor{lime!60}\else\ifdim#1pt>95.0pt\cellcolor{lime!50}\else\ifdim#1pt>95.0pt\cellcolor{lime!40}\else\ifdim#1pt>95.0pt\cellcolor{lime!30}\else\ifdim#1pt>95.0pt\cellcolor{lime!20}\else\ifdim#1pt>95.0pt\cellcolor{lime!10}\else\cellcolor{lime!0}\fi\fi\fi\fi\fi\fi\fi\fi\fi\fi#1}

\def\weightedspearmanzerozerosrcthresholdperadapterMean#1{\ifdim#1pt>85.6pt\cellcolor{lime!100}\else\ifdim#1pt>85.6pt\cellcolor{lime!90}\else\ifdim#1pt>85.6pt\cellcolor{lime!80}\else\ifdim#1pt>85.6pt\cellcolor{lime!70}\else\ifdim#1pt>85.6pt\cellcolor{lime!60}\else\ifdim#1pt>85.6pt\cellcolor{lime!50}\else\ifdim#1pt>85.6pt\cellcolor{lime!40}\else\ifdim#1pt>85.6pt\cellcolor{lime!30}\else\ifdim#1pt>85.6pt\cellcolor{lime!20}\else\ifdim#1pt>85.6pt\cellcolor{lime!10}\else\cellcolor{lime!0}\fi\fi\fi\fi\fi\fi\fi\fi\fi\fi#1}

\def\weightedspearmanzerozerosrcthresholdperadapterStd#1{\ifdim#1pt<7.5pt\cellcolor{lime!100}\else\ifdim#1pt<7.7pt\cellcolor{lime!90}\else\ifdim#1pt<7.8pt\cellcolor{lime!80}\else\ifdim#1pt<8.0pt\cellcolor{lime!70}\else\ifdim#1pt<8.1pt\cellcolor{lime!60}\else\ifdim#1pt<8.2pt\cellcolor{lime!50}\else\ifdim#1pt<8.4pt\cellcolor{lime!40}\else\ifdim#1pt<8.5pt\cellcolor{lime!30}\else\ifdim#1pt<8.7pt\cellcolor{lime!20}\else\ifdim#1pt<8.8pt\cellcolor{lime!10}\else\cellcolor{lime!0}\fi\fi\fi\fi\fi\fi\fi\fi\fi\fi#1}

\begin{table}[H]
\centering
\caption{The weighted Spearman correlation of each validator/task pair for \textbf{CDAN}.}
\label{office_weighted_spearman_0.0_src_threshold_per_adapter_CDAN}
\resizebox{!}{0.22\textheight}{
\begin{tabular}{llrrrrrrrrrrrrrrrrrrrr}
\toprule
 &  & AD & AW & DA & DW & WA & WD & AC & AP & AR & CA & CP & CR & PA & PC & PR & RA & RC & RP & Mean & Std \\
\midrule
\multirow[c]{2}{*}{Accuracy} & Source Train & \weightedspearmanzerozerosrcthresholdperadapterAD{63.9} & \weightedspearmanzerozerosrcthresholdperadapterAW{78.3} & \weightedspearmanzerozerosrcthresholdperadapterDA{80.0} & \weightedspearmanzerozerosrcthresholdperadapterDW{61.7} & \textbf{\weightedspearmanzerozerosrcthresholdperadapterWA{85.6}} & \weightedspearmanzerozerosrcthresholdperadapterWD{48.7} & \weightedspearmanzerozerosrcthresholdperadapterAC{65.5} & \weightedspearmanzerozerosrcthresholdperadapterAP{70.5} & \weightedspearmanzerozerosrcthresholdperadapterAR{59.0} & \weightedspearmanzerozerosrcthresholdperadapterCA{34.7} & \weightedspearmanzerozerosrcthresholdperadapterCP{55.0} & \weightedspearmanzerozerosrcthresholdperadapterCR{59.3} & \weightedspearmanzerozerosrcthresholdperadapterPA{64.0} & \weightedspearmanzerozerosrcthresholdperadapterPC{76.3} & \weightedspearmanzerozerosrcthresholdperadapterPR{76.2} & \weightedspearmanzerozerosrcthresholdperadapterRA{52.6} & \weightedspearmanzerozerosrcthresholdperadapterRC{74.3} & \weightedspearmanzerozerosrcthresholdperadapterRP{57.5} & \weightedspearmanzerozerosrcthresholdperadapterMean{64.6} & \weightedspearmanzerozerosrcthresholdperadapterStd{12.4} \\
\bottomrule
\end{tabular}}
\end{table}

%% file: supp_tables/office31_officehome/weighted_spearman_0.0_src_threshold_per_adapter_DANN.tex
\def\weightedspearmanzerozerosrcthresholdperadapterAD#1{\ifdim#1pt>84.4pt\cellcolor{lime!100}\else\ifdim#1pt>84.4pt\cellcolor{lime!90}\else\ifdim#1pt>84.4pt\cellcolor{lime!80}\else\ifdim#1pt>84.4pt\cellcolor{lime!70}\else\ifdim#1pt>84.4pt\cellcolor{lime!60}\else\ifdim#1pt>84.4pt\cellcolor{lime!50}\else\ifdim#1pt>84.4pt\cellcolor{lime!40}\else\ifdim#1pt>84.4pt\cellcolor{lime!30}\else\ifdim#1pt>84.4pt\cellcolor{lime!20}\else\ifdim#1pt>84.4pt\cellcolor{lime!10}\else\cellcolor{lime!0}\fi\fi\fi\fi\fi\fi\fi\fi\fi\fi#1}

\def\weightedspearmanzerozerosrcthresholdperadapterAW#1{\ifdim#1pt>84.0pt\cellcolor{lime!100}\else\ifdim#1pt>83.8pt\cellcolor{lime!90}\else\ifdim#1pt>83.5pt\cellcolor{lime!80}\else\ifdim#1pt>83.3pt\cellcolor{lime!70}\else\ifdim#1pt>83.1pt\cellcolor{lime!60}\else\ifdim#1pt>82.9pt\cellcolor{lime!50}\else\ifdim#1pt>82.7pt\cellcolor{lime!40}\else\ifdim#1pt>82.4pt\cellcolor{lime!30}\else\ifdim#1pt>82.2pt\cellcolor{lime!20}\else\ifdim#1pt>82.0pt\cellcolor{lime!10}\else\cellcolor{lime!0}\fi\fi\fi\fi\fi\fi\fi\fi\fi\fi#1}

\def\weightedspearmanzerozerosrcthresholdperadapterDA#1{\ifdim#1pt>80.2pt\cellcolor{lime!100}\else\ifdim#1pt>80.2pt\cellcolor{lime!90}\else\ifdim#1pt>80.2pt\cellcolor{lime!80}\else\ifdim#1pt>80.2pt\cellcolor{lime!70}\else\ifdim#1pt>80.2pt\cellcolor{lime!60}\else\ifdim#1pt>80.2pt\cellcolor{lime!50}\else\ifdim#1pt>80.2pt\cellcolor{lime!40}\else\ifdim#1pt>80.2pt\cellcolor{lime!30}\else\ifdim#1pt>80.2pt\cellcolor{lime!20}\else\ifdim#1pt>80.2pt\cellcolor{lime!10}\else\cellcolor{lime!0}\fi\fi\fi\fi\fi\fi\fi\fi\fi\fi#1}

\def\weightedspearmanzerozerosrcthresholdperadapterDW#1{\ifdim#1pt>80.2pt\cellcolor{lime!100}\else\ifdim#1pt>80.2pt\cellcolor{lime!90}\else\ifdim#1pt>80.2pt\cellcolor{lime!80}\else\ifdim#1pt>80.2pt\cellcolor{lime!70}\else\ifdim#1pt>80.2pt\cellcolor{lime!60}\else\ifdim#1pt>80.2pt\cellcolor{lime!50}\else\ifdim#1pt>80.2pt\cellcolor{lime!40}\else\ifdim#1pt>80.2pt\cellcolor{lime!30}\else\ifdim#1pt>80.2pt\cellcolor{lime!20}\else\ifdim#1pt>80.2pt\cellcolor{lime!10}\else\cellcolor{lime!0}\fi\fi\fi\fi\fi\fi\fi\fi\fi\fi#1}

\def\weightedspearmanzerozerosrcthresholdperadapterWA#1{\ifdim#1pt>84.2pt\cellcolor{lime!100}\else\ifdim#1pt>83.9pt\cellcolor{lime!90}\else\ifdim#1pt>83.5pt\cellcolor{lime!80}\else\ifdim#1pt>83.2pt\cellcolor{lime!70}\else\ifdim#1pt>82.8pt\cellcolor{lime!60}\else\ifdim#1pt>82.5pt\cellcolor{lime!50}\else\ifdim#1pt>82.1pt\cellcolor{lime!40}\else\ifdim#1pt>81.8pt\cellcolor{lime!30}\else\ifdim#1pt>81.4pt\cellcolor{lime!20}\else\ifdim#1pt>81.1pt\cellcolor{lime!10}\else\cellcolor{lime!0}\fi\fi\fi\fi\fi\fi\fi\fi\fi\fi#1}

\def\weightedspearmanzerozerosrcthresholdperadapterWD#1{\ifdim#1pt>71.1pt\cellcolor{lime!100}\else\ifdim#1pt>71.1pt\cellcolor{lime!90}\else\ifdim#1pt>71.1pt\cellcolor{lime!80}\else\ifdim#1pt>71.1pt\cellcolor{lime!70}\else\ifdim#1pt>71.1pt\cellcolor{lime!60}\else\ifdim#1pt>71.1pt\cellcolor{lime!50}\else\ifdim#1pt>71.1pt\cellcolor{lime!40}\else\ifdim#1pt>71.1pt\cellcolor{lime!30}\else\ifdim#1pt>71.1pt\cellcolor{lime!20}\else\ifdim#1pt>71.1pt\cellcolor{lime!10}\else\cellcolor{lime!0}\fi\fi\fi\fi\fi\fi\fi\fi\fi\fi#1}

\def\weightedspearmanzerozerosrcthresholdperadapterAC#1{\ifdim#1pt>90.7pt\cellcolor{lime!100}\else\ifdim#1pt>89.3pt\cellcolor{lime!90}\else\ifdim#1pt>88.0pt\cellcolor{lime!80}\else\ifdim#1pt>86.6pt\cellcolor{lime!70}\else\ifdim#1pt>85.2pt\cellcolor{lime!60}\else\ifdim#1pt>83.9pt\cellcolor{lime!50}\else\ifdim#1pt>82.5pt\cellcolor{lime!40}\else\ifdim#1pt>81.2pt\cellcolor{lime!30}\else\ifdim#1pt>79.8pt\cellcolor{lime!20}\else\ifdim#1pt>78.5pt\cellcolor{lime!10}\else\cellcolor{lime!0}\fi\fi\fi\fi\fi\fi\fi\fi\fi\fi#1}

\def\weightedspearmanzerozerosrcthresholdperadapterAP#1{\ifdim#1pt>96.4pt\cellcolor{lime!100}\else\ifdim#1pt>96.4pt\cellcolor{lime!90}\else\ifdim#1pt>96.4pt\cellcolor{lime!80}\else\ifdim#1pt>96.4pt\cellcolor{lime!70}\else\ifdim#1pt>96.4pt\cellcolor{lime!60}\else\ifdim#1pt>96.4pt\cellcolor{lime!50}\else\ifdim#1pt>96.4pt\cellcolor{lime!40}\else\ifdim#1pt>96.4pt\cellcolor{lime!30}\else\ifdim#1pt>96.4pt\cellcolor{lime!20}\else\ifdim#1pt>96.4pt\cellcolor{lime!10}\else\cellcolor{lime!0}\fi\fi\fi\fi\fi\fi\fi\fi\fi\fi#1}

\def\weightedspearmanzerozerosrcthresholdperadapterAR#1{\ifdim#1pt>96.8pt\cellcolor{lime!100}\else\ifdim#1pt>96.8pt\cellcolor{lime!90}\else\ifdim#1pt>96.8pt\cellcolor{lime!80}\else\ifdim#1pt>96.8pt\cellcolor{lime!70}\else\ifdim#1pt>96.8pt\cellcolor{lime!60}\else\ifdim#1pt>96.8pt\cellcolor{lime!50}\else\ifdim#1pt>96.8pt\cellcolor{lime!40}\else\ifdim#1pt>96.8pt\cellcolor{lime!30}\else\ifdim#1pt>96.8pt\cellcolor{lime!20}\else\ifdim#1pt>96.8pt\cellcolor{lime!10}\else\cellcolor{lime!0}\fi\fi\fi\fi\fi\fi\fi\fi\fi\fi#1}

\def\weightedspearmanzerozerosrcthresholdperadapterCA#1{\ifdim#1pt>75.2pt\cellcolor{lime!100}\else\ifdim#1pt>75.2pt\cellcolor{lime!90}\else\ifdim#1pt>75.2pt\cellcolor{lime!80}\else\ifdim#1pt>75.2pt\cellcolor{lime!70}\else\ifdim#1pt>75.2pt\cellcolor{lime!60}\else\ifdim#1pt>75.2pt\cellcolor{lime!50}\else\ifdim#1pt>75.2pt\cellcolor{lime!40}\else\ifdim#1pt>75.2pt\cellcolor{lime!30}\else\ifdim#1pt>75.2pt\cellcolor{lime!20}\else\ifdim#1pt>75.2pt\cellcolor{lime!10}\else\cellcolor{lime!0}\fi\fi\fi\fi\fi\fi\fi\fi\fi\fi#1}

\def\weightedspearmanzerozerosrcthresholdperadapterCP#1{\ifdim#1pt>82.5pt\cellcolor{lime!100}\else\ifdim#1pt>82.5pt\cellcolor{lime!90}\else\ifdim#1pt>82.5pt\cellcolor{lime!80}\else\ifdim#1pt>82.5pt\cellcolor{lime!70}\else\ifdim#1pt>82.5pt\cellcolor{lime!60}\else\ifdim#1pt>82.5pt\cellcolor{lime!50}\else\ifdim#1pt>82.5pt\cellcolor{lime!40}\else\ifdim#1pt>82.5pt\cellcolor{lime!30}\else\ifdim#1pt>82.5pt\cellcolor{lime!20}\else\ifdim#1pt>82.5pt\cellcolor{lime!10}\else\cellcolor{lime!0}\fi\fi\fi\fi\fi\fi\fi\fi\fi\fi#1}

\def\weightedspearmanzerozerosrcthresholdperadapterCR#1{\ifdim#1pt>78.8pt\cellcolor{lime!100}\else\ifdim#1pt>78.8pt\cellcolor{lime!90}\else\ifdim#1pt>78.8pt\cellcolor{lime!80}\else\ifdim#1pt>78.8pt\cellcolor{lime!70}\else\ifdim#1pt>78.8pt\cellcolor{lime!60}\else\ifdim#1pt>78.8pt\cellcolor{lime!50}\else\ifdim#1pt>78.8pt\cellcolor{lime!40}\else\ifdim#1pt>78.8pt\cellcolor{lime!30}\else\ifdim#1pt>78.8pt\cellcolor{lime!20}\else\ifdim#1pt>78.8pt\cellcolor{lime!10}\else\cellcolor{lime!0}\fi\fi\fi\fi\fi\fi\fi\fi\fi\fi#1}

\def\weightedspearmanzerozerosrcthresholdperadapterPA#1{\ifdim#1pt>90.8pt\cellcolor{lime!100}\else\ifdim#1pt>90.8pt\cellcolor{lime!90}\else\ifdim#1pt>90.8pt\cellcolor{lime!80}\else\ifdim#1pt>90.8pt\cellcolor{lime!70}\else\ifdim#1pt>90.8pt\cellcolor{lime!60}\else\ifdim#1pt>90.8pt\cellcolor{lime!50}\else\ifdim#1pt>90.8pt\cellcolor{lime!40}\else\ifdim#1pt>90.8pt\cellcolor{lime!30}\else\ifdim#1pt>90.8pt\cellcolor{lime!20}\else\ifdim#1pt>90.8pt\cellcolor{lime!10}\else\cellcolor{lime!0}\fi\fi\fi\fi\fi\fi\fi\fi\fi\fi#1}

\def\weightedspearmanzerozerosrcthresholdperadapterPC#1{\ifdim#1pt>90.9pt\cellcolor{lime!100}\else\ifdim#1pt>90.6pt\cellcolor{lime!90}\else\ifdim#1pt>90.4pt\cellcolor{lime!80}\else\ifdim#1pt>90.1pt\cellcolor{lime!70}\else\ifdim#1pt>89.9pt\cellcolor{lime!60}\else\ifdim#1pt>89.7pt\cellcolor{lime!50}\else\ifdim#1pt>89.4pt\cellcolor{lime!40}\else\ifdim#1pt>89.2pt\cellcolor{lime!30}\else\ifdim#1pt>88.9pt\cellcolor{lime!20}\else\ifdim#1pt>88.7pt\cellcolor{lime!10}\else\cellcolor{lime!0}\fi\fi\fi\fi\fi\fi\fi\fi\fi\fi#1}

\def\weightedspearmanzerozerosrcthresholdperadapterPR#1{\ifdim#1pt>94.5pt\cellcolor{lime!100}\else\ifdim#1pt>94.5pt\cellcolor{lime!90}\else\ifdim#1pt>94.5pt\cellcolor{lime!80}\else\ifdim#1pt>94.5pt\cellcolor{lime!70}\else\ifdim#1pt>94.5pt\cellcolor{lime!60}\else\ifdim#1pt>94.5pt\cellcolor{lime!50}\else\ifdim#1pt>94.5pt\cellcolor{lime!40}\else\ifdim#1pt>94.5pt\cellcolor{lime!30}\else\ifdim#1pt>94.5pt\cellcolor{lime!20}\else\ifdim#1pt>94.5pt\cellcolor{lime!10}\else\cellcolor{lime!0}\fi\fi\fi\fi\fi\fi\fi\fi\fi\fi#1}

\def\weightedspearmanzerozerosrcthresholdperadapterRA#1{\ifdim#1pt>91.6pt\cellcolor{lime!100}\else\ifdim#1pt>91.6pt\cellcolor{lime!90}\else\ifdim#1pt>91.6pt\cellcolor{lime!80}\else\ifdim#1pt>91.6pt\cellcolor{lime!70}\else\ifdim#1pt>91.6pt\cellcolor{lime!60}\else\ifdim#1pt>91.6pt\cellcolor{lime!50}\else\ifdim#1pt>91.6pt\cellcolor{lime!40}\else\ifdim#1pt>91.6pt\cellcolor{lime!30}\else\ifdim#1pt>91.6pt\cellcolor{lime!20}\else\ifdim#1pt>91.6pt\cellcolor{lime!10}\else\cellcolor{lime!0}\fi\fi\fi\fi\fi\fi\fi\fi\fi\fi#1}

\def\weightedspearmanzerozerosrcthresholdperadapterRC#1{\ifdim#1pt>90.3pt\cellcolor{lime!100}\else\ifdim#1pt>88.1pt\cellcolor{lime!90}\else\ifdim#1pt>85.9pt\cellcolor{lime!80}\else\ifdim#1pt>83.7pt\cellcolor{lime!70}\else\ifdim#1pt>81.5pt\cellcolor{lime!60}\else\ifdim#1pt>79.4pt\cellcolor{lime!50}\else\ifdim#1pt>77.2pt\cellcolor{lime!40}\else\ifdim#1pt>75.0pt\cellcolor{lime!30}\else\ifdim#1pt>72.8pt\cellcolor{lime!20}\else\ifdim#1pt>70.6pt\cellcolor{lime!10}\else\cellcolor{lime!0}\fi\fi\fi\fi\fi\fi\fi\fi\fi\fi#1}

\def\weightedspearmanzerozerosrcthresholdperadapterRP#1{\ifdim#1pt>91.8pt\cellcolor{lime!100}\else\ifdim#1pt>91.8pt\cellcolor{lime!90}\else\ifdim#1pt>91.8pt\cellcolor{lime!80}\else\ifdim#1pt>91.8pt\cellcolor{lime!70}\else\ifdim#1pt>91.8pt\cellcolor{lime!60}\else\ifdim#1pt>91.8pt\cellcolor{lime!50}\else\ifdim#1pt>91.8pt\cellcolor{lime!40}\else\ifdim#1pt>91.8pt\cellcolor{lime!30}\else\ifdim#1pt>91.8pt\cellcolor{lime!20}\else\ifdim#1pt>91.8pt\cellcolor{lime!10}\else\cellcolor{lime!0}\fi\fi\fi\fi\fi\fi\fi\fi\fi\fi#1}

\def\weightedspearmanzerozerosrcthresholdperadapterMean#1{\ifdim#1pt>84.2pt\cellcolor{lime!100}\else\ifdim#1pt>84.2pt\cellcolor{lime!90}\else\ifdim#1pt>84.2pt\cellcolor{lime!80}\else\ifdim#1pt>84.2pt\cellcolor{lime!70}\else\ifdim#1pt>84.2pt\cellcolor{lime!60}\else\ifdim#1pt>84.2pt\cellcolor{lime!50}\else\ifdim#1pt>84.2pt\cellcolor{lime!40}\else\ifdim#1pt>84.2pt\cellcolor{lime!30}\else\ifdim#1pt>84.2pt\cellcolor{lime!20}\else\ifdim#1pt>84.2pt\cellcolor{lime!10}\else\cellcolor{lime!0}\fi\fi\fi\fi\fi\fi\fi\fi\fi\fi#1}

\def\weightedspearmanzerozerosrcthresholdperadapterStd#1{\ifdim#1pt<5.8pt\cellcolor{lime!100}\else\ifdim#1pt<6.0pt\cellcolor{lime!90}\else\ifdim#1pt<6.2pt\cellcolor{lime!80}\else\ifdim#1pt<6.5pt\cellcolor{lime!70}\else\ifdim#1pt<6.8pt\cellcolor{lime!60}\else\ifdim#1pt<7.0pt\cellcolor{lime!50}\else\ifdim#1pt<7.2pt\cellcolor{lime!40}\else\ifdim#1pt<7.5pt\cellcolor{lime!30}\else\ifdim#1pt<7.8pt\cellcolor{lime!20}\else\ifdim#1pt<8.0pt\cellcolor{lime!10}\else\cellcolor{lime!0}\fi\fi\fi\fi\fi\fi\fi\fi\fi\fi#1}

\begin{table}[H]
\centering
\caption{The weighted Spearman correlation of each validator/task pair for \textbf{DANN}.}
\label{office_weighted_spearman_0.0_src_threshold_per_adapter_DANN}
\resizebox{!}{0.22\textheight}{
\begin{tabular}{llrrrrrrrrrrrrrrrrrrrr}
\toprule
 &  & AD & AW & DA & DW & WA & WD & AC & AP & AR & CA & CP & CR & PA & PC & PR & RA & RC & RP & Mean & Std \\
\midrule
\multirow[c]{2}{*}{Accuracy} & Source Train & \weightedspearmanzerozerosrcthresholdperadapterAD{70.6} & \weightedspearmanzerozerosrcthresholdperadapterAW{80.2} & \weightedspearmanzerozerosrcthresholdperadapterDA{78.8} & \weightedspearmanzerozerosrcthresholdperadapterDW{58.8} & \weightedspearmanzerozerosrcthresholdperadapterWA{81.4} & \weightedspearmanzerozerosrcthresholdperadapterWD{48.7} & \weightedspearmanzerozerosrcthresholdperadapterAC{65.8} & \weightedspearmanzerozerosrcthresholdperadapterAP{63.4} & \weightedspearmanzerozerosrcthresholdperadapterAR{58.5} & \weightedspearmanzerozerosrcthresholdperadapterCA{34.2} & \weightedspearmanzerozerosrcthresholdperadapterCP{49.2} & \weightedspearmanzerozerosrcthresholdperadapterCR{44.1} & \weightedspearmanzerozerosrcthresholdperadapterPA{55.2} & \weightedspearmanzerozerosrcthresholdperadapterPC{75.1} & \weightedspearmanzerozerosrcthresholdperadapterPR{71.4} & \weightedspearmanzerozerosrcthresholdperadapterRA{47.7} & \weightedspearmanzerozerosrcthresholdperadapterRC{74.3} & \weightedspearmanzerozerosrcthresholdperadapterRP{54.8} & \weightedspearmanzerozerosrcthresholdperadapterMean{61.8} & \weightedspearmanzerozerosrcthresholdperadapterStd{13.4} \\
\bottomrule
\end{tabular}}
\end{table}

%% file: supp_tables/office31_officehome/weighted_spearman_0.0_src_threshold_per_adapter_GVB.tex
\def\weightedspearmanzerozerosrcthresholdperadapterAD#1{\ifdim#1pt>87.2pt\cellcolor{lime!100}\else\ifdim#1pt>87.2pt\cellcolor{lime!90}\else\ifdim#1pt>87.2pt\cellcolor{lime!80}\else\ifdim#1pt>87.2pt\cellcolor{lime!70}\else\ifdim#1pt>87.2pt\cellcolor{lime!60}\else\ifdim#1pt>87.2pt\cellcolor{lime!50}\else\ifdim#1pt>87.2pt\cellcolor{lime!40}\else\ifdim#1pt>87.2pt\cellcolor{lime!30}\else\ifdim#1pt>87.2pt\cellcolor{lime!20}\else\ifdim#1pt>87.2pt\cellcolor{lime!10}\else\cellcolor{lime!0}\fi\fi\fi\fi\fi\fi\fi\fi\fi\fi#1}

\def\weightedspearmanzerozerosrcthresholdperadapterAW#1{\ifdim#1pt>84.3pt\cellcolor{lime!100}\else\ifdim#1pt>84.3pt\cellcolor{lime!90}\else\ifdim#1pt>84.3pt\cellcolor{lime!80}\else\ifdim#1pt>84.3pt\cellcolor{lime!70}\else\ifdim#1pt>84.3pt\cellcolor{lime!60}\else\ifdim#1pt>84.3pt\cellcolor{lime!50}\else\ifdim#1pt>84.3pt\cellcolor{lime!40}\else\ifdim#1pt>84.3pt\cellcolor{lime!30}\else\ifdim#1pt>84.3pt\cellcolor{lime!20}\else\ifdim#1pt>84.3pt\cellcolor{lime!10}\else\cellcolor{lime!0}\fi\fi\fi\fi\fi\fi\fi\fi\fi\fi#1}

\def\weightedspearmanzerozerosrcthresholdperadapterDA#1{\ifdim#1pt>86.9pt\cellcolor{lime!100}\else\ifdim#1pt>86.9pt\cellcolor{lime!90}\else\ifdim#1pt>86.9pt\cellcolor{lime!80}\else\ifdim#1pt>86.9pt\cellcolor{lime!70}\else\ifdim#1pt>86.9pt\cellcolor{lime!60}\else\ifdim#1pt>86.9pt\cellcolor{lime!50}\else\ifdim#1pt>86.9pt\cellcolor{lime!40}\else\ifdim#1pt>86.9pt\cellcolor{lime!30}\else\ifdim#1pt>86.9pt\cellcolor{lime!20}\else\ifdim#1pt>86.9pt\cellcolor{lime!10}\else\cellcolor{lime!0}\fi\fi\fi\fi\fi\fi\fi\fi\fi\fi#1}

\def\weightedspearmanzerozerosrcthresholdperadapterDW#1{\ifdim#1pt>81.5pt\cellcolor{lime!100}\else\ifdim#1pt>81.5pt\cellcolor{lime!90}\else\ifdim#1pt>81.5pt\cellcolor{lime!80}\else\ifdim#1pt>81.5pt\cellcolor{lime!70}\else\ifdim#1pt>81.5pt\cellcolor{lime!60}\else\ifdim#1pt>81.5pt\cellcolor{lime!50}\else\ifdim#1pt>81.5pt\cellcolor{lime!40}\else\ifdim#1pt>81.5pt\cellcolor{lime!30}\else\ifdim#1pt>81.5pt\cellcolor{lime!20}\else\ifdim#1pt>81.5pt\cellcolor{lime!10}\else\cellcolor{lime!0}\fi\fi\fi\fi\fi\fi\fi\fi\fi\fi#1}

\def\weightedspearmanzerozerosrcthresholdperadapterWA#1{\ifdim#1pt>83.5pt\cellcolor{lime!100}\else\ifdim#1pt>83.5pt\cellcolor{lime!90}\else\ifdim#1pt>83.5pt\cellcolor{lime!80}\else\ifdim#1pt>83.5pt\cellcolor{lime!70}\else\ifdim#1pt>83.5pt\cellcolor{lime!60}\else\ifdim#1pt>83.5pt\cellcolor{lime!50}\else\ifdim#1pt>83.5pt\cellcolor{lime!40}\else\ifdim#1pt>83.5pt\cellcolor{lime!30}\else\ifdim#1pt>83.5pt\cellcolor{lime!20}\else\ifdim#1pt>83.5pt\cellcolor{lime!10}\else\cellcolor{lime!0}\fi\fi\fi\fi\fi\fi\fi\fi\fi\fi#1}

\def\weightedspearmanzerozerosrcthresholdperadapterWD#1{\ifdim#1pt>76.7pt\cellcolor{lime!100}\else\ifdim#1pt>76.7pt\cellcolor{lime!90}\else\ifdim#1pt>76.7pt\cellcolor{lime!80}\else\ifdim#1pt>76.7pt\cellcolor{lime!70}\else\ifdim#1pt>76.7pt\cellcolor{lime!60}\else\ifdim#1pt>76.7pt\cellcolor{lime!50}\else\ifdim#1pt>76.7pt\cellcolor{lime!40}\else\ifdim#1pt>76.7pt\cellcolor{lime!30}\else\ifdim#1pt>76.7pt\cellcolor{lime!20}\else\ifdim#1pt>76.7pt\cellcolor{lime!10}\else\cellcolor{lime!0}\fi\fi\fi\fi\fi\fi\fi\fi\fi\fi#1}

\def\weightedspearmanzerozerosrcthresholdperadapterAC#1{\ifdim#1pt>89.4pt\cellcolor{lime!100}\else\ifdim#1pt>89.4pt\cellcolor{lime!90}\else\ifdim#1pt>89.4pt\cellcolor{lime!80}\else\ifdim#1pt>89.4pt\cellcolor{lime!70}\else\ifdim#1pt>89.4pt\cellcolor{lime!60}\else\ifdim#1pt>89.4pt\cellcolor{lime!50}\else\ifdim#1pt>89.4pt\cellcolor{lime!40}\else\ifdim#1pt>89.4pt\cellcolor{lime!30}\else\ifdim#1pt>89.4pt\cellcolor{lime!20}\else\ifdim#1pt>89.4pt\cellcolor{lime!10}\else\cellcolor{lime!0}\fi\fi\fi\fi\fi\fi\fi\fi\fi\fi#1}

\def\weightedspearmanzerozerosrcthresholdperadapterAP#1{\ifdim#1pt>96.6pt\cellcolor{lime!100}\else\ifdim#1pt>96.6pt\cellcolor{lime!90}\else\ifdim#1pt>96.6pt\cellcolor{lime!80}\else\ifdim#1pt>96.6pt\cellcolor{lime!70}\else\ifdim#1pt>96.6pt\cellcolor{lime!60}\else\ifdim#1pt>96.6pt\cellcolor{lime!50}\else\ifdim#1pt>96.6pt\cellcolor{lime!40}\else\ifdim#1pt>96.6pt\cellcolor{lime!30}\else\ifdim#1pt>96.6pt\cellcolor{lime!20}\else\ifdim#1pt>96.6pt\cellcolor{lime!10}\else\cellcolor{lime!0}\fi\fi\fi\fi\fi\fi\fi\fi\fi\fi#1}

\def\weightedspearmanzerozerosrcthresholdperadapterAR#1{\ifdim#1pt>97.1pt\cellcolor{lime!100}\else\ifdim#1pt>97.1pt\cellcolor{lime!90}\else\ifdim#1pt>97.1pt\cellcolor{lime!80}\else\ifdim#1pt>97.1pt\cellcolor{lime!70}\else\ifdim#1pt>97.1pt\cellcolor{lime!60}\else\ifdim#1pt>97.1pt\cellcolor{lime!50}\else\ifdim#1pt>97.1pt\cellcolor{lime!40}\else\ifdim#1pt>97.1pt\cellcolor{lime!30}\else\ifdim#1pt>97.1pt\cellcolor{lime!20}\else\ifdim#1pt>97.1pt\cellcolor{lime!10}\else\cellcolor{lime!0}\fi\fi\fi\fi\fi\fi\fi\fi\fi\fi#1}

\def\weightedspearmanzerozerosrcthresholdperadapterCA#1{\ifdim#1pt>80.8pt\cellcolor{lime!100}\else\ifdim#1pt>80.8pt\cellcolor{lime!90}\else\ifdim#1pt>80.8pt\cellcolor{lime!80}\else\ifdim#1pt>80.8pt\cellcolor{lime!70}\else\ifdim#1pt>80.8pt\cellcolor{lime!60}\else\ifdim#1pt>80.8pt\cellcolor{lime!50}\else\ifdim#1pt>80.8pt\cellcolor{lime!40}\else\ifdim#1pt>80.8pt\cellcolor{lime!30}\else\ifdim#1pt>80.8pt\cellcolor{lime!20}\else\ifdim#1pt>80.8pt\cellcolor{lime!10}\else\cellcolor{lime!0}\fi\fi\fi\fi\fi\fi\fi\fi\fi\fi#1}

\def\weightedspearmanzerozerosrcthresholdperadapterCP#1{\ifdim#1pt>86.8pt\cellcolor{lime!100}\else\ifdim#1pt>86.8pt\cellcolor{lime!90}\else\ifdim#1pt>86.8pt\cellcolor{lime!80}\else\ifdim#1pt>86.8pt\cellcolor{lime!70}\else\ifdim#1pt>86.8pt\cellcolor{lime!60}\else\ifdim#1pt>86.8pt\cellcolor{lime!50}\else\ifdim#1pt>86.8pt\cellcolor{lime!40}\else\ifdim#1pt>86.8pt\cellcolor{lime!30}\else\ifdim#1pt>86.8pt\cellcolor{lime!20}\else\ifdim#1pt>86.8pt\cellcolor{lime!10}\else\cellcolor{lime!0}\fi\fi\fi\fi\fi\fi\fi\fi\fi\fi#1}

\def\weightedspearmanzerozerosrcthresholdperadapterCR#1{\ifdim#1pt>84.3pt\cellcolor{lime!100}\else\ifdim#1pt>84.3pt\cellcolor{lime!90}\else\ifdim#1pt>84.3pt\cellcolor{lime!80}\else\ifdim#1pt>84.3pt\cellcolor{lime!70}\else\ifdim#1pt>84.3pt\cellcolor{lime!60}\else\ifdim#1pt>84.3pt\cellcolor{lime!50}\else\ifdim#1pt>84.3pt\cellcolor{lime!40}\else\ifdim#1pt>84.3pt\cellcolor{lime!30}\else\ifdim#1pt>84.3pt\cellcolor{lime!20}\else\ifdim#1pt>84.3pt\cellcolor{lime!10}\else\cellcolor{lime!0}\fi\fi\fi\fi\fi\fi\fi\fi\fi\fi#1}

\def\weightedspearmanzerozerosrcthresholdperadapterPA#1{\ifdim#1pt>93.6pt\cellcolor{lime!100}\else\ifdim#1pt>93.6pt\cellcolor{lime!90}\else\ifdim#1pt>93.6pt\cellcolor{lime!80}\else\ifdim#1pt>93.6pt\cellcolor{lime!70}\else\ifdim#1pt>93.6pt\cellcolor{lime!60}\else\ifdim#1pt>93.6pt\cellcolor{lime!50}\else\ifdim#1pt>93.6pt\cellcolor{lime!40}\else\ifdim#1pt>93.6pt\cellcolor{lime!30}\else\ifdim#1pt>93.6pt\cellcolor{lime!20}\else\ifdim#1pt>93.6pt\cellcolor{lime!10}\else\cellcolor{lime!0}\fi\fi\fi\fi\fi\fi\fi\fi\fi\fi#1}

\def\weightedspearmanzerozerosrcthresholdperadapterPC#1{\ifdim#1pt>91.7pt\cellcolor{lime!100}\else\ifdim#1pt>91.7pt\cellcolor{lime!90}\else\ifdim#1pt>91.7pt\cellcolor{lime!80}\else\ifdim#1pt>91.7pt\cellcolor{lime!70}\else\ifdim#1pt>91.7pt\cellcolor{lime!60}\else\ifdim#1pt>91.7pt\cellcolor{lime!50}\else\ifdim#1pt>91.7pt\cellcolor{lime!40}\else\ifdim#1pt>91.7pt\cellcolor{lime!30}\else\ifdim#1pt>91.7pt\cellcolor{lime!20}\else\ifdim#1pt>91.7pt\cellcolor{lime!10}\else\cellcolor{lime!0}\fi\fi\fi\fi\fi\fi\fi\fi\fi\fi#1}

\def\weightedspearmanzerozerosrcthresholdperadapterPR#1{\ifdim#1pt>94.4pt\cellcolor{lime!100}\else\ifdim#1pt>94.4pt\cellcolor{lime!90}\else\ifdim#1pt>94.4pt\cellcolor{lime!80}\else\ifdim#1pt>94.4pt\cellcolor{lime!70}\else\ifdim#1pt>94.4pt\cellcolor{lime!60}\else\ifdim#1pt>94.4pt\cellcolor{lime!50}\else\ifdim#1pt>94.4pt\cellcolor{lime!40}\else\ifdim#1pt>94.4pt\cellcolor{lime!30}\else\ifdim#1pt>94.4pt\cellcolor{lime!20}\else\ifdim#1pt>94.4pt\cellcolor{lime!10}\else\cellcolor{lime!0}\fi\fi\fi\fi\fi\fi\fi\fi\fi\fi#1}

\def\weightedspearmanzerozerosrcthresholdperadapterRA#1{\ifdim#1pt>86.6pt\cellcolor{lime!100}\else\ifdim#1pt>86.6pt\cellcolor{lime!90}\else\ifdim#1pt>86.6pt\cellcolor{lime!80}\else\ifdim#1pt>86.6pt\cellcolor{lime!70}\else\ifdim#1pt>86.6pt\cellcolor{lime!60}\else\ifdim#1pt>86.6pt\cellcolor{lime!50}\else\ifdim#1pt>86.6pt\cellcolor{lime!40}\else\ifdim#1pt>86.6pt\cellcolor{lime!30}\else\ifdim#1pt>86.6pt\cellcolor{lime!20}\else\ifdim#1pt>86.6pt\cellcolor{lime!10}\else\cellcolor{lime!0}\fi\fi\fi\fi\fi\fi\fi\fi\fi\fi#1}

\def\weightedspearmanzerozerosrcthresholdperadapterRC#1{\ifdim#1pt>89.7pt\cellcolor{lime!100}\else\ifdim#1pt>89.1pt\cellcolor{lime!90}\else\ifdim#1pt>88.5pt\cellcolor{lime!80}\else\ifdim#1pt>88.0pt\cellcolor{lime!70}\else\ifdim#1pt>87.5pt\cellcolor{lime!60}\else\ifdim#1pt>86.9pt\cellcolor{lime!50}\else\ifdim#1pt>86.4pt\cellcolor{lime!40}\else\ifdim#1pt>85.8pt\cellcolor{lime!30}\else\ifdim#1pt>85.2pt\cellcolor{lime!20}\else\ifdim#1pt>84.7pt\cellcolor{lime!10}\else\cellcolor{lime!0}\fi\fi\fi\fi\fi\fi\fi\fi\fi\fi#1}

\def\weightedspearmanzerozerosrcthresholdperadapterRP#1{\ifdim#1pt>91.9pt\cellcolor{lime!100}\else\ifdim#1pt>91.9pt\cellcolor{lime!90}\else\ifdim#1pt>91.9pt\cellcolor{lime!80}\else\ifdim#1pt>91.9pt\cellcolor{lime!70}\else\ifdim#1pt>91.9pt\cellcolor{lime!60}\else\ifdim#1pt>91.9pt\cellcolor{lime!50}\else\ifdim#1pt>91.9pt\cellcolor{lime!40}\else\ifdim#1pt>91.9pt\cellcolor{lime!30}\else\ifdim#1pt>91.9pt\cellcolor{lime!20}\else\ifdim#1pt>91.9pt\cellcolor{lime!10}\else\cellcolor{lime!0}\fi\fi\fi\fi\fi\fi\fi\fi\fi\fi#1}

\def\weightedspearmanzerozerosrcthresholdperadapterMean#1{\ifdim#1pt>87.7pt\cellcolor{lime!100}\else\ifdim#1pt>87.7pt\cellcolor{lime!90}\else\ifdim#1pt>87.7pt\cellcolor{lime!80}\else\ifdim#1pt>87.7pt\cellcolor{lime!70}\else\ifdim#1pt>87.7pt\cellcolor{lime!60}\else\ifdim#1pt>87.7pt\cellcolor{lime!50}\else\ifdim#1pt>87.7pt\cellcolor{lime!40}\else\ifdim#1pt>87.7pt\cellcolor{lime!30}\else\ifdim#1pt>87.7pt\cellcolor{lime!20}\else\ifdim#1pt>87.7pt\cellcolor{lime!10}\else\cellcolor{lime!0}\fi\fi\fi\fi\fi\fi\fi\fi\fi\fi#1}

\def\weightedspearmanzerozerosrcthresholdperadapterStd#1{\ifdim#1pt<5.5pt\cellcolor{lime!100}\else\ifdim#1pt<5.5pt\cellcolor{lime!90}\else\ifdim#1pt<5.5pt\cellcolor{lime!80}\else\ifdim#1pt<5.5pt\cellcolor{lime!70}\else\ifdim#1pt<5.5pt\cellcolor{lime!60}\else\ifdim#1pt<5.5pt\cellcolor{lime!50}\else\ifdim#1pt<5.5pt\cellcolor{lime!40}\else\ifdim#1pt<5.5pt\cellcolor{lime!30}\else\ifdim#1pt<5.5pt\cellcolor{lime!20}\else\ifdim#1pt<5.5pt\cellcolor{lime!10}\else\cellcolor{lime!0}\fi\fi\fi\fi\fi\fi\fi\fi\fi\fi#1}

\begin{table}[H]
\centering
\caption{The weighted Spearman correlation of each validator/task pair for \textbf{GVB}.}
\label{office_weighted_spearman_0.0_src_threshold_per_adapter_GVB}
\resizebox{!}{0.22\textheight}{
\begin{tabular}{llrrrrrrrrrrrrrrrrrrrr}
\toprule
 &  & AD & AW & DA & DW & WA & WD & AC & AP & AR & CA & CP & CR & PA & PC & PR & RA & RC & RP & Mean & Std \\
\midrule
\multirow[c]{2}{*}{Accuracy} & Source Train & \weightedspearmanzerozerosrcthresholdperadapterAD{69.3} & \weightedspearmanzerozerosrcthresholdperadapterAW{71.8} & \weightedspearmanzerozerosrcthresholdperadapterDA{82.1} & \weightedspearmanzerozerosrcthresholdperadapterDW{63.0} & \weightedspearmanzerozerosrcthresholdperadapterWA{82.1} & \weightedspearmanzerozerosrcthresholdperadapterWD{67.5} & \weightedspearmanzerozerosrcthresholdperadapterAC{51.7} & \weightedspearmanzerozerosrcthresholdperadapterAP{52.4} & \weightedspearmanzerozerosrcthresholdperadapterAR{50.3} & \weightedspearmanzerozerosrcthresholdperadapterCA{42.9} & \weightedspearmanzerozerosrcthresholdperadapterCP{49.7} & \weightedspearmanzerozerosrcthresholdperadapterCR{45.4} & \weightedspearmanzerozerosrcthresholdperadapterPA{69.8} & \weightedspearmanzerozerosrcthresholdperadapterPC{66.0} & \weightedspearmanzerozerosrcthresholdperadapterPR{61.2} & \weightedspearmanzerozerosrcthresholdperadapterRA{49.7} & \weightedspearmanzerozerosrcthresholdperadapterRC{64.5} & \weightedspearmanzerozerosrcthresholdperadapterRP{52.3} & \weightedspearmanzerozerosrcthresholdperadapterMean{60.6} & \weightedspearmanzerozerosrcthresholdperadapterStd{11.6} \\
\bottomrule
\end{tabular}}
\end{table}

%% file: supp_tables/office31_officehome/weighted_spearman_0.0_src_threshold_per_adapter_IM.tex
\def\weightedspearmanzerozerosrcthresholdperadapterAD#1{\ifdim#1pt>81.8pt\cellcolor{lime!100}\else\ifdim#1pt>81.8pt\cellcolor{lime!90}\else\ifdim#1pt>81.8pt\cellcolor{lime!80}\else\ifdim#1pt>81.8pt\cellcolor{lime!70}\else\ifdim#1pt>81.8pt\cellcolor{lime!60}\else\ifdim#1pt>81.8pt\cellcolor{lime!50}\else\ifdim#1pt>81.8pt\cellcolor{lime!40}\else\ifdim#1pt>81.8pt\cellcolor{lime!30}\else\ifdim#1pt>81.8pt\cellcolor{lime!20}\else\ifdim#1pt>81.8pt\cellcolor{lime!10}\else\cellcolor{lime!0}\fi\fi\fi\fi\fi\fi\fi\fi\fi\fi#1}

\def\weightedspearmanzerozerosrcthresholdperadapterAW#1{\ifdim#1pt>88.2pt\cellcolor{lime!100}\else\ifdim#1pt>88.2pt\cellcolor{lime!90}\else\ifdim#1pt>88.2pt\cellcolor{lime!80}\else\ifdim#1pt>88.2pt\cellcolor{lime!70}\else\ifdim#1pt>88.2pt\cellcolor{lime!60}\else\ifdim#1pt>88.2pt\cellcolor{lime!50}\else\ifdim#1pt>88.2pt\cellcolor{lime!40}\else\ifdim#1pt>88.2pt\cellcolor{lime!30}\else\ifdim#1pt>88.2pt\cellcolor{lime!20}\else\ifdim#1pt>88.2pt\cellcolor{lime!10}\else\cellcolor{lime!0}\fi\fi\fi\fi\fi\fi\fi\fi\fi\fi#1}

\def\weightedspearmanzerozerosrcthresholdperadapterDA#1{\ifdim#1pt>86.8pt\cellcolor{lime!100}\else\ifdim#1pt>86.8pt\cellcolor{lime!90}\else\ifdim#1pt>86.8pt\cellcolor{lime!80}\else\ifdim#1pt>86.8pt\cellcolor{lime!70}\else\ifdim#1pt>86.8pt\cellcolor{lime!60}\else\ifdim#1pt>86.8pt\cellcolor{lime!50}\else\ifdim#1pt>86.8pt\cellcolor{lime!40}\else\ifdim#1pt>86.8pt\cellcolor{lime!30}\else\ifdim#1pt>86.8pt\cellcolor{lime!20}\else\ifdim#1pt>86.8pt\cellcolor{lime!10}\else\cellcolor{lime!0}\fi\fi\fi\fi\fi\fi\fi\fi\fi\fi#1}

\def\weightedspearmanzerozerosrcthresholdperadapterDW#1{\ifdim#1pt>86.8pt\cellcolor{lime!100}\else\ifdim#1pt>86.8pt\cellcolor{lime!90}\else\ifdim#1pt>86.8pt\cellcolor{lime!80}\else\ifdim#1pt>86.8pt\cellcolor{lime!70}\else\ifdim#1pt>86.8pt\cellcolor{lime!60}\else\ifdim#1pt>86.8pt\cellcolor{lime!50}\else\ifdim#1pt>86.8pt\cellcolor{lime!40}\else\ifdim#1pt>86.8pt\cellcolor{lime!30}\else\ifdim#1pt>86.8pt\cellcolor{lime!20}\else\ifdim#1pt>86.8pt\cellcolor{lime!10}\else\cellcolor{lime!0}\fi\fi\fi\fi\fi\fi\fi\fi\fi\fi#1}

\def\weightedspearmanzerozerosrcthresholdperadapterWA#1{\ifdim#1pt>84.8pt\cellcolor{lime!100}\else\ifdim#1pt>84.5pt\cellcolor{lime!90}\else\ifdim#1pt>84.2pt\cellcolor{lime!80}\else\ifdim#1pt>83.9pt\cellcolor{lime!70}\else\ifdim#1pt>83.6pt\cellcolor{lime!60}\else\ifdim#1pt>83.3pt\cellcolor{lime!50}\else\ifdim#1pt>83.0pt\cellcolor{lime!40}\else\ifdim#1pt>82.7pt\cellcolor{lime!30}\else\ifdim#1pt>82.4pt\cellcolor{lime!20}\else\ifdim#1pt>82.1pt\cellcolor{lime!10}\else\cellcolor{lime!0}\fi\fi\fi\fi\fi\fi\fi\fi\fi\fi#1}

\def\weightedspearmanzerozerosrcthresholdperadapterWD#1{\ifdim#1pt>78.4pt\cellcolor{lime!100}\else\ifdim#1pt>78.4pt\cellcolor{lime!90}\else\ifdim#1pt>78.4pt\cellcolor{lime!80}\else\ifdim#1pt>78.4pt\cellcolor{lime!70}\else\ifdim#1pt>78.4pt\cellcolor{lime!60}\else\ifdim#1pt>78.4pt\cellcolor{lime!50}\else\ifdim#1pt>78.4pt\cellcolor{lime!40}\else\ifdim#1pt>78.4pt\cellcolor{lime!30}\else\ifdim#1pt>78.4pt\cellcolor{lime!20}\else\ifdim#1pt>78.4pt\cellcolor{lime!10}\else\cellcolor{lime!0}\fi\fi\fi\fi\fi\fi\fi\fi\fi\fi#1}

\def\weightedspearmanzerozerosrcthresholdperadapterAC#1{\ifdim#1pt>89.0pt\cellcolor{lime!100}\else\ifdim#1pt>87.7pt\cellcolor{lime!90}\else\ifdim#1pt>86.5pt\cellcolor{lime!80}\else\ifdim#1pt>85.3pt\cellcolor{lime!70}\else\ifdim#1pt>84.1pt\cellcolor{lime!60}\else\ifdim#1pt>82.8pt\cellcolor{lime!50}\else\ifdim#1pt>81.6pt\cellcolor{lime!40}\else\ifdim#1pt>80.4pt\cellcolor{lime!30}\else\ifdim#1pt>79.1pt\cellcolor{lime!20}\else\ifdim#1pt>77.9pt\cellcolor{lime!10}\else\cellcolor{lime!0}\fi\fi\fi\fi\fi\fi\fi\fi\fi\fi#1}

\def\weightedspearmanzerozerosrcthresholdperadapterAP#1{\ifdim#1pt>92.8pt\cellcolor{lime!100}\else\ifdim#1pt>92.6pt\cellcolor{lime!90}\else\ifdim#1pt>92.5pt\cellcolor{lime!80}\else\ifdim#1pt>92.3pt\cellcolor{lime!70}\else\ifdim#1pt>92.2pt\cellcolor{lime!60}\else\ifdim#1pt>92.1pt\cellcolor{lime!50}\else\ifdim#1pt>91.9pt\cellcolor{lime!40}\else\ifdim#1pt>91.8pt\cellcolor{lime!30}\else\ifdim#1pt>91.6pt\cellcolor{lime!20}\else\ifdim#1pt>91.5pt\cellcolor{lime!10}\else\cellcolor{lime!0}\fi\fi\fi\fi\fi\fi\fi\fi\fi\fi#1}

\def\weightedspearmanzerozerosrcthresholdperadapterAR#1{\ifdim#1pt>95.7pt\cellcolor{lime!100}\else\ifdim#1pt>95.7pt\cellcolor{lime!90}\else\ifdim#1pt>95.7pt\cellcolor{lime!80}\else\ifdim#1pt>95.7pt\cellcolor{lime!70}\else\ifdim#1pt>95.7pt\cellcolor{lime!60}\else\ifdim#1pt>95.7pt\cellcolor{lime!50}\else\ifdim#1pt>95.7pt\cellcolor{lime!40}\else\ifdim#1pt>95.7pt\cellcolor{lime!30}\else\ifdim#1pt>95.7pt\cellcolor{lime!20}\else\ifdim#1pt>95.7pt\cellcolor{lime!10}\else\cellcolor{lime!0}\fi\fi\fi\fi\fi\fi\fi\fi\fi\fi#1}

\def\weightedspearmanzerozerosrcthresholdperadapterCA#1{\ifdim#1pt>73.4pt\cellcolor{lime!100}\else\ifdim#1pt>73.3pt\cellcolor{lime!90}\else\ifdim#1pt>73.1pt\cellcolor{lime!80}\else\ifdim#1pt>73.0pt\cellcolor{lime!70}\else\ifdim#1pt>72.9pt\cellcolor{lime!60}\else\ifdim#1pt>72.8pt\cellcolor{lime!50}\else\ifdim#1pt>72.7pt\cellcolor{lime!40}\else\ifdim#1pt>72.5pt\cellcolor{lime!30}\else\ifdim#1pt>72.4pt\cellcolor{lime!20}\else\ifdim#1pt>72.3pt\cellcolor{lime!10}\else\cellcolor{lime!0}\fi\fi\fi\fi\fi\fi\fi\fi\fi\fi#1}

\def\weightedspearmanzerozerosrcthresholdperadapterCP#1{\ifdim#1pt>83.1pt\cellcolor{lime!100}\else\ifdim#1pt>83.1pt\cellcolor{lime!90}\else\ifdim#1pt>83.1pt\cellcolor{lime!80}\else\ifdim#1pt>83.1pt\cellcolor{lime!70}\else\ifdim#1pt>83.1pt\cellcolor{lime!60}\else\ifdim#1pt>83.1pt\cellcolor{lime!50}\else\ifdim#1pt>83.1pt\cellcolor{lime!40}\else\ifdim#1pt>83.1pt\cellcolor{lime!30}\else\ifdim#1pt>83.1pt\cellcolor{lime!20}\else\ifdim#1pt>83.1pt\cellcolor{lime!10}\else\cellcolor{lime!0}\fi\fi\fi\fi\fi\fi\fi\fi\fi\fi#1}

\def\weightedspearmanzerozerosrcthresholdperadapterCR#1{\ifdim#1pt>85.2pt\cellcolor{lime!100}\else\ifdim#1pt>85.1pt\cellcolor{lime!90}\else\ifdim#1pt>85.0pt\cellcolor{lime!80}\else\ifdim#1pt>85.0pt\cellcolor{lime!70}\else\ifdim#1pt>85.0pt\cellcolor{lime!60}\else\ifdim#1pt>84.9pt\cellcolor{lime!50}\else\ifdim#1pt>84.9pt\cellcolor{lime!40}\else\ifdim#1pt>84.8pt\cellcolor{lime!30}\else\ifdim#1pt>84.8pt\cellcolor{lime!20}\else\ifdim#1pt>84.7pt\cellcolor{lime!10}\else\cellcolor{lime!0}\fi\fi\fi\fi\fi\fi\fi\fi\fi\fi#1}

\def\weightedspearmanzerozerosrcthresholdperadapterPA#1{\ifdim#1pt>86.5pt\cellcolor{lime!100}\else\ifdim#1pt>86.5pt\cellcolor{lime!90}\else\ifdim#1pt>86.4pt\cellcolor{lime!80}\else\ifdim#1pt>86.4pt\cellcolor{lime!70}\else\ifdim#1pt>86.3pt\cellcolor{lime!60}\else\ifdim#1pt>86.3pt\cellcolor{lime!50}\else\ifdim#1pt>86.2pt\cellcolor{lime!40}\else\ifdim#1pt>86.2pt\cellcolor{lime!30}\else\ifdim#1pt>86.1pt\cellcolor{lime!20}\else\ifdim#1pt>86.1pt\cellcolor{lime!10}\else\cellcolor{lime!0}\fi\fi\fi\fi\fi\fi\fi\fi\fi\fi#1}

\def\weightedspearmanzerozerosrcthresholdperadapterPC#1{\ifdim#1pt>91.4pt\cellcolor{lime!100}\else\ifdim#1pt>91.4pt\cellcolor{lime!90}\else\ifdim#1pt>91.4pt\cellcolor{lime!80}\else\ifdim#1pt>91.4pt\cellcolor{lime!70}\else\ifdim#1pt>91.4pt\cellcolor{lime!60}\else\ifdim#1pt>91.4pt\cellcolor{lime!50}\else\ifdim#1pt>91.4pt\cellcolor{lime!40}\else\ifdim#1pt>91.4pt\cellcolor{lime!30}\else\ifdim#1pt>91.4pt\cellcolor{lime!20}\else\ifdim#1pt>91.4pt\cellcolor{lime!10}\else\cellcolor{lime!0}\fi\fi\fi\fi\fi\fi\fi\fi\fi\fi#1}

\def\weightedspearmanzerozerosrcthresholdperadapterPR#1{\ifdim#1pt>94.8pt\cellcolor{lime!100}\else\ifdim#1pt>94.8pt\cellcolor{lime!90}\else\ifdim#1pt>94.8pt\cellcolor{lime!80}\else\ifdim#1pt>94.8pt\cellcolor{lime!70}\else\ifdim#1pt>94.8pt\cellcolor{lime!60}\else\ifdim#1pt>94.8pt\cellcolor{lime!50}\else\ifdim#1pt>94.8pt\cellcolor{lime!40}\else\ifdim#1pt>94.8pt\cellcolor{lime!30}\else\ifdim#1pt>94.8pt\cellcolor{lime!20}\else\ifdim#1pt>94.8pt\cellcolor{lime!10}\else\cellcolor{lime!0}\fi\fi\fi\fi\fi\fi\fi\fi\fi\fi#1}

\def\weightedspearmanzerozerosrcthresholdperadapterRA#1{\ifdim#1pt>93.0pt\cellcolor{lime!100}\else\ifdim#1pt>93.0pt\cellcolor{lime!90}\else\ifdim#1pt>93.0pt\cellcolor{lime!80}\else\ifdim#1pt>93.0pt\cellcolor{lime!70}\else\ifdim#1pt>93.0pt\cellcolor{lime!60}\else\ifdim#1pt>93.0pt\cellcolor{lime!50}\else\ifdim#1pt>93.0pt\cellcolor{lime!40}\else\ifdim#1pt>93.0pt\cellcolor{lime!30}\else\ifdim#1pt>93.0pt\cellcolor{lime!20}\else\ifdim#1pt>93.0pt\cellcolor{lime!10}\else\cellcolor{lime!0}\fi\fi\fi\fi\fi\fi\fi\fi\fi\fi#1}

\def\weightedspearmanzerozerosrcthresholdperadapterRC#1{\ifdim#1pt>90.8pt\cellcolor{lime!100}\else\ifdim#1pt>89.4pt\cellcolor{lime!90}\else\ifdim#1pt>88.0pt\cellcolor{lime!80}\else\ifdim#1pt>86.7pt\cellcolor{lime!70}\else\ifdim#1pt>85.3pt\cellcolor{lime!60}\else\ifdim#1pt>84.0pt\cellcolor{lime!50}\else\ifdim#1pt>82.6pt\cellcolor{lime!40}\else\ifdim#1pt>81.3pt\cellcolor{lime!30}\else\ifdim#1pt>79.9pt\cellcolor{lime!20}\else\ifdim#1pt>78.6pt\cellcolor{lime!10}\else\cellcolor{lime!0}\fi\fi\fi\fi\fi\fi\fi\fi\fi\fi#1}

\def\weightedspearmanzerozerosrcthresholdperadapterRP#1{\ifdim#1pt>95.2pt\cellcolor{lime!100}\else\ifdim#1pt>95.2pt\cellcolor{lime!90}\else\ifdim#1pt>95.2pt\cellcolor{lime!80}\else\ifdim#1pt>95.2pt\cellcolor{lime!70}\else\ifdim#1pt>95.2pt\cellcolor{lime!60}\else\ifdim#1pt>95.2pt\cellcolor{lime!50}\else\ifdim#1pt>95.2pt\cellcolor{lime!40}\else\ifdim#1pt>95.2pt\cellcolor{lime!30}\else\ifdim#1pt>95.2pt\cellcolor{lime!20}\else\ifdim#1pt>95.2pt\cellcolor{lime!10}\else\cellcolor{lime!0}\fi\fi\fi\fi\fi\fi\fi\fi\fi\fi#1}

\def\weightedspearmanzerozerosrcthresholdperadapterMean#1{\ifdim#1pt>86.0pt\cellcolor{lime!100}\else\ifdim#1pt>86.0pt\cellcolor{lime!90}\else\ifdim#1pt>86.0pt\cellcolor{lime!80}\else\ifdim#1pt>86.0pt\cellcolor{lime!70}\else\ifdim#1pt>86.0pt\cellcolor{lime!60}\else\ifdim#1pt>86.0pt\cellcolor{lime!50}\else\ifdim#1pt>86.0pt\cellcolor{lime!40}\else\ifdim#1pt>86.0pt\cellcolor{lime!30}\else\ifdim#1pt>86.0pt\cellcolor{lime!20}\else\ifdim#1pt>86.0pt\cellcolor{lime!10}\else\cellcolor{lime!0}\fi\fi\fi\fi\fi\fi\fi\fi\fi\fi#1}

\def\weightedspearmanzerozerosrcthresholdperadapterStd#1{\ifdim#1pt<2.9pt\cellcolor{lime!100}\else\ifdim#1pt<3.3pt\cellcolor{lime!90}\else\ifdim#1pt<3.7pt\cellcolor{lime!80}\else\ifdim#1pt<4.1pt\cellcolor{lime!70}\else\ifdim#1pt<4.5pt\cellcolor{lime!60}\else\ifdim#1pt<5.0pt\cellcolor{lime!50}\else\ifdim#1pt<5.4pt\cellcolor{lime!40}\else\ifdim#1pt<5.8pt\cellcolor{lime!30}\else\ifdim#1pt<6.2pt\cellcolor{lime!20}\else\ifdim#1pt<6.6pt\cellcolor{lime!10}\else\cellcolor{lime!0}\fi\fi\fi\fi\fi\fi\fi\fi\fi\fi#1}

\begin{table}[H]
\centering
\caption{The weighted Spearman correlation of each validator/task pair for \textbf{IM}.}
\label{office_weighted_spearman_0.0_src_threshold_per_adapter_IM}
\resizebox{!}{0.22\textheight}{
\begin{tabular}{llrrrrrrrrrrrrrrrrrrrr}
\toprule
 &  & AD & AW & DA & DW & WA & WD & AC & AP & AR & CA & CP & CR & PA & PC & PR & RA & RC & RP & Mean & Std \\
\midrule
\multirow[c]{2}{*}{Accuracy} & Source Train & \weightedspearmanzerozerosrcthresholdperadapterAD{54.7} & \weightedspearmanzerozerosrcthresholdperadapterAW{82.2} & \weightedspearmanzerozerosrcthresholdperadapterDA{83.4} & \weightedspearmanzerozerosrcthresholdperadapterDW{78.8} & \weightedspearmanzerozerosrcthresholdperadapterWA{84.8} & \weightedspearmanzerozerosrcthresholdperadapterWD{68.6} & \weightedspearmanzerozerosrcthresholdperadapterAC{64.7} & \weightedspearmanzerozerosrcthresholdperadapterAP{72.3} & \weightedspearmanzerozerosrcthresholdperadapterAR{65.0} & \weightedspearmanzerozerosrcthresholdperadapterCA{34.4} & \weightedspearmanzerozerosrcthresholdperadapterCP{36.7} & \weightedspearmanzerozerosrcthresholdperadapterCR{47.9} & \weightedspearmanzerozerosrcthresholdperadapterPA{60.4} & \weightedspearmanzerozerosrcthresholdperadapterPC{71.1} & \weightedspearmanzerozerosrcthresholdperadapterPR{72.2} & \weightedspearmanzerozerosrcthresholdperadapterRA{43.1} & \weightedspearmanzerozerosrcthresholdperadapterRC{69.9} & \weightedspearmanzerozerosrcthresholdperadapterRP{47.2} & \weightedspearmanzerozerosrcthresholdperadapterMean{63.2} & \weightedspearmanzerozerosrcthresholdperadapterStd{15.4} \\
\bottomrule
\end{tabular}}
\end{table}

%% file: supp_tables/office31_officehome/weighted_spearman_0.0_src_threshold_per_adapter_MCC.tex
\def\weightedspearmanzerozerosrcthresholdperadapterAD#1{\ifdim#1pt>78.3pt\cellcolor{lime!100}\else\ifdim#1pt>78.3pt\cellcolor{lime!90}\else\ifdim#1pt>78.3pt\cellcolor{lime!80}\else\ifdim#1pt>78.3pt\cellcolor{lime!70}\else\ifdim#1pt>78.3pt\cellcolor{lime!60}\else\ifdim#1pt>78.3pt\cellcolor{lime!50}\else\ifdim#1pt>78.3pt\cellcolor{lime!40}\else\ifdim#1pt>78.3pt\cellcolor{lime!30}\else\ifdim#1pt>78.3pt\cellcolor{lime!20}\else\ifdim#1pt>78.3pt\cellcolor{lime!10}\else\cellcolor{lime!0}\fi\fi\fi\fi\fi\fi\fi\fi\fi\fi#1}

\def\weightedspearmanzerozerosrcthresholdperadapterAW#1{\ifdim#1pt>84.7pt\cellcolor{lime!100}\else\ifdim#1pt>84.7pt\cellcolor{lime!90}\else\ifdim#1pt>84.7pt\cellcolor{lime!80}\else\ifdim#1pt>84.7pt\cellcolor{lime!70}\else\ifdim#1pt>84.7pt\cellcolor{lime!60}\else\ifdim#1pt>84.7pt\cellcolor{lime!50}\else\ifdim#1pt>84.7pt\cellcolor{lime!40}\else\ifdim#1pt>84.7pt\cellcolor{lime!30}\else\ifdim#1pt>84.7pt\cellcolor{lime!20}\else\ifdim#1pt>84.7pt\cellcolor{lime!10}\else\cellcolor{lime!0}\fi\fi\fi\fi\fi\fi\fi\fi\fi\fi#1}

\def\weightedspearmanzerozerosrcthresholdperadapterDA#1{\ifdim#1pt>84.4pt\cellcolor{lime!100}\else\ifdim#1pt>84.4pt\cellcolor{lime!90}\else\ifdim#1pt>84.4pt\cellcolor{lime!80}\else\ifdim#1pt>84.4pt\cellcolor{lime!70}\else\ifdim#1pt>84.3pt\cellcolor{lime!60}\else\ifdim#1pt>84.3pt\cellcolor{lime!50}\else\ifdim#1pt>84.3pt\cellcolor{lime!40}\else\ifdim#1pt>84.3pt\cellcolor{lime!30}\else\ifdim#1pt>84.3pt\cellcolor{lime!20}\else\ifdim#1pt>84.3pt\cellcolor{lime!10}\else\cellcolor{lime!0}\fi\fi\fi\fi\fi\fi\fi\fi\fi\fi#1}

\def\weightedspearmanzerozerosrcthresholdperadapterDW#1{\ifdim#1pt>86.4pt\cellcolor{lime!100}\else\ifdim#1pt>86.1pt\cellcolor{lime!90}\else\ifdim#1pt>85.8pt\cellcolor{lime!80}\else\ifdim#1pt>85.5pt\cellcolor{lime!70}\else\ifdim#1pt>85.2pt\cellcolor{lime!60}\else\ifdim#1pt>84.8pt\cellcolor{lime!50}\else\ifdim#1pt>84.5pt\cellcolor{lime!40}\else\ifdim#1pt>84.2pt\cellcolor{lime!30}\else\ifdim#1pt>83.9pt\cellcolor{lime!20}\else\ifdim#1pt>83.6pt\cellcolor{lime!10}\else\cellcolor{lime!0}\fi\fi\fi\fi\fi\fi\fi\fi\fi\fi#1}

\def\weightedspearmanzerozerosrcthresholdperadapterWA#1{\ifdim#1pt>87.6pt\cellcolor{lime!100}\else\ifdim#1pt>87.4pt\cellcolor{lime!90}\else\ifdim#1pt>87.3pt\cellcolor{lime!80}\else\ifdim#1pt>87.2pt\cellcolor{lime!70}\else\ifdim#1pt>87.1pt\cellcolor{lime!60}\else\ifdim#1pt>86.9pt\cellcolor{lime!50}\else\ifdim#1pt>86.8pt\cellcolor{lime!40}\else\ifdim#1pt>86.7pt\cellcolor{lime!30}\else\ifdim#1pt>86.5pt\cellcolor{lime!20}\else\ifdim#1pt>86.4pt\cellcolor{lime!10}\else\cellcolor{lime!0}\fi\fi\fi\fi\fi\fi\fi\fi\fi\fi#1}

\def\weightedspearmanzerozerosrcthresholdperadapterWD#1{\ifdim#1pt>84.4pt\cellcolor{lime!100}\else\ifdim#1pt>83.7pt\cellcolor{lime!90}\else\ifdim#1pt>83.0pt\cellcolor{lime!80}\else\ifdim#1pt>82.3pt\cellcolor{lime!70}\else\ifdim#1pt>81.5pt\cellcolor{lime!60}\else\ifdim#1pt>80.8pt\cellcolor{lime!50}\else\ifdim#1pt>80.1pt\cellcolor{lime!40}\else\ifdim#1pt>79.4pt\cellcolor{lime!30}\else\ifdim#1pt>78.7pt\cellcolor{lime!20}\else\ifdim#1pt>78.0pt\cellcolor{lime!10}\else\cellcolor{lime!0}\fi\fi\fi\fi\fi\fi\fi\fi\fi\fi#1}

\def\weightedspearmanzerozerosrcthresholdperadapterAC#1{\ifdim#1pt>89.5pt\cellcolor{lime!100}\else\ifdim#1pt>88.9pt\cellcolor{lime!90}\else\ifdim#1pt>88.2pt\cellcolor{lime!80}\else\ifdim#1pt>87.6pt\cellcolor{lime!70}\else\ifdim#1pt>87.0pt\cellcolor{lime!60}\else\ifdim#1pt>86.4pt\cellcolor{lime!50}\else\ifdim#1pt>85.8pt\cellcolor{lime!40}\else\ifdim#1pt>85.1pt\cellcolor{lime!30}\else\ifdim#1pt>84.5pt\cellcolor{lime!20}\else\ifdim#1pt>83.9pt\cellcolor{lime!10}\else\cellcolor{lime!0}\fi\fi\fi\fi\fi\fi\fi\fi\fi\fi#1}

\def\weightedspearmanzerozerosrcthresholdperadapterAP#1{\ifdim#1pt>89.7pt\cellcolor{lime!100}\else\ifdim#1pt>89.7pt\cellcolor{lime!90}\else\ifdim#1pt>89.7pt\cellcolor{lime!80}\else\ifdim#1pt>89.7pt\cellcolor{lime!70}\else\ifdim#1pt>89.7pt\cellcolor{lime!60}\else\ifdim#1pt>89.7pt\cellcolor{lime!50}\else\ifdim#1pt>89.7pt\cellcolor{lime!40}\else\ifdim#1pt>89.7pt\cellcolor{lime!30}\else\ifdim#1pt>89.7pt\cellcolor{lime!20}\else\ifdim#1pt>89.7pt\cellcolor{lime!10}\else\cellcolor{lime!0}\fi\fi\fi\fi\fi\fi\fi\fi\fi\fi#1}

\def\weightedspearmanzerozerosrcthresholdperadapterAR#1{\ifdim#1pt>94.8pt\cellcolor{lime!100}\else\ifdim#1pt>94.8pt\cellcolor{lime!90}\else\ifdim#1pt>94.8pt\cellcolor{lime!80}\else\ifdim#1pt>94.8pt\cellcolor{lime!70}\else\ifdim#1pt>94.8pt\cellcolor{lime!60}\else\ifdim#1pt>94.8pt\cellcolor{lime!50}\else\ifdim#1pt>94.8pt\cellcolor{lime!40}\else\ifdim#1pt>94.8pt\cellcolor{lime!30}\else\ifdim#1pt>94.8pt\cellcolor{lime!20}\else\ifdim#1pt>94.8pt\cellcolor{lime!10}\else\cellcolor{lime!0}\fi\fi\fi\fi\fi\fi\fi\fi\fi\fi#1}

\def\weightedspearmanzerozerosrcthresholdperadapterCA#1{\ifdim#1pt>84.5pt\cellcolor{lime!100}\else\ifdim#1pt>84.5pt\cellcolor{lime!90}\else\ifdim#1pt>84.5pt\cellcolor{lime!80}\else\ifdim#1pt>84.5pt\cellcolor{lime!70}\else\ifdim#1pt>84.5pt\cellcolor{lime!60}\else\ifdim#1pt>84.5pt\cellcolor{lime!50}\else\ifdim#1pt>84.5pt\cellcolor{lime!40}\else\ifdim#1pt>84.5pt\cellcolor{lime!30}\else\ifdim#1pt>84.5pt\cellcolor{lime!20}\else\ifdim#1pt>84.5pt\cellcolor{lime!10}\else\cellcolor{lime!0}\fi\fi\fi\fi\fi\fi\fi\fi\fi\fi#1}

\def\weightedspearmanzerozerosrcthresholdperadapterCP#1{\ifdim#1pt>89.9pt\cellcolor{lime!100}\else\ifdim#1pt>89.9pt\cellcolor{lime!90}\else\ifdim#1pt>89.9pt\cellcolor{lime!80}\else\ifdim#1pt>89.9pt\cellcolor{lime!70}\else\ifdim#1pt>89.9pt\cellcolor{lime!60}\else\ifdim#1pt>89.9pt\cellcolor{lime!50}\else\ifdim#1pt>89.9pt\cellcolor{lime!40}\else\ifdim#1pt>89.9pt\cellcolor{lime!30}\else\ifdim#1pt>89.9pt\cellcolor{lime!20}\else\ifdim#1pt>89.9pt\cellcolor{lime!10}\else\cellcolor{lime!0}\fi\fi\fi\fi\fi\fi\fi\fi\fi\fi#1}

\def\weightedspearmanzerozerosrcthresholdperadapterCR#1{\ifdim#1pt>87.6pt\cellcolor{lime!100}\else\ifdim#1pt>87.6pt\cellcolor{lime!90}\else\ifdim#1pt>87.6pt\cellcolor{lime!80}\else\ifdim#1pt>87.6pt\cellcolor{lime!70}\else\ifdim#1pt>87.6pt\cellcolor{lime!60}\else\ifdim#1pt>87.6pt\cellcolor{lime!50}\else\ifdim#1pt>87.6pt\cellcolor{lime!40}\else\ifdim#1pt>87.6pt\cellcolor{lime!30}\else\ifdim#1pt>87.6pt\cellcolor{lime!20}\else\ifdim#1pt>87.6pt\cellcolor{lime!10}\else\cellcolor{lime!0}\fi\fi\fi\fi\fi\fi\fi\fi\fi\fi#1}

\def\weightedspearmanzerozerosrcthresholdperadapterPA#1{\ifdim#1pt>91.4pt\cellcolor{lime!100}\else\ifdim#1pt>91.4pt\cellcolor{lime!90}\else\ifdim#1pt>91.4pt\cellcolor{lime!80}\else\ifdim#1pt>91.4pt\cellcolor{lime!70}\else\ifdim#1pt>91.4pt\cellcolor{lime!60}\else\ifdim#1pt>91.4pt\cellcolor{lime!50}\else\ifdim#1pt>91.4pt\cellcolor{lime!40}\else\ifdim#1pt>91.4pt\cellcolor{lime!30}\else\ifdim#1pt>91.4pt\cellcolor{lime!20}\else\ifdim#1pt>91.4pt\cellcolor{lime!10}\else\cellcolor{lime!0}\fi\fi\fi\fi\fi\fi\fi\fi\fi\fi#1}

\def\weightedspearmanzerozerosrcthresholdperadapterPC#1{\ifdim#1pt>89.3pt\cellcolor{lime!100}\else\ifdim#1pt>89.3pt\cellcolor{lime!90}\else\ifdim#1pt>89.3pt\cellcolor{lime!80}\else\ifdim#1pt>89.3pt\cellcolor{lime!70}\else\ifdim#1pt>89.3pt\cellcolor{lime!60}\else\ifdim#1pt>89.3pt\cellcolor{lime!50}\else\ifdim#1pt>89.3pt\cellcolor{lime!40}\else\ifdim#1pt>89.3pt\cellcolor{lime!30}\else\ifdim#1pt>89.3pt\cellcolor{lime!20}\else\ifdim#1pt>89.3pt\cellcolor{lime!10}\else\cellcolor{lime!0}\fi\fi\fi\fi\fi\fi\fi\fi\fi\fi#1}

\def\weightedspearmanzerozerosrcthresholdperadapterPR#1{\ifdim#1pt>94.8pt\cellcolor{lime!100}\else\ifdim#1pt>94.8pt\cellcolor{lime!90}\else\ifdim#1pt>94.8pt\cellcolor{lime!80}\else\ifdim#1pt>94.8pt\cellcolor{lime!70}\else\ifdim#1pt>94.8pt\cellcolor{lime!60}\else\ifdim#1pt>94.8pt\cellcolor{lime!50}\else\ifdim#1pt>94.8pt\cellcolor{lime!40}\else\ifdim#1pt>94.8pt\cellcolor{lime!30}\else\ifdim#1pt>94.8pt\cellcolor{lime!20}\else\ifdim#1pt>94.8pt\cellcolor{lime!10}\else\cellcolor{lime!0}\fi\fi\fi\fi\fi\fi\fi\fi\fi\fi#1}

\def\weightedspearmanzerozerosrcthresholdperadapterRA#1{\ifdim#1pt>86.7pt\cellcolor{lime!100}\else\ifdim#1pt>86.7pt\cellcolor{lime!90}\else\ifdim#1pt>86.7pt\cellcolor{lime!80}\else\ifdim#1pt>86.7pt\cellcolor{lime!70}\else\ifdim#1pt>86.7pt\cellcolor{lime!60}\else\ifdim#1pt>86.7pt\cellcolor{lime!50}\else\ifdim#1pt>86.7pt\cellcolor{lime!40}\else\ifdim#1pt>86.7pt\cellcolor{lime!30}\else\ifdim#1pt>86.7pt\cellcolor{lime!20}\else\ifdim#1pt>86.7pt\cellcolor{lime!10}\else\cellcolor{lime!0}\fi\fi\fi\fi\fi\fi\fi\fi\fi\fi#1}

\def\weightedspearmanzerozerosrcthresholdperadapterRC#1{\ifdim#1pt>91.6pt\cellcolor{lime!100}\else\ifdim#1pt>90.5pt\cellcolor{lime!90}\else\ifdim#1pt>89.3pt\cellcolor{lime!80}\else\ifdim#1pt>88.2pt\cellcolor{lime!70}\else\ifdim#1pt>87.0pt\cellcolor{lime!60}\else\ifdim#1pt>85.9pt\cellcolor{lime!50}\else\ifdim#1pt>84.8pt\cellcolor{lime!40}\else\ifdim#1pt>83.6pt\cellcolor{lime!30}\else\ifdim#1pt>82.5pt\cellcolor{lime!20}\else\ifdim#1pt>81.3pt\cellcolor{lime!10}\else\cellcolor{lime!0}\fi\fi\fi\fi\fi\fi\fi\fi\fi\fi#1}

\def\weightedspearmanzerozerosrcthresholdperadapterRP#1{\ifdim#1pt>91.8pt\cellcolor{lime!100}\else\ifdim#1pt>91.8pt\cellcolor{lime!90}\else\ifdim#1pt>91.8pt\cellcolor{lime!80}\else\ifdim#1pt>91.8pt\cellcolor{lime!70}\else\ifdim#1pt>91.8pt\cellcolor{lime!60}\else\ifdim#1pt>91.8pt\cellcolor{lime!50}\else\ifdim#1pt>91.8pt\cellcolor{lime!40}\else\ifdim#1pt>91.8pt\cellcolor{lime!30}\else\ifdim#1pt>91.8pt\cellcolor{lime!20}\else\ifdim#1pt>91.8pt\cellcolor{lime!10}\else\cellcolor{lime!0}\fi\fi\fi\fi\fi\fi\fi\fi\fi\fi#1}

\def\weightedspearmanzerozerosrcthresholdperadapterMean#1{\ifdim#1pt>86.7pt\cellcolor{lime!100}\else\ifdim#1pt>86.7pt\cellcolor{lime!90}\else\ifdim#1pt>86.7pt\cellcolor{lime!80}\else\ifdim#1pt>86.7pt\cellcolor{lime!70}\else\ifdim#1pt>86.7pt\cellcolor{lime!60}\else\ifdim#1pt>86.7pt\cellcolor{lime!50}\else\ifdim#1pt>86.7pt\cellcolor{lime!40}\else\ifdim#1pt>86.7pt\cellcolor{lime!30}\else\ifdim#1pt>86.7pt\cellcolor{lime!20}\else\ifdim#1pt>86.7pt\cellcolor{lime!10}\else\cellcolor{lime!0}\fi\fi\fi\fi\fi\fi\fi\fi\fi\fi#1}

\def\weightedspearmanzerozerosrcthresholdperadapterStd#1{\ifdim#1pt<4.8pt\cellcolor{lime!100}\else\ifdim#1pt<4.8pt\cellcolor{lime!90}\else\ifdim#1pt<4.8pt\cellcolor{lime!80}\else\ifdim#1pt<4.8pt\cellcolor{lime!70}\else\ifdim#1pt<4.8pt\cellcolor{lime!60}\else\ifdim#1pt<4.8pt\cellcolor{lime!50}\else\ifdim#1pt<4.8pt\cellcolor{lime!40}\else\ifdim#1pt<4.8pt\cellcolor{lime!30}\else\ifdim#1pt<4.8pt\cellcolor{lime!20}\else\ifdim#1pt<4.8pt\cellcolor{lime!10}\else\cellcolor{lime!0}\fi\fi\fi\fi\fi\fi\fi\fi\fi\fi#1}

\begin{table}[H]
\centering
\caption{The weighted Spearman correlation of each validator/task pair for \textbf{MCC}.}
\label{office_weighted_spearman_0.0_src_threshold_per_adapter_MCC}
\resizebox{!}{0.22\textheight}{
\begin{tabular}{llrrrrrrrrrrrrrrrrrrrr}
\toprule
 &  & AD & AW & DA & DW & WA & WD & AC & AP & AR & CA & CP & CR & PA & PC & PR & RA & RC & RP & Mean & Std \\
\midrule
\multirow[c]{2}{*}{Accuracy} & Source Train & \weightedspearmanzerozerosrcthresholdperadapterAD{74.8} & \weightedspearmanzerozerosrcthresholdperadapterAW{76.7} & \textbf{\weightedspearmanzerozerosrcthresholdperadapterDA{84.4}} & \weightedspearmanzerozerosrcthresholdperadapterDW{75.3} & \textbf{\weightedspearmanzerozerosrcthresholdperadapterWA{87.7}} & \weightedspearmanzerozerosrcthresholdperadapterWD{70.4} & \weightedspearmanzerozerosrcthresholdperadapterAC{55.9} & \weightedspearmanzerozerosrcthresholdperadapterAP{65.7} & \weightedspearmanzerozerosrcthresholdperadapterAR{56.3} & \weightedspearmanzerozerosrcthresholdperadapterCA{51.7} & \weightedspearmanzerozerosrcthresholdperadapterCP{61.9} & \weightedspearmanzerozerosrcthresholdperadapterCR{55.1} & \weightedspearmanzerozerosrcthresholdperadapterPA{70.4} & \weightedspearmanzerozerosrcthresholdperadapterPC{63.6} & \weightedspearmanzerozerosrcthresholdperadapterPR{72.1} & \weightedspearmanzerozerosrcthresholdperadapterRA{57.7} & \weightedspearmanzerozerosrcthresholdperadapterRC{71.3} & \weightedspearmanzerozerosrcthresholdperadapterRP{61.2} & \weightedspearmanzerozerosrcthresholdperadapterMean{67.3} & \weightedspearmanzerozerosrcthresholdperadapterStd{10.0} \\
\bottomrule
\end{tabular}}
\end{table}

%% file: supp_tables/office31_officehome/weighted_spearman_0.0_src_threshold_per_adapter_MCD.tex
\def\weightedspearmanzerozerosrcthresholdperadapterAD#1{\ifdim#1pt>84.4pt\cellcolor{lime!100}\else\ifdim#1pt>84.4pt\cellcolor{lime!90}\else\ifdim#1pt>84.4pt\cellcolor{lime!80}\else\ifdim#1pt>84.4pt\cellcolor{lime!70}\else\ifdim#1pt>84.4pt\cellcolor{lime!60}\else\ifdim#1pt>84.4pt\cellcolor{lime!50}\else\ifdim#1pt>84.4pt\cellcolor{lime!40}\else\ifdim#1pt>84.4pt\cellcolor{lime!30}\else\ifdim#1pt>84.4pt\cellcolor{lime!20}\else\ifdim#1pt>84.4pt\cellcolor{lime!10}\else\cellcolor{lime!0}\fi\fi\fi\fi\fi\fi\fi\fi\fi\fi#1}

\def\weightedspearmanzerozerosrcthresholdperadapterAW#1{\ifdim#1pt>86.7pt\cellcolor{lime!100}\else\ifdim#1pt>86.6pt\cellcolor{lime!90}\else\ifdim#1pt>86.5pt\cellcolor{lime!80}\else\ifdim#1pt>86.5pt\cellcolor{lime!70}\else\ifdim#1pt>86.5pt\cellcolor{lime!60}\else\ifdim#1pt>86.4pt\cellcolor{lime!50}\else\ifdim#1pt>86.4pt\cellcolor{lime!40}\else\ifdim#1pt>86.3pt\cellcolor{lime!30}\else\ifdim#1pt>86.2pt\cellcolor{lime!20}\else\ifdim#1pt>86.2pt\cellcolor{lime!10}\else\cellcolor{lime!0}\fi\fi\fi\fi\fi\fi\fi\fi\fi\fi#1}

\def\weightedspearmanzerozerosrcthresholdperadapterDA#1{\ifdim#1pt>88.5pt\cellcolor{lime!100}\else\ifdim#1pt>88.3pt\cellcolor{lime!90}\else\ifdim#1pt>88.2pt\cellcolor{lime!80}\else\ifdim#1pt>88.1pt\cellcolor{lime!70}\else\ifdim#1pt>87.9pt\cellcolor{lime!60}\else\ifdim#1pt>87.8pt\cellcolor{lime!50}\else\ifdim#1pt>87.7pt\cellcolor{lime!40}\else\ifdim#1pt>87.6pt\cellcolor{lime!30}\else\ifdim#1pt>87.4pt\cellcolor{lime!20}\else\ifdim#1pt>87.3pt\cellcolor{lime!10}\else\cellcolor{lime!0}\fi\fi\fi\fi\fi\fi\fi\fi\fi\fi#1}

\def\weightedspearmanzerozerosrcthresholdperadapterDW#1{\ifdim#1pt>82.5pt\cellcolor{lime!100}\else\ifdim#1pt>81.0pt\cellcolor{lime!90}\else\ifdim#1pt>79.5pt\cellcolor{lime!80}\else\ifdim#1pt>78.0pt\cellcolor{lime!70}\else\ifdim#1pt>76.5pt\cellcolor{lime!60}\else\ifdim#1pt>75.1pt\cellcolor{lime!50}\else\ifdim#1pt>73.6pt\cellcolor{lime!40}\else\ifdim#1pt>72.1pt\cellcolor{lime!30}\else\ifdim#1pt>70.6pt\cellcolor{lime!20}\else\ifdim#1pt>69.1pt\cellcolor{lime!10}\else\cellcolor{lime!0}\fi\fi\fi\fi\fi\fi\fi\fi\fi\fi#1}

\def\weightedspearmanzerozerosrcthresholdperadapterWA#1{\ifdim#1pt>85.4pt\cellcolor{lime!100}\else\ifdim#1pt>85.4pt\cellcolor{lime!90}\else\ifdim#1pt>85.4pt\cellcolor{lime!80}\else\ifdim#1pt>85.4pt\cellcolor{lime!70}\else\ifdim#1pt>85.4pt\cellcolor{lime!60}\else\ifdim#1pt>85.4pt\cellcolor{lime!50}\else\ifdim#1pt>85.4pt\cellcolor{lime!40}\else\ifdim#1pt>85.4pt\cellcolor{lime!30}\else\ifdim#1pt>85.4pt\cellcolor{lime!20}\else\ifdim#1pt>85.4pt\cellcolor{lime!10}\else\cellcolor{lime!0}\fi\fi\fi\fi\fi\fi\fi\fi\fi\fi#1}

\def\weightedspearmanzerozerosrcthresholdperadapterWD#1{\ifdim#1pt>88.7pt\cellcolor{lime!100}\else\ifdim#1pt>87.8pt\cellcolor{lime!90}\else\ifdim#1pt>86.9pt\cellcolor{lime!80}\else\ifdim#1pt>86.0pt\cellcolor{lime!70}\else\ifdim#1pt>85.0pt\cellcolor{lime!60}\else\ifdim#1pt>84.1pt\cellcolor{lime!50}\else\ifdim#1pt>83.2pt\cellcolor{lime!40}\else\ifdim#1pt>82.3pt\cellcolor{lime!30}\else\ifdim#1pt>81.4pt\cellcolor{lime!20}\else\ifdim#1pt>80.5pt\cellcolor{lime!10}\else\cellcolor{lime!0}\fi\fi\fi\fi\fi\fi\fi\fi\fi\fi#1}

\def\weightedspearmanzerozerosrcthresholdperadapterAC#1{\ifdim#1pt>97.3pt\cellcolor{lime!100}\else\ifdim#1pt>97.2pt\cellcolor{lime!90}\else\ifdim#1pt>97.2pt\cellcolor{lime!80}\else\ifdim#1pt>97.2pt\cellcolor{lime!70}\else\ifdim#1pt>97.2pt\cellcolor{lime!60}\else\ifdim#1pt>97.1pt\cellcolor{lime!50}\else\ifdim#1pt>97.1pt\cellcolor{lime!40}\else\ifdim#1pt>97.1pt\cellcolor{lime!30}\else\ifdim#1pt>97.0pt\cellcolor{lime!20}\else\ifdim#1pt>97.0pt\cellcolor{lime!10}\else\cellcolor{lime!0}\fi\fi\fi\fi\fi\fi\fi\fi\fi\fi#1}

\def\weightedspearmanzerozerosrcthresholdperadapterAP#1{\ifdim#1pt>98.9pt\cellcolor{lime!100}\else\ifdim#1pt>98.9pt\cellcolor{lime!90}\else\ifdim#1pt>98.9pt\cellcolor{lime!80}\else\ifdim#1pt>98.9pt\cellcolor{lime!70}\else\ifdim#1pt>98.9pt\cellcolor{lime!60}\else\ifdim#1pt>98.9pt\cellcolor{lime!50}\else\ifdim#1pt>98.9pt\cellcolor{lime!40}\else\ifdim#1pt>98.9pt\cellcolor{lime!30}\else\ifdim#1pt>98.9pt\cellcolor{lime!20}\else\ifdim#1pt>98.9pt\cellcolor{lime!10}\else\cellcolor{lime!0}\fi\fi\fi\fi\fi\fi\fi\fi\fi\fi#1}

\def\weightedspearmanzerozerosrcthresholdperadapterAR#1{\ifdim#1pt>98.8pt\cellcolor{lime!100}\else\ifdim#1pt>98.8pt\cellcolor{lime!90}\else\ifdim#1pt>98.8pt\cellcolor{lime!80}\else\ifdim#1pt>98.8pt\cellcolor{lime!70}\else\ifdim#1pt>98.8pt\cellcolor{lime!60}\else\ifdim#1pt>98.8pt\cellcolor{lime!50}\else\ifdim#1pt>98.8pt\cellcolor{lime!40}\else\ifdim#1pt>98.8pt\cellcolor{lime!30}\else\ifdim#1pt>98.8pt\cellcolor{lime!20}\else\ifdim#1pt>98.8pt\cellcolor{lime!10}\else\cellcolor{lime!0}\fi\fi\fi\fi\fi\fi\fi\fi\fi\fi#1}

\def\weightedspearmanzerozerosrcthresholdperadapterCA#1{\ifdim#1pt>90.5pt\cellcolor{lime!100}\else\ifdim#1pt>90.0pt\cellcolor{lime!90}\else\ifdim#1pt>89.6pt\cellcolor{lime!80}\else\ifdim#1pt>89.2pt\cellcolor{lime!70}\else\ifdim#1pt>88.8pt\cellcolor{lime!60}\else\ifdim#1pt>88.3pt\cellcolor{lime!50}\else\ifdim#1pt>87.9pt\cellcolor{lime!40}\else\ifdim#1pt>87.5pt\cellcolor{lime!30}\else\ifdim#1pt>87.0pt\cellcolor{lime!20}\else\ifdim#1pt>86.6pt\cellcolor{lime!10}\else\cellcolor{lime!0}\fi\fi\fi\fi\fi\fi\fi\fi\fi\fi#1}

\def\weightedspearmanzerozerosrcthresholdperadapterCP#1{\ifdim#1pt>87.7pt\cellcolor{lime!100}\else\ifdim#1pt>87.7pt\cellcolor{lime!90}\else\ifdim#1pt>87.7pt\cellcolor{lime!80}\else\ifdim#1pt>87.7pt\cellcolor{lime!70}\else\ifdim#1pt>87.7pt\cellcolor{lime!60}\else\ifdim#1pt>87.7pt\cellcolor{lime!50}\else\ifdim#1pt>87.7pt\cellcolor{lime!40}\else\ifdim#1pt>87.7pt\cellcolor{lime!30}\else\ifdim#1pt>87.7pt\cellcolor{lime!20}\else\ifdim#1pt>87.7pt\cellcolor{lime!10}\else\cellcolor{lime!0}\fi\fi\fi\fi\fi\fi\fi\fi\fi\fi#1}

\def\weightedspearmanzerozerosrcthresholdperadapterCR#1{\ifdim#1pt>86.5pt\cellcolor{lime!100}\else\ifdim#1pt>86.5pt\cellcolor{lime!90}\else\ifdim#1pt>86.5pt\cellcolor{lime!80}\else\ifdim#1pt>86.5pt\cellcolor{lime!70}\else\ifdim#1pt>86.5pt\cellcolor{lime!60}\else\ifdim#1pt>86.5pt\cellcolor{lime!50}\else\ifdim#1pt>86.5pt\cellcolor{lime!40}\else\ifdim#1pt>86.5pt\cellcolor{lime!30}\else\ifdim#1pt>86.5pt\cellcolor{lime!20}\else\ifdim#1pt>86.5pt\cellcolor{lime!10}\else\cellcolor{lime!0}\fi\fi\fi\fi\fi\fi\fi\fi\fi\fi#1}

\def\weightedspearmanzerozerosrcthresholdperadapterPA#1{\ifdim#1pt>96.6pt\cellcolor{lime!100}\else\ifdim#1pt>96.6pt\cellcolor{lime!90}\else\ifdim#1pt>96.5pt\cellcolor{lime!80}\else\ifdim#1pt>96.4pt\cellcolor{lime!70}\else\ifdim#1pt>96.3pt\cellcolor{lime!60}\else\ifdim#1pt>96.3pt\cellcolor{lime!50}\else\ifdim#1pt>96.2pt\cellcolor{lime!40}\else\ifdim#1pt>96.1pt\cellcolor{lime!30}\else\ifdim#1pt>96.1pt\cellcolor{lime!20}\else\ifdim#1pt>96.0pt\cellcolor{lime!10}\else\cellcolor{lime!0}\fi\fi\fi\fi\fi\fi\fi\fi\fi\fi#1}

\def\weightedspearmanzerozerosrcthresholdperadapterPC#1{\ifdim#1pt>94.9pt\cellcolor{lime!100}\else\ifdim#1pt>94.9pt\cellcolor{lime!90}\else\ifdim#1pt>94.9pt\cellcolor{lime!80}\else\ifdim#1pt>94.9pt\cellcolor{lime!70}\else\ifdim#1pt>94.9pt\cellcolor{lime!60}\else\ifdim#1pt>94.9pt\cellcolor{lime!50}\else\ifdim#1pt>94.9pt\cellcolor{lime!40}\else\ifdim#1pt>94.9pt\cellcolor{lime!30}\else\ifdim#1pt>94.9pt\cellcolor{lime!20}\else\ifdim#1pt>94.9pt\cellcolor{lime!10}\else\cellcolor{lime!0}\fi\fi\fi\fi\fi\fi\fi\fi\fi\fi#1}

\def\weightedspearmanzerozerosrcthresholdperadapterPR#1{\ifdim#1pt>94.1pt\cellcolor{lime!100}\else\ifdim#1pt>94.1pt\cellcolor{lime!90}\else\ifdim#1pt>94.1pt\cellcolor{lime!80}\else\ifdim#1pt>94.1pt\cellcolor{lime!70}\else\ifdim#1pt>94.1pt\cellcolor{lime!60}\else\ifdim#1pt>94.1pt\cellcolor{lime!50}\else\ifdim#1pt>94.1pt\cellcolor{lime!40}\else\ifdim#1pt>94.1pt\cellcolor{lime!30}\else\ifdim#1pt>94.1pt\cellcolor{lime!20}\else\ifdim#1pt>94.1pt\cellcolor{lime!10}\else\cellcolor{lime!0}\fi\fi\fi\fi\fi\fi\fi\fi\fi\fi#1}

\def\weightedspearmanzerozerosrcthresholdperadapterRA#1{\ifdim#1pt>95.5pt\cellcolor{lime!100}\else\ifdim#1pt>95.5pt\cellcolor{lime!90}\else\ifdim#1pt>95.5pt\cellcolor{lime!80}\else\ifdim#1pt>95.5pt\cellcolor{lime!70}\else\ifdim#1pt>95.5pt\cellcolor{lime!60}\else\ifdim#1pt>95.5pt\cellcolor{lime!50}\else\ifdim#1pt>95.5pt\cellcolor{lime!40}\else\ifdim#1pt>95.5pt\cellcolor{lime!30}\else\ifdim#1pt>95.5pt\cellcolor{lime!20}\else\ifdim#1pt>95.5pt\cellcolor{lime!10}\else\cellcolor{lime!0}\fi\fi\fi\fi\fi\fi\fi\fi\fi\fi#1}

\def\weightedspearmanzerozerosrcthresholdperadapterRC#1{\ifdim#1pt>94.2pt\cellcolor{lime!100}\else\ifdim#1pt>94.2pt\cellcolor{lime!90}\else\ifdim#1pt>94.1pt\cellcolor{lime!80}\else\ifdim#1pt>94.1pt\cellcolor{lime!70}\else\ifdim#1pt>94.0pt\cellcolor{lime!60}\else\ifdim#1pt>94.0pt\cellcolor{lime!50}\else\ifdim#1pt>94.0pt\cellcolor{lime!40}\else\ifdim#1pt>93.9pt\cellcolor{lime!30}\else\ifdim#1pt>93.8pt\cellcolor{lime!20}\else\ifdim#1pt>93.8pt\cellcolor{lime!10}\else\cellcolor{lime!0}\fi\fi\fi\fi\fi\fi\fi\fi\fi\fi#1}

\def\weightedspearmanzerozerosrcthresholdperadapterRP#1{\ifdim#1pt>95.5pt\cellcolor{lime!100}\else\ifdim#1pt>95.5pt\cellcolor{lime!90}\else\ifdim#1pt>95.5pt\cellcolor{lime!80}\else\ifdim#1pt>95.5pt\cellcolor{lime!70}\else\ifdim#1pt>95.5pt\cellcolor{lime!60}\else\ifdim#1pt>95.5pt\cellcolor{lime!50}\else\ifdim#1pt>95.5pt\cellcolor{lime!40}\else\ifdim#1pt>95.5pt\cellcolor{lime!30}\else\ifdim#1pt>95.5pt\cellcolor{lime!20}\else\ifdim#1pt>95.5pt\cellcolor{lime!10}\else\cellcolor{lime!0}\fi\fi\fi\fi\fi\fi\fi\fi\fi\fi#1}

\def\weightedspearmanzerozerosrcthresholdperadapterMean#1{\ifdim#1pt>89.9pt\cellcolor{lime!100}\else\ifdim#1pt>89.9pt\cellcolor{lime!90}\else\ifdim#1pt>89.9pt\cellcolor{lime!80}\else\ifdim#1pt>89.9pt\cellcolor{lime!70}\else\ifdim#1pt>89.9pt\cellcolor{lime!60}\else\ifdim#1pt>89.9pt\cellcolor{lime!50}\else\ifdim#1pt>89.9pt\cellcolor{lime!40}\else\ifdim#1pt>89.9pt\cellcolor{lime!30}\else\ifdim#1pt>89.9pt\cellcolor{lime!20}\else\ifdim#1pt>89.9pt\cellcolor{lime!10}\else\cellcolor{lime!0}\fi\fi\fi\fi\fi\fi\fi\fi\fi\fi#1}

\def\weightedspearmanzerozerosrcthresholdperadapterStd#1{\ifdim#1pt<4.0pt\cellcolor{lime!100}\else\ifdim#1pt<4.4pt\cellcolor{lime!90}\else\ifdim#1pt<4.7pt\cellcolor{lime!80}\else\ifdim#1pt<5.1pt\cellcolor{lime!70}\else\ifdim#1pt<5.5pt\cellcolor{lime!60}\else\ifdim#1pt<5.9pt\cellcolor{lime!50}\else\ifdim#1pt<6.3pt\cellcolor{lime!40}\else\ifdim#1pt<6.6pt\cellcolor{lime!30}\else\ifdim#1pt<7.0pt\cellcolor{lime!20}\else\ifdim#1pt<7.4pt\cellcolor{lime!10}\else\cellcolor{lime!0}\fi\fi\fi\fi\fi\fi\fi\fi\fi\fi#1}

\begin{table}[H]
\centering
\caption{The weighted Spearman correlation of each validator/task pair for \textbf{MCD}.}
\label{office_weighted_spearman_0.0_src_threshold_per_adapter_MCD}
\resizebox{!}{0.22\textheight}{
\begin{tabular}{llrrrrrrrrrrrrrrrrrrrr}
\toprule
 &  & AD & AW & DA & DW & WA & WD & AC & AP & AR & CA & CP & CR & PA & PC & PR & RA & RC & RP & Mean & Std \\
\midrule
\multirow[c]{2}{*}{Accuracy} & Source Train & \weightedspearmanzerozerosrcthresholdperadapterAD{79.3} & \weightedspearmanzerozerosrcthresholdperadapterAW{77.0} & \weightedspearmanzerozerosrcthresholdperadapterDA{75.5} & \weightedspearmanzerozerosrcthresholdperadapterDW{73.0} & \weightedspearmanzerozerosrcthresholdperadapterWA{83.4} & \weightedspearmanzerozerosrcthresholdperadapterWD{76.3} & \weightedspearmanzerozerosrcthresholdperadapterAC{79.8} & \weightedspearmanzerozerosrcthresholdperadapterAP{74.2} & \weightedspearmanzerozerosrcthresholdperadapterAR{51.9} & \weightedspearmanzerozerosrcthresholdperadapterCA{65.5} & \weightedspearmanzerozerosrcthresholdperadapterCP{54.6} & \weightedspearmanzerozerosrcthresholdperadapterCR{52.0} & \weightedspearmanzerozerosrcthresholdperadapterPA{79.1} & \weightedspearmanzerozerosrcthresholdperadapterPC{74.1} & \weightedspearmanzerozerosrcthresholdperadapterPR{67.4} & \weightedspearmanzerozerosrcthresholdperadapterRA{63.6} & \weightedspearmanzerozerosrcthresholdperadapterRC{82.2} & \weightedspearmanzerozerosrcthresholdperadapterRP{66.4} & \weightedspearmanzerozerosrcthresholdperadapterMean{70.9} & \weightedspearmanzerozerosrcthresholdperadapterStd{9.7} \\
\bottomrule
\end{tabular}}
\end{table}

%% file: supp_tables/office31_officehome/weighted_spearman_0.0_src_threshold_per_adapter_MMD.tex
\def\weightedspearmanzerozerosrcthresholdperadapterAD#1{\ifdim#1pt>81.0pt\cellcolor{lime!100}\else\ifdim#1pt>81.0pt\cellcolor{lime!90}\else\ifdim#1pt>81.0pt\cellcolor{lime!80}\else\ifdim#1pt>81.0pt\cellcolor{lime!70}\else\ifdim#1pt>81.0pt\cellcolor{lime!60}\else\ifdim#1pt>80.9pt\cellcolor{lime!50}\else\ifdim#1pt>80.9pt\cellcolor{lime!40}\else\ifdim#1pt>80.9pt\cellcolor{lime!30}\else\ifdim#1pt>80.9pt\cellcolor{lime!20}\else\ifdim#1pt>80.9pt\cellcolor{lime!10}\else\cellcolor{lime!0}\fi\fi\fi\fi\fi\fi\fi\fi\fi\fi#1}

\def\weightedspearmanzerozerosrcthresholdperadapterAW#1{\ifdim#1pt>88.0pt\cellcolor{lime!100}\else\ifdim#1pt>88.0pt\cellcolor{lime!90}\else\ifdim#1pt>88.0pt\cellcolor{lime!80}\else\ifdim#1pt>88.0pt\cellcolor{lime!70}\else\ifdim#1pt>88.0pt\cellcolor{lime!60}\else\ifdim#1pt>88.0pt\cellcolor{lime!50}\else\ifdim#1pt>88.0pt\cellcolor{lime!40}\else\ifdim#1pt>88.0pt\cellcolor{lime!30}\else\ifdim#1pt>88.0pt\cellcolor{lime!20}\else\ifdim#1pt>88.0pt\cellcolor{lime!10}\else\cellcolor{lime!0}\fi\fi\fi\fi\fi\fi\fi\fi\fi\fi#1}

\def\weightedspearmanzerozerosrcthresholdperadapterDA#1{\ifdim#1pt>84.8pt\cellcolor{lime!100}\else\ifdim#1pt>84.8pt\cellcolor{lime!90}\else\ifdim#1pt>84.8pt\cellcolor{lime!80}\else\ifdim#1pt>84.8pt\cellcolor{lime!70}\else\ifdim#1pt>84.8pt\cellcolor{lime!60}\else\ifdim#1pt>84.8pt\cellcolor{lime!50}\else\ifdim#1pt>84.8pt\cellcolor{lime!40}\else\ifdim#1pt>84.8pt\cellcolor{lime!30}\else\ifdim#1pt>84.8pt\cellcolor{lime!20}\else\ifdim#1pt>84.8pt\cellcolor{lime!10}\else\cellcolor{lime!0}\fi\fi\fi\fi\fi\fi\fi\fi\fi\fi#1}

\def\weightedspearmanzerozerosrcthresholdperadapterDW#1{\ifdim#1pt>77.8pt\cellcolor{lime!100}\else\ifdim#1pt>77.8pt\cellcolor{lime!90}\else\ifdim#1pt>77.8pt\cellcolor{lime!80}\else\ifdim#1pt>77.8pt\cellcolor{lime!70}\else\ifdim#1pt>77.8pt\cellcolor{lime!60}\else\ifdim#1pt>77.8pt\cellcolor{lime!50}\else\ifdim#1pt>77.8pt\cellcolor{lime!40}\else\ifdim#1pt>77.8pt\cellcolor{lime!30}\else\ifdim#1pt>77.8pt\cellcolor{lime!20}\else\ifdim#1pt>77.8pt\cellcolor{lime!10}\else\cellcolor{lime!0}\fi\fi\fi\fi\fi\fi\fi\fi\fi\fi#1}

\def\weightedspearmanzerozerosrcthresholdperadapterWA#1{\ifdim#1pt>81.4pt\cellcolor{lime!100}\else\ifdim#1pt>81.4pt\cellcolor{lime!90}\else\ifdim#1pt>81.4pt\cellcolor{lime!80}\else\ifdim#1pt>81.4pt\cellcolor{lime!70}\else\ifdim#1pt>81.4pt\cellcolor{lime!60}\else\ifdim#1pt>81.4pt\cellcolor{lime!50}\else\ifdim#1pt>81.4pt\cellcolor{lime!40}\else\ifdim#1pt>81.4pt\cellcolor{lime!30}\else\ifdim#1pt>81.4pt\cellcolor{lime!20}\else\ifdim#1pt>81.4pt\cellcolor{lime!10}\else\cellcolor{lime!0}\fi\fi\fi\fi\fi\fi\fi\fi\fi\fi#1}

\def\weightedspearmanzerozerosrcthresholdperadapterWD#1{\ifdim#1pt>77.6pt\cellcolor{lime!100}\else\ifdim#1pt>77.5pt\cellcolor{lime!90}\else\ifdim#1pt>77.3pt\cellcolor{lime!80}\else\ifdim#1pt>77.2pt\cellcolor{lime!70}\else\ifdim#1pt>77.1pt\cellcolor{lime!60}\else\ifdim#1pt>77.0pt\cellcolor{lime!50}\else\ifdim#1pt>76.9pt\cellcolor{lime!40}\else\ifdim#1pt>76.7pt\cellcolor{lime!30}\else\ifdim#1pt>76.6pt\cellcolor{lime!20}\else\ifdim#1pt>76.5pt\cellcolor{lime!10}\else\cellcolor{lime!0}\fi\fi\fi\fi\fi\fi\fi\fi\fi\fi#1}

\def\weightedspearmanzerozerosrcthresholdperadapterAC#1{\ifdim#1pt>92.0pt\cellcolor{lime!100}\else\ifdim#1pt>92.0pt\cellcolor{lime!90}\else\ifdim#1pt>92.0pt\cellcolor{lime!80}\else\ifdim#1pt>92.0pt\cellcolor{lime!70}\else\ifdim#1pt>92.0pt\cellcolor{lime!60}\else\ifdim#1pt>92.0pt\cellcolor{lime!50}\else\ifdim#1pt>92.0pt\cellcolor{lime!40}\else\ifdim#1pt>92.0pt\cellcolor{lime!30}\else\ifdim#1pt>92.0pt\cellcolor{lime!20}\else\ifdim#1pt>92.0pt\cellcolor{lime!10}\else\cellcolor{lime!0}\fi\fi\fi\fi\fi\fi\fi\fi\fi\fi#1}

\def\weightedspearmanzerozerosrcthresholdperadapterAP#1{\ifdim#1pt>95.5pt\cellcolor{lime!100}\else\ifdim#1pt>95.5pt\cellcolor{lime!90}\else\ifdim#1pt>95.5pt\cellcolor{lime!80}\else\ifdim#1pt>95.5pt\cellcolor{lime!70}\else\ifdim#1pt>95.5pt\cellcolor{lime!60}\else\ifdim#1pt>95.5pt\cellcolor{lime!50}\else\ifdim#1pt>95.5pt\cellcolor{lime!40}\else\ifdim#1pt>95.5pt\cellcolor{lime!30}\else\ifdim#1pt>95.5pt\cellcolor{lime!20}\else\ifdim#1pt>95.5pt\cellcolor{lime!10}\else\cellcolor{lime!0}\fi\fi\fi\fi\fi\fi\fi\fi\fi\fi#1}

\def\weightedspearmanzerozerosrcthresholdperadapterAR#1{\ifdim#1pt>97.1pt\cellcolor{lime!100}\else\ifdim#1pt>97.1pt\cellcolor{lime!90}\else\ifdim#1pt>97.1pt\cellcolor{lime!80}\else\ifdim#1pt>97.1pt\cellcolor{lime!70}\else\ifdim#1pt>97.1pt\cellcolor{lime!60}\else\ifdim#1pt>97.1pt\cellcolor{lime!50}\else\ifdim#1pt>97.1pt\cellcolor{lime!40}\else\ifdim#1pt>97.1pt\cellcolor{lime!30}\else\ifdim#1pt>97.1pt\cellcolor{lime!20}\else\ifdim#1pt>97.1pt\cellcolor{lime!10}\else\cellcolor{lime!0}\fi\fi\fi\fi\fi\fi\fi\fi\fi\fi#1}

\def\weightedspearmanzerozerosrcthresholdperadapterCA#1{\ifdim#1pt>78.8pt\cellcolor{lime!100}\else\ifdim#1pt>78.8pt\cellcolor{lime!90}\else\ifdim#1pt>78.8pt\cellcolor{lime!80}\else\ifdim#1pt>78.8pt\cellcolor{lime!70}\else\ifdim#1pt>78.8pt\cellcolor{lime!60}\else\ifdim#1pt>78.8pt\cellcolor{lime!50}\else\ifdim#1pt>78.8pt\cellcolor{lime!40}\else\ifdim#1pt>78.8pt\cellcolor{lime!30}\else\ifdim#1pt>78.8pt\cellcolor{lime!20}\else\ifdim#1pt>78.8pt\cellcolor{lime!10}\else\cellcolor{lime!0}\fi\fi\fi\fi\fi\fi\fi\fi\fi\fi#1}

\def\weightedspearmanzerozerosrcthresholdperadapterCP#1{\ifdim#1pt>84.3pt\cellcolor{lime!100}\else\ifdim#1pt>84.3pt\cellcolor{lime!90}\else\ifdim#1pt>84.3pt\cellcolor{lime!80}\else\ifdim#1pt>84.3pt\cellcolor{lime!70}\else\ifdim#1pt>84.3pt\cellcolor{lime!60}\else\ifdim#1pt>84.3pt\cellcolor{lime!50}\else\ifdim#1pt>84.3pt\cellcolor{lime!40}\else\ifdim#1pt>84.3pt\cellcolor{lime!30}\else\ifdim#1pt>84.3pt\cellcolor{lime!20}\else\ifdim#1pt>84.3pt\cellcolor{lime!10}\else\cellcolor{lime!0}\fi\fi\fi\fi\fi\fi\fi\fi\fi\fi#1}

\def\weightedspearmanzerozerosrcthresholdperadapterCR#1{\ifdim#1pt>82.1pt\cellcolor{lime!100}\else\ifdim#1pt>82.1pt\cellcolor{lime!90}\else\ifdim#1pt>82.1pt\cellcolor{lime!80}\else\ifdim#1pt>82.1pt\cellcolor{lime!70}\else\ifdim#1pt>82.1pt\cellcolor{lime!60}\else\ifdim#1pt>82.1pt\cellcolor{lime!50}\else\ifdim#1pt>82.1pt\cellcolor{lime!40}\else\ifdim#1pt>82.1pt\cellcolor{lime!30}\else\ifdim#1pt>82.1pt\cellcolor{lime!20}\else\ifdim#1pt>82.1pt\cellcolor{lime!10}\else\cellcolor{lime!0}\fi\fi\fi\fi\fi\fi\fi\fi\fi\fi#1}

\def\weightedspearmanzerozerosrcthresholdperadapterPA#1{\ifdim#1pt>92.7pt\cellcolor{lime!100}\else\ifdim#1pt>92.7pt\cellcolor{lime!90}\else\ifdim#1pt>92.7pt\cellcolor{lime!80}\else\ifdim#1pt>92.7pt\cellcolor{lime!70}\else\ifdim#1pt>92.7pt\cellcolor{lime!60}\else\ifdim#1pt>92.7pt\cellcolor{lime!50}\else\ifdim#1pt>92.7pt\cellcolor{lime!40}\else\ifdim#1pt>92.7pt\cellcolor{lime!30}\else\ifdim#1pt>92.7pt\cellcolor{lime!20}\else\ifdim#1pt>92.7pt\cellcolor{lime!10}\else\cellcolor{lime!0}\fi\fi\fi\fi\fi\fi\fi\fi\fi\fi#1}

\def\weightedspearmanzerozerosrcthresholdperadapterPC#1{\ifdim#1pt>94.4pt\cellcolor{lime!100}\else\ifdim#1pt>94.4pt\cellcolor{lime!90}\else\ifdim#1pt>94.4pt\cellcolor{lime!80}\else\ifdim#1pt>94.4pt\cellcolor{lime!70}\else\ifdim#1pt>94.4pt\cellcolor{lime!60}\else\ifdim#1pt>94.4pt\cellcolor{lime!50}\else\ifdim#1pt>94.4pt\cellcolor{lime!40}\else\ifdim#1pt>94.4pt\cellcolor{lime!30}\else\ifdim#1pt>94.4pt\cellcolor{lime!20}\else\ifdim#1pt>94.4pt\cellcolor{lime!10}\else\cellcolor{lime!0}\fi\fi\fi\fi\fi\fi\fi\fi\fi\fi#1}

\def\weightedspearmanzerozerosrcthresholdperadapterPR#1{\ifdim#1pt>94.7pt\cellcolor{lime!100}\else\ifdim#1pt>94.7pt\cellcolor{lime!90}\else\ifdim#1pt>94.7pt\cellcolor{lime!80}\else\ifdim#1pt>94.7pt\cellcolor{lime!70}\else\ifdim#1pt>94.7pt\cellcolor{lime!60}\else\ifdim#1pt>94.7pt\cellcolor{lime!50}\else\ifdim#1pt>94.7pt\cellcolor{lime!40}\else\ifdim#1pt>94.7pt\cellcolor{lime!30}\else\ifdim#1pt>94.7pt\cellcolor{lime!20}\else\ifdim#1pt>94.7pt\cellcolor{lime!10}\else\cellcolor{lime!0}\fi\fi\fi\fi\fi\fi\fi\fi\fi\fi#1}

\def\weightedspearmanzerozerosrcthresholdperadapterRA#1{\ifdim#1pt>94.0pt\cellcolor{lime!100}\else\ifdim#1pt>94.0pt\cellcolor{lime!90}\else\ifdim#1pt>94.0pt\cellcolor{lime!80}\else\ifdim#1pt>94.0pt\cellcolor{lime!70}\else\ifdim#1pt>94.0pt\cellcolor{lime!60}\else\ifdim#1pt>94.0pt\cellcolor{lime!50}\else\ifdim#1pt>94.0pt\cellcolor{lime!40}\else\ifdim#1pt>94.0pt\cellcolor{lime!30}\else\ifdim#1pt>94.0pt\cellcolor{lime!20}\else\ifdim#1pt>94.0pt\cellcolor{lime!10}\else\cellcolor{lime!0}\fi\fi\fi\fi\fi\fi\fi\fi\fi\fi#1}

\def\weightedspearmanzerozerosrcthresholdperadapterRC#1{\ifdim#1pt>90.9pt\cellcolor{lime!100}\else\ifdim#1pt>90.9pt\cellcolor{lime!90}\else\ifdim#1pt>90.9pt\cellcolor{lime!80}\else\ifdim#1pt>90.9pt\cellcolor{lime!70}\else\ifdim#1pt>90.9pt\cellcolor{lime!60}\else\ifdim#1pt>90.9pt\cellcolor{lime!50}\else\ifdim#1pt>90.9pt\cellcolor{lime!40}\else\ifdim#1pt>90.9pt\cellcolor{lime!30}\else\ifdim#1pt>90.9pt\cellcolor{lime!20}\else\ifdim#1pt>90.9pt\cellcolor{lime!10}\else\cellcolor{lime!0}\fi\fi\fi\fi\fi\fi\fi\fi\fi\fi#1}

\def\weightedspearmanzerozerosrcthresholdperadapterRP#1{\ifdim#1pt>96.6pt\cellcolor{lime!100}\else\ifdim#1pt>96.6pt\cellcolor{lime!90}\else\ifdim#1pt>96.6pt\cellcolor{lime!80}\else\ifdim#1pt>96.6pt\cellcolor{lime!70}\else\ifdim#1pt>96.6pt\cellcolor{lime!60}\else\ifdim#1pt>96.6pt\cellcolor{lime!50}\else\ifdim#1pt>96.6pt\cellcolor{lime!40}\else\ifdim#1pt>96.6pt\cellcolor{lime!30}\else\ifdim#1pt>96.6pt\cellcolor{lime!20}\else\ifdim#1pt>96.6pt\cellcolor{lime!10}\else\cellcolor{lime!0}\fi\fi\fi\fi\fi\fi\fi\fi\fi\fi#1}

\def\weightedspearmanzerozerosrcthresholdperadapterMean#1{\ifdim#1pt>87.9pt\cellcolor{lime!100}\else\ifdim#1pt>87.9pt\cellcolor{lime!90}\else\ifdim#1pt>87.9pt\cellcolor{lime!80}\else\ifdim#1pt>87.9pt\cellcolor{lime!70}\else\ifdim#1pt>87.9pt\cellcolor{lime!60}\else\ifdim#1pt>87.9pt\cellcolor{lime!50}\else\ifdim#1pt>87.9pt\cellcolor{lime!40}\else\ifdim#1pt>87.9pt\cellcolor{lime!30}\else\ifdim#1pt>87.9pt\cellcolor{lime!20}\else\ifdim#1pt>87.9pt\cellcolor{lime!10}\else\cellcolor{lime!0}\fi\fi\fi\fi\fi\fi\fi\fi\fi\fi#1}

\def\weightedspearmanzerozerosrcthresholdperadapterStd#1{\ifdim#1pt<6.3pt\cellcolor{lime!100}\else\ifdim#1pt<6.3pt\cellcolor{lime!90}\else\ifdim#1pt<6.4pt\cellcolor{lime!80}\else\ifdim#1pt<6.5pt\cellcolor{lime!70}\else\ifdim#1pt<6.6pt\cellcolor{lime!60}\else\ifdim#1pt<6.6pt\cellcolor{lime!50}\else\ifdim#1pt<6.7pt\cellcolor{lime!40}\else\ifdim#1pt<6.8pt\cellcolor{lime!30}\else\ifdim#1pt<6.8pt\cellcolor{lime!20}\else\ifdim#1pt<6.9pt\cellcolor{lime!10}\else\cellcolor{lime!0}\fi\fi\fi\fi\fi\fi\fi\fi\fi\fi#1}

\begin{table}[H]
\centering
\caption{The weighted Spearman correlation of each validator/task pair for \textbf{MMD}.}
\label{office_weighted_spearman_0.0_src_threshold_per_adapter_MMD}
\resizebox{!}{0.22\textheight}{
\begin{tabular}{llrrrrrrrrrrrrrrrrrrrr}
\toprule
 &  & AD & AW & DA & DW & WA & WD & AC & AP & AR & CA & CP & CR & PA & PC & PR & RA & RC & RP & Mean & Std \\
\midrule
\multirow[c]{2}{*}{Accuracy} & Source Train & \weightedspearmanzerozerosrcthresholdperadapterAD{52.4} & \weightedspearmanzerozerosrcthresholdperadapterAW{55.8} & \weightedspearmanzerozerosrcthresholdperadapterDA{80.5} & \weightedspearmanzerozerosrcthresholdperadapterDW{62.2} & \weightedspearmanzerozerosrcthresholdperadapterWA{78.5} & \weightedspearmanzerozerosrcthresholdperadapterWD{66.8} & \weightedspearmanzerozerosrcthresholdperadapterAC{50.1} & \weightedspearmanzerozerosrcthresholdperadapterAP{55.9} & \weightedspearmanzerozerosrcthresholdperadapterAR{46.4} & \weightedspearmanzerozerosrcthresholdperadapterCA{35.3} & \weightedspearmanzerozerosrcthresholdperadapterCP{44.8} & \weightedspearmanzerozerosrcthresholdperadapterCR{40.4} & \weightedspearmanzerozerosrcthresholdperadapterPA{59.3} & \weightedspearmanzerozerosrcthresholdperadapterPC{58.5} & \weightedspearmanzerozerosrcthresholdperadapterPR{67.6} & \weightedspearmanzerozerosrcthresholdperadapterRA{47.5} & \weightedspearmanzerozerosrcthresholdperadapterRC{57.2} & \weightedspearmanzerozerosrcthresholdperadapterRP{51.0} & \weightedspearmanzerozerosrcthresholdperadapterMean{56.1} & \weightedspearmanzerozerosrcthresholdperadapterStd{11.7} \\
\bottomrule
\end{tabular}}
\end{table}

%% file: supp_tables/domainnet126/weighted_spearman_0.0_src_threshold.tex
\def\weightedspearmanzerozerosrcthresholdDCP#1{\ifdim#1pt>83.6pt\cellcolor{lime!100}\else\ifdim#1pt>82.7pt\cellcolor{lime!90}\else\ifdim#1pt>81.9pt\cellcolor{lime!80}\else\ifdim#1pt>81.0pt\cellcolor{lime!70}\else\ifdim#1pt>80.1pt\cellcolor{lime!60}\else\ifdim#1pt>79.2pt\cellcolor{lime!50}\else\ifdim#1pt>78.3pt\cellcolor{lime!40}\else\ifdim#1pt>77.5pt\cellcolor{lime!30}\else\ifdim#1pt>76.6pt\cellcolor{lime!20}\else\ifdim#1pt>75.7pt\cellcolor{lime!10}\else\cellcolor{lime!0}\fi\fi\fi\fi\fi\fi\fi\fi\fi\fi#1}

\def\weightedspearmanzerozerosrcthresholdDCR#1{\ifdim#1pt>86.4pt\cellcolor{lime!100}\else\ifdim#1pt>85.9pt\cellcolor{lime!90}\else\ifdim#1pt>85.3pt\cellcolor{lime!80}\else\ifdim#1pt>84.7pt\cellcolor{lime!70}\else\ifdim#1pt>84.2pt\cellcolor{lime!60}\else\ifdim#1pt>83.6pt\cellcolor{lime!50}\else\ifdim#1pt>83.0pt\cellcolor{lime!40}\else\ifdim#1pt>82.4pt\cellcolor{lime!30}\else\ifdim#1pt>81.9pt\cellcolor{lime!20}\else\ifdim#1pt>81.3pt\cellcolor{lime!10}\else\cellcolor{lime!0}\fi\fi\fi\fi\fi\fi\fi\fi\fi\fi#1}

\def\weightedspearmanzerozerosrcthresholdDCS#1{\ifdim#1pt>84.8pt\cellcolor{lime!100}\else\ifdim#1pt>84.5pt\cellcolor{lime!90}\else\ifdim#1pt>84.2pt\cellcolor{lime!80}\else\ifdim#1pt>83.9pt\cellcolor{lime!70}\else\ifdim#1pt>83.7pt\cellcolor{lime!60}\else\ifdim#1pt>83.4pt\cellcolor{lime!50}\else\ifdim#1pt>83.1pt\cellcolor{lime!40}\else\ifdim#1pt>82.8pt\cellcolor{lime!30}\else\ifdim#1pt>82.5pt\cellcolor{lime!20}\else\ifdim#1pt>82.2pt\cellcolor{lime!10}\else\cellcolor{lime!0}\fi\fi\fi\fi\fi\fi\fi\fi\fi\fi#1}

\def\weightedspearmanzerozerosrcthresholdDPC#1{\ifdim#1pt>84.0pt\cellcolor{lime!100}\else\ifdim#1pt>82.9pt\cellcolor{lime!90}\else\ifdim#1pt>81.8pt\cellcolor{lime!80}\else\ifdim#1pt>80.7pt\cellcolor{lime!70}\else\ifdim#1pt>79.7pt\cellcolor{lime!60}\else\ifdim#1pt>78.6pt\cellcolor{lime!50}\else\ifdim#1pt>77.5pt\cellcolor{lime!40}\else\ifdim#1pt>76.4pt\cellcolor{lime!30}\else\ifdim#1pt>75.3pt\cellcolor{lime!20}\else\ifdim#1pt>74.2pt\cellcolor{lime!10}\else\cellcolor{lime!0}\fi\fi\fi\fi\fi\fi\fi\fi\fi\fi#1}

\def\weightedspearmanzerozerosrcthresholdDPR#1{\ifdim#1pt>88.3pt\cellcolor{lime!100}\else\ifdim#1pt>88.3pt\cellcolor{lime!90}\else\ifdim#1pt>88.3pt\cellcolor{lime!80}\else\ifdim#1pt>88.3pt\cellcolor{lime!70}\else\ifdim#1pt>88.3pt\cellcolor{lime!60}\else\ifdim#1pt>88.3pt\cellcolor{lime!50}\else\ifdim#1pt>88.3pt\cellcolor{lime!40}\else\ifdim#1pt>88.3pt\cellcolor{lime!30}\else\ifdim#1pt>88.3pt\cellcolor{lime!20}\else\ifdim#1pt>88.3pt\cellcolor{lime!10}\else\cellcolor{lime!0}\fi\fi\fi\fi\fi\fi\fi\fi\fi\fi#1}

\def\weightedspearmanzerozerosrcthresholdDPS#1{\ifdim#1pt>83.1pt\cellcolor{lime!100}\else\ifdim#1pt>81.0pt\cellcolor{lime!90}\else\ifdim#1pt>79.0pt\cellcolor{lime!80}\else\ifdim#1pt>77.0pt\cellcolor{lime!70}\else\ifdim#1pt>74.9pt\cellcolor{lime!60}\else\ifdim#1pt>72.9pt\cellcolor{lime!50}\else\ifdim#1pt>70.9pt\cellcolor{lime!40}\else\ifdim#1pt>68.9pt\cellcolor{lime!30}\else\ifdim#1pt>66.8pt\cellcolor{lime!20}\else\ifdim#1pt>64.8pt\cellcolor{lime!10}\else\cellcolor{lime!0}\fi\fi\fi\fi\fi\fi\fi\fi\fi\fi#1}

\def\weightedspearmanzerozerosrcthresholdDRC#1{\ifdim#1pt>85.0pt\cellcolor{lime!100}\else\ifdim#1pt>84.0pt\cellcolor{lime!90}\else\ifdim#1pt>83.1pt\cellcolor{lime!80}\else\ifdim#1pt>82.1pt\cellcolor{lime!70}\else\ifdim#1pt>81.1pt\cellcolor{lime!60}\else\ifdim#1pt>80.1pt\cellcolor{lime!50}\else\ifdim#1pt>79.1pt\cellcolor{lime!40}\else\ifdim#1pt>78.2pt\cellcolor{lime!30}\else\ifdim#1pt>77.2pt\cellcolor{lime!20}\else\ifdim#1pt>76.2pt\cellcolor{lime!10}\else\cellcolor{lime!0}\fi\fi\fi\fi\fi\fi\fi\fi\fi\fi#1}

\def\weightedspearmanzerozerosrcthresholdDRP#1{\ifdim#1pt>89.6pt\cellcolor{lime!100}\else\ifdim#1pt>89.6pt\cellcolor{lime!90}\else\ifdim#1pt>89.6pt\cellcolor{lime!80}\else\ifdim#1pt>89.6pt\cellcolor{lime!70}\else\ifdim#1pt>89.6pt\cellcolor{lime!60}\else\ifdim#1pt>89.6pt\cellcolor{lime!50}\else\ifdim#1pt>89.6pt\cellcolor{lime!40}\else\ifdim#1pt>89.6pt\cellcolor{lime!30}\else\ifdim#1pt>89.6pt\cellcolor{lime!20}\else\ifdim#1pt>89.6pt\cellcolor{lime!10}\else\cellcolor{lime!0}\fi\fi\fi\fi\fi\fi\fi\fi\fi\fi#1}

\def\weightedspearmanzerozerosrcthresholdDRS#1{\ifdim#1pt>85.4pt\cellcolor{lime!100}\else\ifdim#1pt>83.8pt\cellcolor{lime!90}\else\ifdim#1pt>82.1pt\cellcolor{lime!80}\else\ifdim#1pt>80.5pt\cellcolor{lime!70}\else\ifdim#1pt>78.9pt\cellcolor{lime!60}\else\ifdim#1pt>77.3pt\cellcolor{lime!50}\else\ifdim#1pt>75.7pt\cellcolor{lime!40}\else\ifdim#1pt>74.0pt\cellcolor{lime!30}\else\ifdim#1pt>72.4pt\cellcolor{lime!20}\else\ifdim#1pt>70.8pt\cellcolor{lime!10}\else\cellcolor{lime!0}\fi\fi\fi\fi\fi\fi\fi\fi\fi\fi#1}

\def\weightedspearmanzerozerosrcthresholdDSC#1{\ifdim#1pt>84.0pt\cellcolor{lime!100}\else\ifdim#1pt>82.5pt\cellcolor{lime!90}\else\ifdim#1pt>81.0pt\cellcolor{lime!80}\else\ifdim#1pt>79.5pt\cellcolor{lime!70}\else\ifdim#1pt>78.0pt\cellcolor{lime!60}\else\ifdim#1pt>76.6pt\cellcolor{lime!50}\else\ifdim#1pt>75.1pt\cellcolor{lime!40}\else\ifdim#1pt>73.6pt\cellcolor{lime!30}\else\ifdim#1pt>72.1pt\cellcolor{lime!20}\else\ifdim#1pt>70.6pt\cellcolor{lime!10}\else\cellcolor{lime!0}\fi\fi\fi\fi\fi\fi\fi\fi\fi\fi#1}

\def\weightedspearmanzerozerosrcthresholdDSP#1{\ifdim#1pt>83.3pt\cellcolor{lime!100}\else\ifdim#1pt>81.6pt\cellcolor{lime!90}\else\ifdim#1pt>79.9pt\cellcolor{lime!80}\else\ifdim#1pt>78.2pt\cellcolor{lime!70}\else\ifdim#1pt>76.5pt\cellcolor{lime!60}\else\ifdim#1pt>74.8pt\cellcolor{lime!50}\else\ifdim#1pt>73.1pt\cellcolor{lime!40}\else\ifdim#1pt>71.4pt\cellcolor{lime!30}\else\ifdim#1pt>69.7pt\cellcolor{lime!20}\else\ifdim#1pt>68.0pt\cellcolor{lime!10}\else\cellcolor{lime!0}\fi\fi\fi\fi\fi\fi\fi\fi\fi\fi#1}

\def\weightedspearmanzerozerosrcthresholdDSR#1{\ifdim#1pt>89.5pt\cellcolor{lime!100}\else\ifdim#1pt>86.6pt\cellcolor{lime!90}\else\ifdim#1pt>83.8pt\cellcolor{lime!80}\else\ifdim#1pt>80.9pt\cellcolor{lime!70}\else\ifdim#1pt>78.1pt\cellcolor{lime!60}\else\ifdim#1pt>75.3pt\cellcolor{lime!50}\else\ifdim#1pt>72.4pt\cellcolor{lime!40}\else\ifdim#1pt>69.6pt\cellcolor{lime!30}\else\ifdim#1pt>66.7pt\cellcolor{lime!20}\else\ifdim#1pt>63.9pt\cellcolor{lime!10}\else\cellcolor{lime!0}\fi\fi\fi\fi\fi\fi\fi\fi\fi\fi#1}

\def\weightedspearmanzerozerosrcthresholdMean#1{\ifdim#1pt>83.6pt\cellcolor{lime!100}\else\ifdim#1pt>82.7pt\cellcolor{lime!90}\else\ifdim#1pt>81.8pt\cellcolor{lime!80}\else\ifdim#1pt>80.9pt\cellcolor{lime!70}\else\ifdim#1pt>80.0pt\cellcolor{lime!60}\else\ifdim#1pt>79.1pt\cellcolor{lime!50}\else\ifdim#1pt>78.2pt\cellcolor{lime!40}\else\ifdim#1pt>77.3pt\cellcolor{lime!30}\else\ifdim#1pt>76.4pt\cellcolor{lime!20}\else\ifdim#1pt>75.5pt\cellcolor{lime!10}\else\cellcolor{lime!0}\fi\fi\fi\fi\fi\fi\fi\fi\fi\fi#1}

\def\weightedspearmanzerozerosrcthresholdStd#1{\ifdim#1pt<1.9pt\cellcolor{lime!100}\else\ifdim#1pt<2.6pt\cellcolor{lime!90}\else\ifdim#1pt<3.3pt\cellcolor{lime!80}\else\ifdim#1pt<4.0pt\cellcolor{lime!70}\else\ifdim#1pt<4.6pt\cellcolor{lime!60}\else\ifdim#1pt<5.3pt\cellcolor{lime!50}\else\ifdim#1pt<6.0pt\cellcolor{lime!40}\else\ifdim#1pt<6.7pt\cellcolor{lime!30}\else\ifdim#1pt<7.4pt\cellcolor{lime!20}\else\ifdim#1pt<8.1pt\cellcolor{lime!10}\else\cellcolor{lime!0}\fi\fi\fi\fi\fi\fi\fi\fi\fi\fi#1}

\begin{table}[H]
\centering
\caption{The weighted Spearman correlation of each validator/task pair, using the checkpoints of all algorithms.}
\label{domainnet126_weighted_spearman_0.0_src_threshold}
\resizebox{!}{0.21\textheight}{
\begin{tabular}{llrrrrrrrrrrrrrr}
\toprule
 &  & CP & CR & CS & PC & PR & PS & RC & RP & RS & SC & SP & SR & Mean & Std \\
\midrule
\multirow[c]{2}{*}{Accuracy} & Source Train & \weightedspearmanzerozerosrcthresholdDCP{84.3} & \weightedspearmanzerozerosrcthresholdDCR{82.9} & \weightedspearmanzerozerosrcthresholdDCS{81.9} & \weightedspearmanzerozerosrcthresholdDPC{74.6} & \weightedspearmanzerozerosrcthresholdDPR{67.6} & \weightedspearmanzerozerosrcthresholdDPS{65.2} & \weightedspearmanzerozerosrcthresholdDRC{75.5} & \weightedspearmanzerozerosrcthresholdDRP{84.9} & \weightedspearmanzerozerosrcthresholdDRS{68.1} & \weightedspearmanzerozerosrcthresholdDSC{76.3} & \weightedspearmanzerozerosrcthresholdDSP{78.7} & \weightedspearmanzerozerosrcthresholdDSR{76.0} & \weightedspearmanzerozerosrcthresholdMean{76.3} & \weightedspearmanzerozerosrcthresholdStd{6.4} \\
 & Source Val & \weightedspearmanzerozerosrcthresholdDCP{75.7} & \weightedspearmanzerozerosrcthresholdDCR{81.3} & \weightedspearmanzerozerosrcthresholdDCS{82.2} & \weightedspearmanzerozerosrcthresholdDPC{74.2} & \textbf{\weightedspearmanzerozerosrcthresholdDPR{88.3}} & \weightedspearmanzerozerosrcthresholdDPS{64.8} & \weightedspearmanzerozerosrcthresholdDRC{76.2} & \textbf{\weightedspearmanzerozerosrcthresholdDRP{89.6}} & \weightedspearmanzerozerosrcthresholdDRS{70.8} & \weightedspearmanzerozerosrcthresholdDSC{70.6} & \weightedspearmanzerozerosrcthresholdDSP{68.0} & \weightedspearmanzerozerosrcthresholdDSR{63.9} & \weightedspearmanzerozerosrcthresholdMean{75.5} & \weightedspearmanzerozerosrcthresholdStd{8.1} \\
\bottomrule
\end{tabular}}
\end{table}

%% file: supp_tables/domainnet126/weighted_spearman_0.0_src_threshold_per_adapter_ATDOC.tex
\def\weightedspearmanzerozerosrcthresholdperadapterDCP#1{\ifdim#1pt>96.7pt\cellcolor{lime!100}\else\ifdim#1pt>96.1pt\cellcolor{lime!90}\else\ifdim#1pt>95.5pt\cellcolor{lime!80}\else\ifdim#1pt>94.9pt\cellcolor{lime!70}\else\ifdim#1pt>94.2pt\cellcolor{lime!60}\else\ifdim#1pt>93.6pt\cellcolor{lime!50}\else\ifdim#1pt>93.0pt\cellcolor{lime!40}\else\ifdim#1pt>92.4pt\cellcolor{lime!30}\else\ifdim#1pt>91.8pt\cellcolor{lime!20}\else\ifdim#1pt>91.2pt\cellcolor{lime!10}\else\cellcolor{lime!0}\fi\fi\fi\fi\fi\fi\fi\fi\fi\fi#1}

\def\weightedspearmanzerozerosrcthresholdperadapterDCR#1{\ifdim#1pt>97.8pt\cellcolor{lime!100}\else\ifdim#1pt>97.7pt\cellcolor{lime!90}\else\ifdim#1pt>97.6pt\cellcolor{lime!80}\else\ifdim#1pt>97.5pt\cellcolor{lime!70}\else\ifdim#1pt>97.3pt\cellcolor{lime!60}\else\ifdim#1pt>97.2pt\cellcolor{lime!50}\else\ifdim#1pt>97.1pt\cellcolor{lime!40}\else\ifdim#1pt>97.0pt\cellcolor{lime!30}\else\ifdim#1pt>96.9pt\cellcolor{lime!20}\else\ifdim#1pt>96.8pt\cellcolor{lime!10}\else\cellcolor{lime!0}\fi\fi\fi\fi\fi\fi\fi\fi\fi\fi#1}

\def\weightedspearmanzerozerosrcthresholdperadapterDCS#1{\ifdim#1pt>98.4pt\cellcolor{lime!100}\else\ifdim#1pt>98.3pt\cellcolor{lime!90}\else\ifdim#1pt>98.2pt\cellcolor{lime!80}\else\ifdim#1pt>98.2pt\cellcolor{lime!70}\else\ifdim#1pt>98.2pt\cellcolor{lime!60}\else\ifdim#1pt>98.1pt\cellcolor{lime!50}\else\ifdim#1pt>98.1pt\cellcolor{lime!40}\else\ifdim#1pt>98.0pt\cellcolor{lime!30}\else\ifdim#1pt>98.0pt\cellcolor{lime!20}\else\ifdim#1pt>97.9pt\cellcolor{lime!10}\else\cellcolor{lime!0}\fi\fi\fi\fi\fi\fi\fi\fi\fi\fi#1}

\def\weightedspearmanzerozerosrcthresholdperadapterDPC#1{\ifdim#1pt>95.1pt\cellcolor{lime!100}\else\ifdim#1pt>94.9pt\cellcolor{lime!90}\else\ifdim#1pt>94.8pt\cellcolor{lime!80}\else\ifdim#1pt>94.6pt\cellcolor{lime!70}\else\ifdim#1pt>94.4pt\cellcolor{lime!60}\else\ifdim#1pt>94.2pt\cellcolor{lime!50}\else\ifdim#1pt>94.0pt\cellcolor{lime!40}\else\ifdim#1pt>93.9pt\cellcolor{lime!30}\else\ifdim#1pt>93.7pt\cellcolor{lime!20}\else\ifdim#1pt>93.5pt\cellcolor{lime!10}\else\cellcolor{lime!0}\fi\fi\fi\fi\fi\fi\fi\fi\fi\fi#1}

\def\weightedspearmanzerozerosrcthresholdperadapterDPR#1{\ifdim#1pt>97.7pt\cellcolor{lime!100}\else\ifdim#1pt>97.6pt\cellcolor{lime!90}\else\ifdim#1pt>97.5pt\cellcolor{lime!80}\else\ifdim#1pt>97.5pt\cellcolor{lime!70}\else\ifdim#1pt>97.5pt\cellcolor{lime!60}\else\ifdim#1pt>97.4pt\cellcolor{lime!50}\else\ifdim#1pt>97.4pt\cellcolor{lime!40}\else\ifdim#1pt>97.3pt\cellcolor{lime!30}\else\ifdim#1pt>97.2pt\cellcolor{lime!20}\else\ifdim#1pt>97.2pt\cellcolor{lime!10}\else\cellcolor{lime!0}\fi\fi\fi\fi\fi\fi\fi\fi\fi\fi#1}

\def\weightedspearmanzerozerosrcthresholdperadapterDPS#1{\ifdim#1pt>96.5pt\cellcolor{lime!100}\else\ifdim#1pt>95.9pt\cellcolor{lime!90}\else\ifdim#1pt>95.3pt\cellcolor{lime!80}\else\ifdim#1pt>94.8pt\cellcolor{lime!70}\else\ifdim#1pt>94.2pt\cellcolor{lime!60}\else\ifdim#1pt>93.7pt\cellcolor{lime!50}\else\ifdim#1pt>93.2pt\cellcolor{lime!40}\else\ifdim#1pt>92.6pt\cellcolor{lime!30}\else\ifdim#1pt>92.0pt\cellcolor{lime!20}\else\ifdim#1pt>91.5pt\cellcolor{lime!10}\else\cellcolor{lime!0}\fi\fi\fi\fi\fi\fi\fi\fi\fi\fi#1}

\def\weightedspearmanzerozerosrcthresholdperadapterDRC#1{\ifdim#1pt>98.3pt\cellcolor{lime!100}\else\ifdim#1pt>98.0pt\cellcolor{lime!90}\else\ifdim#1pt>97.7pt\cellcolor{lime!80}\else\ifdim#1pt>97.4pt\cellcolor{lime!70}\else\ifdim#1pt>97.0pt\cellcolor{lime!60}\else\ifdim#1pt>96.7pt\cellcolor{lime!50}\else\ifdim#1pt>96.4pt\cellcolor{lime!40}\else\ifdim#1pt>96.1pt\cellcolor{lime!30}\else\ifdim#1pt>95.8pt\cellcolor{lime!20}\else\ifdim#1pt>95.5pt\cellcolor{lime!10}\else\cellcolor{lime!0}\fi\fi\fi\fi\fi\fi\fi\fi\fi\fi#1}

\def\weightedspearmanzerozerosrcthresholdperadapterDRP#1{\ifdim#1pt>96.9pt\cellcolor{lime!100}\else\ifdim#1pt>96.8pt\cellcolor{lime!90}\else\ifdim#1pt>96.6pt\cellcolor{lime!80}\else\ifdim#1pt>96.4pt\cellcolor{lime!70}\else\ifdim#1pt>96.2pt\cellcolor{lime!60}\else\ifdim#1pt>96.1pt\cellcolor{lime!50}\else\ifdim#1pt>95.9pt\cellcolor{lime!40}\else\ifdim#1pt>95.7pt\cellcolor{lime!30}\else\ifdim#1pt>95.6pt\cellcolor{lime!20}\else\ifdim#1pt>95.4pt\cellcolor{lime!10}\else\cellcolor{lime!0}\fi\fi\fi\fi\fi\fi\fi\fi\fi\fi#1}

\def\weightedspearmanzerozerosrcthresholdperadapterDRS#1{\ifdim#1pt>97.8pt\cellcolor{lime!100}\else\ifdim#1pt>97.4pt\cellcolor{lime!90}\else\ifdim#1pt>97.0pt\cellcolor{lime!80}\else\ifdim#1pt>96.5pt\cellcolor{lime!70}\else\ifdim#1pt>96.0pt\cellcolor{lime!60}\else\ifdim#1pt>95.6pt\cellcolor{lime!50}\else\ifdim#1pt>95.1pt\cellcolor{lime!40}\else\ifdim#1pt>94.7pt\cellcolor{lime!30}\else\ifdim#1pt>94.2pt\cellcolor{lime!20}\else\ifdim#1pt>93.8pt\cellcolor{lime!10}\else\cellcolor{lime!0}\fi\fi\fi\fi\fi\fi\fi\fi\fi\fi#1}

\def\weightedspearmanzerozerosrcthresholdperadapterDSC#1{\ifdim#1pt>97.7pt\cellcolor{lime!100}\else\ifdim#1pt>96.8pt\cellcolor{lime!90}\else\ifdim#1pt>96.0pt\cellcolor{lime!80}\else\ifdim#1pt>95.1pt\cellcolor{lime!70}\else\ifdim#1pt>94.2pt\cellcolor{lime!60}\else\ifdim#1pt>93.4pt\cellcolor{lime!50}\else\ifdim#1pt>92.5pt\cellcolor{lime!40}\else\ifdim#1pt>91.7pt\cellcolor{lime!30}\else\ifdim#1pt>90.8pt\cellcolor{lime!20}\else\ifdim#1pt>90.0pt\cellcolor{lime!10}\else\cellcolor{lime!0}\fi\fi\fi\fi\fi\fi\fi\fi\fi\fi#1}

\def\weightedspearmanzerozerosrcthresholdperadapterDSP#1{\ifdim#1pt>97.7pt\cellcolor{lime!100}\else\ifdim#1pt>96.5pt\cellcolor{lime!90}\else\ifdim#1pt>95.4pt\cellcolor{lime!80}\else\ifdim#1pt>94.2pt\cellcolor{lime!70}\else\ifdim#1pt>93.0pt\cellcolor{lime!60}\else\ifdim#1pt>91.8pt\cellcolor{lime!50}\else\ifdim#1pt>90.6pt\cellcolor{lime!40}\else\ifdim#1pt>89.5pt\cellcolor{lime!30}\else\ifdim#1pt>88.3pt\cellcolor{lime!20}\else\ifdim#1pt>87.1pt\cellcolor{lime!10}\else\cellcolor{lime!0}\fi\fi\fi\fi\fi\fi\fi\fi\fi\fi#1}

\def\weightedspearmanzerozerosrcthresholdperadapterDSR#1{\ifdim#1pt>97.9pt\cellcolor{lime!100}\else\ifdim#1pt>97.3pt\cellcolor{lime!90}\else\ifdim#1pt>96.7pt\cellcolor{lime!80}\else\ifdim#1pt>96.1pt\cellcolor{lime!70}\else\ifdim#1pt>95.5pt\cellcolor{lime!60}\else\ifdim#1pt>94.8pt\cellcolor{lime!50}\else\ifdim#1pt>94.2pt\cellcolor{lime!40}\else\ifdim#1pt>93.6pt\cellcolor{lime!30}\else\ifdim#1pt>93.0pt\cellcolor{lime!20}\else\ifdim#1pt>92.4pt\cellcolor{lime!10}\else\cellcolor{lime!0}\fi\fi\fi\fi\fi\fi\fi\fi\fi\fi#1}

\def\weightedspearmanzerozerosrcthresholdperadapterMean#1{\ifdim#1pt>97.1pt\cellcolor{lime!100}\else\ifdim#1pt>96.7pt\cellcolor{lime!90}\else\ifdim#1pt>96.3pt\cellcolor{lime!80}\else\ifdim#1pt>95.9pt\cellcolor{lime!70}\else\ifdim#1pt>95.5pt\cellcolor{lime!60}\else\ifdim#1pt>95.1pt\cellcolor{lime!50}\else\ifdim#1pt>94.7pt\cellcolor{lime!40}\else\ifdim#1pt>94.3pt\cellcolor{lime!30}\else\ifdim#1pt>93.9pt\cellcolor{lime!20}\else\ifdim#1pt>93.5pt\cellcolor{lime!10}\else\cellcolor{lime!0}\fi\fi\fi\fi\fi\fi\fi\fi\fi\fi#1}

\def\weightedspearmanzerozerosrcthresholdperadapterStd#1{\ifdim#1pt<1.0pt\cellcolor{lime!100}\else\ifdim#1pt<1.3pt\cellcolor{lime!90}\else\ifdim#1pt<1.5pt\cellcolor{lime!80}\else\ifdim#1pt<1.7pt\cellcolor{lime!70}\else\ifdim#1pt<1.9pt\cellcolor{lime!60}\else\ifdim#1pt<2.2pt\cellcolor{lime!50}\else\ifdim#1pt<2.4pt\cellcolor{lime!40}\else\ifdim#1pt<2.6pt\cellcolor{lime!30}\else\ifdim#1pt<2.9pt\cellcolor{lime!20}\else\ifdim#1pt<3.1pt\cellcolor{lime!10}\else\cellcolor{lime!0}\fi\fi\fi\fi\fi\fi\fi\fi\fi\fi#1}

\begin{table}[H]
\centering
\caption{The weighted Spearman correlation of each validator/task pair for \textbf{ATDOC}.}
\label{domainnet126_weighted_spearman_0.0_src_threshold_per_adapter_ATDOC}
\resizebox{!}{0.22\textheight}{
\begin{tabular}{llrrrrrrrrrrrrrr}
\toprule
 &  & CP & CR & CS & PC & PR & PS & RC & RP & RS & SC & SP & SR & Mean & Std \\
\midrule
\multirow[c]{2}{*}{Accuracy} & Source Train & \weightedspearmanzerozerosrcthresholdperadapterDCP{91.7} & \weightedspearmanzerozerosrcthresholdperadapterDCR{95.4} & \weightedspearmanzerozerosrcthresholdperadapterDCS{94.2} & \weightedspearmanzerozerosrcthresholdperadapterDPC{93.8} & \weightedspearmanzerozerosrcthresholdperadapterDPR{87.7} & \weightedspearmanzerozerosrcthresholdperadapterDPS{94.4} & \weightedspearmanzerozerosrcthresholdperadapterDRC{95.9} & \weightedspearmanzerozerosrcthresholdperadapterDRP{94.0} & \weightedspearmanzerozerosrcthresholdperadapterDRS{93.4} & \weightedspearmanzerozerosrcthresholdperadapterDSC{91.7} & \weightedspearmanzerozerosrcthresholdperadapterDSP{92.8} & \weightedspearmanzerozerosrcthresholdperadapterDSR{95.0} & \weightedspearmanzerozerosrcthresholdperadapterMean{93.3} & \weightedspearmanzerozerosrcthresholdperadapterStd{2.1} \\
 & Source Val & \weightedspearmanzerozerosrcthresholdperadapterDCP{91.2} & \weightedspearmanzerozerosrcthresholdperadapterDCR{96.8} & \weightedspearmanzerozerosrcthresholdperadapterDCS{97.9} & \weightedspearmanzerozerosrcthresholdperadapterDPC{93.5} & \weightedspearmanzerozerosrcthresholdperadapterDPR{97.2} & \weightedspearmanzerozerosrcthresholdperadapterDPS{91.5} & \weightedspearmanzerozerosrcthresholdperadapterDRC{95.5} & \weightedspearmanzerozerosrcthresholdperadapterDRP{95.4} & \weightedspearmanzerozerosrcthresholdperadapterDRS{93.8} & \weightedspearmanzerozerosrcthresholdperadapterDSC{90.0} & \weightedspearmanzerozerosrcthresholdperadapterDSP{87.1} & \weightedspearmanzerozerosrcthresholdperadapterDSR{92.4} & \weightedspearmanzerozerosrcthresholdperadapterMean{93.5} & \weightedspearmanzerozerosrcthresholdperadapterStd{3.1} \\
\bottomrule
\end{tabular}}
\end{table}

%% file: supp_tables/domainnet126/weighted_spearman_0.0_src_threshold_per_adapter_BNM.tex
\def\weightedspearmanzerozerosrcthresholdperadapterDCP#1{\ifdim#1pt>92.6pt\cellcolor{lime!100}\else\ifdim#1pt>89.9pt\cellcolor{lime!90}\else\ifdim#1pt>87.1pt\cellcolor{lime!80}\else\ifdim#1pt>84.4pt\cellcolor{lime!70}\else\ifdim#1pt>81.7pt\cellcolor{lime!60}\else\ifdim#1pt>79.0pt\cellcolor{lime!50}\else\ifdim#1pt>76.3pt\cellcolor{lime!40}\else\ifdim#1pt>73.5pt\cellcolor{lime!30}\else\ifdim#1pt>70.8pt\cellcolor{lime!20}\else\ifdim#1pt>68.1pt\cellcolor{lime!10}\else\cellcolor{lime!0}\fi\fi\fi\fi\fi\fi\fi\fi\fi\fi#1}

\def\weightedspearmanzerozerosrcthresholdperadapterDCR#1{\ifdim#1pt>93.3pt\cellcolor{lime!100}\else\ifdim#1pt>91.2pt\cellcolor{lime!90}\else\ifdim#1pt>89.0pt\cellcolor{lime!80}\else\ifdim#1pt>86.9pt\cellcolor{lime!70}\else\ifdim#1pt>84.7pt\cellcolor{lime!60}\else\ifdim#1pt>82.5pt\cellcolor{lime!50}\else\ifdim#1pt>80.4pt\cellcolor{lime!40}\else\ifdim#1pt>78.2pt\cellcolor{lime!30}\else\ifdim#1pt>76.1pt\cellcolor{lime!20}\else\ifdim#1pt>73.9pt\cellcolor{lime!10}\else\cellcolor{lime!0}\fi\fi\fi\fi\fi\fi\fi\fi\fi\fi#1}

\def\weightedspearmanzerozerosrcthresholdperadapterDCS#1{\ifdim#1pt>89.1pt\cellcolor{lime!100}\else\ifdim#1pt>86.0pt\cellcolor{lime!90}\else\ifdim#1pt>83.0pt\cellcolor{lime!80}\else\ifdim#1pt>79.9pt\cellcolor{lime!70}\else\ifdim#1pt>76.9pt\cellcolor{lime!60}\else\ifdim#1pt>73.9pt\cellcolor{lime!50}\else\ifdim#1pt>70.8pt\cellcolor{lime!40}\else\ifdim#1pt>67.8pt\cellcolor{lime!30}\else\ifdim#1pt>64.7pt\cellcolor{lime!20}\else\ifdim#1pt>61.7pt\cellcolor{lime!10}\else\cellcolor{lime!0}\fi\fi\fi\fi\fi\fi\fi\fi\fi\fi#1}

\def\weightedspearmanzerozerosrcthresholdperadapterDPC#1{\ifdim#1pt>81.6pt\cellcolor{lime!100}\else\ifdim#1pt>78.4pt\cellcolor{lime!90}\else\ifdim#1pt>75.3pt\cellcolor{lime!80}\else\ifdim#1pt>72.2pt\cellcolor{lime!70}\else\ifdim#1pt>69.0pt\cellcolor{lime!60}\else\ifdim#1pt>65.9pt\cellcolor{lime!50}\else\ifdim#1pt>62.8pt\cellcolor{lime!40}\else\ifdim#1pt>59.7pt\cellcolor{lime!30}\else\ifdim#1pt>56.5pt\cellcolor{lime!20}\else\ifdim#1pt>53.4pt\cellcolor{lime!10}\else\cellcolor{lime!0}\fi\fi\fi\fi\fi\fi\fi\fi\fi\fi#1}

\def\weightedspearmanzerozerosrcthresholdperadapterDPR#1{\ifdim#1pt>91.6pt\cellcolor{lime!100}\else\ifdim#1pt>88.7pt\cellcolor{lime!90}\else\ifdim#1pt>85.8pt\cellcolor{lime!80}\else\ifdim#1pt>82.9pt\cellcolor{lime!70}\else\ifdim#1pt>80.0pt\cellcolor{lime!60}\else\ifdim#1pt>77.0pt\cellcolor{lime!50}\else\ifdim#1pt>74.1pt\cellcolor{lime!40}\else\ifdim#1pt>71.2pt\cellcolor{lime!30}\else\ifdim#1pt>68.3pt\cellcolor{lime!20}\else\ifdim#1pt>65.4pt\cellcolor{lime!10}\else\cellcolor{lime!0}\fi\fi\fi\fi\fi\fi\fi\fi\fi\fi#1}

\def\weightedspearmanzerozerosrcthresholdperadapterDPS#1{\ifdim#1pt>78.9pt\cellcolor{lime!100}\else\ifdim#1pt>75.9pt\cellcolor{lime!90}\else\ifdim#1pt>72.8pt\cellcolor{lime!80}\else\ifdim#1pt>69.8pt\cellcolor{lime!70}\else\ifdim#1pt>66.8pt\cellcolor{lime!60}\else\ifdim#1pt>63.8pt\cellcolor{lime!50}\else\ifdim#1pt>60.8pt\cellcolor{lime!40}\else\ifdim#1pt>57.7pt\cellcolor{lime!30}\else\ifdim#1pt>54.7pt\cellcolor{lime!20}\else\ifdim#1pt>51.7pt\cellcolor{lime!10}\else\cellcolor{lime!0}\fi\fi\fi\fi\fi\fi\fi\fi\fi\fi#1}

\def\weightedspearmanzerozerosrcthresholdperadapterDRC#1{\ifdim#1pt>91.3pt\cellcolor{lime!100}\else\ifdim#1pt>89.0pt\cellcolor{lime!90}\else\ifdim#1pt>86.7pt\cellcolor{lime!80}\else\ifdim#1pt>84.4pt\cellcolor{lime!70}\else\ifdim#1pt>82.2pt\cellcolor{lime!60}\else\ifdim#1pt>79.9pt\cellcolor{lime!50}\else\ifdim#1pt>77.6pt\cellcolor{lime!40}\else\ifdim#1pt>75.3pt\cellcolor{lime!30}\else\ifdim#1pt>73.0pt\cellcolor{lime!20}\else\ifdim#1pt>70.7pt\cellcolor{lime!10}\else\cellcolor{lime!0}\fi\fi\fi\fi\fi\fi\fi\fi\fi\fi#1}

\def\weightedspearmanzerozerosrcthresholdperadapterDRP#1{\ifdim#1pt>86.2pt\cellcolor{lime!100}\else\ifdim#1pt>85.1pt\cellcolor{lime!90}\else\ifdim#1pt>84.1pt\cellcolor{lime!80}\else\ifdim#1pt>83.0pt\cellcolor{lime!70}\else\ifdim#1pt>81.9pt\cellcolor{lime!60}\else\ifdim#1pt>80.8pt\cellcolor{lime!50}\else\ifdim#1pt>79.7pt\cellcolor{lime!40}\else\ifdim#1pt>78.7pt\cellcolor{lime!30}\else\ifdim#1pt>77.6pt\cellcolor{lime!20}\else\ifdim#1pt>76.5pt\cellcolor{lime!10}\else\cellcolor{lime!0}\fi\fi\fi\fi\fi\fi\fi\fi\fi\fi#1}

\def\weightedspearmanzerozerosrcthresholdperadapterDRS#1{\ifdim#1pt>91.1pt\cellcolor{lime!100}\else\ifdim#1pt>87.4pt\cellcolor{lime!90}\else\ifdim#1pt>83.6pt\cellcolor{lime!80}\else\ifdim#1pt>79.9pt\cellcolor{lime!70}\else\ifdim#1pt>76.1pt\cellcolor{lime!60}\else\ifdim#1pt>72.3pt\cellcolor{lime!50}\else\ifdim#1pt>68.6pt\cellcolor{lime!40}\else\ifdim#1pt>64.8pt\cellcolor{lime!30}\else\ifdim#1pt>61.1pt\cellcolor{lime!20}\else\ifdim#1pt>57.3pt\cellcolor{lime!10}\else\cellcolor{lime!0}\fi\fi\fi\fi\fi\fi\fi\fi\fi\fi#1}

\def\weightedspearmanzerozerosrcthresholdperadapterDSC#1{\ifdim#1pt>84.2pt\cellcolor{lime!100}\else\ifdim#1pt>81.5pt\cellcolor{lime!90}\else\ifdim#1pt>78.8pt\cellcolor{lime!80}\else\ifdim#1pt>76.1pt\cellcolor{lime!70}\else\ifdim#1pt>73.4pt\cellcolor{lime!60}\else\ifdim#1pt>70.7pt\cellcolor{lime!50}\else\ifdim#1pt>68.0pt\cellcolor{lime!40}\else\ifdim#1pt>65.3pt\cellcolor{lime!30}\else\ifdim#1pt>62.6pt\cellcolor{lime!20}\else\ifdim#1pt>59.9pt\cellcolor{lime!10}\else\cellcolor{lime!0}\fi\fi\fi\fi\fi\fi\fi\fi\fi\fi#1}

\def\weightedspearmanzerozerosrcthresholdperadapterDSP#1{\ifdim#1pt>82.6pt\cellcolor{lime!100}\else\ifdim#1pt>77.9pt\cellcolor{lime!90}\else\ifdim#1pt>73.2pt\cellcolor{lime!80}\else\ifdim#1pt>68.5pt\cellcolor{lime!70}\else\ifdim#1pt>63.8pt\cellcolor{lime!60}\else\ifdim#1pt>59.1pt\cellcolor{lime!50}\else\ifdim#1pt>54.4pt\cellcolor{lime!40}\else\ifdim#1pt>49.7pt\cellcolor{lime!30}\else\ifdim#1pt>45.0pt\cellcolor{lime!20}\else\ifdim#1pt>40.3pt\cellcolor{lime!10}\else\cellcolor{lime!0}\fi\fi\fi\fi\fi\fi\fi\fi\fi\fi#1}

\def\weightedspearmanzerozerosrcthresholdperadapterDSR#1{\ifdim#1pt>86.9pt\cellcolor{lime!100}\else\ifdim#1pt>82.1pt\cellcolor{lime!90}\else\ifdim#1pt>77.3pt\cellcolor{lime!80}\else\ifdim#1pt>72.5pt\cellcolor{lime!70}\else\ifdim#1pt>67.8pt\cellcolor{lime!60}\else\ifdim#1pt>63.0pt\cellcolor{lime!50}\else\ifdim#1pt>58.2pt\cellcolor{lime!40}\else\ifdim#1pt>53.4pt\cellcolor{lime!30}\else\ifdim#1pt>48.6pt\cellcolor{lime!20}\else\ifdim#1pt>43.8pt\cellcolor{lime!10}\else\cellcolor{lime!0}\fi\fi\fi\fi\fi\fi\fi\fi\fi\fi#1}

\def\weightedspearmanzerozerosrcthresholdperadapterMean#1{\ifdim#1pt>87.4pt\cellcolor{lime!100}\else\ifdim#1pt>84.4pt\cellcolor{lime!90}\else\ifdim#1pt>81.3pt\cellcolor{lime!80}\else\ifdim#1pt>78.3pt\cellcolor{lime!70}\else\ifdim#1pt>75.3pt\cellcolor{lime!60}\else\ifdim#1pt>72.3pt\cellcolor{lime!50}\else\ifdim#1pt>69.3pt\cellcolor{lime!40}\else\ifdim#1pt>66.2pt\cellcolor{lime!30}\else\ifdim#1pt>63.2pt\cellcolor{lime!20}\else\ifdim#1pt>60.2pt\cellcolor{lime!10}\else\cellcolor{lime!0}\fi\fi\fi\fi\fi\fi\fi\fi\fi\fi#1}

\def\weightedspearmanzerozerosrcthresholdperadapterStd#1{\ifdim#1pt<2.9pt\cellcolor{lime!100}\else\ifdim#1pt<3.8pt\cellcolor{lime!90}\else\ifdim#1pt<4.7pt\cellcolor{lime!80}\else\ifdim#1pt<5.6pt\cellcolor{lime!70}\else\ifdim#1pt<6.5pt\cellcolor{lime!60}\else\ifdim#1pt<7.4pt\cellcolor{lime!50}\else\ifdim#1pt<8.3pt\cellcolor{lime!40}\else\ifdim#1pt<9.2pt\cellcolor{lime!30}\else\ifdim#1pt<10.1pt\cellcolor{lime!20}\else\ifdim#1pt<11.0pt\cellcolor{lime!10}\else\cellcolor{lime!0}\fi\fi\fi\fi\fi\fi\fi\fi\fi\fi#1}

\begin{table}[H]
\centering
\caption{The weighted Spearman correlation of each validator/task pair for \textbf{BNM}.}
\label{domainnet126_weighted_spearman_0.0_src_threshold_per_adapter_BNM}
\resizebox{!}{0.22\textheight}{
\begin{tabular}{llrrrrrrrrrrrrrr}
\toprule
 &  & CP & CR & CS & PC & PR & PS & RC & RP & RS & SC & SP & SR & Mean & Std \\
\midrule
\multirow[c]{2}{*}{Accuracy} & Source Train & \weightedspearmanzerozerosrcthresholdperadapterDCP{78.8} & \weightedspearmanzerozerosrcthresholdperadapterDCR{78.0} & \weightedspearmanzerozerosrcthresholdperadapterDCS{68.6} & \weightedspearmanzerozerosrcthresholdperadapterDPC{66.6} & \weightedspearmanzerozerosrcthresholdperadapterDPR{77.6} & \weightedspearmanzerozerosrcthresholdperadapterDPS{48.2} & \weightedspearmanzerozerosrcthresholdperadapterDRC{75.3} & \weightedspearmanzerozerosrcthresholdperadapterDRP{73.6} & \weightedspearmanzerozerosrcthresholdperadapterDRS{52.8} & \weightedspearmanzerozerosrcthresholdperadapterDSC{71.0} & \weightedspearmanzerozerosrcthresholdperadapterDSP{57.0} & \weightedspearmanzerozerosrcthresholdperadapterDSR{73.2} & \weightedspearmanzerozerosrcthresholdperadapterMean{68.4} & \weightedspearmanzerozerosrcthresholdperadapterStd{9.9} \\
 & Source Val & \weightedspearmanzerozerosrcthresholdperadapterDCP{68.1} & \weightedspearmanzerozerosrcthresholdperadapterDCR{73.9} & \weightedspearmanzerozerosrcthresholdperadapterDCS{61.7} & \weightedspearmanzerozerosrcthresholdperadapterDPC{53.4} & \weightedspearmanzerozerosrcthresholdperadapterDPR{65.4} & \weightedspearmanzerozerosrcthresholdperadapterDPS{51.7} & \weightedspearmanzerozerosrcthresholdperadapterDRC{70.7} & \weightedspearmanzerozerosrcthresholdperadapterDRP{76.5} & \weightedspearmanzerozerosrcthresholdperadapterDRS{57.3} & \weightedspearmanzerozerosrcthresholdperadapterDSC{59.9} & \weightedspearmanzerozerosrcthresholdperadapterDSP{40.3} & \weightedspearmanzerozerosrcthresholdperadapterDSR{43.8} & \weightedspearmanzerozerosrcthresholdperadapterMean{60.2} & \weightedspearmanzerozerosrcthresholdperadapterStd{11.0} \\
\bottomrule
\end{tabular}}
\end{table}

%% file: supp_tables/domainnet126/weighted_spearman_0.0_src_threshold_per_adapter_BSP.tex
\def\weightedspearmanzerozerosrcthresholdperadapterDCP#1{\ifdim#1pt>95.2pt\cellcolor{lime!100}\else\ifdim#1pt>95.2pt\cellcolor{lime!90}\else\ifdim#1pt>95.1pt\cellcolor{lime!80}\else\ifdim#1pt>95.1pt\cellcolor{lime!70}\else\ifdim#1pt>95.1pt\cellcolor{lime!60}\else\ifdim#1pt>95.1pt\cellcolor{lime!50}\else\ifdim#1pt>95.1pt\cellcolor{lime!40}\else\ifdim#1pt>95.0pt\cellcolor{lime!30}\else\ifdim#1pt>95.0pt\cellcolor{lime!20}\else\ifdim#1pt>95.0pt\cellcolor{lime!10}\else\cellcolor{lime!0}\fi\fi\fi\fi\fi\fi\fi\fi\fi\fi#1}

\def\weightedspearmanzerozerosrcthresholdperadapterDCR#1{\ifdim#1pt>98.9pt\cellcolor{lime!100}\else\ifdim#1pt>98.9pt\cellcolor{lime!90}\else\ifdim#1pt>98.9pt\cellcolor{lime!80}\else\ifdim#1pt>98.9pt\cellcolor{lime!70}\else\ifdim#1pt>98.8pt\cellcolor{lime!60}\else\ifdim#1pt>98.8pt\cellcolor{lime!50}\else\ifdim#1pt>98.8pt\cellcolor{lime!40}\else\ifdim#1pt>98.8pt\cellcolor{lime!30}\else\ifdim#1pt>98.8pt\cellcolor{lime!20}\else\ifdim#1pt>98.8pt\cellcolor{lime!10}\else\cellcolor{lime!0}\fi\fi\fi\fi\fi\fi\fi\fi\fi\fi#1}

\def\weightedspearmanzerozerosrcthresholdperadapterDCS#1{\ifdim#1pt>98.3pt\cellcolor{lime!100}\else\ifdim#1pt>98.2pt\cellcolor{lime!90}\else\ifdim#1pt>98.2pt\cellcolor{lime!80}\else\ifdim#1pt>98.1pt\cellcolor{lime!70}\else\ifdim#1pt>98.1pt\cellcolor{lime!60}\else\ifdim#1pt>98.1pt\cellcolor{lime!50}\else\ifdim#1pt>98.0pt\cellcolor{lime!40}\else\ifdim#1pt>98.0pt\cellcolor{lime!30}\else\ifdim#1pt>97.9pt\cellcolor{lime!20}\else\ifdim#1pt>97.9pt\cellcolor{lime!10}\else\cellcolor{lime!0}\fi\fi\fi\fi\fi\fi\fi\fi\fi\fi#1}

\def\weightedspearmanzerozerosrcthresholdperadapterDPC#1{\ifdim#1pt>98.4pt\cellcolor{lime!100}\else\ifdim#1pt>98.4pt\cellcolor{lime!90}\else\ifdim#1pt>98.3pt\cellcolor{lime!80}\else\ifdim#1pt>98.3pt\cellcolor{lime!70}\else\ifdim#1pt>98.3pt\cellcolor{lime!60}\else\ifdim#1pt>98.3pt\cellcolor{lime!50}\else\ifdim#1pt>98.3pt\cellcolor{lime!40}\else\ifdim#1pt>98.2pt\cellcolor{lime!30}\else\ifdim#1pt>98.2pt\cellcolor{lime!20}\else\ifdim#1pt>98.2pt\cellcolor{lime!10}\else\cellcolor{lime!0}\fi\fi\fi\fi\fi\fi\fi\fi\fi\fi#1}

\def\weightedspearmanzerozerosrcthresholdperadapterDPR#1{\ifdim#1pt>98.5pt\cellcolor{lime!100}\else\ifdim#1pt>98.5pt\cellcolor{lime!90}\else\ifdim#1pt>98.4pt\cellcolor{lime!80}\else\ifdim#1pt>98.4pt\cellcolor{lime!70}\else\ifdim#1pt>98.3pt\cellcolor{lime!60}\else\ifdim#1pt>98.3pt\cellcolor{lime!50}\else\ifdim#1pt>98.2pt\cellcolor{lime!40}\else\ifdim#1pt>98.2pt\cellcolor{lime!30}\else\ifdim#1pt>98.1pt\cellcolor{lime!20}\else\ifdim#1pt>98.1pt\cellcolor{lime!10}\else\cellcolor{lime!0}\fi\fi\fi\fi\fi\fi\fi\fi\fi\fi#1}

\def\weightedspearmanzerozerosrcthresholdperadapterDPS#1{\ifdim#1pt>91.6pt\cellcolor{lime!100}\else\ifdim#1pt>91.6pt\cellcolor{lime!90}\else\ifdim#1pt>91.5pt\cellcolor{lime!80}\else\ifdim#1pt>91.4pt\cellcolor{lime!70}\else\ifdim#1pt>91.3pt\cellcolor{lime!60}\else\ifdim#1pt>91.3pt\cellcolor{lime!50}\else\ifdim#1pt>91.2pt\cellcolor{lime!40}\else\ifdim#1pt>91.1pt\cellcolor{lime!30}\else\ifdim#1pt>91.1pt\cellcolor{lime!20}\else\ifdim#1pt>91.0pt\cellcolor{lime!10}\else\cellcolor{lime!0}\fi\fi\fi\fi\fi\fi\fi\fi\fi\fi#1}

\def\weightedspearmanzerozerosrcthresholdperadapterDRC#1{\ifdim#1pt>97.6pt\cellcolor{lime!100}\else\ifdim#1pt>97.5pt\cellcolor{lime!90}\else\ifdim#1pt>97.5pt\cellcolor{lime!80}\else\ifdim#1pt>97.5pt\cellcolor{lime!70}\else\ifdim#1pt>97.4pt\cellcolor{lime!60}\else\ifdim#1pt>97.4pt\cellcolor{lime!50}\else\ifdim#1pt>97.4pt\cellcolor{lime!40}\else\ifdim#1pt>97.4pt\cellcolor{lime!30}\else\ifdim#1pt>97.3pt\cellcolor{lime!20}\else\ifdim#1pt>97.3pt\cellcolor{lime!10}\else\cellcolor{lime!0}\fi\fi\fi\fi\fi\fi\fi\fi\fi\fi#1}

\def\weightedspearmanzerozerosrcthresholdperadapterDRP#1{\ifdim#1pt>98.8pt\cellcolor{lime!100}\else\ifdim#1pt>98.8pt\cellcolor{lime!90}\else\ifdim#1pt>98.7pt\cellcolor{lime!80}\else\ifdim#1pt>98.6pt\cellcolor{lime!70}\else\ifdim#1pt>98.6pt\cellcolor{lime!60}\else\ifdim#1pt>98.5pt\cellcolor{lime!50}\else\ifdim#1pt>98.4pt\cellcolor{lime!40}\else\ifdim#1pt>98.3pt\cellcolor{lime!30}\else\ifdim#1pt>98.3pt\cellcolor{lime!20}\else\ifdim#1pt>98.2pt\cellcolor{lime!10}\else\cellcolor{lime!0}\fi\fi\fi\fi\fi\fi\fi\fi\fi\fi#1}

\def\weightedspearmanzerozerosrcthresholdperadapterDRS#1{\ifdim#1pt>96.4pt\cellcolor{lime!100}\else\ifdim#1pt>96.2pt\cellcolor{lime!90}\else\ifdim#1pt>95.9pt\cellcolor{lime!80}\else\ifdim#1pt>95.6pt\cellcolor{lime!70}\else\ifdim#1pt>95.3pt\cellcolor{lime!60}\else\ifdim#1pt>95.1pt\cellcolor{lime!50}\else\ifdim#1pt>94.8pt\cellcolor{lime!40}\else\ifdim#1pt>94.5pt\cellcolor{lime!30}\else\ifdim#1pt>94.3pt\cellcolor{lime!20}\else\ifdim#1pt>94.0pt\cellcolor{lime!10}\else\cellcolor{lime!0}\fi\fi\fi\fi\fi\fi\fi\fi\fi\fi#1}

\def\weightedspearmanzerozerosrcthresholdperadapterDSC#1{\ifdim#1pt>94.6pt\cellcolor{lime!100}\else\ifdim#1pt>94.6pt\cellcolor{lime!90}\else\ifdim#1pt>94.5pt\cellcolor{lime!80}\else\ifdim#1pt>94.5pt\cellcolor{lime!70}\else\ifdim#1pt>94.4pt\cellcolor{lime!60}\else\ifdim#1pt>94.3pt\cellcolor{lime!50}\else\ifdim#1pt>94.3pt\cellcolor{lime!40}\else\ifdim#1pt>94.2pt\cellcolor{lime!30}\else\ifdim#1pt>94.2pt\cellcolor{lime!20}\else\ifdim#1pt>94.1pt\cellcolor{lime!10}\else\cellcolor{lime!0}\fi\fi\fi\fi\fi\fi\fi\fi\fi\fi#1}

\def\weightedspearmanzerozerosrcthresholdperadapterDSP#1{\ifdim#1pt>95.1pt\cellcolor{lime!100}\else\ifdim#1pt>95.1pt\cellcolor{lime!90}\else\ifdim#1pt>95.1pt\cellcolor{lime!80}\else\ifdim#1pt>95.1pt\cellcolor{lime!70}\else\ifdim#1pt>95.1pt\cellcolor{lime!60}\else\ifdim#1pt>95.1pt\cellcolor{lime!50}\else\ifdim#1pt>95.1pt\cellcolor{lime!40}\else\ifdim#1pt>95.1pt\cellcolor{lime!30}\else\ifdim#1pt>95.1pt\cellcolor{lime!20}\else\ifdim#1pt>95.1pt\cellcolor{lime!10}\else\cellcolor{lime!0}\fi\fi\fi\fi\fi\fi\fi\fi\fi\fi#1}

\def\weightedspearmanzerozerosrcthresholdperadapterDSR#1{\ifdim#1pt>95.7pt\cellcolor{lime!100}\else\ifdim#1pt>95.7pt\cellcolor{lime!90}\else\ifdim#1pt>95.7pt\cellcolor{lime!80}\else\ifdim#1pt>95.7pt\cellcolor{lime!70}\else\ifdim#1pt>95.7pt\cellcolor{lime!60}\else\ifdim#1pt>95.6pt\cellcolor{lime!50}\else\ifdim#1pt>95.6pt\cellcolor{lime!40}\else\ifdim#1pt>95.6pt\cellcolor{lime!30}\else\ifdim#1pt>95.6pt\cellcolor{lime!20}\else\ifdim#1pt>95.6pt\cellcolor{lime!10}\else\cellcolor{lime!0}\fi\fi\fi\fi\fi\fi\fi\fi\fi\fi#1}

\def\weightedspearmanzerozerosrcthresholdperadapterMean#1{\ifdim#1pt>96.3pt\cellcolor{lime!100}\else\ifdim#1pt>96.3pt\cellcolor{lime!90}\else\ifdim#1pt>96.2pt\cellcolor{lime!80}\else\ifdim#1pt>96.2pt\cellcolor{lime!70}\else\ifdim#1pt>96.2pt\cellcolor{lime!60}\else\ifdim#1pt>96.2pt\cellcolor{lime!50}\else\ifdim#1pt>96.2pt\cellcolor{lime!40}\else\ifdim#1pt>96.1pt\cellcolor{lime!30}\else\ifdim#1pt>96.1pt\cellcolor{lime!20}\else\ifdim#1pt>96.1pt\cellcolor{lime!10}\else\cellcolor{lime!0}\fi\fi\fi\fi\fi\fi\fi\fi\fi\fi#1}

\def\weightedspearmanzerozerosrcthresholdperadapterStd#1{\ifdim#1pt<2.3pt\cellcolor{lime!100}\else\ifdim#1pt<2.3pt\cellcolor{lime!90}\else\ifdim#1pt<2.3pt\cellcolor{lime!80}\else\ifdim#1pt<2.3pt\cellcolor{lime!70}\else\ifdim#1pt<2.3pt\cellcolor{lime!60}\else\ifdim#1pt<2.3pt\cellcolor{lime!50}\else\ifdim#1pt<2.3pt\cellcolor{lime!40}\else\ifdim#1pt<2.3pt\cellcolor{lime!30}\else\ifdim#1pt<2.3pt\cellcolor{lime!20}\else\ifdim#1pt<2.3pt\cellcolor{lime!10}\else\cellcolor{lime!0}\fi\fi\fi\fi\fi\fi\fi\fi\fi\fi#1}

\begin{table}[H]
\centering
\caption{The weighted Spearman correlation of each validator/task pair for \textbf{BSP}.}
\label{domainnet126_weighted_spearman_0.0_src_threshold_per_adapter_BSP}
\resizebox{!}{0.22\textheight}{
\begin{tabular}{llrrrrrrrrrrrrrr}
\toprule
 &  & CP & CR & CS & PC & PR & PS & RC & RP & RS & SC & SP & SR & Mean & Std \\
\midrule
\multirow[c]{2}{*}{Accuracy} & Source Train & \weightedspearmanzerozerosrcthresholdperadapterDCP{92.5} & \weightedspearmanzerozerosrcthresholdperadapterDCR{89.5} & \weightedspearmanzerozerosrcthresholdperadapterDCS{79.1} & \weightedspearmanzerozerosrcthresholdperadapterDPC{93.7} & \weightedspearmanzerozerosrcthresholdperadapterDPR{60.6} & \weightedspearmanzerozerosrcthresholdperadapterDPS{88.9} & \weightedspearmanzerozerosrcthresholdperadapterDRC{93.2} & \weightedspearmanzerozerosrcthresholdperadapterDRP{87.7} & \weightedspearmanzerozerosrcthresholdperadapterDRS{89.8} & \weightedspearmanzerozerosrcthresholdperadapterDSC{83.4} & \weightedspearmanzerozerosrcthresholdperadapterDSP{92.4} & \weightedspearmanzerozerosrcthresholdperadapterDSR{85.8} & \weightedspearmanzerozerosrcthresholdperadapterMean{86.4} & \weightedspearmanzerozerosrcthresholdperadapterStd{8.8} \\
 & Source Val & \weightedspearmanzerozerosrcthresholdperadapterDCP{95.0} & \weightedspearmanzerozerosrcthresholdperadapterDCR{98.8} & \weightedspearmanzerozerosrcthresholdperadapterDCS{97.9} & \weightedspearmanzerozerosrcthresholdperadapterDPC{98.2} & \weightedspearmanzerozerosrcthresholdperadapterDPR{98.1} & \weightedspearmanzerozerosrcthresholdperadapterDPS{91.0} & \weightedspearmanzerozerosrcthresholdperadapterDRC{97.3} & \weightedspearmanzerozerosrcthresholdperadapterDRP{98.2} & \weightedspearmanzerozerosrcthresholdperadapterDRS{94.0} & \weightedspearmanzerozerosrcthresholdperadapterDSC{94.1} & \textbf{\weightedspearmanzerozerosrcthresholdperadapterDSP{95.1}} & \weightedspearmanzerozerosrcthresholdperadapterDSR{95.6} & \weightedspearmanzerozerosrcthresholdperadapterMean{96.1} & \textbf{\weightedspearmanzerozerosrcthresholdperadapterStd{2.3}} \\
\bottomrule
\end{tabular}}
\end{table}

%% file: supp_tables/domainnet126/weighted_spearman_0.0_src_threshold_per_adapter_CDAN.tex
\def\weightedspearmanzerozerosrcthresholdperadapterDCP#1{\ifdim#1pt>82.8pt\cellcolor{lime!100}\else\ifdim#1pt>80.9pt\cellcolor{lime!90}\else\ifdim#1pt>78.9pt\cellcolor{lime!80}\else\ifdim#1pt>77.0pt\cellcolor{lime!70}\else\ifdim#1pt>75.0pt\cellcolor{lime!60}\else\ifdim#1pt>73.0pt\cellcolor{lime!50}\else\ifdim#1pt>71.1pt\cellcolor{lime!40}\else\ifdim#1pt>69.1pt\cellcolor{lime!30}\else\ifdim#1pt>67.2pt\cellcolor{lime!20}\else\ifdim#1pt>65.2pt\cellcolor{lime!10}\else\cellcolor{lime!0}\fi\fi\fi\fi\fi\fi\fi\fi\fi\fi#1}

\def\weightedspearmanzerozerosrcthresholdperadapterDCR#1{\ifdim#1pt>84.4pt\cellcolor{lime!100}\else\ifdim#1pt>81.8pt\cellcolor{lime!90}\else\ifdim#1pt>79.1pt\cellcolor{lime!80}\else\ifdim#1pt>76.4pt\cellcolor{lime!70}\else\ifdim#1pt>73.8pt\cellcolor{lime!60}\else\ifdim#1pt>71.1pt\cellcolor{lime!50}\else\ifdim#1pt>68.4pt\cellcolor{lime!40}\else\ifdim#1pt>65.7pt\cellcolor{lime!30}\else\ifdim#1pt>63.1pt\cellcolor{lime!20}\else\ifdim#1pt>60.4pt\cellcolor{lime!10}\else\cellcolor{lime!0}\fi\fi\fi\fi\fi\fi\fi\fi\fi\fi#1}

\def\weightedspearmanzerozerosrcthresholdperadapterDCS#1{\ifdim#1pt>82.5pt\cellcolor{lime!100}\else\ifdim#1pt>81.5pt\cellcolor{lime!90}\else\ifdim#1pt>80.6pt\cellcolor{lime!80}\else\ifdim#1pt>79.6pt\cellcolor{lime!70}\else\ifdim#1pt>78.7pt\cellcolor{lime!60}\else\ifdim#1pt>77.8pt\cellcolor{lime!50}\else\ifdim#1pt>76.8pt\cellcolor{lime!40}\else\ifdim#1pt>75.9pt\cellcolor{lime!30}\else\ifdim#1pt>74.9pt\cellcolor{lime!20}\else\ifdim#1pt>74.0pt\cellcolor{lime!10}\else\cellcolor{lime!0}\fi\fi\fi\fi\fi\fi\fi\fi\fi\fi#1}

\def\weightedspearmanzerozerosrcthresholdperadapterDPC#1{\ifdim#1pt>86.4pt\cellcolor{lime!100}\else\ifdim#1pt>84.6pt\cellcolor{lime!90}\else\ifdim#1pt>82.7pt\cellcolor{lime!80}\else\ifdim#1pt>80.8pt\cellcolor{lime!70}\else\ifdim#1pt>78.9pt\cellcolor{lime!60}\else\ifdim#1pt>77.1pt\cellcolor{lime!50}\else\ifdim#1pt>75.2pt\cellcolor{lime!40}\else\ifdim#1pt>73.3pt\cellcolor{lime!30}\else\ifdim#1pt>71.5pt\cellcolor{lime!20}\else\ifdim#1pt>69.6pt\cellcolor{lime!10}\else\cellcolor{lime!0}\fi\fi\fi\fi\fi\fi\fi\fi\fi\fi#1}

\def\weightedspearmanzerozerosrcthresholdperadapterDPR#1{\ifdim#1pt>96.9pt\cellcolor{lime!100}\else\ifdim#1pt>96.9pt\cellcolor{lime!90}\else\ifdim#1pt>96.9pt\cellcolor{lime!80}\else\ifdim#1pt>96.9pt\cellcolor{lime!70}\else\ifdim#1pt>96.9pt\cellcolor{lime!60}\else\ifdim#1pt>96.9pt\cellcolor{lime!50}\else\ifdim#1pt>96.9pt\cellcolor{lime!40}\else\ifdim#1pt>96.9pt\cellcolor{lime!30}\else\ifdim#1pt>96.9pt\cellcolor{lime!20}\else\ifdim#1pt>96.9pt\cellcolor{lime!10}\else\cellcolor{lime!0}\fi\fi\fi\fi\fi\fi\fi\fi\fi\fi#1}

\def\weightedspearmanzerozerosrcthresholdperadapterDPS#1{\ifdim#1pt>73.4pt\cellcolor{lime!100}\else\ifdim#1pt>69.9pt\cellcolor{lime!90}\else\ifdim#1pt>66.5pt\cellcolor{lime!80}\else\ifdim#1pt>63.0pt\cellcolor{lime!70}\else\ifdim#1pt>59.5pt\cellcolor{lime!60}\else\ifdim#1pt>56.0pt\cellcolor{lime!50}\else\ifdim#1pt>52.5pt\cellcolor{lime!40}\else\ifdim#1pt>49.1pt\cellcolor{lime!30}\else\ifdim#1pt>45.6pt\cellcolor{lime!20}\else\ifdim#1pt>42.1pt\cellcolor{lime!10}\else\cellcolor{lime!0}\fi\fi\fi\fi\fi\fi\fi\fi\fi\fi#1}

\def\weightedspearmanzerozerosrcthresholdperadapterDRC#1{\ifdim#1pt>91.0pt\cellcolor{lime!100}\else\ifdim#1pt>88.5pt\cellcolor{lime!90}\else\ifdim#1pt>86.1pt\cellcolor{lime!80}\else\ifdim#1pt>83.6pt\cellcolor{lime!70}\else\ifdim#1pt>81.2pt\cellcolor{lime!60}\else\ifdim#1pt>78.7pt\cellcolor{lime!50}\else\ifdim#1pt>76.2pt\cellcolor{lime!40}\else\ifdim#1pt>73.8pt\cellcolor{lime!30}\else\ifdim#1pt>71.4pt\cellcolor{lime!20}\else\ifdim#1pt>68.9pt\cellcolor{lime!10}\else\cellcolor{lime!0}\fi\fi\fi\fi\fi\fi\fi\fi\fi\fi#1}

\def\weightedspearmanzerozerosrcthresholdperadapterDRP#1{\ifdim#1pt>85.6pt\cellcolor{lime!100}\else\ifdim#1pt>85.6pt\cellcolor{lime!90}\else\ifdim#1pt>85.6pt\cellcolor{lime!80}\else\ifdim#1pt>85.6pt\cellcolor{lime!70}\else\ifdim#1pt>85.6pt\cellcolor{lime!60}\else\ifdim#1pt>85.6pt\cellcolor{lime!50}\else\ifdim#1pt>85.6pt\cellcolor{lime!40}\else\ifdim#1pt>85.6pt\cellcolor{lime!30}\else\ifdim#1pt>85.6pt\cellcolor{lime!20}\else\ifdim#1pt>85.6pt\cellcolor{lime!10}\else\cellcolor{lime!0}\fi\fi\fi\fi\fi\fi\fi\fi\fi\fi#1}

\def\weightedspearmanzerozerosrcthresholdperadapterDRS#1{\ifdim#1pt>87.1pt\cellcolor{lime!100}\else\ifdim#1pt>82.9pt\cellcolor{lime!90}\else\ifdim#1pt>78.8pt\cellcolor{lime!80}\else\ifdim#1pt>74.6pt\cellcolor{lime!70}\else\ifdim#1pt>70.5pt\cellcolor{lime!60}\else\ifdim#1pt>66.4pt\cellcolor{lime!50}\else\ifdim#1pt>62.2pt\cellcolor{lime!40}\else\ifdim#1pt>58.1pt\cellcolor{lime!30}\else\ifdim#1pt>53.9pt\cellcolor{lime!20}\else\ifdim#1pt>49.8pt\cellcolor{lime!10}\else\cellcolor{lime!0}\fi\fi\fi\fi\fi\fi\fi\fi\fi\fi#1}

\def\weightedspearmanzerozerosrcthresholdperadapterDSC#1{\ifdim#1pt>79.8pt\cellcolor{lime!100}\else\ifdim#1pt>77.6pt\cellcolor{lime!90}\else\ifdim#1pt>75.5pt\cellcolor{lime!80}\else\ifdim#1pt>73.3pt\cellcolor{lime!70}\else\ifdim#1pt>71.2pt\cellcolor{lime!60}\else\ifdim#1pt>69.1pt\cellcolor{lime!50}\else\ifdim#1pt>66.9pt\cellcolor{lime!40}\else\ifdim#1pt>64.8pt\cellcolor{lime!30}\else\ifdim#1pt>62.6pt\cellcolor{lime!20}\else\ifdim#1pt>60.5pt\cellcolor{lime!10}\else\cellcolor{lime!0}\fi\fi\fi\fi\fi\fi\fi\fi\fi\fi#1}

\def\weightedspearmanzerozerosrcthresholdperadapterDSP#1{\ifdim#1pt>85.7pt\cellcolor{lime!100}\else\ifdim#1pt>82.2pt\cellcolor{lime!90}\else\ifdim#1pt>78.7pt\cellcolor{lime!80}\else\ifdim#1pt>75.2pt\cellcolor{lime!70}\else\ifdim#1pt>71.7pt\cellcolor{lime!60}\else\ifdim#1pt>68.2pt\cellcolor{lime!50}\else\ifdim#1pt>64.7pt\cellcolor{lime!40}\else\ifdim#1pt>61.2pt\cellcolor{lime!30}\else\ifdim#1pt>57.7pt\cellcolor{lime!20}\else\ifdim#1pt>54.2pt\cellcolor{lime!10}\else\cellcolor{lime!0}\fi\fi\fi\fi\fi\fi\fi\fi\fi\fi#1}

\def\weightedspearmanzerozerosrcthresholdperadapterDSR#1{\ifdim#1pt>89.5pt\cellcolor{lime!100}\else\ifdim#1pt>85.3pt\cellcolor{lime!90}\else\ifdim#1pt>81.2pt\cellcolor{lime!80}\else\ifdim#1pt>77.0pt\cellcolor{lime!70}\else\ifdim#1pt>72.9pt\cellcolor{lime!60}\else\ifdim#1pt>68.8pt\cellcolor{lime!50}\else\ifdim#1pt>64.6pt\cellcolor{lime!40}\else\ifdim#1pt>60.5pt\cellcolor{lime!30}\else\ifdim#1pt>56.3pt\cellcolor{lime!20}\else\ifdim#1pt>52.2pt\cellcolor{lime!10}\else\cellcolor{lime!0}\fi\fi\fi\fi\fi\fi\fi\fi\fi\fi#1}

\def\weightedspearmanzerozerosrcthresholdperadapterMean#1{\ifdim#1pt>79.2pt\cellcolor{lime!100}\else\ifdim#1pt>77.6pt\cellcolor{lime!90}\else\ifdim#1pt>76.1pt\cellcolor{lime!80}\else\ifdim#1pt>74.5pt\cellcolor{lime!70}\else\ifdim#1pt>72.9pt\cellcolor{lime!60}\else\ifdim#1pt>71.3pt\cellcolor{lime!50}\else\ifdim#1pt>69.7pt\cellcolor{lime!40}\else\ifdim#1pt>68.2pt\cellcolor{lime!30}\else\ifdim#1pt>66.6pt\cellcolor{lime!20}\else\ifdim#1pt>65.0pt\cellcolor{lime!10}\else\cellcolor{lime!0}\fi\fi\fi\fi\fi\fi\fi\fi\fi\fi#1}

\def\weightedspearmanzerozerosrcthresholdperadapterStd#1{\ifdim#1pt<2.2pt\cellcolor{lime!100}\else\ifdim#1pt<3.6pt\cellcolor{lime!90}\else\ifdim#1pt<5.0pt\cellcolor{lime!80}\else\ifdim#1pt<6.4pt\cellcolor{lime!70}\else\ifdim#1pt<7.8pt\cellcolor{lime!60}\else\ifdim#1pt<9.2pt\cellcolor{lime!50}\else\ifdim#1pt<10.6pt\cellcolor{lime!40}\else\ifdim#1pt<12.0pt\cellcolor{lime!30}\else\ifdim#1pt<13.4pt\cellcolor{lime!20}\else\ifdim#1pt<14.8pt\cellcolor{lime!10}\else\cellcolor{lime!0}\fi\fi\fi\fi\fi\fi\fi\fi\fi\fi#1}

\begin{table}[H]
\centering
\caption{The weighted Spearman correlation of each validator/task pair for \textbf{CDAN}.}
\label{domainnet126_weighted_spearman_0.0_src_threshold_per_adapter_CDAN}
\resizebox{!}{0.22\textheight}{
\begin{tabular}{llrrrrrrrrrrrrrr}
\toprule
 &  & CP & CR & CS & PC & PR & PS & RC & RP & RS & SC & SP & SR & Mean & Std \\
\midrule
\multirow[c]{2}{*}{Accuracy} & Source Train & \weightedspearmanzerozerosrcthresholdperadapterDCP{83.6} & \weightedspearmanzerozerosrcthresholdperadapterDCR{83.8} & \textbf{\weightedspearmanzerozerosrcthresholdperadapterDCS{83.4}} & \weightedspearmanzerozerosrcthresholdperadapterDPC{84.4} & \weightedspearmanzerozerosrcthresholdperadapterDPR{50.5} & \weightedspearmanzerozerosrcthresholdperadapterDPS{61.7} & \weightedspearmanzerozerosrcthresholdperadapterDRC{72.1} & \weightedspearmanzerozerosrcthresholdperadapterDRP{80.4} & \weightedspearmanzerozerosrcthresholdperadapterDRS{62.0} & \weightedspearmanzerozerosrcthresholdperadapterDSC{81.8} & \weightedspearmanzerozerosrcthresholdperadapterDSP{83.9} & \weightedspearmanzerozerosrcthresholdperadapterDSR{78.8} & \weightedspearmanzerozerosrcthresholdperadapterMean{75.5} & \weightedspearmanzerozerosrcthresholdperadapterStd{10.9} \\
 & Source Val & \weightedspearmanzerozerosrcthresholdperadapterDCP{65.2} & \weightedspearmanzerozerosrcthresholdperadapterDCR{60.4} & \weightedspearmanzerozerosrcthresholdperadapterDCS{74.0} & \weightedspearmanzerozerosrcthresholdperadapterDPC{69.6} & \textbf{\weightedspearmanzerozerosrcthresholdperadapterDPR{96.9}} & \weightedspearmanzerozerosrcthresholdperadapterDPS{42.1} & \weightedspearmanzerozerosrcthresholdperadapterDRC{68.9} & \textbf{\weightedspearmanzerozerosrcthresholdperadapterDRP{85.6}} & \weightedspearmanzerozerosrcthresholdperadapterDRS{49.8} & \weightedspearmanzerozerosrcthresholdperadapterDSC{60.5} & \weightedspearmanzerozerosrcthresholdperadapterDSP{54.2} & \weightedspearmanzerozerosrcthresholdperadapterDSR{52.2} & \weightedspearmanzerozerosrcthresholdperadapterMean{65.0} & \weightedspearmanzerozerosrcthresholdperadapterStd{14.8} \\
\bottomrule
\end{tabular}}
\end{table}

%% file: supp_tables/domainnet126/weighted_spearman_0.0_src_threshold_per_adapter_DANN.tex
\def\weightedspearmanzerozerosrcthresholdperadapterDCP#1{\ifdim#1pt>74.6pt\cellcolor{lime!100}\else\ifdim#1pt>72.2pt\cellcolor{lime!90}\else\ifdim#1pt>69.9pt\cellcolor{lime!80}\else\ifdim#1pt>67.5pt\cellcolor{lime!70}\else\ifdim#1pt>65.2pt\cellcolor{lime!60}\else\ifdim#1pt>62.9pt\cellcolor{lime!50}\else\ifdim#1pt>60.5pt\cellcolor{lime!40}\else\ifdim#1pt>58.2pt\cellcolor{lime!30}\else\ifdim#1pt>55.8pt\cellcolor{lime!20}\else\ifdim#1pt>53.5pt\cellcolor{lime!10}\else\cellcolor{lime!0}\fi\fi\fi\fi\fi\fi\fi\fi\fi\fi#1}

\def\weightedspearmanzerozerosrcthresholdperadapterDCR#1{\ifdim#1pt>89.9pt\cellcolor{lime!100}\else\ifdim#1pt>87.1pt\cellcolor{lime!90}\else\ifdim#1pt>84.3pt\cellcolor{lime!80}\else\ifdim#1pt>81.5pt\cellcolor{lime!70}\else\ifdim#1pt>78.7pt\cellcolor{lime!60}\else\ifdim#1pt>75.8pt\cellcolor{lime!50}\else\ifdim#1pt>73.0pt\cellcolor{lime!40}\else\ifdim#1pt>70.2pt\cellcolor{lime!30}\else\ifdim#1pt>67.4pt\cellcolor{lime!20}\else\ifdim#1pt>64.6pt\cellcolor{lime!10}\else\cellcolor{lime!0}\fi\fi\fi\fi\fi\fi\fi\fi\fi\fi#1}

\def\weightedspearmanzerozerosrcthresholdperadapterDCS#1{\ifdim#1pt>83.6pt\cellcolor{lime!100}\else\ifdim#1pt>82.6pt\cellcolor{lime!90}\else\ifdim#1pt>81.5pt\cellcolor{lime!80}\else\ifdim#1pt>80.5pt\cellcolor{lime!70}\else\ifdim#1pt>79.4pt\cellcolor{lime!60}\else\ifdim#1pt>78.3pt\cellcolor{lime!50}\else\ifdim#1pt>77.3pt\cellcolor{lime!40}\else\ifdim#1pt>76.2pt\cellcolor{lime!30}\else\ifdim#1pt>75.2pt\cellcolor{lime!20}\else\ifdim#1pt>74.1pt\cellcolor{lime!10}\else\cellcolor{lime!0}\fi\fi\fi\fi\fi\fi\fi\fi\fi\fi#1}

\def\weightedspearmanzerozerosrcthresholdperadapterDPC#1{\ifdim#1pt>77.6pt\cellcolor{lime!100}\else\ifdim#1pt>76.0pt\cellcolor{lime!90}\else\ifdim#1pt>74.4pt\cellcolor{lime!80}\else\ifdim#1pt>72.8pt\cellcolor{lime!70}\else\ifdim#1pt>71.2pt\cellcolor{lime!60}\else\ifdim#1pt>69.5pt\cellcolor{lime!50}\else\ifdim#1pt>67.9pt\cellcolor{lime!40}\else\ifdim#1pt>66.3pt\cellcolor{lime!30}\else\ifdim#1pt>64.7pt\cellcolor{lime!20}\else\ifdim#1pt>63.1pt\cellcolor{lime!10}\else\cellcolor{lime!0}\fi\fi\fi\fi\fi\fi\fi\fi\fi\fi#1}

\def\weightedspearmanzerozerosrcthresholdperadapterDPR#1{\ifdim#1pt>94.7pt\cellcolor{lime!100}\else\ifdim#1pt>94.7pt\cellcolor{lime!90}\else\ifdim#1pt>94.7pt\cellcolor{lime!80}\else\ifdim#1pt>94.7pt\cellcolor{lime!70}\else\ifdim#1pt>94.7pt\cellcolor{lime!60}\else\ifdim#1pt>94.7pt\cellcolor{lime!50}\else\ifdim#1pt>94.7pt\cellcolor{lime!40}\else\ifdim#1pt>94.7pt\cellcolor{lime!30}\else\ifdim#1pt>94.7pt\cellcolor{lime!20}\else\ifdim#1pt>94.7pt\cellcolor{lime!10}\else\cellcolor{lime!0}\fi\fi\fi\fi\fi\fi\fi\fi\fi\fi#1}

\def\weightedspearmanzerozerosrcthresholdperadapterDPS#1{\ifdim#1pt>83.0pt\cellcolor{lime!100}\else\ifdim#1pt>80.2pt\cellcolor{lime!90}\else\ifdim#1pt>77.3pt\cellcolor{lime!80}\else\ifdim#1pt>74.5pt\cellcolor{lime!70}\else\ifdim#1pt>71.7pt\cellcolor{lime!60}\else\ifdim#1pt>68.9pt\cellcolor{lime!50}\else\ifdim#1pt>66.1pt\cellcolor{lime!40}\else\ifdim#1pt>63.2pt\cellcolor{lime!30}\else\ifdim#1pt>60.4pt\cellcolor{lime!20}\else\ifdim#1pt>57.6pt\cellcolor{lime!10}\else\cellcolor{lime!0}\fi\fi\fi\fi\fi\fi\fi\fi\fi\fi#1}

\def\weightedspearmanzerozerosrcthresholdperadapterDRC#1{\ifdim#1pt>90.8pt\cellcolor{lime!100}\else\ifdim#1pt>88.4pt\cellcolor{lime!90}\else\ifdim#1pt>85.9pt\cellcolor{lime!80}\else\ifdim#1pt>83.5pt\cellcolor{lime!70}\else\ifdim#1pt>81.0pt\cellcolor{lime!60}\else\ifdim#1pt>78.5pt\cellcolor{lime!50}\else\ifdim#1pt>76.1pt\cellcolor{lime!40}\else\ifdim#1pt>73.6pt\cellcolor{lime!30}\else\ifdim#1pt>71.2pt\cellcolor{lime!20}\else\ifdim#1pt>68.7pt\cellcolor{lime!10}\else\cellcolor{lime!0}\fi\fi\fi\fi\fi\fi\fi\fi\fi\fi#1}

\def\weightedspearmanzerozerosrcthresholdperadapterDRP#1{\ifdim#1pt>91.1pt\cellcolor{lime!100}\else\ifdim#1pt>91.1pt\cellcolor{lime!90}\else\ifdim#1pt>91.1pt\cellcolor{lime!80}\else\ifdim#1pt>91.1pt\cellcolor{lime!70}\else\ifdim#1pt>91.1pt\cellcolor{lime!60}\else\ifdim#1pt>91.1pt\cellcolor{lime!50}\else\ifdim#1pt>91.1pt\cellcolor{lime!40}\else\ifdim#1pt>91.1pt\cellcolor{lime!30}\else\ifdim#1pt>91.1pt\cellcolor{lime!20}\else\ifdim#1pt>91.1pt\cellcolor{lime!10}\else\cellcolor{lime!0}\fi\fi\fi\fi\fi\fi\fi\fi\fi\fi#1}

\def\weightedspearmanzerozerosrcthresholdperadapterDRS#1{\ifdim#1pt>84.2pt\cellcolor{lime!100}\else\ifdim#1pt>80.5pt\cellcolor{lime!90}\else\ifdim#1pt>76.7pt\cellcolor{lime!80}\else\ifdim#1pt>73.0pt\cellcolor{lime!70}\else\ifdim#1pt>69.2pt\cellcolor{lime!60}\else\ifdim#1pt>65.4pt\cellcolor{lime!50}\else\ifdim#1pt>61.7pt\cellcolor{lime!40}\else\ifdim#1pt>57.9pt\cellcolor{lime!30}\else\ifdim#1pt>54.2pt\cellcolor{lime!20}\else\ifdim#1pt>50.4pt\cellcolor{lime!10}\else\cellcolor{lime!0}\fi\fi\fi\fi\fi\fi\fi\fi\fi\fi#1}

\def\weightedspearmanzerozerosrcthresholdperadapterDSC#1{\ifdim#1pt>69.5pt\cellcolor{lime!100}\else\ifdim#1pt>67.9pt\cellcolor{lime!90}\else\ifdim#1pt>66.4pt\cellcolor{lime!80}\else\ifdim#1pt>64.8pt\cellcolor{lime!70}\else\ifdim#1pt>63.2pt\cellcolor{lime!60}\else\ifdim#1pt>61.6pt\cellcolor{lime!50}\else\ifdim#1pt>60.0pt\cellcolor{lime!40}\else\ifdim#1pt>58.5pt\cellcolor{lime!30}\else\ifdim#1pt>56.9pt\cellcolor{lime!20}\else\ifdim#1pt>55.3pt\cellcolor{lime!10}\else\cellcolor{lime!0}\fi\fi\fi\fi\fi\fi\fi\fi\fi\fi#1}

\def\weightedspearmanzerozerosrcthresholdperadapterDSP#1{\ifdim#1pt>85.7pt\cellcolor{lime!100}\else\ifdim#1pt>82.0pt\cellcolor{lime!90}\else\ifdim#1pt>78.4pt\cellcolor{lime!80}\else\ifdim#1pt>74.7pt\cellcolor{lime!70}\else\ifdim#1pt>71.0pt\cellcolor{lime!60}\else\ifdim#1pt>67.3pt\cellcolor{lime!50}\else\ifdim#1pt>63.6pt\cellcolor{lime!40}\else\ifdim#1pt>60.0pt\cellcolor{lime!30}\else\ifdim#1pt>56.3pt\cellcolor{lime!20}\else\ifdim#1pt>52.6pt\cellcolor{lime!10}\else\cellcolor{lime!0}\fi\fi\fi\fi\fi\fi\fi\fi\fi\fi#1}

\def\weightedspearmanzerozerosrcthresholdperadapterDSR#1{\ifdim#1pt>88.9pt\cellcolor{lime!100}\else\ifdim#1pt>83.6pt\cellcolor{lime!90}\else\ifdim#1pt>78.2pt\cellcolor{lime!80}\else\ifdim#1pt>72.9pt\cellcolor{lime!70}\else\ifdim#1pt>67.5pt\cellcolor{lime!60}\else\ifdim#1pt>62.1pt\cellcolor{lime!50}\else\ifdim#1pt>56.8pt\cellcolor{lime!40}\else\ifdim#1pt>51.4pt\cellcolor{lime!30}\else\ifdim#1pt>46.1pt\cellcolor{lime!20}\else\ifdim#1pt>40.7pt\cellcolor{lime!10}\else\cellcolor{lime!0}\fi\fi\fi\fi\fi\fi\fi\fi\fi\fi#1}

\def\weightedspearmanzerozerosrcthresholdperadapterMean#1{\ifdim#1pt>79.5pt\cellcolor{lime!100}\else\ifdim#1pt>77.7pt\cellcolor{lime!90}\else\ifdim#1pt>76.0pt\cellcolor{lime!80}\else\ifdim#1pt>74.3pt\cellcolor{lime!70}\else\ifdim#1pt>72.5pt\cellcolor{lime!60}\else\ifdim#1pt>70.8pt\cellcolor{lime!50}\else\ifdim#1pt>69.1pt\cellcolor{lime!40}\else\ifdim#1pt>67.4pt\cellcolor{lime!30}\else\ifdim#1pt>65.6pt\cellcolor{lime!20}\else\ifdim#1pt>63.9pt\cellcolor{lime!10}\else\cellcolor{lime!0}\fi\fi\fi\fi\fi\fi\fi\fi\fi\fi#1}

\def\weightedspearmanzerozerosrcthresholdperadapterStd#1{\ifdim#1pt<2.4pt\cellcolor{lime!100}\else\ifdim#1pt<3.8pt\cellcolor{lime!90}\else\ifdim#1pt<5.3pt\cellcolor{lime!80}\else\ifdim#1pt<6.7pt\cellcolor{lime!70}\else\ifdim#1pt<8.2pt\cellcolor{lime!60}\else\ifdim#1pt<9.7pt\cellcolor{lime!50}\else\ifdim#1pt<11.1pt\cellcolor{lime!40}\else\ifdim#1pt<12.6pt\cellcolor{lime!30}\else\ifdim#1pt<14.0pt\cellcolor{lime!20}\else\ifdim#1pt<15.5pt\cellcolor{lime!10}\else\cellcolor{lime!0}\fi\fi\fi\fi\fi\fi\fi\fi\fi\fi#1}

\begin{table}[H]
\centering
\caption{The weighted Spearman correlation of each validator/task pair for \textbf{DANN}.}
\label{domainnet126_weighted_spearman_0.0_src_threshold_per_adapter_DANN}
\resizebox{!}{0.22\textheight}{
\begin{tabular}{llrrrrrrrrrrrrrr}
\toprule
 &  & CP & CR & CS & PC & PR & PS & RC & RP & RS & SC & SP & SR & Mean & Std \\
\midrule
\multirow[c]{2}{*}{Accuracy} & Source Train & \textbf{\weightedspearmanzerozerosrcthresholdperadapterDCP{76.9}} & \weightedspearmanzerozerosrcthresholdperadapterDCR{87.2} & \weightedspearmanzerozerosrcthresholdperadapterDCS{76.1} & \weightedspearmanzerozerosrcthresholdperadapterDPC{60.6} & \weightedspearmanzerozerosrcthresholdperadapterDPR{67.4} & \weightedspearmanzerozerosrcthresholdperadapterDPS{66.9} & \weightedspearmanzerozerosrcthresholdperadapterDRC{74.3} & \weightedspearmanzerozerosrcthresholdperadapterDRP{68.7} & \weightedspearmanzerozerosrcthresholdperadapterDRS{58.4} & \weightedspearmanzerozerosrcthresholdperadapterDSC{56.5} & \weightedspearmanzerozerosrcthresholdperadapterDSP{86.3} & \weightedspearmanzerozerosrcthresholdperadapterDSR{82.7} & \weightedspearmanzerozerosrcthresholdperadapterMean{71.8} & \weightedspearmanzerozerosrcthresholdperadapterStd{10.0} \\
 & Source Val & \weightedspearmanzerozerosrcthresholdperadapterDCP{53.5} & \weightedspearmanzerozerosrcthresholdperadapterDCR{64.6} & \weightedspearmanzerozerosrcthresholdperadapterDCS{74.1} & \weightedspearmanzerozerosrcthresholdperadapterDPC{63.1} & \textbf{\weightedspearmanzerozerosrcthresholdperadapterDPR{94.7}} & \weightedspearmanzerozerosrcthresholdperadapterDPS{57.6} & \weightedspearmanzerozerosrcthresholdperadapterDRC{68.7} & \textbf{\weightedspearmanzerozerosrcthresholdperadapterDRP{91.1}} & \weightedspearmanzerozerosrcthresholdperadapterDRS{50.4} & \weightedspearmanzerozerosrcthresholdperadapterDSC{55.3} & \weightedspearmanzerozerosrcthresholdperadapterDSP{52.6} & \weightedspearmanzerozerosrcthresholdperadapterDSR{40.7} & \weightedspearmanzerozerosrcthresholdperadapterMean{63.9} & \weightedspearmanzerozerosrcthresholdperadapterStd{15.5} \\
\bottomrule
\end{tabular}}
\end{table}

%% file: supp_tables/domainnet126/weighted_spearman_0.0_src_threshold_per_adapter_GVB.tex
\def\weightedspearmanzerozerosrcthresholdperadapterDCP#1{\ifdim#1pt>84.3pt\cellcolor{lime!100}\else\ifdim#1pt>81.7pt\cellcolor{lime!90}\else\ifdim#1pt>79.0pt\cellcolor{lime!80}\else\ifdim#1pt>76.4pt\cellcolor{lime!70}\else\ifdim#1pt>73.8pt\cellcolor{lime!60}\else\ifdim#1pt>71.2pt\cellcolor{lime!50}\else\ifdim#1pt>68.6pt\cellcolor{lime!40}\else\ifdim#1pt>65.9pt\cellcolor{lime!30}\else\ifdim#1pt>63.3pt\cellcolor{lime!20}\else\ifdim#1pt>60.7pt\cellcolor{lime!10}\else\cellcolor{lime!0}\fi\fi\fi\fi\fi\fi\fi\fi\fi\fi#1}

\def\weightedspearmanzerozerosrcthresholdperadapterDCR#1{\ifdim#1pt>88.8pt\cellcolor{lime!100}\else\ifdim#1pt>85.9pt\cellcolor{lime!90}\else\ifdim#1pt>83.0pt\cellcolor{lime!80}\else\ifdim#1pt>80.2pt\cellcolor{lime!70}\else\ifdim#1pt>77.3pt\cellcolor{lime!60}\else\ifdim#1pt>74.5pt\cellcolor{lime!50}\else\ifdim#1pt>71.7pt\cellcolor{lime!40}\else\ifdim#1pt>68.8pt\cellcolor{lime!30}\else\ifdim#1pt>66.0pt\cellcolor{lime!20}\else\ifdim#1pt>63.1pt\cellcolor{lime!10}\else\cellcolor{lime!0}\fi\fi\fi\fi\fi\fi\fi\fi\fi\fi#1}

\def\weightedspearmanzerozerosrcthresholdperadapterDCS#1{\ifdim#1pt>89.2pt\cellcolor{lime!100}\else\ifdim#1pt>87.9pt\cellcolor{lime!90}\else\ifdim#1pt>86.7pt\cellcolor{lime!80}\else\ifdim#1pt>85.4pt\cellcolor{lime!70}\else\ifdim#1pt>84.2pt\cellcolor{lime!60}\else\ifdim#1pt>82.9pt\cellcolor{lime!50}\else\ifdim#1pt>81.7pt\cellcolor{lime!40}\else\ifdim#1pt>80.4pt\cellcolor{lime!30}\else\ifdim#1pt>79.2pt\cellcolor{lime!20}\else\ifdim#1pt>77.9pt\cellcolor{lime!10}\else\cellcolor{lime!0}\fi\fi\fi\fi\fi\fi\fi\fi\fi\fi#1}

\def\weightedspearmanzerozerosrcthresholdperadapterDPC#1{\ifdim#1pt>80.5pt\cellcolor{lime!100}\else\ifdim#1pt>78.3pt\cellcolor{lime!90}\else\ifdim#1pt>76.2pt\cellcolor{lime!80}\else\ifdim#1pt>74.0pt\cellcolor{lime!70}\else\ifdim#1pt>71.9pt\cellcolor{lime!60}\else\ifdim#1pt>69.8pt\cellcolor{lime!50}\else\ifdim#1pt>67.6pt\cellcolor{lime!40}\else\ifdim#1pt>65.5pt\cellcolor{lime!30}\else\ifdim#1pt>63.3pt\cellcolor{lime!20}\else\ifdim#1pt>61.2pt\cellcolor{lime!10}\else\cellcolor{lime!0}\fi\fi\fi\fi\fi\fi\fi\fi\fi\fi#1}

\def\weightedspearmanzerozerosrcthresholdperadapterDPR#1{\ifdim#1pt>93.6pt\cellcolor{lime!100}\else\ifdim#1pt>93.6pt\cellcolor{lime!90}\else\ifdim#1pt>93.6pt\cellcolor{lime!80}\else\ifdim#1pt>93.6pt\cellcolor{lime!70}\else\ifdim#1pt>93.6pt\cellcolor{lime!60}\else\ifdim#1pt>93.6pt\cellcolor{lime!50}\else\ifdim#1pt>93.6pt\cellcolor{lime!40}\else\ifdim#1pt>93.6pt\cellcolor{lime!30}\else\ifdim#1pt>93.6pt\cellcolor{lime!20}\else\ifdim#1pt>93.6pt\cellcolor{lime!10}\else\cellcolor{lime!0}\fi\fi\fi\fi\fi\fi\fi\fi\fi\fi#1}

\def\weightedspearmanzerozerosrcthresholdperadapterDPS#1{\ifdim#1pt>88.2pt\cellcolor{lime!100}\else\ifdim#1pt>84.1pt\cellcolor{lime!90}\else\ifdim#1pt>79.9pt\cellcolor{lime!80}\else\ifdim#1pt>75.7pt\cellcolor{lime!70}\else\ifdim#1pt>71.6pt\cellcolor{lime!60}\else\ifdim#1pt>67.4pt\cellcolor{lime!50}\else\ifdim#1pt>63.2pt\cellcolor{lime!40}\else\ifdim#1pt>59.0pt\cellcolor{lime!30}\else\ifdim#1pt>54.9pt\cellcolor{lime!20}\else\ifdim#1pt>50.7pt\cellcolor{lime!10}\else\cellcolor{lime!0}\fi\fi\fi\fi\fi\fi\fi\fi\fi\fi#1}

\def\weightedspearmanzerozerosrcthresholdperadapterDRC#1{\ifdim#1pt>83.9pt\cellcolor{lime!100}\else\ifdim#1pt>80.2pt\cellcolor{lime!90}\else\ifdim#1pt>76.6pt\cellcolor{lime!80}\else\ifdim#1pt>72.9pt\cellcolor{lime!70}\else\ifdim#1pt>69.3pt\cellcolor{lime!60}\else\ifdim#1pt>65.7pt\cellcolor{lime!50}\else\ifdim#1pt>62.0pt\cellcolor{lime!40}\else\ifdim#1pt>58.4pt\cellcolor{lime!30}\else\ifdim#1pt>54.7pt\cellcolor{lime!20}\else\ifdim#1pt>51.1pt\cellcolor{lime!10}\else\cellcolor{lime!0}\fi\fi\fi\fi\fi\fi\fi\fi\fi\fi#1}

\def\weightedspearmanzerozerosrcthresholdperadapterDRP#1{\ifdim#1pt>83.5pt\cellcolor{lime!100}\else\ifdim#1pt>83.4pt\cellcolor{lime!90}\else\ifdim#1pt>83.2pt\cellcolor{lime!80}\else\ifdim#1pt>83.1pt\cellcolor{lime!70}\else\ifdim#1pt>83.0pt\cellcolor{lime!60}\else\ifdim#1pt>82.8pt\cellcolor{lime!50}\else\ifdim#1pt>82.7pt\cellcolor{lime!40}\else\ifdim#1pt>82.5pt\cellcolor{lime!30}\else\ifdim#1pt>82.4pt\cellcolor{lime!20}\else\ifdim#1pt>82.2pt\cellcolor{lime!10}\else\cellcolor{lime!0}\fi\fi\fi\fi\fi\fi\fi\fi\fi\fi#1}

\def\weightedspearmanzerozerosrcthresholdperadapterDRS#1{\ifdim#1pt>88.1pt\cellcolor{lime!100}\else\ifdim#1pt>83.2pt\cellcolor{lime!90}\else\ifdim#1pt>78.4pt\cellcolor{lime!80}\else\ifdim#1pt>73.6pt\cellcolor{lime!70}\else\ifdim#1pt>68.8pt\cellcolor{lime!60}\else\ifdim#1pt>63.9pt\cellcolor{lime!50}\else\ifdim#1pt>59.1pt\cellcolor{lime!40}\else\ifdim#1pt>54.3pt\cellcolor{lime!30}\else\ifdim#1pt>49.4pt\cellcolor{lime!20}\else\ifdim#1pt>44.6pt\cellcolor{lime!10}\else\cellcolor{lime!0}\fi\fi\fi\fi\fi\fi\fi\fi\fi\fi#1}

\def\weightedspearmanzerozerosrcthresholdperadapterDSC#1{\ifdim#1pt>81.2pt\cellcolor{lime!100}\else\ifdim#1pt>79.1pt\cellcolor{lime!90}\else\ifdim#1pt>77.1pt\cellcolor{lime!80}\else\ifdim#1pt>75.0pt\cellcolor{lime!70}\else\ifdim#1pt>73.0pt\cellcolor{lime!60}\else\ifdim#1pt>71.0pt\cellcolor{lime!50}\else\ifdim#1pt>68.9pt\cellcolor{lime!40}\else\ifdim#1pt>66.9pt\cellcolor{lime!30}\else\ifdim#1pt>64.8pt\cellcolor{lime!20}\else\ifdim#1pt>62.8pt\cellcolor{lime!10}\else\cellcolor{lime!0}\fi\fi\fi\fi\fi\fi\fi\fi\fi\fi#1}

\def\weightedspearmanzerozerosrcthresholdperadapterDSP#1{\ifdim#1pt>88.8pt\cellcolor{lime!100}\else\ifdim#1pt>85.6pt\cellcolor{lime!90}\else\ifdim#1pt>82.3pt\cellcolor{lime!80}\else\ifdim#1pt>79.1pt\cellcolor{lime!70}\else\ifdim#1pt>75.8pt\cellcolor{lime!60}\else\ifdim#1pt>72.6pt\cellcolor{lime!50}\else\ifdim#1pt>69.3pt\cellcolor{lime!40}\else\ifdim#1pt>66.1pt\cellcolor{lime!30}\else\ifdim#1pt>62.9pt\cellcolor{lime!20}\else\ifdim#1pt>59.6pt\cellcolor{lime!10}\else\cellcolor{lime!0}\fi\fi\fi\fi\fi\fi\fi\fi\fi\fi#1}

\def\weightedspearmanzerozerosrcthresholdperadapterDSR#1{\ifdim#1pt>85.8pt\cellcolor{lime!100}\else\ifdim#1pt>80.6pt\cellcolor{lime!90}\else\ifdim#1pt>75.3pt\cellcolor{lime!80}\else\ifdim#1pt>70.1pt\cellcolor{lime!70}\else\ifdim#1pt>64.8pt\cellcolor{lime!60}\else\ifdim#1pt>59.6pt\cellcolor{lime!50}\else\ifdim#1pt>54.3pt\cellcolor{lime!40}\else\ifdim#1pt>49.1pt\cellcolor{lime!30}\else\ifdim#1pt>43.9pt\cellcolor{lime!20}\else\ifdim#1pt>38.6pt\cellcolor{lime!10}\else\cellcolor{lime!0}\fi\fi\fi\fi\fi\fi\fi\fi\fi\fi#1}

\def\weightedspearmanzerozerosrcthresholdperadapterMean#1{\ifdim#1pt>81.5pt\cellcolor{lime!100}\else\ifdim#1pt>79.3pt\cellcolor{lime!90}\else\ifdim#1pt>77.2pt\cellcolor{lime!80}\else\ifdim#1pt>75.0pt\cellcolor{lime!70}\else\ifdim#1pt>72.9pt\cellcolor{lime!60}\else\ifdim#1pt>70.8pt\cellcolor{lime!50}\else\ifdim#1pt>68.6pt\cellcolor{lime!40}\else\ifdim#1pt>66.5pt\cellcolor{lime!30}\else\ifdim#1pt>64.3pt\cellcolor{lime!20}\else\ifdim#1pt>62.2pt\cellcolor{lime!10}\else\cellcolor{lime!0}\fi\fi\fi\fi\fi\fi\fi\fi\fi\fi#1}

\def\weightedspearmanzerozerosrcthresholdperadapterStd#1{\ifdim#1pt<2.5pt\cellcolor{lime!100}\else\ifdim#1pt<3.9pt\cellcolor{lime!90}\else\ifdim#1pt<5.3pt\cellcolor{lime!80}\else\ifdim#1pt<6.7pt\cellcolor{lime!70}\else\ifdim#1pt<8.2pt\cellcolor{lime!60}\else\ifdim#1pt<9.6pt\cellcolor{lime!50}\else\ifdim#1pt<11.0pt\cellcolor{lime!40}\else\ifdim#1pt<12.4pt\cellcolor{lime!30}\else\ifdim#1pt<13.8pt\cellcolor{lime!20}\else\ifdim#1pt<15.2pt\cellcolor{lime!10}\else\cellcolor{lime!0}\fi\fi\fi\fi\fi\fi\fi\fi\fi\fi#1}

\begin{table}[H]
\centering
\caption{The weighted Spearman correlation of each validator/task pair for \textbf{GVB}.}
\label{domainnet126_weighted_spearman_0.0_src_threshold_per_adapter_GVB}
\resizebox{!}{0.22\textheight}{
\begin{tabular}{llrrrrrrrrrrrrrr}
\toprule
 &  & CP & CR & CS & PC & PR & PS & RC & RP & RS & SC & SP & SR & Mean & Std \\
\midrule
\multirow[c]{2}{*}{Accuracy} & Source Train & \weightedspearmanzerozerosrcthresholdperadapterDCP{79.2} & \weightedspearmanzerozerosrcthresholdperadapterDCR{76.0} & \weightedspearmanzerozerosrcthresholdperadapterDCS{81.2} & \weightedspearmanzerozerosrcthresholdperadapterDPC{72.1} & \weightedspearmanzerozerosrcthresholdperadapterDPR{52.8} & \weightedspearmanzerozerosrcthresholdperadapterDPS{56.4} & \weightedspearmanzerozerosrcthresholdperadapterDRC{52.1} & \weightedspearmanzerozerosrcthresholdperadapterDRP{78.4} & \weightedspearmanzerozerosrcthresholdperadapterDRS{49.8} & \weightedspearmanzerozerosrcthresholdperadapterDSC{82.1} & \weightedspearmanzerozerosrcthresholdperadapterDSP{78.4} & \weightedspearmanzerozerosrcthresholdperadapterDSR{56.6} & \weightedspearmanzerozerosrcthresholdperadapterMean{67.9} & \weightedspearmanzerozerosrcthresholdperadapterStd{12.5} \\
 & Source Val & \weightedspearmanzerozerosrcthresholdperadapterDCP{60.7} & \weightedspearmanzerozerosrcthresholdperadapterDCR{63.1} & \weightedspearmanzerozerosrcthresholdperadapterDCS{77.9} & \weightedspearmanzerozerosrcthresholdperadapterDPC{61.2} & \textbf{\weightedspearmanzerozerosrcthresholdperadapterDPR{93.6}} & \weightedspearmanzerozerosrcthresholdperadapterDPS{50.7} & \weightedspearmanzerozerosrcthresholdperadapterDRC{51.1} & \weightedspearmanzerozerosrcthresholdperadapterDRP{82.2} & \weightedspearmanzerozerosrcthresholdperadapterDRS{44.6} & \weightedspearmanzerozerosrcthresholdperadapterDSC{62.8} & \weightedspearmanzerozerosrcthresholdperadapterDSP{59.6} & \weightedspearmanzerozerosrcthresholdperadapterDSR{38.6} & \weightedspearmanzerozerosrcthresholdperadapterMean{62.2} & \weightedspearmanzerozerosrcthresholdperadapterStd{15.2} \\
\bottomrule
\end{tabular}}
\end{table}

%% file: supp_tables/domainnet126/weighted_spearman_0.0_src_threshold_per_adapter_IM.tex
\def\weightedspearmanzerozerosrcthresholdperadapterDCP#1{\ifdim#1pt>90.6pt\cellcolor{lime!100}\else\ifdim#1pt>87.0pt\cellcolor{lime!90}\else\ifdim#1pt>83.3pt\cellcolor{lime!80}\else\ifdim#1pt>79.6pt\cellcolor{lime!70}\else\ifdim#1pt>76.0pt\cellcolor{lime!60}\else\ifdim#1pt>72.3pt\cellcolor{lime!50}\else\ifdim#1pt>68.6pt\cellcolor{lime!40}\else\ifdim#1pt>64.9pt\cellcolor{lime!30}\else\ifdim#1pt>61.3pt\cellcolor{lime!20}\else\ifdim#1pt>57.6pt\cellcolor{lime!10}\else\cellcolor{lime!0}\fi\fi\fi\fi\fi\fi\fi\fi\fi\fi#1}

\def\weightedspearmanzerozerosrcthresholdperadapterDCR#1{\ifdim#1pt>85.4pt\cellcolor{lime!100}\else\ifdim#1pt>82.9pt\cellcolor{lime!90}\else\ifdim#1pt>80.4pt\cellcolor{lime!80}\else\ifdim#1pt>77.9pt\cellcolor{lime!70}\else\ifdim#1pt>75.3pt\cellcolor{lime!60}\else\ifdim#1pt>72.8pt\cellcolor{lime!50}\else\ifdim#1pt>70.3pt\cellcolor{lime!40}\else\ifdim#1pt>67.8pt\cellcolor{lime!30}\else\ifdim#1pt>65.3pt\cellcolor{lime!20}\else\ifdim#1pt>62.8pt\cellcolor{lime!10}\else\cellcolor{lime!0}\fi\fi\fi\fi\fi\fi\fi\fi\fi\fi#1}

\def\weightedspearmanzerozerosrcthresholdperadapterDCS#1{\ifdim#1pt>93.4pt\cellcolor{lime!100}\else\ifdim#1pt>91.1pt\cellcolor{lime!90}\else\ifdim#1pt>88.9pt\cellcolor{lime!80}\else\ifdim#1pt>86.6pt\cellcolor{lime!70}\else\ifdim#1pt>84.4pt\cellcolor{lime!60}\else\ifdim#1pt>82.2pt\cellcolor{lime!50}\else\ifdim#1pt>79.9pt\cellcolor{lime!40}\else\ifdim#1pt>77.7pt\cellcolor{lime!30}\else\ifdim#1pt>75.4pt\cellcolor{lime!20}\else\ifdim#1pt>73.2pt\cellcolor{lime!10}\else\cellcolor{lime!0}\fi\fi\fi\fi\fi\fi\fi\fi\fi\fi#1}

\def\weightedspearmanzerozerosrcthresholdperadapterDPC#1{\ifdim#1pt>86.9pt\cellcolor{lime!100}\else\ifdim#1pt>82.9pt\cellcolor{lime!90}\else\ifdim#1pt>78.8pt\cellcolor{lime!80}\else\ifdim#1pt>74.8pt\cellcolor{lime!70}\else\ifdim#1pt>70.7pt\cellcolor{lime!60}\else\ifdim#1pt>66.6pt\cellcolor{lime!50}\else\ifdim#1pt>62.6pt\cellcolor{lime!40}\else\ifdim#1pt>58.5pt\cellcolor{lime!30}\else\ifdim#1pt>54.5pt\cellcolor{lime!20}\else\ifdim#1pt>50.4pt\cellcolor{lime!10}\else\cellcolor{lime!0}\fi\fi\fi\fi\fi\fi\fi\fi\fi\fi#1}

\def\weightedspearmanzerozerosrcthresholdperadapterDPR#1{\ifdim#1pt>92.3pt\cellcolor{lime!100}\else\ifdim#1pt>89.0pt\cellcolor{lime!90}\else\ifdim#1pt>85.7pt\cellcolor{lime!80}\else\ifdim#1pt>82.4pt\cellcolor{lime!70}\else\ifdim#1pt>79.0pt\cellcolor{lime!60}\else\ifdim#1pt>75.7pt\cellcolor{lime!50}\else\ifdim#1pt>72.4pt\cellcolor{lime!40}\else\ifdim#1pt>69.1pt\cellcolor{lime!30}\else\ifdim#1pt>65.8pt\cellcolor{lime!20}\else\ifdim#1pt>62.5pt\cellcolor{lime!10}\else\cellcolor{lime!0}\fi\fi\fi\fi\fi\fi\fi\fi\fi\fi#1}

\def\weightedspearmanzerozerosrcthresholdperadapterDPS#1{\ifdim#1pt>67.5pt\cellcolor{lime!100}\else\ifdim#1pt>64.2pt\cellcolor{lime!90}\else\ifdim#1pt>61.0pt\cellcolor{lime!80}\else\ifdim#1pt>57.7pt\cellcolor{lime!70}\else\ifdim#1pt>54.5pt\cellcolor{lime!60}\else\ifdim#1pt>51.3pt\cellcolor{lime!50}\else\ifdim#1pt>48.0pt\cellcolor{lime!40}\else\ifdim#1pt>44.8pt\cellcolor{lime!30}\else\ifdim#1pt>41.5pt\cellcolor{lime!20}\else\ifdim#1pt>38.3pt\cellcolor{lime!10}\else\cellcolor{lime!0}\fi\fi\fi\fi\fi\fi\fi\fi\fi\fi#1}

\def\weightedspearmanzerozerosrcthresholdperadapterDRC#1{\ifdim#1pt>88.8pt\cellcolor{lime!100}\else\ifdim#1pt>84.8pt\cellcolor{lime!90}\else\ifdim#1pt>80.7pt\cellcolor{lime!80}\else\ifdim#1pt>76.7pt\cellcolor{lime!70}\else\ifdim#1pt>72.7pt\cellcolor{lime!60}\else\ifdim#1pt>68.7pt\cellcolor{lime!50}\else\ifdim#1pt>64.7pt\cellcolor{lime!40}\else\ifdim#1pt>60.6pt\cellcolor{lime!30}\else\ifdim#1pt>56.6pt\cellcolor{lime!20}\else\ifdim#1pt>52.6pt\cellcolor{lime!10}\else\cellcolor{lime!0}\fi\fi\fi\fi\fi\fi\fi\fi\fi\fi#1}

\def\weightedspearmanzerozerosrcthresholdperadapterDRP#1{\ifdim#1pt>90.4pt\cellcolor{lime!100}\else\ifdim#1pt>89.8pt\cellcolor{lime!90}\else\ifdim#1pt>89.2pt\cellcolor{lime!80}\else\ifdim#1pt>88.6pt\cellcolor{lime!70}\else\ifdim#1pt>88.0pt\cellcolor{lime!60}\else\ifdim#1pt>87.4pt\cellcolor{lime!50}\else\ifdim#1pt>86.8pt\cellcolor{lime!40}\else\ifdim#1pt>86.2pt\cellcolor{lime!30}\else\ifdim#1pt>85.6pt\cellcolor{lime!20}\else\ifdim#1pt>85.0pt\cellcolor{lime!10}\else\cellcolor{lime!0}\fi\fi\fi\fi\fi\fi\fi\fi\fi\fi#1}

\def\weightedspearmanzerozerosrcthresholdperadapterDRS#1{\ifdim#1pt>92.5pt\cellcolor{lime!100}\else\ifdim#1pt>89.2pt\cellcolor{lime!90}\else\ifdim#1pt>85.9pt\cellcolor{lime!80}\else\ifdim#1pt>82.6pt\cellcolor{lime!70}\else\ifdim#1pt>79.2pt\cellcolor{lime!60}\else\ifdim#1pt>75.9pt\cellcolor{lime!50}\else\ifdim#1pt>72.6pt\cellcolor{lime!40}\else\ifdim#1pt>69.3pt\cellcolor{lime!30}\else\ifdim#1pt>66.0pt\cellcolor{lime!20}\else\ifdim#1pt>62.7pt\cellcolor{lime!10}\else\cellcolor{lime!0}\fi\fi\fi\fi\fi\fi\fi\fi\fi\fi#1}

\def\weightedspearmanzerozerosrcthresholdperadapterDSC#1{\ifdim#1pt>86.8pt\cellcolor{lime!100}\else\ifdim#1pt>83.2pt\cellcolor{lime!90}\else\ifdim#1pt>79.6pt\cellcolor{lime!80}\else\ifdim#1pt>76.1pt\cellcolor{lime!70}\else\ifdim#1pt>72.5pt\cellcolor{lime!60}\else\ifdim#1pt>69.0pt\cellcolor{lime!50}\else\ifdim#1pt>65.4pt\cellcolor{lime!40}\else\ifdim#1pt>61.9pt\cellcolor{lime!30}\else\ifdim#1pt>58.3pt\cellcolor{lime!20}\else\ifdim#1pt>54.8pt\cellcolor{lime!10}\else\cellcolor{lime!0}\fi\fi\fi\fi\fi\fi\fi\fi\fi\fi#1}

\def\weightedspearmanzerozerosrcthresholdperadapterDSP#1{\ifdim#1pt>88.6pt\cellcolor{lime!100}\else\ifdim#1pt>84.6pt\cellcolor{lime!90}\else\ifdim#1pt>80.6pt\cellcolor{lime!80}\else\ifdim#1pt>76.6pt\cellcolor{lime!70}\else\ifdim#1pt>72.6pt\cellcolor{lime!60}\else\ifdim#1pt>68.6pt\cellcolor{lime!50}\else\ifdim#1pt>64.6pt\cellcolor{lime!40}\else\ifdim#1pt>60.6pt\cellcolor{lime!30}\else\ifdim#1pt>56.6pt\cellcolor{lime!20}\else\ifdim#1pt>52.6pt\cellcolor{lime!10}\else\cellcolor{lime!0}\fi\fi\fi\fi\fi\fi\fi\fi\fi\fi#1}

\def\weightedspearmanzerozerosrcthresholdperadapterDSR#1{\ifdim#1pt>89.5pt\cellcolor{lime!100}\else\ifdim#1pt>84.2pt\cellcolor{lime!90}\else\ifdim#1pt>78.8pt\cellcolor{lime!80}\else\ifdim#1pt>73.4pt\cellcolor{lime!70}\else\ifdim#1pt>68.1pt\cellcolor{lime!60}\else\ifdim#1pt>62.7pt\cellcolor{lime!50}\else\ifdim#1pt>57.3pt\cellcolor{lime!40}\else\ifdim#1pt>51.9pt\cellcolor{lime!30}\else\ifdim#1pt>46.6pt\cellcolor{lime!20}\else\ifdim#1pt>41.2pt\cellcolor{lime!10}\else\cellcolor{lime!0}\fi\fi\fi\fi\fi\fi\fi\fi\fi\fi#1}

\def\weightedspearmanzerozerosrcthresholdperadapterMean#1{\ifdim#1pt>87.4pt\cellcolor{lime!100}\else\ifdim#1pt>84.1pt\cellcolor{lime!90}\else\ifdim#1pt>80.8pt\cellcolor{lime!80}\else\ifdim#1pt>77.5pt\cellcolor{lime!70}\else\ifdim#1pt>74.2pt\cellcolor{lime!60}\else\ifdim#1pt>71.0pt\cellcolor{lime!50}\else\ifdim#1pt>67.7pt\cellcolor{lime!40}\else\ifdim#1pt>64.4pt\cellcolor{lime!30}\else\ifdim#1pt>61.1pt\cellcolor{lime!20}\else\ifdim#1pt>57.8pt\cellcolor{lime!10}\else\cellcolor{lime!0}\fi\fi\fi\fi\fi\fi\fi\fi\fi\fi#1}

\def\weightedspearmanzerozerosrcthresholdperadapterStd#1{\ifdim#1pt<4.7pt\cellcolor{lime!100}\else\ifdim#1pt<5.6pt\cellcolor{lime!90}\else\ifdim#1pt<6.4pt\cellcolor{lime!80}\else\ifdim#1pt<7.3pt\cellcolor{lime!70}\else\ifdim#1pt<8.1pt\cellcolor{lime!60}\else\ifdim#1pt<8.9pt\cellcolor{lime!50}\else\ifdim#1pt<9.8pt\cellcolor{lime!40}\else\ifdim#1pt<10.6pt\cellcolor{lime!30}\else\ifdim#1pt<11.5pt\cellcolor{lime!20}\else\ifdim#1pt<12.3pt\cellcolor{lime!10}\else\cellcolor{lime!0}\fi\fi\fi\fi\fi\fi\fi\fi\fi\fi#1}

\begin{table}[H]
\centering
\caption{The weighted Spearman correlation of each validator/task pair for \textbf{IM}.}
\label{domainnet126_weighted_spearman_0.0_src_threshold_per_adapter_IM}
\resizebox{!}{0.22\textheight}{
\begin{tabular}{llrrrrrrrrrrrrrr}
\toprule
 &  & CP & CR & CS & PC & PR & PS & RC & RP & RS & SC & SP & SR & Mean & Std \\
\midrule
\multirow[c]{2}{*}{Accuracy} & Source Train & \weightedspearmanzerozerosrcthresholdperadapterDCP{70.4} & \weightedspearmanzerozerosrcthresholdperadapterDCR{78.7} & \weightedspearmanzerozerosrcthresholdperadapterDCS{82.0} & \weightedspearmanzerozerosrcthresholdperadapterDPC{57.8} & \weightedspearmanzerozerosrcthresholdperadapterDPR{76.0} & \weightedspearmanzerozerosrcthresholdperadapterDPS{29.4} & \weightedspearmanzerozerosrcthresholdperadapterDRC{53.2} & \weightedspearmanzerozerosrcthresholdperadapterDRP{85.2} & \weightedspearmanzerozerosrcthresholdperadapterDRS{62.1} & \weightedspearmanzerozerosrcthresholdperadapterDSC{77.2} & \weightedspearmanzerozerosrcthresholdperadapterDSP{71.9} & \weightedspearmanzerozerosrcthresholdperadapterDSR{69.9} & \weightedspearmanzerozerosrcthresholdperadapterMean{67.8} & \weightedspearmanzerozerosrcthresholdperadapterStd{14.8} \\
 & Source Val & \weightedspearmanzerozerosrcthresholdperadapterDCP{57.6} & \weightedspearmanzerozerosrcthresholdperadapterDCR{62.8} & \weightedspearmanzerozerosrcthresholdperadapterDCS{73.2} & \weightedspearmanzerozerosrcthresholdperadapterDPC{50.4} & \weightedspearmanzerozerosrcthresholdperadapterDPR{62.5} & \weightedspearmanzerozerosrcthresholdperadapterDPS{38.3} & \weightedspearmanzerozerosrcthresholdperadapterDRC{52.6} & \weightedspearmanzerozerosrcthresholdperadapterDRP{85.0} & \weightedspearmanzerozerosrcthresholdperadapterDRS{62.7} & \weightedspearmanzerozerosrcthresholdperadapterDSC{54.8} & \weightedspearmanzerozerosrcthresholdperadapterDSP{52.6} & \weightedspearmanzerozerosrcthresholdperadapterDSR{41.2} & \weightedspearmanzerozerosrcthresholdperadapterMean{57.8} & \weightedspearmanzerozerosrcthresholdperadapterStd{12.3} \\
\bottomrule
\end{tabular}}
\end{table}

%% file: supp_tables/domainnet126/weighted_spearman_0.0_src_threshold_per_adapter_MCC.tex
\def\weightedspearmanzerozerosrcthresholdperadapterDCP#1{\ifdim#1pt>90.5pt\cellcolor{lime!100}\else\ifdim#1pt>85.9pt\cellcolor{lime!90}\else\ifdim#1pt>81.3pt\cellcolor{lime!80}\else\ifdim#1pt>76.7pt\cellcolor{lime!70}\else\ifdim#1pt>72.0pt\cellcolor{lime!60}\else\ifdim#1pt>67.4pt\cellcolor{lime!50}\else\ifdim#1pt>62.8pt\cellcolor{lime!40}\else\ifdim#1pt>58.2pt\cellcolor{lime!30}\else\ifdim#1pt>53.6pt\cellcolor{lime!20}\else\ifdim#1pt>49.0pt\cellcolor{lime!10}\else\cellcolor{lime!0}\fi\fi\fi\fi\fi\fi\fi\fi\fi\fi#1}

\def\weightedspearmanzerozerosrcthresholdperadapterDCR#1{\ifdim#1pt>88.9pt\cellcolor{lime!100}\else\ifdim#1pt>86.5pt\cellcolor{lime!90}\else\ifdim#1pt>84.1pt\cellcolor{lime!80}\else\ifdim#1pt>81.7pt\cellcolor{lime!70}\else\ifdim#1pt>79.3pt\cellcolor{lime!60}\else\ifdim#1pt>77.0pt\cellcolor{lime!50}\else\ifdim#1pt>74.6pt\cellcolor{lime!40}\else\ifdim#1pt>72.2pt\cellcolor{lime!30}\else\ifdim#1pt>69.8pt\cellcolor{lime!20}\else\ifdim#1pt>67.4pt\cellcolor{lime!10}\else\cellcolor{lime!0}\fi\fi\fi\fi\fi\fi\fi\fi\fi\fi#1}

\def\weightedspearmanzerozerosrcthresholdperadapterDCS#1{\ifdim#1pt>90.5pt\cellcolor{lime!100}\else\ifdim#1pt>86.5pt\cellcolor{lime!90}\else\ifdim#1pt>82.6pt\cellcolor{lime!80}\else\ifdim#1pt>78.6pt\cellcolor{lime!70}\else\ifdim#1pt>74.7pt\cellcolor{lime!60}\else\ifdim#1pt>70.8pt\cellcolor{lime!50}\else\ifdim#1pt>66.8pt\cellcolor{lime!40}\else\ifdim#1pt>62.9pt\cellcolor{lime!30}\else\ifdim#1pt>58.9pt\cellcolor{lime!20}\else\ifdim#1pt>55.0pt\cellcolor{lime!10}\else\cellcolor{lime!0}\fi\fi\fi\fi\fi\fi\fi\fi\fi\fi#1}

\def\weightedspearmanzerozerosrcthresholdperadapterDPC#1{\ifdim#1pt>81.3pt\cellcolor{lime!100}\else\ifdim#1pt>76.3pt\cellcolor{lime!90}\else\ifdim#1pt>71.3pt\cellcolor{lime!80}\else\ifdim#1pt>66.3pt\cellcolor{lime!70}\else\ifdim#1pt>61.4pt\cellcolor{lime!60}\else\ifdim#1pt>56.4pt\cellcolor{lime!50}\else\ifdim#1pt>51.4pt\cellcolor{lime!40}\else\ifdim#1pt>46.4pt\cellcolor{lime!30}\else\ifdim#1pt>41.4pt\cellcolor{lime!20}\else\ifdim#1pt>36.4pt\cellcolor{lime!10}\else\cellcolor{lime!0}\fi\fi\fi\fi\fi\fi\fi\fi\fi\fi#1}

\def\weightedspearmanzerozerosrcthresholdperadapterDPR#1{\ifdim#1pt>88.9pt\cellcolor{lime!100}\else\ifdim#1pt>85.1pt\cellcolor{lime!90}\else\ifdim#1pt>81.3pt\cellcolor{lime!80}\else\ifdim#1pt>77.5pt\cellcolor{lime!70}\else\ifdim#1pt>73.7pt\cellcolor{lime!60}\else\ifdim#1pt>69.8pt\cellcolor{lime!50}\else\ifdim#1pt>66.0pt\cellcolor{lime!40}\else\ifdim#1pt>62.2pt\cellcolor{lime!30}\else\ifdim#1pt>58.4pt\cellcolor{lime!20}\else\ifdim#1pt>54.6pt\cellcolor{lime!10}\else\cellcolor{lime!0}\fi\fi\fi\fi\fi\fi\fi\fi\fi\fi#1}

\def\weightedspearmanzerozerosrcthresholdperadapterDPS#1{\ifdim#1pt>85.8pt\cellcolor{lime!100}\else\ifdim#1pt>82.2pt\cellcolor{lime!90}\else\ifdim#1pt>78.7pt\cellcolor{lime!80}\else\ifdim#1pt>75.1pt\cellcolor{lime!70}\else\ifdim#1pt>71.6pt\cellcolor{lime!60}\else\ifdim#1pt>68.1pt\cellcolor{lime!50}\else\ifdim#1pt>64.5pt\cellcolor{lime!40}\else\ifdim#1pt>61.0pt\cellcolor{lime!30}\else\ifdim#1pt>57.4pt\cellcolor{lime!20}\else\ifdim#1pt>53.9pt\cellcolor{lime!10}\else\cellcolor{lime!0}\fi\fi\fi\fi\fi\fi\fi\fi\fi\fi#1}

\def\weightedspearmanzerozerosrcthresholdperadapterDRC#1{\ifdim#1pt>81.4pt\cellcolor{lime!100}\else\ifdim#1pt>78.3pt\cellcolor{lime!90}\else\ifdim#1pt>75.3pt\cellcolor{lime!80}\else\ifdim#1pt>72.3pt\cellcolor{lime!70}\else\ifdim#1pt>69.2pt\cellcolor{lime!60}\else\ifdim#1pt>66.2pt\cellcolor{lime!50}\else\ifdim#1pt>63.2pt\cellcolor{lime!40}\else\ifdim#1pt>60.2pt\cellcolor{lime!30}\else\ifdim#1pt>57.1pt\cellcolor{lime!20}\else\ifdim#1pt>54.1pt\cellcolor{lime!10}\else\cellcolor{lime!0}\fi\fi\fi\fi\fi\fi\fi\fi\fi\fi#1}

\def\weightedspearmanzerozerosrcthresholdperadapterDRP#1{\ifdim#1pt>86.0pt\cellcolor{lime!100}\else\ifdim#1pt>84.4pt\cellcolor{lime!90}\else\ifdim#1pt>82.8pt\cellcolor{lime!80}\else\ifdim#1pt>81.2pt\cellcolor{lime!70}\else\ifdim#1pt>79.5pt\cellcolor{lime!60}\else\ifdim#1pt>77.9pt\cellcolor{lime!50}\else\ifdim#1pt>76.3pt\cellcolor{lime!40}\else\ifdim#1pt>74.7pt\cellcolor{lime!30}\else\ifdim#1pt>73.1pt\cellcolor{lime!20}\else\ifdim#1pt>71.5pt\cellcolor{lime!10}\else\cellcolor{lime!0}\fi\fi\fi\fi\fi\fi\fi\fi\fi\fi#1}

\def\weightedspearmanzerozerosrcthresholdperadapterDRS#1{\ifdim#1pt>86.3pt\cellcolor{lime!100}\else\ifdim#1pt>82.3pt\cellcolor{lime!90}\else\ifdim#1pt>78.2pt\cellcolor{lime!80}\else\ifdim#1pt>74.2pt\cellcolor{lime!70}\else\ifdim#1pt>70.1pt\cellcolor{lime!60}\else\ifdim#1pt>66.0pt\cellcolor{lime!50}\else\ifdim#1pt>62.0pt\cellcolor{lime!40}\else\ifdim#1pt>57.9pt\cellcolor{lime!30}\else\ifdim#1pt>53.9pt\cellcolor{lime!20}\else\ifdim#1pt>49.8pt\cellcolor{lime!10}\else\cellcolor{lime!0}\fi\fi\fi\fi\fi\fi\fi\fi\fi\fi#1}

\def\weightedspearmanzerozerosrcthresholdperadapterDSC#1{\ifdim#1pt>91.1pt\cellcolor{lime!100}\else\ifdim#1pt>86.8pt\cellcolor{lime!90}\else\ifdim#1pt>82.6pt\cellcolor{lime!80}\else\ifdim#1pt>78.4pt\cellcolor{lime!70}\else\ifdim#1pt>74.2pt\cellcolor{lime!60}\else\ifdim#1pt>69.9pt\cellcolor{lime!50}\else\ifdim#1pt>65.7pt\cellcolor{lime!40}\else\ifdim#1pt>61.5pt\cellcolor{lime!30}\else\ifdim#1pt>57.2pt\cellcolor{lime!20}\else\ifdim#1pt>53.0pt\cellcolor{lime!10}\else\cellcolor{lime!0}\fi\fi\fi\fi\fi\fi\fi\fi\fi\fi#1}

\def\weightedspearmanzerozerosrcthresholdperadapterDSP#1{\ifdim#1pt>86.4pt\cellcolor{lime!100}\else\ifdim#1pt>81.2pt\cellcolor{lime!90}\else\ifdim#1pt>75.9pt\cellcolor{lime!80}\else\ifdim#1pt>70.7pt\cellcolor{lime!70}\else\ifdim#1pt>65.4pt\cellcolor{lime!60}\else\ifdim#1pt>60.1pt\cellcolor{lime!50}\else\ifdim#1pt>54.9pt\cellcolor{lime!40}\else\ifdim#1pt>49.6pt\cellcolor{lime!30}\else\ifdim#1pt>44.4pt\cellcolor{lime!20}\else\ifdim#1pt>39.1pt\cellcolor{lime!10}\else\cellcolor{lime!0}\fi\fi\fi\fi\fi\fi\fi\fi\fi\fi#1}

\def\weightedspearmanzerozerosrcthresholdperadapterDSR#1{\ifdim#1pt>87.4pt\cellcolor{lime!100}\else\ifdim#1pt>82.0pt\cellcolor{lime!90}\else\ifdim#1pt>76.6pt\cellcolor{lime!80}\else\ifdim#1pt>71.2pt\cellcolor{lime!70}\else\ifdim#1pt>65.8pt\cellcolor{lime!60}\else\ifdim#1pt>60.5pt\cellcolor{lime!50}\else\ifdim#1pt>55.1pt\cellcolor{lime!40}\else\ifdim#1pt>49.7pt\cellcolor{lime!30}\else\ifdim#1pt>44.3pt\cellcolor{lime!20}\else\ifdim#1pt>38.9pt\cellcolor{lime!10}\else\cellcolor{lime!0}\fi\fi\fi\fi\fi\fi\fi\fi\fi\fi#1}

\def\weightedspearmanzerozerosrcthresholdperadapterMean#1{\ifdim#1pt>86.6pt\cellcolor{lime!100}\else\ifdim#1pt>82.8pt\cellcolor{lime!90}\else\ifdim#1pt>78.9pt\cellcolor{lime!80}\else\ifdim#1pt>75.1pt\cellcolor{lime!70}\else\ifdim#1pt>71.2pt\cellcolor{lime!60}\else\ifdim#1pt>67.3pt\cellcolor{lime!50}\else\ifdim#1pt>63.5pt\cellcolor{lime!40}\else\ifdim#1pt>59.6pt\cellcolor{lime!30}\else\ifdim#1pt>55.8pt\cellcolor{lime!20}\else\ifdim#1pt>51.9pt\cellcolor{lime!10}\else\cellcolor{lime!0}\fi\fi\fi\fi\fi\fi\fi\fi\fi\fi#1}

\def\weightedspearmanzerozerosrcthresholdperadapterStd#1{\ifdim#1pt<1.9pt\cellcolor{lime!100}\else\ifdim#1pt<2.8pt\cellcolor{lime!90}\else\ifdim#1pt<3.8pt\cellcolor{lime!80}\else\ifdim#1pt<4.7pt\cellcolor{lime!70}\else\ifdim#1pt<5.6pt\cellcolor{lime!60}\else\ifdim#1pt<6.5pt\cellcolor{lime!50}\else\ifdim#1pt<7.4pt\cellcolor{lime!40}\else\ifdim#1pt<8.4pt\cellcolor{lime!30}\else\ifdim#1pt<9.3pt\cellcolor{lime!20}\else\ifdim#1pt<10.2pt\cellcolor{lime!10}\else\cellcolor{lime!0}\fi\fi\fi\fi\fi\fi\fi\fi\fi\fi#1}

\begin{table}[H]
\centering
\caption{The weighted Spearman correlation of each validator/task pair for \textbf{MCC}.}
\label{domainnet126_weighted_spearman_0.0_src_threshold_per_adapter_MCC}
\resizebox{!}{0.22\textheight}{
\begin{tabular}{llrrrrrrrrrrrrrr}
\toprule
 &  & CP & CR & CS & PC & PR & PS & RC & RP & RS & SC & SP & SR & Mean & Std \\
\midrule
\multirow[c]{2}{*}{Accuracy} & Source Train & \weightedspearmanzerozerosrcthresholdperadapterDCP{59.3} & \weightedspearmanzerozerosrcthresholdperadapterDCR{76.0} & \weightedspearmanzerozerosrcthresholdperadapterDCS{59.6} & \weightedspearmanzerozerosrcthresholdperadapterDPC{43.9} & \weightedspearmanzerozerosrcthresholdperadapterDPR{59.1} & \weightedspearmanzerozerosrcthresholdperadapterDPS{58.8} & \weightedspearmanzerozerosrcthresholdperadapterDRC{64.0} & \weightedspearmanzerozerosrcthresholdperadapterDRP{77.0} & \weightedspearmanzerozerosrcthresholdperadapterDRS{44.7} & \weightedspearmanzerozerosrcthresholdperadapterDSC{66.3} & \weightedspearmanzerozerosrcthresholdperadapterDSP{46.8} & \weightedspearmanzerozerosrcthresholdperadapterDSR{53.9} & \weightedspearmanzerozerosrcthresholdperadapterMean{59.1} & \weightedspearmanzerozerosrcthresholdperadapterStd{10.4} \\
 & Source Val & \weightedspearmanzerozerosrcthresholdperadapterDCP{49.0} & \weightedspearmanzerozerosrcthresholdperadapterDCR{67.4} & \weightedspearmanzerozerosrcthresholdperadapterDCS{55.0} & \weightedspearmanzerozerosrcthresholdperadapterDPC{36.4} & \weightedspearmanzerozerosrcthresholdperadapterDPR{54.6} & \weightedspearmanzerozerosrcthresholdperadapterDPS{53.9} & \weightedspearmanzerozerosrcthresholdperadapterDRC{54.1} & \weightedspearmanzerozerosrcthresholdperadapterDRP{71.5} & \weightedspearmanzerozerosrcthresholdperadapterDRS{49.8} & \weightedspearmanzerozerosrcthresholdperadapterDSC{53.0} & \weightedspearmanzerozerosrcthresholdperadapterDSP{39.1} & \weightedspearmanzerozerosrcthresholdperadapterDSR{38.9} & \weightedspearmanzerozerosrcthresholdperadapterMean{51.9} & \weightedspearmanzerozerosrcthresholdperadapterStd{10.2} \\
\bottomrule
\end{tabular}}
\end{table}

%% file: supp_tables/domainnet126/weighted_spearman_0.0_src_threshold_per_adapter_MCD.tex
\def\weightedspearmanzerozerosrcthresholdperadapterDCP#1{\ifdim#1pt>95.7pt\cellcolor{lime!100}\else\ifdim#1pt>94.9pt\cellcolor{lime!90}\else\ifdim#1pt>94.0pt\cellcolor{lime!80}\else\ifdim#1pt>93.2pt\cellcolor{lime!70}\else\ifdim#1pt>92.3pt\cellcolor{lime!60}\else\ifdim#1pt>91.4pt\cellcolor{lime!50}\else\ifdim#1pt>90.6pt\cellcolor{lime!40}\else\ifdim#1pt>89.7pt\cellcolor{lime!30}\else\ifdim#1pt>88.9pt\cellcolor{lime!20}\else\ifdim#1pt>88.0pt\cellcolor{lime!10}\else\cellcolor{lime!0}\fi\fi\fi\fi\fi\fi\fi\fi\fi\fi#1}

\def\weightedspearmanzerozerosrcthresholdperadapterDCR#1{\ifdim#1pt>89.9pt\cellcolor{lime!100}\else\ifdim#1pt>89.9pt\cellcolor{lime!90}\else\ifdim#1pt>89.9pt\cellcolor{lime!80}\else\ifdim#1pt>89.9pt\cellcolor{lime!70}\else\ifdim#1pt>89.9pt\cellcolor{lime!60}\else\ifdim#1pt>89.9pt\cellcolor{lime!50}\else\ifdim#1pt>89.9pt\cellcolor{lime!40}\else\ifdim#1pt>89.9pt\cellcolor{lime!30}\else\ifdim#1pt>89.9pt\cellcolor{lime!20}\else\ifdim#1pt>89.9pt\cellcolor{lime!10}\else\cellcolor{lime!0}\fi\fi\fi\fi\fi\fi\fi\fi\fi\fi#1}

\def\weightedspearmanzerozerosrcthresholdperadapterDCS#1{\ifdim#1pt>93.2pt\cellcolor{lime!100}\else\ifdim#1pt>93.2pt\cellcolor{lime!90}\else\ifdim#1pt>93.2pt\cellcolor{lime!80}\else\ifdim#1pt>93.2pt\cellcolor{lime!70}\else\ifdim#1pt>93.2pt\cellcolor{lime!60}\else\ifdim#1pt>93.2pt\cellcolor{lime!50}\else\ifdim#1pt>93.2pt\cellcolor{lime!40}\else\ifdim#1pt>93.2pt\cellcolor{lime!30}\else\ifdim#1pt>93.2pt\cellcolor{lime!20}\else\ifdim#1pt>93.2pt\cellcolor{lime!10}\else\cellcolor{lime!0}\fi\fi\fi\fi\fi\fi\fi\fi\fi\fi#1}

\def\weightedspearmanzerozerosrcthresholdperadapterDPC#1{\ifdim#1pt>93.1pt\cellcolor{lime!100}\else\ifdim#1pt>91.5pt\cellcolor{lime!90}\else\ifdim#1pt>90.0pt\cellcolor{lime!80}\else\ifdim#1pt>88.4pt\cellcolor{lime!70}\else\ifdim#1pt>86.8pt\cellcolor{lime!60}\else\ifdim#1pt>85.2pt\cellcolor{lime!50}\else\ifdim#1pt>83.6pt\cellcolor{lime!40}\else\ifdim#1pt>82.1pt\cellcolor{lime!30}\else\ifdim#1pt>80.5pt\cellcolor{lime!20}\else\ifdim#1pt>78.9pt\cellcolor{lime!10}\else\cellcolor{lime!0}\fi\fi\fi\fi\fi\fi\fi\fi\fi\fi#1}

\def\weightedspearmanzerozerosrcthresholdperadapterDPR#1{\ifdim#1pt>95.8pt\cellcolor{lime!100}\else\ifdim#1pt>95.6pt\cellcolor{lime!90}\else\ifdim#1pt>95.4pt\cellcolor{lime!80}\else\ifdim#1pt>95.2pt\cellcolor{lime!70}\else\ifdim#1pt>95.0pt\cellcolor{lime!60}\else\ifdim#1pt>94.8pt\cellcolor{lime!50}\else\ifdim#1pt>94.6pt\cellcolor{lime!40}\else\ifdim#1pt>94.4pt\cellcolor{lime!30}\else\ifdim#1pt>94.2pt\cellcolor{lime!20}\else\ifdim#1pt>94.0pt\cellcolor{lime!10}\else\cellcolor{lime!0}\fi\fi\fi\fi\fi\fi\fi\fi\fi\fi#1}

\def\weightedspearmanzerozerosrcthresholdperadapterDPS#1{\ifdim#1pt>94.6pt\cellcolor{lime!100}\else\ifdim#1pt>92.5pt\cellcolor{lime!90}\else\ifdim#1pt>90.4pt\cellcolor{lime!80}\else\ifdim#1pt>88.3pt\cellcolor{lime!70}\else\ifdim#1pt>86.2pt\cellcolor{lime!60}\else\ifdim#1pt>84.0pt\cellcolor{lime!50}\else\ifdim#1pt>81.9pt\cellcolor{lime!40}\else\ifdim#1pt>79.8pt\cellcolor{lime!30}\else\ifdim#1pt>77.7pt\cellcolor{lime!20}\else\ifdim#1pt>75.6pt\cellcolor{lime!10}\else\cellcolor{lime!0}\fi\fi\fi\fi\fi\fi\fi\fi\fi\fi#1}

\def\weightedspearmanzerozerosrcthresholdperadapterDRC#1{\ifdim#1pt>94.9pt\cellcolor{lime!100}\else\ifdim#1pt>93.8pt\cellcolor{lime!90}\else\ifdim#1pt>92.6pt\cellcolor{lime!80}\else\ifdim#1pt>91.5pt\cellcolor{lime!70}\else\ifdim#1pt>90.3pt\cellcolor{lime!60}\else\ifdim#1pt>89.2pt\cellcolor{lime!50}\else\ifdim#1pt>88.0pt\cellcolor{lime!40}\else\ifdim#1pt>86.9pt\cellcolor{lime!30}\else\ifdim#1pt>85.8pt\cellcolor{lime!20}\else\ifdim#1pt>84.6pt\cellcolor{lime!10}\else\cellcolor{lime!0}\fi\fi\fi\fi\fi\fi\fi\fi\fi\fi#1}

\def\weightedspearmanzerozerosrcthresholdperadapterDRP#1{\ifdim#1pt>94.0pt\cellcolor{lime!100}\else\ifdim#1pt>92.9pt\cellcolor{lime!90}\else\ifdim#1pt>91.8pt\cellcolor{lime!80}\else\ifdim#1pt>90.8pt\cellcolor{lime!70}\else\ifdim#1pt>89.8pt\cellcolor{lime!60}\else\ifdim#1pt>88.7pt\cellcolor{lime!50}\else\ifdim#1pt>87.7pt\cellcolor{lime!40}\else\ifdim#1pt>86.6pt\cellcolor{lime!30}\else\ifdim#1pt>85.5pt\cellcolor{lime!20}\else\ifdim#1pt>84.5pt\cellcolor{lime!10}\else\cellcolor{lime!0}\fi\fi\fi\fi\fi\fi\fi\fi\fi\fi#1}

\def\weightedspearmanzerozerosrcthresholdperadapterDRS#1{\ifdim#1pt>96.4pt\cellcolor{lime!100}\else\ifdim#1pt>96.0pt\cellcolor{lime!90}\else\ifdim#1pt>95.6pt\cellcolor{lime!80}\else\ifdim#1pt>95.2pt\cellcolor{lime!70}\else\ifdim#1pt>94.8pt\cellcolor{lime!60}\else\ifdim#1pt>94.3pt\cellcolor{lime!50}\else\ifdim#1pt>93.9pt\cellcolor{lime!40}\else\ifdim#1pt>93.5pt\cellcolor{lime!30}\else\ifdim#1pt>93.1pt\cellcolor{lime!20}\else\ifdim#1pt>92.7pt\cellcolor{lime!10}\else\cellcolor{lime!0}\fi\fi\fi\fi\fi\fi\fi\fi\fi\fi#1}

\def\weightedspearmanzerozerosrcthresholdperadapterDSC#1{\ifdim#1pt>92.1pt\cellcolor{lime!100}\else\ifdim#1pt>88.3pt\cellcolor{lime!90}\else\ifdim#1pt>84.5pt\cellcolor{lime!80}\else\ifdim#1pt>80.7pt\cellcolor{lime!70}\else\ifdim#1pt>76.8pt\cellcolor{lime!60}\else\ifdim#1pt>73.0pt\cellcolor{lime!50}\else\ifdim#1pt>69.2pt\cellcolor{lime!40}\else\ifdim#1pt>65.4pt\cellcolor{lime!30}\else\ifdim#1pt>61.6pt\cellcolor{lime!20}\else\ifdim#1pt>57.8pt\cellcolor{lime!10}\else\cellcolor{lime!0}\fi\fi\fi\fi\fi\fi\fi\fi\fi\fi#1}

\def\weightedspearmanzerozerosrcthresholdperadapterDSP#1{\ifdim#1pt>94.6pt\cellcolor{lime!100}\else\ifdim#1pt>92.3pt\cellcolor{lime!90}\else\ifdim#1pt>89.9pt\cellcolor{lime!80}\else\ifdim#1pt>87.6pt\cellcolor{lime!70}\else\ifdim#1pt>85.2pt\cellcolor{lime!60}\else\ifdim#1pt>82.8pt\cellcolor{lime!50}\else\ifdim#1pt>80.5pt\cellcolor{lime!40}\else\ifdim#1pt>78.1pt\cellcolor{lime!30}\else\ifdim#1pt>75.8pt\cellcolor{lime!20}\else\ifdim#1pt>73.4pt\cellcolor{lime!10}\else\cellcolor{lime!0}\fi\fi\fi\fi\fi\fi\fi\fi\fi\fi#1}

\def\weightedspearmanzerozerosrcthresholdperadapterDSR#1{\ifdim#1pt>92.2pt\cellcolor{lime!100}\else\ifdim#1pt>89.6pt\cellcolor{lime!90}\else\ifdim#1pt>86.9pt\cellcolor{lime!80}\else\ifdim#1pt>84.2pt\cellcolor{lime!70}\else\ifdim#1pt>81.6pt\cellcolor{lime!60}\else\ifdim#1pt>78.9pt\cellcolor{lime!50}\else\ifdim#1pt>76.2pt\cellcolor{lime!40}\else\ifdim#1pt>73.5pt\cellcolor{lime!30}\else\ifdim#1pt>70.9pt\cellcolor{lime!20}\else\ifdim#1pt>68.2pt\cellcolor{lime!10}\else\cellcolor{lime!0}\fi\fi\fi\fi\fi\fi\fi\fi\fi\fi#1}

\def\weightedspearmanzerozerosrcthresholdperadapterMean#1{\ifdim#1pt>92.1pt\cellcolor{lime!100}\else\ifdim#1pt>91.0pt\cellcolor{lime!90}\else\ifdim#1pt>89.8pt\cellcolor{lime!80}\else\ifdim#1pt>88.7pt\cellcolor{lime!70}\else\ifdim#1pt>87.5pt\cellcolor{lime!60}\else\ifdim#1pt>86.3pt\cellcolor{lime!50}\else\ifdim#1pt>85.2pt\cellcolor{lime!40}\else\ifdim#1pt>84.0pt\cellcolor{lime!30}\else\ifdim#1pt>82.9pt\cellcolor{lime!20}\else\ifdim#1pt>81.7pt\cellcolor{lime!10}\else\cellcolor{lime!0}\fi\fi\fi\fi\fi\fi\fi\fi\fi\fi#1}

\def\weightedspearmanzerozerosrcthresholdperadapterStd#1{\ifdim#1pt<3.1pt\cellcolor{lime!100}\else\ifdim#1pt<3.9pt\cellcolor{lime!90}\else\ifdim#1pt<4.8pt\cellcolor{lime!80}\else\ifdim#1pt<5.6pt\cellcolor{lime!70}\else\ifdim#1pt<6.5pt\cellcolor{lime!60}\else\ifdim#1pt<7.4pt\cellcolor{lime!50}\else\ifdim#1pt<8.2pt\cellcolor{lime!40}\else\ifdim#1pt<9.1pt\cellcolor{lime!30}\else\ifdim#1pt<9.9pt\cellcolor{lime!20}\else\ifdim#1pt<10.8pt\cellcolor{lime!10}\else\cellcolor{lime!0}\fi\fi\fi\fi\fi\fi\fi\fi\fi\fi#1}

\begin{table}[H]
\centering
\caption{The weighted Spearman correlation of each validator/task pair for \textbf{MCD}.}
\label{domainnet126_weighted_spearman_0.0_src_threshold_per_adapter_MCD}
\resizebox{!}{0.22\textheight}{
\begin{tabular}{llrrrrrrrrrrrrrr}
\toprule
 &  & CP & CR & CS & PC & PR & PS & RC & RP & RS & SC & SP & SR & Mean & Std \\
\midrule
\multirow[c]{2}{*}{Accuracy} & Source Train & \weightedspearmanzerozerosrcthresholdperadapterDCP{95.8} & \weightedspearmanzerozerosrcthresholdperadapterDCR{82.8} & \weightedspearmanzerozerosrcthresholdperadapterDCS{92.6} & \weightedspearmanzerozerosrcthresholdperadapterDPC{87.9} & \weightedspearmanzerozerosrcthresholdperadapterDPR{71.7} & \weightedspearmanzerozerosrcthresholdperadapterDPS{84.7} & \weightedspearmanzerozerosrcthresholdperadapterDRC{87.3} & \weightedspearmanzerozerosrcthresholdperadapterDRP{93.8} & \weightedspearmanzerozerosrcthresholdperadapterDRS{92.8} & \weightedspearmanzerozerosrcthresholdperadapterDSC{92.5} & \weightedspearmanzerozerosrcthresholdperadapterDSP{92.9} & \weightedspearmanzerozerosrcthresholdperadapterDSR{88.7} & \weightedspearmanzerozerosrcthresholdperadapterMean{88.6} & \weightedspearmanzerozerosrcthresholdperadapterStd{6.3} \\
 & Source Val & \weightedspearmanzerozerosrcthresholdperadapterDCP{88.0} & \textbf{\weightedspearmanzerozerosrcthresholdperadapterDCR{89.9}} & \textbf{\weightedspearmanzerozerosrcthresholdperadapterDCS{93.2}} & \weightedspearmanzerozerosrcthresholdperadapterDPC{78.9} & \weightedspearmanzerozerosrcthresholdperadapterDPR{94.0} & \weightedspearmanzerozerosrcthresholdperadapterDPS{75.6} & \weightedspearmanzerozerosrcthresholdperadapterDRC{84.6} & \weightedspearmanzerozerosrcthresholdperadapterDRP{84.5} & \weightedspearmanzerozerosrcthresholdperadapterDRS{92.7} & \weightedspearmanzerozerosrcthresholdperadapterDSC{57.8} & \weightedspearmanzerozerosrcthresholdperadapterDSP{73.4} & \weightedspearmanzerozerosrcthresholdperadapterDSR{68.2} & \weightedspearmanzerozerosrcthresholdperadapterMean{81.7} & \weightedspearmanzerozerosrcthresholdperadapterStd{10.8} \\
\bottomrule
\end{tabular}}
\end{table}

%% file: supp_tables/domainnet126/weighted_spearman_0.0_src_threshold_per_adapter_MMD.tex
\def\weightedspearmanzerozerosrcthresholdperadapterDCP#1{\ifdim#1pt>89.3pt\cellcolor{lime!100}\else\ifdim#1pt>88.1pt\cellcolor{lime!90}\else\ifdim#1pt>87.0pt\cellcolor{lime!80}\else\ifdim#1pt>85.8pt\cellcolor{lime!70}\else\ifdim#1pt>84.6pt\cellcolor{lime!60}\else\ifdim#1pt>83.4pt\cellcolor{lime!50}\else\ifdim#1pt>82.2pt\cellcolor{lime!40}\else\ifdim#1pt>81.1pt\cellcolor{lime!30}\else\ifdim#1pt>79.9pt\cellcolor{lime!20}\else\ifdim#1pt>78.7pt\cellcolor{lime!10}\else\cellcolor{lime!0}\fi\fi\fi\fi\fi\fi\fi\fi\fi\fi#1}

\def\weightedspearmanzerozerosrcthresholdperadapterDCR#1{\ifdim#1pt>89.7pt\cellcolor{lime!100}\else\ifdim#1pt>88.7pt\cellcolor{lime!90}\else\ifdim#1pt>87.8pt\cellcolor{lime!80}\else\ifdim#1pt>86.9pt\cellcolor{lime!70}\else\ifdim#1pt>85.9pt\cellcolor{lime!60}\else\ifdim#1pt>85.0pt\cellcolor{lime!50}\else\ifdim#1pt>84.1pt\cellcolor{lime!40}\else\ifdim#1pt>83.2pt\cellcolor{lime!30}\else\ifdim#1pt>82.2pt\cellcolor{lime!20}\else\ifdim#1pt>81.3pt\cellcolor{lime!10}\else\cellcolor{lime!0}\fi\fi\fi\fi\fi\fi\fi\fi\fi\fi#1}

\def\weightedspearmanzerozerosrcthresholdperadapterDCS#1{\ifdim#1pt>87.9pt\cellcolor{lime!100}\else\ifdim#1pt>87.0pt\cellcolor{lime!90}\else\ifdim#1pt>86.2pt\cellcolor{lime!80}\else\ifdim#1pt>85.4pt\cellcolor{lime!70}\else\ifdim#1pt>84.6pt\cellcolor{lime!60}\else\ifdim#1pt>83.7pt\cellcolor{lime!50}\else\ifdim#1pt>82.9pt\cellcolor{lime!40}\else\ifdim#1pt>82.1pt\cellcolor{lime!30}\else\ifdim#1pt>81.2pt\cellcolor{lime!20}\else\ifdim#1pt>80.4pt\cellcolor{lime!10}\else\cellcolor{lime!0}\fi\fi\fi\fi\fi\fi\fi\fi\fi\fi#1}

\def\weightedspearmanzerozerosrcthresholdperadapterDPC#1{\ifdim#1pt>63.4pt\cellcolor{lime!100}\else\ifdim#1pt>63.4pt\cellcolor{lime!90}\else\ifdim#1pt>63.4pt\cellcolor{lime!80}\else\ifdim#1pt>63.4pt\cellcolor{lime!70}\else\ifdim#1pt>63.4pt\cellcolor{lime!60}\else\ifdim#1pt>63.4pt\cellcolor{lime!50}\else\ifdim#1pt>63.4pt\cellcolor{lime!40}\else\ifdim#1pt>63.4pt\cellcolor{lime!30}\else\ifdim#1pt>63.4pt\cellcolor{lime!20}\else\ifdim#1pt>63.4pt\cellcolor{lime!10}\else\cellcolor{lime!0}\fi\fi\fi\fi\fi\fi\fi\fi\fi\fi#1}

\def\weightedspearmanzerozerosrcthresholdperadapterDPR#1{\ifdim#1pt>90.2pt\cellcolor{lime!100}\else\ifdim#1pt>90.2pt\cellcolor{lime!90}\else\ifdim#1pt>90.2pt\cellcolor{lime!80}\else\ifdim#1pt>90.2pt\cellcolor{lime!70}\else\ifdim#1pt>90.2pt\cellcolor{lime!60}\else\ifdim#1pt>90.2pt\cellcolor{lime!50}\else\ifdim#1pt>90.2pt\cellcolor{lime!40}\else\ifdim#1pt>90.2pt\cellcolor{lime!30}\else\ifdim#1pt>90.2pt\cellcolor{lime!20}\else\ifdim#1pt>90.2pt\cellcolor{lime!10}\else\cellcolor{lime!0}\fi\fi\fi\fi\fi\fi\fi\fi\fi\fi#1}

\def\weightedspearmanzerozerosrcthresholdperadapterDPS#1{\ifdim#1pt>64.7pt\cellcolor{lime!100}\else\ifdim#1pt>63.6pt\cellcolor{lime!90}\else\ifdim#1pt>62.6pt\cellcolor{lime!80}\else\ifdim#1pt>61.5pt\cellcolor{lime!70}\else\ifdim#1pt>60.5pt\cellcolor{lime!60}\else\ifdim#1pt>59.5pt\cellcolor{lime!50}\else\ifdim#1pt>58.4pt\cellcolor{lime!40}\else\ifdim#1pt>57.4pt\cellcolor{lime!30}\else\ifdim#1pt>56.3pt\cellcolor{lime!20}\else\ifdim#1pt>55.3pt\cellcolor{lime!10}\else\cellcolor{lime!0}\fi\fi\fi\fi\fi\fi\fi\fi\fi\fi#1}

\def\weightedspearmanzerozerosrcthresholdperadapterDRC#1{\ifdim#1pt>87.8pt\cellcolor{lime!100}\else\ifdim#1pt>86.2pt\cellcolor{lime!90}\else\ifdim#1pt>84.6pt\cellcolor{lime!80}\else\ifdim#1pt>83.0pt\cellcolor{lime!70}\else\ifdim#1pt>81.4pt\cellcolor{lime!60}\else\ifdim#1pt>79.8pt\cellcolor{lime!50}\else\ifdim#1pt>78.2pt\cellcolor{lime!40}\else\ifdim#1pt>76.6pt\cellcolor{lime!30}\else\ifdim#1pt>75.0pt\cellcolor{lime!20}\else\ifdim#1pt>73.4pt\cellcolor{lime!10}\else\cellcolor{lime!0}\fi\fi\fi\fi\fi\fi\fi\fi\fi\fi#1}

\def\weightedspearmanzerozerosrcthresholdperadapterDRP#1{\ifdim#1pt>86.3pt\cellcolor{lime!100}\else\ifdim#1pt>85.7pt\cellcolor{lime!90}\else\ifdim#1pt>85.0pt\cellcolor{lime!80}\else\ifdim#1pt>84.3pt\cellcolor{lime!70}\else\ifdim#1pt>83.7pt\cellcolor{lime!60}\else\ifdim#1pt>83.0pt\cellcolor{lime!50}\else\ifdim#1pt>82.3pt\cellcolor{lime!40}\else\ifdim#1pt>81.6pt\cellcolor{lime!30}\else\ifdim#1pt>81.0pt\cellcolor{lime!20}\else\ifdim#1pt>80.3pt\cellcolor{lime!10}\else\cellcolor{lime!0}\fi\fi\fi\fi\fi\fi\fi\fi\fi\fi#1}

\def\weightedspearmanzerozerosrcthresholdperadapterDRS#1{\ifdim#1pt>83.7pt\cellcolor{lime!100}\else\ifdim#1pt>80.5pt\cellcolor{lime!90}\else\ifdim#1pt>77.3pt\cellcolor{lime!80}\else\ifdim#1pt>74.1pt\cellcolor{lime!70}\else\ifdim#1pt>71.0pt\cellcolor{lime!60}\else\ifdim#1pt>67.8pt\cellcolor{lime!50}\else\ifdim#1pt>64.6pt\cellcolor{lime!40}\else\ifdim#1pt>61.4pt\cellcolor{lime!30}\else\ifdim#1pt>58.2pt\cellcolor{lime!20}\else\ifdim#1pt>55.0pt\cellcolor{lime!10}\else\cellcolor{lime!0}\fi\fi\fi\fi\fi\fi\fi\fi\fi\fi#1}

\def\weightedspearmanzerozerosrcthresholdperadapterDSC#1{\ifdim#1pt>73.6pt\cellcolor{lime!100}\else\ifdim#1pt>72.9pt\cellcolor{lime!90}\else\ifdim#1pt>72.3pt\cellcolor{lime!80}\else\ifdim#1pt>71.6pt\cellcolor{lime!70}\else\ifdim#1pt>71.0pt\cellcolor{lime!60}\else\ifdim#1pt>70.4pt\cellcolor{lime!50}\else\ifdim#1pt>69.7pt\cellcolor{lime!40}\else\ifdim#1pt>69.1pt\cellcolor{lime!30}\else\ifdim#1pt>68.4pt\cellcolor{lime!20}\else\ifdim#1pt>67.8pt\cellcolor{lime!10}\else\cellcolor{lime!0}\fi\fi\fi\fi\fi\fi\fi\fi\fi\fi#1}

\def\weightedspearmanzerozerosrcthresholdperadapterDSP#1{\ifdim#1pt>75.7pt\cellcolor{lime!100}\else\ifdim#1pt>74.4pt\cellcolor{lime!90}\else\ifdim#1pt>73.2pt\cellcolor{lime!80}\else\ifdim#1pt>72.0pt\cellcolor{lime!70}\else\ifdim#1pt>70.8pt\cellcolor{lime!60}\else\ifdim#1pt>69.5pt\cellcolor{lime!50}\else\ifdim#1pt>68.3pt\cellcolor{lime!40}\else\ifdim#1pt>67.1pt\cellcolor{lime!30}\else\ifdim#1pt>65.8pt\cellcolor{lime!20}\else\ifdim#1pt>64.6pt\cellcolor{lime!10}\else\cellcolor{lime!0}\fi\fi\fi\fi\fi\fi\fi\fi\fi\fi#1}

\def\weightedspearmanzerozerosrcthresholdperadapterDSR#1{\ifdim#1pt>85.5pt\cellcolor{lime!100}\else\ifdim#1pt>83.8pt\cellcolor{lime!90}\else\ifdim#1pt>82.0pt\cellcolor{lime!80}\else\ifdim#1pt>80.3pt\cellcolor{lime!70}\else\ifdim#1pt>78.6pt\cellcolor{lime!60}\else\ifdim#1pt>76.9pt\cellcolor{lime!50}\else\ifdim#1pt>75.2pt\cellcolor{lime!40}\else\ifdim#1pt>73.4pt\cellcolor{lime!30}\else\ifdim#1pt>71.7pt\cellcolor{lime!20}\else\ifdim#1pt>70.0pt\cellcolor{lime!10}\else\cellcolor{lime!0}\fi\fi\fi\fi\fi\fi\fi\fi\fi\fi#1}

\def\weightedspearmanzerozerosrcthresholdperadapterMean#1{\ifdim#1pt>75.7pt\cellcolor{lime!100}\else\ifdim#1pt>75.2pt\cellcolor{lime!90}\else\ifdim#1pt>74.8pt\cellcolor{lime!80}\else\ifdim#1pt>74.3pt\cellcolor{lime!70}\else\ifdim#1pt>73.9pt\cellcolor{lime!60}\else\ifdim#1pt>73.5pt\cellcolor{lime!50}\else\ifdim#1pt>73.0pt\cellcolor{lime!40}\else\ifdim#1pt>72.6pt\cellcolor{lime!30}\else\ifdim#1pt>72.1pt\cellcolor{lime!20}\else\ifdim#1pt>71.7pt\cellcolor{lime!10}\else\cellcolor{lime!0}\fi\fi\fi\fi\fi\fi\fi\fi\fi\fi#1}

\def\weightedspearmanzerozerosrcthresholdperadapterStd#1{\ifdim#1pt<2.0pt\cellcolor{lime!100}\else\ifdim#1pt<3.0pt\cellcolor{lime!90}\else\ifdim#1pt<3.9pt\cellcolor{lime!80}\else\ifdim#1pt<4.9pt\cellcolor{lime!70}\else\ifdim#1pt<5.8pt\cellcolor{lime!60}\else\ifdim#1pt<6.7pt\cellcolor{lime!50}\else\ifdim#1pt<7.7pt\cellcolor{lime!40}\else\ifdim#1pt<8.6pt\cellcolor{lime!30}\else\ifdim#1pt<9.6pt\cellcolor{lime!20}\else\ifdim#1pt<10.5pt\cellcolor{lime!10}\else\cellcolor{lime!0}\fi\fi\fi\fi\fi\fi\fi\fi\fi\fi#1}

\begin{table}[H]
\centering
\caption{The weighted Spearman correlation of each validator/task pair for \textbf{MMD}.}
\label{domainnet126_weighted_spearman_0.0_src_threshold_per_adapter_MMD}
\resizebox{!}{0.22\textheight}{
\begin{tabular}{llrrrrrrrrrrrrrr}
\toprule
 &  & CP & CR & CS & PC & PR & PS & RC & RP & RS & SC & SP & SR & Mean & Std \\
\midrule
\multirow[c]{2}{*}{Accuracy} & Source Train & \weightedspearmanzerozerosrcthresholdperadapterDCP{79.6} & \weightedspearmanzerozerosrcthresholdperadapterDCR{78.2} & \weightedspearmanzerozerosrcthresholdperadapterDCS{67.0} & \weightedspearmanzerozerosrcthresholdperadapterDPC{40.6} & \weightedspearmanzerozerosrcthresholdperadapterDPR{60.6} & \weightedspearmanzerozerosrcthresholdperadapterDPS{47.0} & \weightedspearmanzerozerosrcthresholdperadapterDRC{73.7} & \weightedspearmanzerozerosrcthresholdperadapterDRP{70.0} & \weightedspearmanzerozerosrcthresholdperadapterDRS{52.5} & \weightedspearmanzerozerosrcthresholdperadapterDSC{66.2} & \weightedspearmanzerozerosrcthresholdperadapterDSP{69.2} & \weightedspearmanzerozerosrcthresholdperadapterDSR{78.1} & \weightedspearmanzerozerosrcthresholdperadapterMean{65.2} & \weightedspearmanzerozerosrcthresholdperadapterStd{12.2} \\
 & Source Val & \weightedspearmanzerozerosrcthresholdperadapterDCP{78.7} & \weightedspearmanzerozerosrcthresholdperadapterDCR{81.3} & \weightedspearmanzerozerosrcthresholdperadapterDCS{80.4} & \textbf{\weightedspearmanzerozerosrcthresholdperadapterDPC{63.4}} & \textbf{\weightedspearmanzerozerosrcthresholdperadapterDPR{90.2}} & \weightedspearmanzerozerosrcthresholdperadapterDPS{55.3} & \weightedspearmanzerozerosrcthresholdperadapterDRC{73.4} & \weightedspearmanzerozerosrcthresholdperadapterDRP{80.3} & \weightedspearmanzerozerosrcthresholdperadapterDRS{55.0} & \weightedspearmanzerozerosrcthresholdperadapterDSC{67.8} & \weightedspearmanzerozerosrcthresholdperadapterDSP{64.6} & \weightedspearmanzerozerosrcthresholdperadapterDSR{70.0} & \weightedspearmanzerozerosrcthresholdperadapterMean{71.7} & \weightedspearmanzerozerosrcthresholdperadapterStd{10.5} \\
\bottomrule
\end{tabular}}
\end{table}

%% file: supp_tables/mnist/weighted_spearman_0.0_src_threshold.tex
\def\weightedspearmanzerozerosrcthresholdMM#1{\ifdim#1pt>48.8pt\cellcolor{lime!100}\else\ifdim#1pt>42.9pt\cellcolor{lime!90}\else\ifdim#1pt>37.0pt\cellcolor{lime!80}\else\ifdim#1pt>31.1pt\cellcolor{lime!70}\else\ifdim#1pt>25.2pt\cellcolor{lime!60}\else\ifdim#1pt>19.3pt\cellcolor{lime!50}\else\ifdim#1pt>13.4pt\cellcolor{lime!40}\else\ifdim#1pt>7.5pt\cellcolor{lime!30}\else\ifdim#1pt>1.6pt\cellcolor{lime!20}\else\ifdim#1pt>-4.3pt\cellcolor{lime!10}\else\cellcolor{lime!0}\fi\fi\fi\fi\fi\fi\fi\fi\fi\fi#1}

\begin{table}[H]
\centering
\caption{The weighted Spearman correlation for the MNIST $\rightarrow$ MNISTM task, using the checkpoints of all algorithms.}
\label{mnist_weighted_spearman_0.0_src_threshold}
\resizebox{!}{0.21\textheight}{\begin{tabular}{llr}
\toprule
 &  & MM \\
\midrule
\multirow[c]{2}{*}{Accuracy} & Source Train & \weightedspearmanzerozerosrcthresholdMM{7.7} \\
 & Source Val & \weightedspearmanzerozerosrcthresholdMM{-4.3} \\
\cline{1-3}
\multirow[c]{5}{*}{BNM} & Source Train & \weightedspearmanzerozerosrcthresholdMM{-18.9} \\
 & Source Train + Target & \weightedspearmanzerozerosrcthresholdMM{53.3} \\
 & Source Val & \weightedspearmanzerozerosrcthresholdMM{-23.1} \\
 & Source Val + Target & \weightedspearmanzerozerosrcthresholdMM{49.7} \\
 & Target & \textbf{\weightedspearmanzerozerosrcthresholdMM{54.7}} \\
\cline{1-3}
\multirow[c]{4}{*}{ClassAMI} & Source + Target Features & \weightedspearmanzerozerosrcthresholdMM{-31.0} \\
 & Source + Target Logits & \weightedspearmanzerozerosrcthresholdMM{-21.2} \\
 & Target Features & \weightedspearmanzerozerosrcthresholdMM{-3.0} \\
 & Target Logits & \weightedspearmanzerozerosrcthresholdMM{-16.1} \\
\cline{1-3}
\multirow[c]{4}{*}{ClassSS} & Source + Target Features & \weightedspearmanzerozerosrcthresholdMM{-71.5} \\
 & Source + Target Logits & \weightedspearmanzerozerosrcthresholdMM{-61.6} \\
 & Target Features & \weightedspearmanzerozerosrcthresholdMM{-59.1} \\
 & Target Logits & \weightedspearmanzerozerosrcthresholdMM{-48.0} \\
\cline{1-3}
\multirow[c]{3}{*}{DEV} & Features & \weightedspearmanzerozerosrcthresholdMM{-30.6} \\
 & Logits & \weightedspearmanzerozerosrcthresholdMM{-14.1} \\
 & Preds & \weightedspearmanzerozerosrcthresholdMM{-31.3} \\
\cline{1-3}
\multirow[c]{3}{*}{DEVN} & Features, max normalization & \weightedspearmanzerozerosrcthresholdMM{-43.7} \\
 & Logits, max normalization & \weightedspearmanzerozerosrcthresholdMM{-44.1} \\
 & Preds, max normalization & \weightedspearmanzerozerosrcthresholdMM{-54.7} \\
\cline{1-3}
\multirow[c]{5}{*}{Entropy} & Source Train & \weightedspearmanzerozerosrcthresholdMM{-31.5} \\
 & Source Train + Target & \weightedspearmanzerozerosrcthresholdMM{-34.3} \\
 & Source Val & \weightedspearmanzerozerosrcthresholdMM{-34.8} \\
 & Source Val + Target & \weightedspearmanzerozerosrcthresholdMM{-35.1} \\
 & Target & \weightedspearmanzerozerosrcthresholdMM{-38.5} \\
\cline{1-3}
\multirow[c]{9}{*}{SND} & Features, $\tau=0.05$ & \weightedspearmanzerozerosrcthresholdMM{-78.1} \\
 & Features, $\tau=0.1$ & \weightedspearmanzerozerosrcthresholdMM{-78.3} \\
 & Features, $\tau=0.5$ & \weightedspearmanzerozerosrcthresholdMM{-79.2} \\
 & Logits, $\tau=0.05$ & \weightedspearmanzerozerosrcthresholdMM{-80.2} \\
 & Logits, $\tau=0.1$ & \weightedspearmanzerozerosrcthresholdMM{-86.3} \\
 & Logits, $\tau=0.5$ & \weightedspearmanzerozerosrcthresholdMM{-87.5} \\
 & Preds, $\tau=0.05$ & \weightedspearmanzerozerosrcthresholdMM{-82.5} \\
 & Preds, $\tau=0.1$ & \weightedspearmanzerozerosrcthresholdMM{-86.3} \\
 & Preds, $\tau=0.5$ & \weightedspearmanzerozerosrcthresholdMM{-85.1} \\
\cline{1-3}
\bottomrule
\end{tabular}}
\end{table}

%% file: supp_tables/mnist/weighted_spearman_0.0_src_threshold_per_adapter.tex
\def\weightedspearmanzerozerosrcthresholdperadapterATDOC#1{\ifdim#1pt>70.9pt\cellcolor{lime!100}\else\ifdim#1pt>63.6pt\cellcolor{lime!90}\else\ifdim#1pt>56.4pt\cellcolor{lime!80}\else\ifdim#1pt>49.2pt\cellcolor{lime!70}\else\ifdim#1pt>41.9pt\cellcolor{lime!60}\else\ifdim#1pt>34.7pt\cellcolor{lime!50}\else\ifdim#1pt>27.5pt\cellcolor{lime!40}\else\ifdim#1pt>20.3pt\cellcolor{lime!30}\else\ifdim#1pt>13.0pt\cellcolor{lime!20}\else\ifdim#1pt>5.8pt\cellcolor{lime!10}\else\cellcolor{lime!0}\fi\fi\fi\fi\fi\fi\fi\fi\fi\fi#1}

\def\weightedspearmanzerozerosrcthresholdperadapterBNM#1{\ifdim#1pt>51.0pt\cellcolor{lime!100}\else\ifdim#1pt>48.7pt\cellcolor{lime!90}\else\ifdim#1pt>46.5pt\cellcolor{lime!80}\else\ifdim#1pt>44.2pt\cellcolor{lime!70}\else\ifdim#1pt>42.0pt\cellcolor{lime!60}\else\ifdim#1pt>39.7pt\cellcolor{lime!50}\else\ifdim#1pt>37.5pt\cellcolor{lime!40}\else\ifdim#1pt>35.2pt\cellcolor{lime!30}\else\ifdim#1pt>33.0pt\cellcolor{lime!20}\else\ifdim#1pt>30.7pt\cellcolor{lime!10}\else\cellcolor{lime!0}\fi\fi\fi\fi\fi\fi\fi\fi\fi\fi#1}

\def\weightedspearmanzerozerosrcthresholdperadapterBSP#1{\ifdim#1pt>69.9pt\cellcolor{lime!100}\else\ifdim#1pt>62.7pt\cellcolor{lime!90}\else\ifdim#1pt>55.6pt\cellcolor{lime!80}\else\ifdim#1pt>48.5pt\cellcolor{lime!70}\else\ifdim#1pt>41.4pt\cellcolor{lime!60}\else\ifdim#1pt>34.2pt\cellcolor{lime!50}\else\ifdim#1pt>27.1pt\cellcolor{lime!40}\else\ifdim#1pt>20.0pt\cellcolor{lime!30}\else\ifdim#1pt>12.8pt\cellcolor{lime!20}\else\ifdim#1pt>5.7pt\cellcolor{lime!10}\else\cellcolor{lime!0}\fi\fi\fi\fi\fi\fi\fi\fi\fi\fi#1}

\def\weightedspearmanzerozerosrcthresholdperadapterCDAN#1{\ifdim#1pt>31.7pt\cellcolor{lime!100}\else\ifdim#1pt>25.4pt\cellcolor{lime!90}\else\ifdim#1pt>19.2pt\cellcolor{lime!80}\else\ifdim#1pt>12.9pt\cellcolor{lime!70}\else\ifdim#1pt>6.7pt\cellcolor{lime!60}\else\ifdim#1pt>0.5pt\cellcolor{lime!50}\else\ifdim#1pt>-5.8pt\cellcolor{lime!40}\else\ifdim#1pt>-12.0pt\cellcolor{lime!30}\else\ifdim#1pt>-18.3pt\cellcolor{lime!20}\else\ifdim#1pt>-24.5pt\cellcolor{lime!10}\else\cellcolor{lime!0}\fi\fi\fi\fi\fi\fi\fi\fi\fi\fi#1}

\def\weightedspearmanzerozerosrcthresholdperadapterDANN#1{\ifdim#1pt>48.8pt\cellcolor{lime!100}\else\ifdim#1pt>40.7pt\cellcolor{lime!90}\else\ifdim#1pt>32.7pt\cellcolor{lime!80}\else\ifdim#1pt>24.6pt\cellcolor{lime!70}\else\ifdim#1pt>16.6pt\cellcolor{lime!60}\else\ifdim#1pt>8.6pt\cellcolor{lime!50}\else\ifdim#1pt>0.5pt\cellcolor{lime!40}\else\ifdim#1pt>-7.5pt\cellcolor{lime!30}\else\ifdim#1pt>-15.6pt\cellcolor{lime!20}\else\ifdim#1pt>-23.6pt\cellcolor{lime!10}\else\cellcolor{lime!0}\fi\fi\fi\fi\fi\fi\fi\fi\fi\fi#1}

\def\weightedspearmanzerozerosrcthresholdperadapterGVB#1{\ifdim#1pt>23.1pt\cellcolor{lime!100}\else\ifdim#1pt>20.7pt\cellcolor{lime!90}\else\ifdim#1pt>18.4pt\cellcolor{lime!80}\else\ifdim#1pt>16.0pt\cellcolor{lime!70}\else\ifdim#1pt>13.7pt\cellcolor{lime!60}\else\ifdim#1pt>11.4pt\cellcolor{lime!50}\else\ifdim#1pt>9.0pt\cellcolor{lime!40}\else\ifdim#1pt>6.7pt\cellcolor{lime!30}\else\ifdim#1pt>4.3pt\cellcolor{lime!20}\else\ifdim#1pt>2.0pt\cellcolor{lime!10}\else\cellcolor{lime!0}\fi\fi\fi\fi\fi\fi\fi\fi\fi\fi#1}

\def\weightedspearmanzerozerosrcthresholdperadapterIM#1{\ifdim#1pt>60.2pt\cellcolor{lime!100}\else\ifdim#1pt>56.3pt\cellcolor{lime!90}\else\ifdim#1pt>52.3pt\cellcolor{lime!80}\else\ifdim#1pt>48.4pt\cellcolor{lime!70}\else\ifdim#1pt>44.5pt\cellcolor{lime!60}\else\ifdim#1pt>40.6pt\cellcolor{lime!50}\else\ifdim#1pt>36.7pt\cellcolor{lime!40}\else\ifdim#1pt>32.7pt\cellcolor{lime!30}\else\ifdim#1pt>28.8pt\cellcolor{lime!20}\else\ifdim#1pt>24.9pt\cellcolor{lime!10}\else\cellcolor{lime!0}\fi\fi\fi\fi\fi\fi\fi\fi\fi\fi#1}

\def\weightedspearmanzerozerosrcthresholdperadapterMCC#1{\ifdim#1pt>47.7pt\cellcolor{lime!100}\else\ifdim#1pt>42.7pt\cellcolor{lime!90}\else\ifdim#1pt>37.6pt\cellcolor{lime!80}\else\ifdim#1pt>32.6pt\cellcolor{lime!70}\else\ifdim#1pt>27.5pt\cellcolor{lime!60}\else\ifdim#1pt>22.4pt\cellcolor{lime!50}\else\ifdim#1pt>17.4pt\cellcolor{lime!40}\else\ifdim#1pt>12.3pt\cellcolor{lime!30}\else\ifdim#1pt>7.3pt\cellcolor{lime!20}\else\ifdim#1pt>2.2pt\cellcolor{lime!10}\else\cellcolor{lime!0}\fi\fi\fi\fi\fi\fi\fi\fi\fi\fi#1}

\def\weightedspearmanzerozerosrcthresholdperadapterMCD#1{\ifdim#1pt>78.3pt\cellcolor{lime!100}\else\ifdim#1pt>68.5pt\cellcolor{lime!90}\else\ifdim#1pt>58.7pt\cellcolor{lime!80}\else\ifdim#1pt>48.9pt\cellcolor{lime!70}\else\ifdim#1pt>39.1pt\cellcolor{lime!60}\else\ifdim#1pt>29.4pt\cellcolor{lime!50}\else\ifdim#1pt>19.6pt\cellcolor{lime!40}\else\ifdim#1pt>9.8pt\cellcolor{lime!30}\else\ifdim#1pt>-0.0pt\cellcolor{lime!20}\else\ifdim#1pt>-9.8pt\cellcolor{lime!10}\else\cellcolor{lime!0}\fi\fi\fi\fi\fi\fi\fi\fi\fi\fi#1}

\def\weightedspearmanzerozerosrcthresholdperadapterMMD#1{\ifdim#1pt>17.2pt\cellcolor{lime!100}\else\ifdim#1pt>17.2pt\cellcolor{lime!90}\else\ifdim#1pt>17.2pt\cellcolor{lime!80}\else\ifdim#1pt>17.2pt\cellcolor{lime!70}\else\ifdim#1pt>17.1pt\cellcolor{lime!60}\else\ifdim#1pt>17.1pt\cellcolor{lime!50}\else\ifdim#1pt>17.1pt\cellcolor{lime!40}\else\ifdim#1pt>17.1pt\cellcolor{lime!30}\else\ifdim#1pt>17.1pt\cellcolor{lime!20}\else\ifdim#1pt>17.1pt\cellcolor{lime!10}\else\cellcolor{lime!0}\fi\fi\fi\fi\fi\fi\fi\fi\fi\fi#1}

\def\weightedspearmanzerozerosrcthresholdperadapterMean#1{\ifdim#1pt>45.0pt\cellcolor{lime!100}\else\ifdim#1pt>40.4pt\cellcolor{lime!90}\else\ifdim#1pt>35.7pt\cellcolor{lime!80}\else\ifdim#1pt>31.0pt\cellcolor{lime!70}\else\ifdim#1pt>26.4pt\cellcolor{lime!60}\else\ifdim#1pt>21.7pt\cellcolor{lime!50}\else\ifdim#1pt>17.0pt\cellcolor{lime!40}\else\ifdim#1pt>12.3pt\cellcolor{lime!30}\else\ifdim#1pt>7.7pt\cellcolor{lime!20}\else\ifdim#1pt>3.0pt\cellcolor{lime!10}\else\cellcolor{lime!0}\fi\fi\fi\fi\fi\fi\fi\fi\fi\fi#1}

\def\weightedspearmanzerozerosrcthresholdperadapterStd#1{\ifdim#1pt<11.2pt\cellcolor{lime!100}\else\ifdim#1pt<11.9pt\cellcolor{lime!90}\else\ifdim#1pt<12.6pt\cellcolor{lime!80}\else\ifdim#1pt<13.3pt\cellcolor{lime!70}\else\ifdim#1pt<14.1pt\cellcolor{lime!60}\else\ifdim#1pt<14.8pt\cellcolor{lime!50}\else\ifdim#1pt<15.5pt\cellcolor{lime!40}\else\ifdim#1pt<16.2pt\cellcolor{lime!30}\else\ifdim#1pt<16.9pt\cellcolor{lime!20}\else\ifdim#1pt<17.6pt\cellcolor{lime!10}\else\cellcolor{lime!0}\fi\fi\fi\fi\fi\fi\fi\fi\fi\fi#1}

\begin{table}[H]
\centering
\caption{The weighted Spearman correlation of each validator/algorithm pair, for the MNIST $\rightarrow$ MNISTM task.}
\label{mnist_weighted_spearman_0.0_src_threshold_per_adapter}
\resizebox{!}{0.22\textheight}{
\begin{tabular}{llrrrrrrrrrrrr}
\toprule
 &  & ATDOC & BNM & BSP & CDAN & DANN & GVB & IM & MCC & MCD & MMD & Mean & Std \\
\midrule
\multirow[c]{2}{*}{Accuracy} & Source Train & \weightedspearmanzerozerosrcthresholdperadapterATDOC{-13.8} & \weightedspearmanzerozerosrcthresholdperadapterBNM{38.0} & \weightedspearmanzerozerosrcthresholdperadapterBSP{-11.5} & \weightedspearmanzerozerosrcthresholdperadapterCDAN{-24.1} & \weightedspearmanzerozerosrcthresholdperadapterDANN{-2.1} & \weightedspearmanzerozerosrcthresholdperadapterGVB{5.5} & \weightedspearmanzerozerosrcthresholdperadapterIM{24.0} & \weightedspearmanzerozerosrcthresholdperadapterMCC{11.2} & \weightedspearmanzerozerosrcthresholdperadapterMCD{-16.2} & \textbf{\weightedspearmanzerozerosrcthresholdperadapterMMD{17.2}} & \weightedspearmanzerozerosrcthresholdperadapterMean{2.8} & \weightedspearmanzerozerosrcthresholdperadapterStd{18.9} \\
 & Source Val & \weightedspearmanzerozerosrcthresholdperadapterATDOC{5.8} & \weightedspearmanzerozerosrcthresholdperadapterBNM{30.7} & \weightedspearmanzerozerosrcthresholdperadapterBSP{5.7} & \weightedspearmanzerozerosrcthresholdperadapterCDAN{-24.5} & \weightedspearmanzerozerosrcthresholdperadapterDANN{-23.6} & \weightedspearmanzerozerosrcthresholdperadapterGVB{2.0} & \weightedspearmanzerozerosrcthresholdperadapterIM{24.9} & \weightedspearmanzerozerosrcthresholdperadapterMCC{2.2} & \weightedspearmanzerozerosrcthresholdperadapterMCD{-9.8} & \weightedspearmanzerozerosrcthresholdperadapterMMD{17.1} & \weightedspearmanzerozerosrcthresholdperadapterMean{3.0} & \weightedspearmanzerozerosrcthresholdperadapterStd{17.6} \\
\bottomrule
\end{tabular}}
\end{table}

%% file: main.bbl
\begin{thebibliography}{10}\itemsep=-1pt

\bibitem{optunaPaper}
Takuya Akiba, Shotaro Sano, Toshihiko Yanase, Takeru Ohta, and Masanori Koyama.
\newblock Optuna.
\newblock {\em Proceedings of the 25th ACM SIGKDD International Conference on
  Knowledge Discovery \& Data Mining}, Jul 2019.

\bibitem{bailey2018weighted}
Paul Bailey, Ahmad Emad, Ting Zhang, Qingshu Xie, and Emmanuel Sikali.
\newblock Weighted and unweighted correlation methods for large-scale
  educational assessment: wcorr formulas. air--naep working paper no. 2018-01.
  nces data r project series\# 02.
\newblock {\em American Institutes for Research}, 2018.

\bibitem{Brown2020LanguageMA}
Tom~B. Brown, Benjamin Mann, Nick Ryder, Melanie Subbiah, Jared Kaplan,
  Prafulla Dhariwal, Arvind Neelakantan, Pranav Shyam, Girish Sastry, Amanda
  Askell, Sandhini Agarwal, Ariel Herbert-Voss, Gretchen Krueger, T.~J.
  Henighan, Rewon Child, Aditya Ramesh, Daniel~M. Ziegler, Jeff Wu, Clemens
  Winter, Christopher Hesse, Mark Chen, Eric Sigler, Mateusz Litwin, Scott
  Gray, Benjamin Chess, Jack Clark, Christopher Berner, Sam McCandlish, Alec
  Radford, Ilya Sutskever, and Dario Amodei.
\newblock Language models are few-shot learners.
\newblock {\em ArXiv}, abs/2005.14165, 2020.

\bibitem{cao2018partial}
Zhangjie Cao, Lijia Ma, Mingsheng Long, and Jianmin Wang.
\newblock Partial adversarial domain adaptation.
\newblock In {\em Proceedings of the European Conference on Computer Vision
  (ECCV)}, pages 135--150, 2018.

\bibitem{chang2019domain}
Woong-Gi Chang, Tackgeun You, Seonguk Seo, Suha Kwak, and Bohyung Han.
\newblock Domain-specific batch normalization for unsupervised domain
  adaptation.
\newblock In {\em Proceedings of the IEEE/CVF Conference on Computer Vision and
  Pattern Recognition}, pages 7354--7362, 2019.

\bibitem{chen2019transferability}
Xinyang Chen, Sinan Wang, Mingsheng Long, and Jianmin Wang.
\newblock Transferability vs. discriminability: Batch spectral penalization for
  adversarial domain adaptation.
\newblock In {\em International conference on machine learning}, pages
  1081--1090. PMLR, 2019.

\bibitem{chuang2020estimating}
Ching-Yao Chuang, Antonio Torralba, and Stefanie Jegelka.
\newblock Estimating generalization under distribution shifts via
  domain-invariant representations.
\newblock {\em International conference on machine learning}, 2020.

\bibitem{cui2020towards}
Shuhao Cui, Shuhui Wang, Junbao Zhuo, Liang Li, Qingming Huang, and Qi Tian.
\newblock Towards discriminability and diversity: Batch nuclear-norm
  maximization under label insufficient situations.
\newblock In {\em Proceedings of the IEEE/CVF Conference on Computer Vision and
  Pattern Recognition}, pages 3941--3950, 2020.

\bibitem{Cui2021FastBN}
Shuhao Cui, Shuhui Wang, Junbao Zhuo, Liang Li, Qingming Huang, and Qi Tian.
\newblock Fast batch nuclear-norm maximization and minimization for robust
  domain adaptation.
\newblock {\em ArXiv}, abs/2107.06154, 2021.

\bibitem{cui2020gradually}
Shuhao Cui, Shuhui Wang, Junbao Zhuo, Chi Su, Qingming Huang, and Qi Tian.
\newblock Gradually vanishing bridge for adversarial domain adaptation.
\newblock In {\em Proceedings of the IEEE/CVF Conference on Computer Vision and
  Pattern Recognition}, pages 12455--12464, 2020.

\bibitem{ganin2016domain}
Yaroslav Ganin, Evgeniya Ustinova, Hana Ajakan, Pascal Germain, Hugo
  Larochelle, Fran{\c{c}}ois Laviolette, Mario Marchand, and Victor Lempitsky.
\newblock Domain-adversarial training of neural networks.
\newblock {\em The journal of machine learning research}, 17(1):2096--2030,
  2016.

\bibitem{ghifary2016deep}
Muhammad Ghifary, W~Bastiaan Kleijn, Mengjie Zhang, David Balduzzi, and Wen Li.
\newblock Deep reconstruction-classification networks for unsupervised domain
  adaptation.
\newblock In {\em European conference on computer vision}, pages 597--613.
  Springer, 2016.

\bibitem{goodfellow2020generative}
Ian Goodfellow, Jean Pouget-Abadie, Mehdi Mirza, Bing Xu, David Warde-Farley,
  Sherjil Ozair, Aaron Courville, and Yoshua Bengio.
\newblock Generative adversarial networks.
\newblock {\em Communications of the ACM}, 63(11):139--144, 2020.

\bibitem{He_2016_CVPR}
Kaiming He, Xiangyu Zhang, Shaoqing Ren, and Jian Sun.
\newblock Deep residual learning for image recognition.
\newblock In {\em Proceedings of the IEEE Conference on Computer Vision and
  Pattern Recognition (CVPR)}, June 2016.

\bibitem{ho2020denoising}
Jonathan Ho, Ajay Jain, and Pieter Abbeel.
\newblock Denoising diffusion probabilistic models.
\newblock {\em Advances in Neural Information Processing Systems},
  33:6840--6851, 2020.

\bibitem{jin2020minimum}
Ying Jin, Ximei Wang, Mingsheng Long, and Jianmin Wang.
\newblock Minimum class confusion for versatile domain adaptation.
\newblock In {\em European Conference on Computer Vision}, pages 464--480.
  Springer, 2020.

\bibitem{kingma2014adam}
Diederik~P. Kingma and Jimmy Ba.
\newblock Adam: A method for stochastic optimization, 2014.

\bibitem{lee2019sliced}
Chen-Yu Lee, Tanmay Batra, Mohammad~Haris Baig, and Daniel Ulbricht.
\newblock Sliced wasserstein discrepancy for unsupervised domain adaptation.
\newblock In {\em Proceedings of the IEEE/CVF Conference on Computer Vision and
  Pattern Recognition}, pages 10285--10295, 2019.

\bibitem{liang2020shot}
Jian Liang, Dapeng Hu, and Jiashi Feng.
\newblock Do we really need to access the source data? source hypothesis
  transfer for unsupervised domain adaptation.
\newblock In {\em International Conference on Machine Learning (ICML)}, pages
  6028--6039, July 13--18 2020.

\bibitem{liang2021domain}
Jian Liang, Dapeng Hu, and Jiashi Feng.
\newblock Domain adaptation with auxiliary target domain-oriented classifier.
\newblock In {\em Proceedings of the IEEE/CVF Conference on Computer Vision and
  Pattern Recognition}, pages 16632--16642, 2021.

\bibitem{long2015learning}
Mingsheng Long, Yue Cao, Jianmin Wang, and Michael Jordan.
\newblock Learning transferable features with deep adaptation networks.
\newblock In {\em International conference on machine learning}, pages 97--105.
  PMLR, 2015.

\bibitem{long2017conditional}
Mingsheng Long, Zhangjie Cao, Jianmin Wang, and Michael~I Jordan.
\newblock Conditional adversarial domain adaptation.
\newblock {\em arXiv preprint arXiv:1705.10667}, 2017.

\bibitem{long2016unsupervised}
Mingsheng Long, Han Zhu, Jianmin Wang, and Michael~I Jordan.
\newblock Unsupervised domain adaptation with residual transfer networks.
\newblock {\em arXiv preprint arXiv:1602.04433}, 2016.

\bibitem{long2017deep}
Mingsheng Long, Han Zhu, Jianmin Wang, and Michael~I Jordan.
\newblock Deep transfer learning with joint adaptation networks.
\newblock In {\em International conference on machine learning}, pages
  2208--2217. PMLR, 2017.

\bibitem{Loshchilov2016SGDRSG}
Ilya Loshchilov and Frank Hutter.
\newblock Sgdr: Stochastic gradient descent with warm restarts.
\newblock {\em arXiv: Learning}, 2016.

\bibitem{lu2020stochastic}
Zhihe Lu, Yongxin Yang, Xiatian Zhu, Cong Liu, Yi-Zhe Song, and Tao Xiang.
\newblock Stochastic classifiers for unsupervised domain adaptation.
\newblock In {\em Proceedings of the IEEE/CVF Conference on Computer Vision and
  Pattern Recognition}, pages 9111--9120, 2020.

\bibitem{Musgrave2021UnsupervisedDA}
Kevin Musgrave, Serge~J. Belongie, and Ser~Nam Lim.
\newblock Unsupervised domain adaptation: A reality check.
\newblock {\em ArXiv}, abs/2111.15672, 2021.

\bibitem{oza2021unsupervised}
Poojan Oza, Vishwanath~A Sindagi, Vibashan VS, and Vishal~M Patel.
\newblock Unsupervised domain adaption of object detectors: A survey.
\newblock {\em arXiv preprint arXiv:2105.13502}, 2021.

\bibitem{panareda2017open}
Pau Panareda~Busto and Juergen Gall.
\newblock Open set domain adaptation.
\newblock In {\em Proceedings of the IEEE International Conference on Computer
  Vision}, pages 754--763, 2017.

\bibitem{pytorchPaper}
Adam Paszke, Sam Gross, Francisco Massa, Adam Lerer, James Bradbury, Gregory
  Chanan, Trevor Killeen, Zeming Lin, Natalia Gimelshein, Luca Antiga, Alban
  Desmaison, Andreas Kopf, Edward Yang, Zachary DeVito, Martin Raison, Alykhan
  Tejani, Sasank Chilamkurthy, Benoit Steiner, Lu Fang, Junjie Bai, and Soumith
  Chintala.
\newblock Pytorch: An imperative style, high-performance deep learning library.
\newblock In H. Wallach, H. Larochelle, A. Beygelzimer, F. d\textquotesingle
  Alch\'{e}-Buc, E. Fox, and R. Garnett, editors, {\em Advances in Neural
  Information Processing Systems 32}, pages 8024--8035. Curran Associates,
  Inc., 2019.

\bibitem{peng2019moment}
Xingchao Peng, Qinxun Bai, Xide Xia, Zijun Huang, Kate Saenko, and Bo Wang.
\newblock Moment matching for multi-source domain adaptation.
\newblock In {\em Proceedings of the IEEE/CVF International Conference on
  Computer Vision}, pages 1406--1415, 2019.

\bibitem{prabhu2021sentry}
Viraj Prabhu, Shivam Khare, Deeksha Kartik, and Judy Hoffman.
\newblock Sentry: Selective entropy optimization via committee consistency for
  unsupervised domain adaptation.
\newblock In {\em Proceedings of the IEEE/CVF International Conference on
  Computer Vision}, pages 8558--8567, 2021.

\bibitem{ramponi-plank-2020-neural}
Alan Ramponi and Barbara Plank.
\newblock Neural unsupervised domain adaptation in {NLP}{---}{A} survey.
\newblock In {\em Proceedings of the 28th International Conference on
  Computational Linguistics}, pages 6838--6855, Barcelona, Spain (Online), Dec.
  2020. International Committee on Computational Linguistics.

\bibitem{robbiano2021adversarial}
Luca Robbiano, Muhammad Rameez~Ur Rahman, Fabio Galasso, Barbara Caputo, and
  Fabio~Maria Carlucci.
\newblock Adversarial branch architecture search for unsupervised domain
  adaptation, 2021.

\bibitem{russakovsky2015imagenet}
Olga Russakovsky, Jia Deng, Hao Su, Jonathan Krause, Sanjeev Satheesh, Sean Ma,
  Zhiheng Huang, Andrej Karpathy, Aditya Khosla, Michael Bernstein, et~al.
\newblock Imagenet large scale visual recognition challenge.
\newblock {\em International journal of computer vision}, 115(3):211--252,
  2015.

\bibitem{saenko2010adapting}
Kate Saenko, Brian Kulis, Mario Fritz, and Trevor Darrell.
\newblock Adapting visual category models to new domains.
\newblock In {\em European conference on computer vision}, pages 213--226.
  Springer, 2010.

\bibitem{saito2019semi}
Kuniaki Saito, Donghyun Kim, Stan Sclaroff, Trevor Darrell, and Kate Saenko.
\newblock Semi-supervised domain adaptation via minimax entropy.
\newblock {\em ICCV}, 2019.

\bibitem{saito2021tune}
Kuniaki Saito, Donghyun Kim, Piotr Teterwak, Stan Sclaroff, Trevor Darrell, and
  Kate Saenko.
\newblock Tune it the right way: Unsupervised validation of domain adaptation
  via soft neighborhood density, 2021.

\bibitem{saito2017asymmetric}
Kuniaki Saito, Yoshitaka Ushiku, and Tatsuya Harada.
\newblock Asymmetric tri-training for unsupervised domain adaptation.
\newblock In {\em International Conference on Machine Learning}, pages
  2988--2997. PMLR, 2017.

\bibitem{saito2018maximum}
Kuniaki Saito, Kohei Watanabe, Yoshitaka Ushiku, and Tatsuya Harada.
\newblock Maximum classifier discrepancy for unsupervised domain adaptation.
\newblock In {\em Proceedings of the IEEE conference on computer vision and
  pattern recognition}, pages 3723--3732, 2018.

\bibitem{Saito_2018_ECCV}
Kuniaki Saito, Shohei Yamamoto, Yoshitaka Ushiku, and Tatsuya Harada.
\newblock Open set domain adaptation by backpropagation.
\newblock In {\em Proceedings of the European Conference on Computer Vision
  (ECCV)}, September 2018.

\bibitem{sankaranarayanan2018generate}
Swami Sankaranarayanan, Yogesh Balaji, Carlos~D Castillo, and Rama Chellappa.
\newblock Generate to adapt: Aligning domains using generative adversarial
  networks.
\newblock In {\em Proceedings of the IEEE Conference on Computer Vision and
  Pattern Recognition}, pages 8503--8512, 2018.

\bibitem{ICML2012Shi_566}
Yuan Shi and Fei Sha.
\newblock Information-theoretical learning of discriminative clusters for
  unsupervised domain adaptation.
\newblock In John Langford and Joelle Pineau, editors, {\em Proceedings of the
  29th International Conference on Machine Learning (ICML-12)}, ICML '12, pages
  1079--1086, New York, NY, USA, July 2012. Omnipress.

\bibitem{shu2018dirt}
Rui Shu, Hung~H Bui, Hirokazu Narui, and Stefano Ermon.
\newblock A dirt-t approach to unsupervised domain adaptation.
\newblock {\em arXiv preprint arXiv:1802.08735}, 2018.

\bibitem{onelr2019}
Leslie~N. Smith and Nicholay Topin.
\newblock Super-convergence: very fast training of neural networks using large
  learning rates.
\newblock {\em Artificial Intelligence and Machine Learning for Multi-Domain
  Operations Applications}, May 2019.

\bibitem{sun2016return}
Baochen Sun, Jiashi Feng, and Kate Saenko.
\newblock Return of frustratingly easy domain adaptation.
\newblock In {\em Proceedings of the AAAI Conference on Artificial
  Intelligence}, volume~30, 2016.

\bibitem{toldo2020unsupervised}
Marco Toldo, Andrea Maracani, Umberto Michieli, and Pietro Zanuttigh.
\newblock Unsupervised domain adaptation in semantic segmentation: a review.
\newblock {\em Technologies}, 8(2):35, 2020.

\bibitem{tzeng2015simultaneous}
Eric Tzeng, Judy Hoffman, Trevor Darrell, and Kate Saenko.
\newblock Simultaneous deep transfer across domains and tasks.
\newblock In {\em Proceedings of the IEEE international conference on computer
  vision}, pages 4068--4076, 2015.

\bibitem{tzeng2017adversarial}
Eric Tzeng, Judy Hoffman, Kate Saenko, and Trevor Darrell.
\newblock Adversarial discriminative domain adaptation.
\newblock In {\em Proceedings of the IEEE conference on computer vision and
  pattern recognition}, pages 7167--7176, 2017.

\bibitem{Vaswani2017AttentionIA}
Ashish Vaswani, Noam~M. Shazeer, Niki Parmar, Jakob Uszkoreit, Llion Jones,
  Aidan~N. Gomez, Lukasz Kaiser, and Illia Polosukhin.
\newblock Attention is all you need.
\newblock {\em ArXiv}, abs/1706.03762, 2017.

\bibitem{venkateswara2017deep}
Hemanth Venkateswara, Jose Eusebio, Shayok Chakraborty, and Sethuraman
  Panchanathan.
\newblock Deep hashing network for unsupervised domain adaptation.
\newblock In {\em Proceedings of the IEEE conference on computer vision and
  pattern recognition}, pages 5018--5027, 2017.

\bibitem{rw2019timm}
Ross Wightman.
\newblock Pytorch image models.
\newblock \url{https://github.com/rwightman/pytorch-image-models}, 2019.

\bibitem{wu2020dual}
Yuan Wu, Diana Inkpen, and Ahmed El-Roby.
\newblock Dual mixup regularized learning for adversarial domain adaptation.
\newblock In {\em European Conference on Computer Vision}, pages 540--555.
  Springer, 2020.

\bibitem{xu2020adversarial}
Minghao Xu, Jian Zhang, Bingbing Ni, Teng Li, Chengjie Wang, Qi Tian, and
  Wenjun Zhang.
\newblock Adversarial domain adaptation with domain mixup.
\newblock In {\em Proceedings of the AAAI Conference on Artificial
  Intelligence}, volume~34, pages 6502--6509, 2020.

\bibitem{xu2019larger}
Ruijia Xu, Guanbin Li, Jihan Yang, and Liang Lin.
\newblock Larger norm more transferable: An adaptive feature norm approach for
  unsupervised domain adaptation.
\newblock In {\em Proceedings of the IEEE/CVF International Conference on
  Computer Vision}, pages 1426--1435, 2019.

\bibitem{you2019universal}
Kaichao You, Mingsheng Long, Zhangjie Cao, Jianmin Wang, and Michael~I Jordan.
\newblock Universal domain adaptation.
\newblock In {\em Proceedings of the IEEE/CVF conference on computer vision and
  pattern recognition}, pages 2720--2729, 2019.

\bibitem{pmlr-v97-you19a}
Kaichao You, Ximei Wang, Mingsheng Long, and Michael Jordan.
\newblock Towards accurate model selection in deep unsupervised domain
  adaptation.
\newblock In Kamalika Chaudhuri and Ruslan Salakhutdinov, editors, {\em
  Proceedings of the 36th International Conference on Machine Learning},
  volume~97 of {\em Proceedings of Machine Learning Research}, pages
  7124--7133. PMLR, 09--15 Jun 2019.

\bibitem{zhang2019domain}
Yabin Zhang, Hui Tang, Kui Jia, and Mingkui Tan.
\newblock Domain-symmetric networks for adversarial domain adaptation.
\newblock In {\em Proceedings of the IEEE/CVF Conference on Computer Vision and
  Pattern Recognition}, pages 5031--5040, 2019.

\bibitem{reverseValidation}
Erheng Zhong, Wei Fan, Qiang Yang, Olivier Verscheure, and Jiangtao Ren.
\newblock Cross validation framework to choose amongst models and datasets for
  transfer learning.
\newblock In Jos{\'e}~Luis Balc{\'a}zar, Francesco Bonchi, Aristides Gionis,
  and Mich{\`e}le Sebag, editors, {\em Machine Learning and Knowledge Discovery
  in Databases}, pages 547--562, Berlin, Heidelberg, 2010. Springer Berlin
  Heidelberg.

\end{thebibliography}
